\documentclass{article}
\pdfoutput=1

   \usepackage[square, comma, sort&compress, numbers]{natbib}

\usepackage{neurips_2021}

\usepackage[utf8]{inputenc} 
\usepackage[T1]{fontenc}    
\usepackage[colorlinks,linkcolor=black,citecolor=blue,urlcolor=blue,bookmarks=true]{hyperref}
\usepackage{url}            
\usepackage{booktabs}       
\usepackage{amsfonts}       
\usepackage{nicefrac}       
\usepackage{microtype}      
\usepackage{xcolor}         

\usepackage{times}
\usepackage{graphicx}
\usepackage{amsmath}
\usepackage{amssymb}
\usepackage{xcolor}
\usepackage{adjustbox}
\usepackage{caption}
\usepackage{wrapfig}
\usepackage{pifont}
\usepackage{amsmath}
\usepackage{multirow}

\newcommand{\SEUNG}[1]{{\color{red} {[Seung: #1]}}}
\newcommand{\DL}[1]{{\color{blue} {[Daiqing: #1]}}}

\usepackage{amsmath,amsfonts,bm}

\def\eqref#1{equation~\ref{#1}}

\def\1{\bm{1}}

\def\rvw{{\mathbf{w}}}
\def\rvx{{\mathbf{x}}}
\def\rvy{{\mathbf{y}}}
\def\rvz{{\mathbf{z}}}

\DeclareMathAlphabet{\mathsfit}{\encodingdefault}{\sfdefault}{m}{sl}
\SetMathAlphabet{\mathsfit}{bold}{\encodingdefault}{\sfdefault}{bx}{n}

\newcommand{\bzero}{\mathbf{0}}
\newcommand{\beye}{\mathbf{I}}
\newcommand{\bphi}{{\boldsymbol{\phi}}}
\newcommand{\bpsi}{{\boldsymbol{\psi}}}

\title{EditGAN: High-Precision Semantic Image Editing}

\author{\quad Huan Ling$^{1,2,3,}$\thanks{These authors contributed equally.}  \quad\quad\quad Karsten Kreis$^{1,*}$  \quad\quad\quad Daiqing Li $^{1}$  \\
\AND
  \quad \quad \quad \; Seung Wook Kim$^{1,2,3}$   \quad\quad\quad Antonio Torralba$^{4}$    \quad\quad\quad Sanja Fidler$^{1,2,3}$ \\
  \\
  \; \small{\textsuperscript{1}NVIDIA \quad \textsuperscript{2}University of Toronto \quad \textsuperscript{3}Vector Institute \quad \textsuperscript{4}MIT \vspace{3pt}}\\
  \; \quad \texttt{\scriptsize \{huling,kkreis,daiqingl,seungwookk,sfidler\}@nvidia.com,} \texttt{\scriptsize   torralba@mit.edu}\\
}

\begin{document}

\maketitle



\begin{abstract}
  Generative adversarial networks (GANs) have recently found applications in image editing.
  However, most GAN-based image editing methods often require large-scale datasets with semantic segmentation annotations for training, only provide high level control, or merely interpolate between different images. 
  Here, we propose \textit{EditGAN}, a novel method for high-quality, high-precision semantic image editing, allowing users to edit images by modifying their highly detailed part segmentation masks, e.g., drawing a new mask for the headlight of a car.  EditGAN builds on a GAN framework that jointly models images and their semantic segmentations~\cite{zhang2021datasetgan,li2021semantic}, requiring only a handful of labeled examples -- making it a scalable tool for editing. 
  Specifically, we embed an image into the GAN's latent space and perform  conditional  latent code optimization according to the segmentation edit, which effectively also modifies the image. 
  To amortize optimization, we find ``editing vectors'' in latent space that realize the edits. The framework allows us to learn an arbitrary number of editing vectors, which can then be directly applied on other images at interactive rates. 
  We experimentally show that EditGAN can manipulate images with an unprecedented level of detail and freedom, while preserving full image quality.We can also easily combine multiple edits and perform plausible edits beyond EditGAN's training data. We demonstrate EditGAN on a wide variety of image types and quantitatively outperform several previous editing methods on standard editing benchmark tasks. Project page: {\footnotesize\url{https://nv-tlabs.github.io/editGAN}}.
\end{abstract}

\vspace{-2mm}
\section{Introduction}
\vspace{-1mm}
\begin{wrapfigure}[10]{r}{0.64\textwidth}
\vspace{-8mm}

 \vspace{-5mm}
{\small
\begin{tabular}{p{1.9cm}p{1.7cm}p{1.6cm}p{2cm}}
&{\scriptsize Change Shape
} & {\scriptsize Enlarge Wheels
} & {\scriptsize  Shrink Front-Light
} \\
\end{tabular}}

\includegraphics[width=1\linewidth, trim=0 50 0 50,clip]{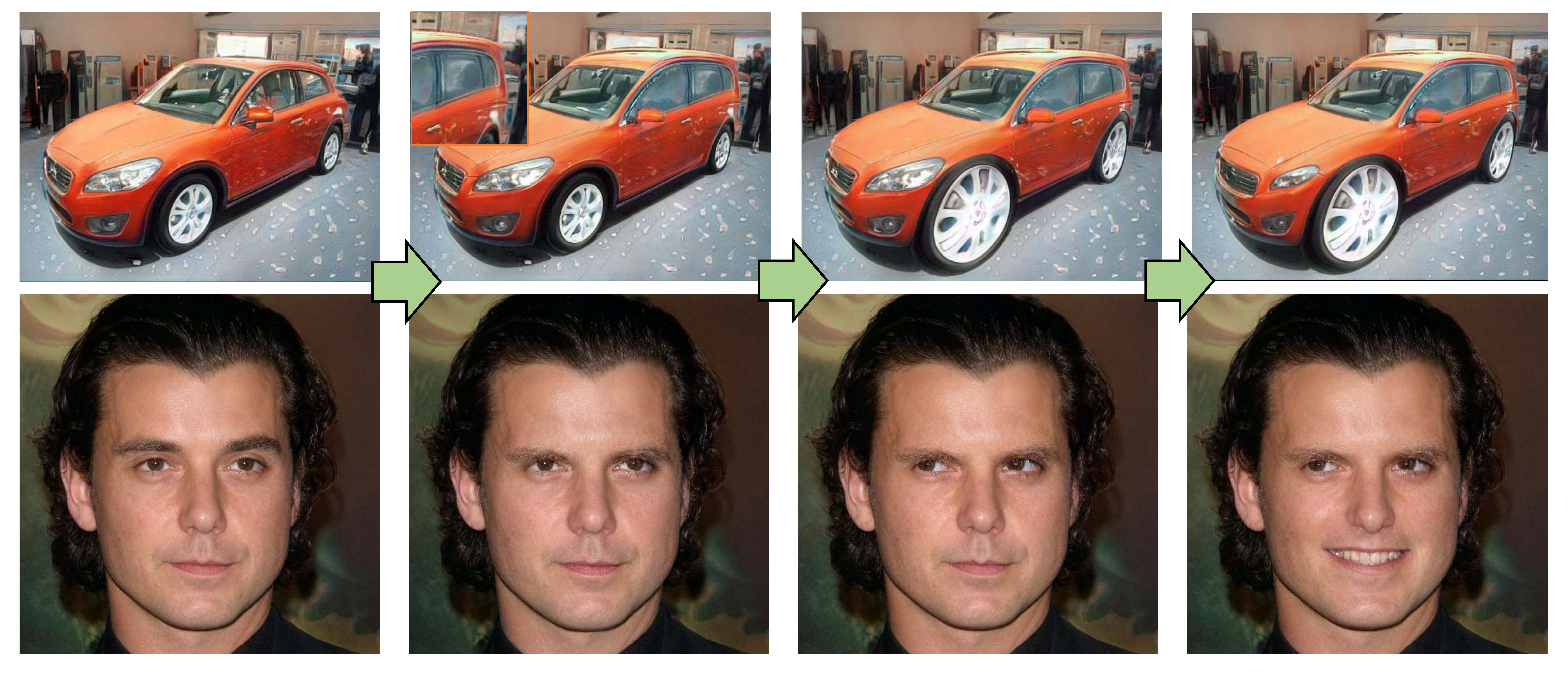}\\
 \vspace{-5mm}
{\small
\begin{tabular}{p{2.2cm}p{1.6cm}p{1.9cm}p{1.5cm}}
&{\scriptsize Frown
} & {\scriptsize Look Left
} & {\scriptsize  Smile
} \\[2mm]
\end{tabular}}

\caption{\footnotesize High-precision semantic image editing with EditGAN.} 

\end{wrapfigure}

AI-driven photo and image editing has the potential to streamline the workflow of photographers and content creators and to enable new levels of creativity and digital artistry~\cite{Bailey2020thetools}.
AI-based image editing tools have already found their way into consumer software in the form of neural photo editing filters, and the deep learning research community is actively developing further techniques. A particularly promising line of research uses generative adversarial networks (GANs)~\cite{goodfellow2014generative,radford2015unsupervised,karras2017progressive,karras2019style,karras2020analyzing} and either embeds images into the GAN's latent space or works directly with GAN-generated images. Careful modifications of the latent embeddings then translate to desired changes in generated output, allowing, for example, to coherently change facial expressions in portraits~\cite{choi2018stargan,lee2020maskgan,wu2020cascade,shen2020interpreting,shen2020interfacegan,alharbi2020disentangled,hou2020guidedstyle,Navigan_CVPR_2021}, change viewpoint or shapes and textures of cars~\cite{zhang2020image}, or to interpolate between different images in a semantically meaningful manner~\cite{collins2020editing,zhu2020sean,lewis2020vogue,kim2021stylemapgan}.

Most GAN-based image editing methods fall into few categories. Some works rely on GANs conditioning on class labels or pixel-wise semantic segmentation annotations~\cite{zhu2020sean,lee2020maskgan,chen2020deepfacedrawing,wu2020cascade}, where different conditionings lead to modifications in the output, while others use auxiliary attribute classifiers~\cite{he2019attgan,hou2020guidedstyle} 
to guide synthesis and edit images. However, training such conditional GANs or external classifiers requires large labeled datasets. 
Therefore, these methods are currently limited to image types for which large annotated datasets are available, like portraits~\cite{lee2020maskgan}. Furthermore, even if annotations are available, most techniques offer only limited editing control, since these annotations usually consist only of high-level global attributes or relatively coarse pixel-wise segmentations. 
Another line of work focuses on mixing and interpolating features from different images~\cite{collins2020editing,zhu2020sean,lewis2020vogue,kim2021stylemapgan}, thereby requiring reference images as editing targets and usually also not offering fine control. Other approaches carefully analyze and dissect GANs' latent spaces, finding disentangled latent variables suitable for editing~\cite{bau2019gandissect,bau2019semantic,shen2020interpreting,shen2020interfacegan,alharbi2020disentangled,plumerault2020Controlling,harkonen2020ganspace}, or control the GANs' network parameters~\cite{bau2019semantic,bau2020rewriting,Navigan_CVPR_2021}. Usually, these methods do not enable detailed editing and are often slow.

In this work, we are addressing
these limitations and propose \textit{EditGAN}, a novel GAN-based image editing framework that enables high-precision semantic image editing by allowing users to modify detailed object part segmentations. 
EditGAN builds on a recently proposed  GAN that jointly models both images and their semantic segmentations based on the same underlying latent code~\cite{zhang2021datasetgan,li2021semantic}, and requires as few as 16 labeled examples -- allowing it to scale to many object classes and choices of part labels.
We achieve editing by modifying the segmentation mask according to a desired edit and optimizing the latent code to be consistent with the new segmentation, thus effectively changing the RGB image. 
To achieve efficiency, we learn  \textit{editing vectors} in latent space that realize the edits, and that can be directly applied on other images, without any or only few additional  optimization steps. We can thus pre-train a library of interesting edits that a user can directly utilize in an interactive tool. 

We apply EditGAN on a wide range of images, including images of cars, cats, birds, and human faces, demonstrating unprecedented high-precision editing. We perform quantitative comparisons to multiple baselines and outperform them in metrics such as identity preservation, quality preservation, and target attribute accuracy, while requiring orders of magnitude less annotated training data. EditGAN is the first GAN-driven image editing framework, which simultaneously \textbf{(i)} offers very high-precision editing, \textbf{(ii)} requires only very little annotated training data (and does not rely on external classifiers), \textbf{(iii)} can be run interactively in real time, \textbf{(iv)} allows for straightforward compositionality of multiple edits, \textbf{(v)} and works on real embedded, GAN-generated, and even out-of-domain images.

\begin{figure*}[t!]
\vspace{-2.5mm}

\vspace{-4mm}
\begin{center}
\includegraphics[width=0.99\linewidth,clip]{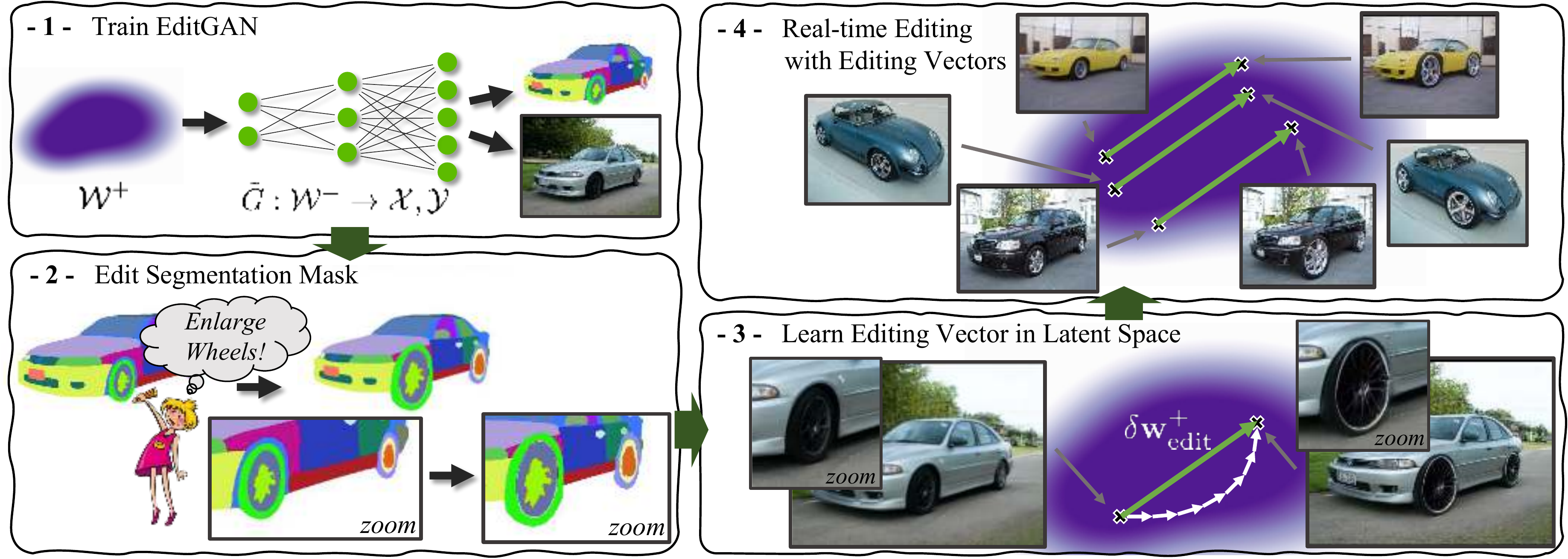}
\end{center}

\vspace{-2mm}
\caption{\footnotesize (\textbf{1}) EditGAN builds on a GAN framework that jointly models images and their semantic segmentations. (\textbf{2 \& 3}) Users can modify segmentation masks, based on which we perform optimization in the GAN’s latent space to realize the edit. (\textbf{4}) Users can perform editing simply by applying previously learnt editing vectors and manipulate images at interactive rates.}

\vspace{-5mm}
\end{figure*}


\vspace{-3mm}
\section{Related Work}
\vspace{-2mm}

\textbf{Image Editing and Manipulation.} Image Editing has a long history in computer vision and graphics, as well as machine learning~\cite{wolberg1994digital,efros2001image,hertzmann2001image,reinhard2001color,perez2003poisson,schaefer2006image,barnes2009patchmatch,tao2010error,gatys2016imagestyel,zhu2017unpaired,portenier2018faceshop,ling2020variational,collins2020editing,wu2020cascade,bau2020rewriting,park2020swapping,kim2021grivegan,Navigan_CVPR_2021}. Recently, deep generative models~\cite{goodfellow2014generative,kingma2014vae,rezende2014stochastic}, in particular modern GANs~\cite{karras2017progressive,brock2018large,karras2019style,park2019semantic,karras2020analyzing}, received much attention as a promising tool for efficient image editing, as it was found that latent space manipulations often lead to interpretable and predictable changes in output~\cite{goetschalckx2019ganalyze,bau2019gandissect,jahanian2020Osteerability,voynov2020unsupervised,plumerault2020Controlling,harkonen2020ganspace,wang2021geometric}.

GAN-based image editing methods can be broadly sorted into a number of categories. \textbf{(i)} One line of work relies on the careful dissection of the GAN's latent space, aiming to find interpretable and disentangled latent variables, which can be leveraged for image editing, in a fully unsupervised manner~\cite{goetschalckx2019ganalyze,bau2019gandissect,bau2019semantic,shen2020interpreting,shen2020interfacegan,alharbi2020disentangled,jahanian2020Osteerability,voynov2020unsupervised,plumerault2020Controlling,harkonen2020ganspace,wang2021geometric,shen2021closedform}. Although powerful, these approaches usually do not result in any high-precision editing capabilities. The editing vectors we are learning in EditGAN would be too hard to find independently without segmentation-based guidance. \textbf{(ii)} Other works utilize GANs that condition on class or pixel-wise semantic segmentation labels to control synthesis and achieve editing~\cite{choi2018stargan,wang2018high,park2019semantic,zhu2020sean,lee2020maskgan,chen2020deepfacedrawing,wu2020cascade}. Hence, these works usually rely on large annotated datasets, which are often not available, and even if available, the possible editing operations are tied to whatever labels are available. This stands in stark contrast to EditGAN, which can be trained in a semi-supervised fashion with very little labeled data and where an arbitrary number of high-precision edits can be learnt. \textbf{(iii)} Furthermore, auxiliary attribute classifiers have been used for image manipulation~\cite{he2019attgan,hou2020guidedstyle}, thereby still relying on annotated data and usually only providing high-level control. \textbf{(iv)} Image editing is often explored in the context of ``interpolating'' between a target and different reference image in sophisticated ways, for example by replacing certain features in a given image with features from a reference images~\cite{collins2020editing,zhu2020sean,lewis2020vogue,kim2021stylemapgan}. From the general image editing perspective, the requirement of reference images limits the broad applicability of these techniques and prevents the user from performing specific, detailed edits for which potentially no reference images are available.
\textbf{(v)} Recently, different works proposed to directly operate in the parameter space of the GAN instead of the latent space to realize different edits~\cite{bau2019semantic,bau2020rewriting,Navigan_CVPR_2021}. For example,~\cite{bau2019semantic,bau2020rewriting} essentially specialize the generator network for certain images at test time to aid image embedding or ``rewrite'' the network to achieve desired semantic changes in output. The drawback is that such specializations prevent the model from being used in real-time on different images and with different edits. 
~\cite{Navigan_CVPR_2021} proposed an approach that more directly analyses the parameter space of a GAN and treats it as a latent space in which to apply edits. 
However, the method still merely discovers edits in the network's parameter space, rather than actively defining them like we do. It remains unclear whether their method can combine multiple such edits, as we can, considering that they change the GAN parameters themselves. \textbf{(vi)} Finally, another line of research targets primarily very high-level image and photo stylization and global appearance modifications~\cite{gatys2016imagestyel,luan2017deepphoto,liu2017unsupervised,li2018eccv,wang2018high,kazemi2019style,park2019semantic,yoo2019photorealistic,park2020swapping}.

Generally, most works only do relatively high-level and not the detailed, high-precision editing, which EditGAN targets. 
Hence, we consider EditGAN as complementary to this body of work.

\textbf{GANs and Latent Space Image Embedding.} EditGAN builds on top of DatasetGAN~\cite{zhang2021datasetgan} and SemanticGAN~\cite{li2021semantic}, which  proposed to jointly model images and their semantic segmentations using shared latent codes. However, these works leveraged this model design only for semi-supervised learning, not for editing. EditGAN also relies on an encoder, together with optimization, to embed new images to be edited into the GAN's latent space. This task in itself has been studied extensively in different contexts before, and we are building on these works. Previous papers studied encoder-based methods~\cite{perarnau2016invertible,donahue2016adversarial,Brock2017neural,dumoulin2017adversarially,richardson2020encoding}, used primarily optimization-based techniques~\cite{zhu2016generative,Yeh2017,lipton2017precise,abdal2019image2stylegan,huh2020transforming,creswell2019inverting,Raj2019gan,plumerault2020Controlling}, and developed hybrid approaches~\cite{zhu2016generative,bau2019gandissect,bau2019semantic,bau2019seeing,zhu2020domain}. 

Finally, a concurrent paper~\cite{xu2021linear} shares similarities with DatasetGAN~\cite{zhang2021datasetgan}, on which our method builds, and explores an editing approach related to our EditGAN as one of its applications. However, our editing approach is methodologically different and leverages editing vectors, and also demonstrates significantly more diverse and stronger experimental results. Furthermore, \cite{bau2021paint} shares some high-level ideas with EditGAN; however, it leverages the CLIP~\cite{radford2021learning} model and targets text-driven editing. 
\vspace{-4mm}
\section{High-Precision Semantic Image Editing with EditGAN} \label{sec:method}
\vspace{-2mm}


\subsection{Background} \label{sec:background}
\vspace{-1mm}

EditGAN's image generation component is StyleGAN2~\cite{karras2019style, karras2020analyzing}, currently the state-of-the-art GAN for image synthesis. The StyleGAN2 generator maps latent codes $\rvz \in \mathcal{Z}$,
drawn from a multivariate Normal distribution, 
into realistic images. A latent code $\rvz$ is first transformed into an intermediate code $\rvw \in \mathcal{W}$ by a non-linear mapping function 
and then further transformed into $K+1$ vectors, $\rvw^0,...,\rvw^K$, through 
learned affine transformations. These 
transformed latent codes are fed into synthesis blocks, whose outputs are deep feature maps.

Deep generative models such as StyleGAN2, which are trained to synthesize highly realistic images, acquire a semantic understanding of the modeled images in their high-dimensional feature space. Recently, DatasetGAN~\cite{zhang2021datasetgan} and SemanticGAN~\cite{li2021semantic} built on this insight to learn a joint distribution $p(\rvx, \rvy)$ over images $\rvx$ and pixel-wise semantic segmentation labels $\rvy$, while requiring only a handful of labeled examples. EditGAN utilizes this joint distribution $p(\rvx, \rvy)$ to perform high-precision semantic image editing of real and synthesized images. 

Both methods~\cite{zhang2021datasetgan,li2021semantic} model $p(\rvx, \rvy)$ by adding an additional segmentation branch 
to the image generator, which is a pre-trained StyleGAN~\cite{zhang2021datasetgan}. We follow DatasetGAN~\cite{zhang2021datasetgan}, which applies a simple three-layer multi-layer perceptron classifier on the layer-wise concatenated and appropriately upsampled feature maps.
This classifier operates on the concatenated feature maps in a per-pixel fashion and predicts the segmentation label of each pixel. 

\vspace{-3mm}
\subsection{Segmentation Training and Inference by Embedding Images into GAN's Latent Space} \label{sec:embedding}
\vspace{-2mm}

To both train the segmentation branch and perform segmentation on a new image, we embed an image into the GAN's latent space using an encoder and optimization. 
To this end, we build on previous works~\cite{abdal2019image2stylegan,richardson2020encoding,li2021semantic} and train an encoder that embeds images into $\mathcal{W}^+$ space, which is defined as $\mathcal{W}$ but where the $\rvw$’s are modeled independently~\cite{abdal2019image2stylegan,richardson2020encoding}. Our objectives to train this encoder consist of standard pixel-wise L2 and perceptual LPIPS reconstruction losses using both the real training data as well as samples from the GAN itself. For the GAN samples, we also explicitly regularize the encoder with the known underlying latent codes. 
In practice, we use the encoder to initialize images' latent space embeddings and then iteratively refine the latent code $\rvw^+$ via optimization, again using standard reconstruction objectives. 

In that way, we embed the annotated images $\rvx$ from a dataset labeled with semantic segmentations into latent space, and train the segmentation branch of the generator using standard supervised learning objectives, i.e., the cross entropy loss. We keep the image generator's weights frozen and only backpropagate the loss to the segmentation branch~\cite{zhang2021datasetgan}. After training the segmentation branch, we can formally define a generator $\tilde{G} : \mathcal{W}^+ \rightarrow  \mathcal{X}, \mathcal{Y} $ that models the joint distribution $p(\rvx, \rvy)$ of images $\rvx$ and semantic segmentations $\rvy$. Details about encoder and segmentation branch training as well as optimization for image embedding can be found in the Appendix.

 \vspace{-2mm}
\subsection{Finding Semantics in Latent Space via Segmentation Editing} \label{sec:editing}
\begin{wrapfigure}[13]{r}{0.62\textwidth}
\vspace{-8mm}
\begin{center}
\includegraphics[width=1\linewidth,clip]{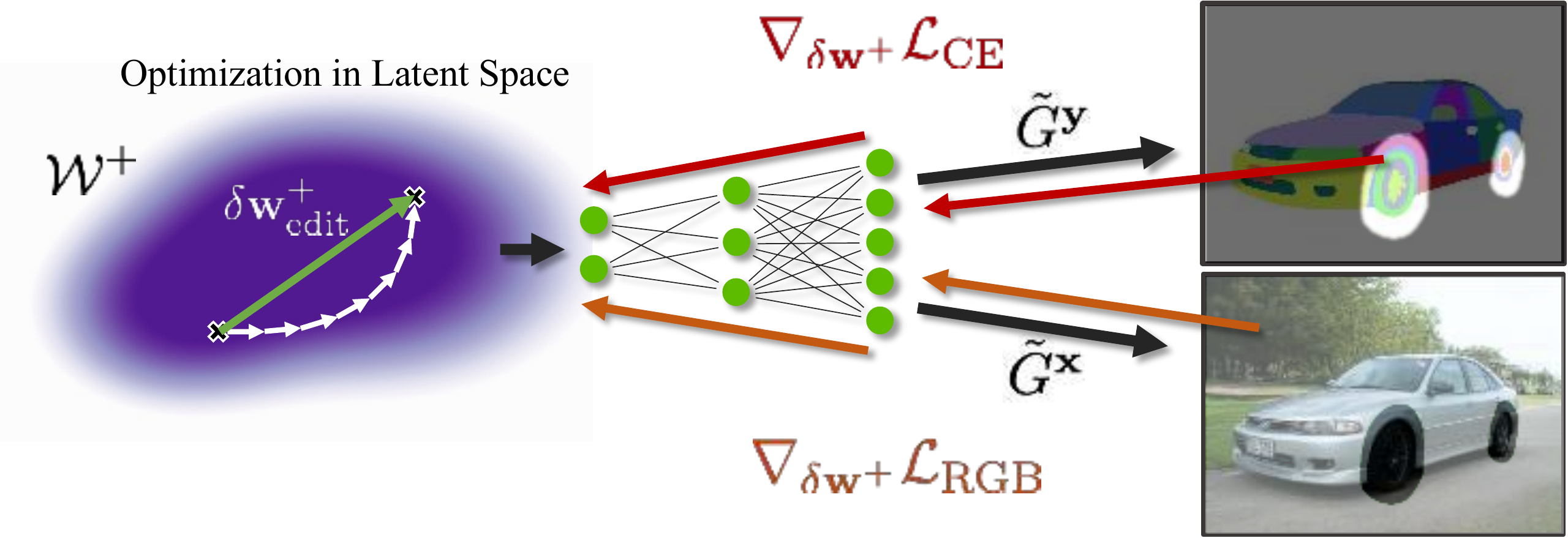}
\end{center}
\vspace{-2mm}
\caption{\footnotesize We modify semantic segmentations and optimize the shared latent code for consistency with the new segmentation \textit{within} the editing region, and with the RGB appearance \textit{outside} the editing region. Corresponding gradients are backpropagated through the shared generator. The result is a latent space editing vector $\delta \rvw^+_\textrm{edit}$.} 
\label{fig:editinggradient}
\vspace{-4mm}
\end{wrapfigure}

The key idea of EditGAN lies in leveraging the joint distribution $p(\rvx, \rvy)$ of images and semantic segmentations for high-precision image editing. Given a new image $\rvx$ to be edited, we can embed it into EditGAN's $\mathcal{W}^+$ latent space, as described above (alternatively, we can also sample images from the model itself and use those). The segmentation branch will then generate the corresponding segmentation $\rvy$, since segmentations and RGB images share the same latent codes $\rvw^+$. Using simple interactive digital painting or labeling tools, we can now manually modify the segmentation according to a desired edit. We denote the edited segmentation mask by $\rvy_\textrm{edited}$.
Starting from the embedding $\rvw^+$ of the unedited image $\rvx$ and segmentation $\rvy$, we can then perform optimization within $\mathcal{W}^+$ to find a new $\rvw^+_\textrm{edited}=\rvw^+ + \delta \rvw^+_\textrm{edit}$ consistent with the new segmentation $\rvy_\textrm{edited}$, while allowing the RGB output $\rvx$ to change within the editing region.

Formally, we are seeking an \textit{editing vector} $\delta \rvw^+_\textrm{edit}\in\mathcal{W}^+$ such that $(\rvx_\textrm{edited}, \rvy_\textrm{edited}) = \tilde{G}( \rvw^+ + \delta \rvw^+_\textrm{edit})$, where $\tilde{G}$ denotes the fixed generator that synthesizes both images and segmentations. Defining $(\rvx', \rvy') = \tilde{G}( \rvw^+ + \delta \rvw^+)$, we perform optimization to approximate $\delta \rvw^+_\textrm{edit}$ by $\delta \rvw^+$. The region of interest $r$ within which we expect the image to change due to the edit is formally given by 
\begin{gather}
r = \left\{ p : c^\rvy_p \in Q_\textrm{edit}  \right\} \cup \left\{ p : c^{\rvy_\textrm{edited}}_p \in Q_\textrm{edit}  \right\}
\end{gather}
which means that $r$ is defined by all pixels $p$ whose part segmentation labels $c_p^{\{\rvy,\rvy_\textrm{edited}\}}$ according to either the initial segmentation $\rvy$ or the edited one $\rvy_\textrm{edited}$ are within an edit-specific pre-specified list $Q_\textrm{edit}$ of part labels relevant for the edit. 
For example, when modifying the wheel in a photo of a car $Q_\textrm{edit}$ would contain all part labels related to the wheels, such as tire, spoke, and wheelhub (see Fig.~\ref{fig:editinggradient}). 
We use a further buffer of $5$ pixels to give the GAN freedom in modeling the transition between the edited and non-edited area. In practice, $r$ acts as a binary pixel-wise mask (see Eqs.~\ref{eq:editing_rgb} and \ref{eq:editing_ce} below).

Note that $\rvx_\textrm{edited}$ is not available during optimization. After all, $\rvx_\textrm{edited}$ is the edited image we are ultimately intested in. It emerges indirectly when optimizing for the segmentation modification, since images and segmentations are closely tied together in the joint distribution $p(\rvx, \rvy)$ modeled by $\tilde{G}$.
We further define $\rvx' = \tilde{G}^\rvx( \rvw^+ + \delta \rvw^+)$ as $\tilde{G}$'s image generation and $\rvy' = \tilde{G}^\rvy( \rvw^+ + \delta \rvw^+)$ as $\tilde{G}$'s segmentation generation branch. 


To find $\delta \rvw^+$, approximating $\delta \rvw^+_\textrm{edit}$, we use the following losses as minimization targets:
\begin{equation}\label{eq:editing_rgb}
\begin{split}
\mathcal{L}_\textrm{RGB} (\delta \rvw^+)  &=  L_\textrm{LPIPS}( \tilde{G}^\rvx( \rvw^+ + \delta \rvw^+) \odot(1 - r), \; \rvx \odot(1 - r) ) \\  &+  L_{L2}( \tilde{G}^\rvx( \rvw^+ + \delta \rvw^+) \odot(1 - r), \; \rvx  \odot(1 - r) )
\end{split}
\end{equation}
\begin{gather} \label{eq:editing_ce}
\mathcal{L}_\textrm{CE} (\delta \rvw^+) =  H( \tilde{G}^\rvy( \rvw^+ + \delta \rvw^+) \odot r, \; \rvy_\textrm{edited} \odot r  )
\end{gather}
where $H$ denotes the pixel-wise cross-entropy, $L_\textrm{LPIPS}$ loss is based on the Learned Perceptual Image Patch Similarity (LPIPS) distance~\cite{zhang2018unreasonable}, and $L_{L2}$ is a regular pixel-wise L2 loss.
$\mathcal{L}_\textrm{RGB} (\delta \rvw^+)$ ensures that the image appearance does not change \textit{outside} the region of interest, while $\mathcal{L}_\textrm{CE} (\delta \rvw^+)$ ensures that the target segmentation $\rvy_\textrm{edited}$ is enforced \textit{within} the editing region (see visualization in Fig.~\ref{fig:editinggradient}).
When editing human faces, we also apply the identity loss~\cite{richardson2020encoding}:
\begin{gather}\label{eq:idloss}
\mathcal{L}_\textrm{ID} (\delta \rvw^+) =  \langle R(\tilde{G}^\rvx( \rvw^+ + \delta \rvw^+)), R(\rvx) \rangle
\end{gather}
with $R$ denoting the pretrained ArcFace feature extraction network~\cite{deng2019arcface} and $\langle\cdot, \cdot\rangle$ cosine-similiarity.

The final objective function for optimization then becomes:
\begin{gather}
\mathcal{L}_\textrm{editing}(\delta \rvw^+) =  \lambda^{\textrm{editing}}_1 \mathcal{L}_\textrm{RGB} (\delta \rvw^+) + \lambda^{\textrm{editing}}_2 \mathcal{L}_\textrm{CE} (\delta \rvw^+)   + \lambda^{\textrm{editing}}_3 \mathcal{L}_\textrm{ID} (\delta \rvw^+)
\end{gather}
with hyperparameters $\lambda^{\textrm{editing}}_{1,...,3}$. The only ``learnable'' variable is the editing vector $\delta \rvw^+$; all neural networks are kept fixed. After optimizing $\delta \rvw^+$ with the objective function, we can use $\delta \rvw^+\approx\delta \rvw^+_\textrm{edit}$. Note that there is a certain amount of ambiguity in how the segmentation modification is realized in RGB output. We rely on the GAN generator, trained to synthesize realistic images, to modify the RGB values in the editing region in a plausible way consistent with the segmentation edit.

 \vspace{-2mm}
\subsection{Different Ways of Editing during Inference} \label{sec:inference}
The latent space editing vectors $\delta \rvw^+_\textrm{edit}$ obtained by optimization as described are semantically meaningful and often disentangled with other attributes. Therefore, for new images $\rvx$ to be edited, we can embed the images into the $\mathcal{W}^+$ latent space and the same editing operations can be directly performed by applying the previously learnt $\delta \rvw^+_\textrm{edit}$ as $(\rvx', \rvy') = G( \rvw^+ +  s_\textrm{edit}\,\delta \rvw^+_\textrm{edit})$ without doing any optimization from scratch again. In other words, the learnt editing vectors $\delta \rvw^+$ amortize the iterative optimization that was necessary to achieve the edit initially. For well-disentangled editing operations, $\rvx'$ can be used directly as the edited image $\rvx_\textrm{edited}$. Note that we introduced $s_\textrm{edit}$, a scalar editing coefficient, which effectively scales and controls the editing magnitude during inference. For $s_\textrm{edit}=0$, we do not do any editing at all, while for $s_\textrm{edit}>1$ we manipulate the images with an effectively larger editing operation in latent space, leading to exaggerated effects. 

Unfortunately, disentanglement is not always perfect and the editing vectors $\delta \rvw^+_\textrm{edit}$ do not always translate perfectly to other images. We can remove editing artifacts in other regions of the image by a few additional optimization steps at test time. Specifically, we can use the exact same minimization objectives as above, using the initial prediction $\rvy'$, obtained after applying the editing vector $\delta \rvw^+_\textrm{edit}$, as $\rvy_\textrm{edited}$. This assumes that the editing vector still induces a plausible segmentation change when applied on other images and that artifacts only arise in RGB output. The RGB objective $\mathcal{L}_\textrm{RGB}$ then removes these editing artifacts outside the editing region, while $\mathcal{L}_\textrm{CE}$ ensures that the modified segmentation stays as predicted by the editing vector.

Summarizing, we can perform image editing with EditGAN in three different modes:
\begin{itemize}
\item \textbf{Real-time Editing with Editing Vectors.} For localized, well-disentangled edits we perform editing purely by applying previously learnt editing vectors with varying scales $s_\textrm{edit}$ and manipulate images at interactive rates.
\item \textbf{Vector-based Editing with Self-Supervised Refinement.} For localized edits that are not perfectly disentangled with other parts of the image, we can remove editing artifacts by additional optimization at test time, while initializing the edit using the learnt editing vectors.
\item \textbf{Optimization-based Editing.} Image-specific and very large edits do not transfer to other images via editing vectors. For such operations, we perform optimization from scratch.
\end{itemize}

\begin{figure*}[t!]
\includegraphics[width=1\linewidth, trim=40 20 35 8,clip]{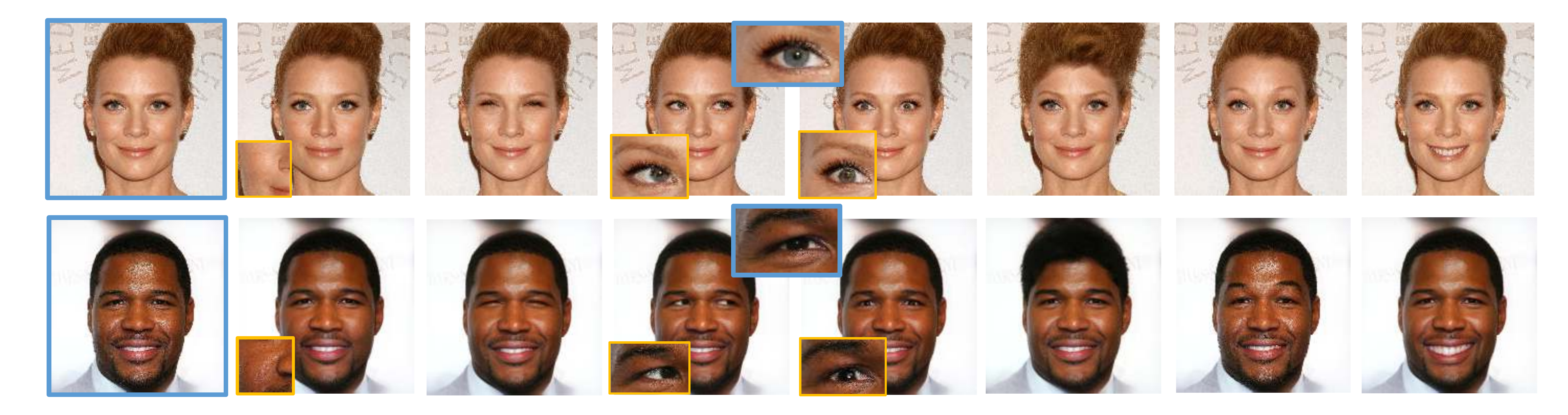}\\[-1mm]
 \begin{adjustbox}{width=0.965\linewidth}
 \vspace{-5mm}
{\small
\begin{tabular}{cccccccc}
\quad\quad\quad\ \ &{\scriptsize $\quad\ \ $De-wrinkle
} & {\scriptsize $\ \ \ $Close Eyes
} & {\scriptsize  Gaze Position
} & {\scriptsize  $\ $Gaze Pos. 2
} & {\scriptsize  $\ $Hairstyle
} & {\scriptsize $\ \ $Raise Eyebrows
} & {\scriptsize $$Smile

}\\
\end{tabular}}
\end{adjustbox}
\includegraphics[width=1\linewidth, trim=30 210 20 0,clip]{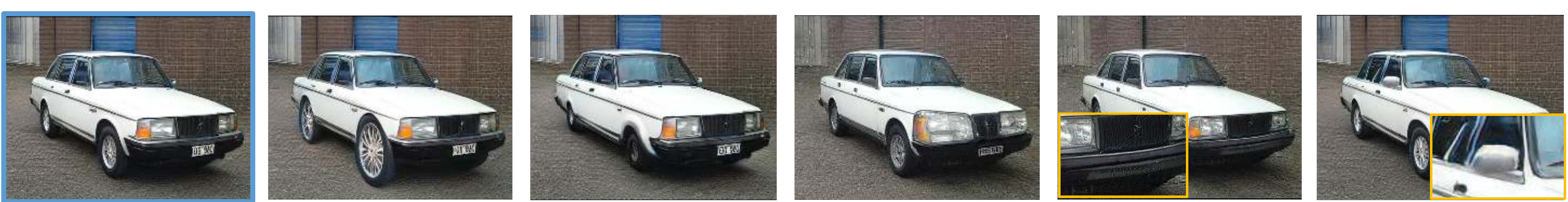}
\includegraphics[width=1\linewidth, trim=30 30 20 205,clip]{figs/plot/one_apply_all_car.pdf}\\[-0.5mm]
 \vspace{-5mm}
{\small
\begin{tabular}{p{2.1cm}p{2.0cm}p{1.6cm}p{2.0cm}p{1.85cm}p{2.5cm}}
\quad\quad\quad\quad\quad\quad\quad\ &{\scriptsize Enlarge Wheels
} & {\scriptsize Shrink Wheels
} & {\scriptsize  Enlarge Front Light
} & {\scriptsize  Add License-Plate
} & {\scriptsize  Remove Side Mirror
}\\[5mm]
\end{tabular}}
\includegraphics[width=1\linewidth, trim=45 10 45 10,clip]{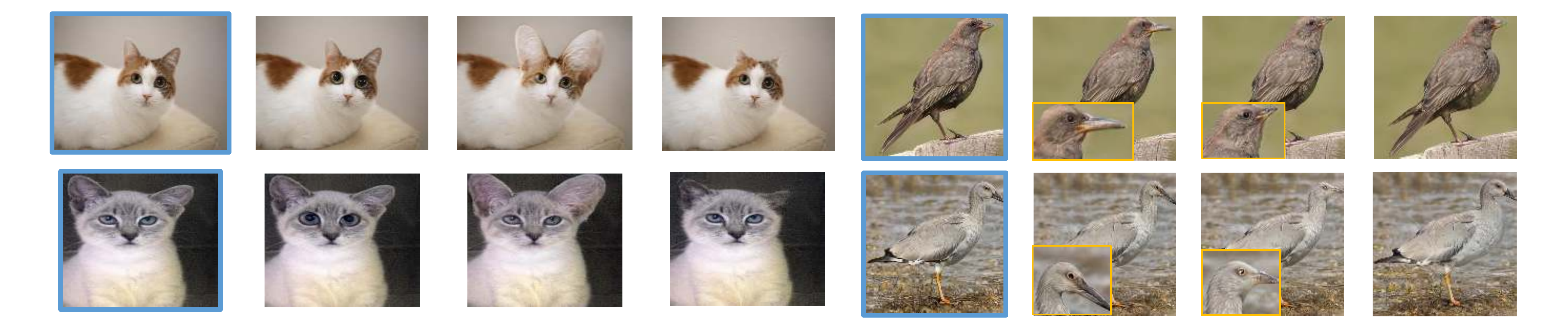}\\[-0mm]
\begin{minipage}{1\linewidth}
 \vspace{-3mm}
\begin{tabular}{p{1.55cm}p{1.55cm}p{1.6cm}p{2.0cm}}
&{\scriptsize Enlarge Eyes
} & {\scriptsize Enlarge Ear
} & {\scriptsize  Smaller Ear}
\end{tabular}
\hspace{-9.5mm}\begin{tabular}{p{1.35cm}p{1.35cm}p{1.0cm}p{2.0cm}}
& {\scriptsize  Longer Beak
} & {\scriptsize Head Up
} & {\scriptsize Bigger Belly}
\end{tabular}
\end{minipage}
\vspace{-6mm}
\caption{\footnotesize Examples of segmentation-driven edits with EditGAN. Results are based on editing with editing vectors and 30 steps self-supervised refinement. \textit{Blue boxes}: Original images. \textit{Orange boxes}: Zoom-in views.}
\label{fig:vis_apply_all}
\vspace{-3mm}
\end{figure*}

\begin{figure}[t!]
\vspace{-9mm}
\begin{center}
 \vspace{-2mm}
\includegraphics[width=0.85\linewidth, trim=0 0 0 0,clip]{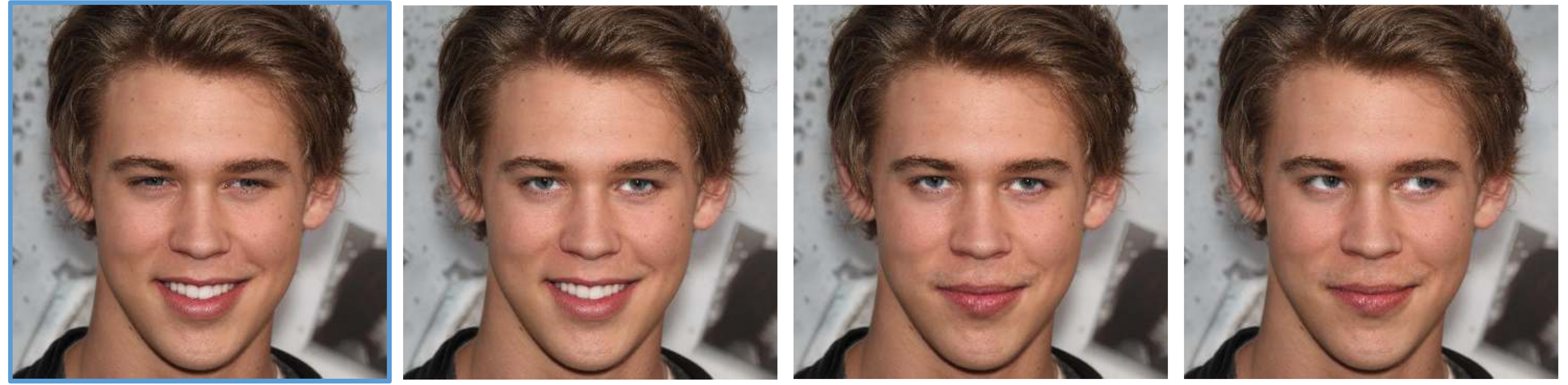}
\end{center}
 \vspace{-5mm}
\begin{center}
 \begin{adjustbox}{width=0.85\linewidth}
 \vspace{-3mm}
{\small
\begin{tabular}{cccc}
\quad\quad\quad\quad\quad\quad& {\scriptsize  + Open Eyes
} & {\scriptsize  + Close Mouth
} & {\scriptsize  + Look Right
}\\
\end{tabular}}
\end{adjustbox}
\end{center}
 \vspace{-5mm}
\begin{center}
\includegraphics[width=0.85\linewidth, trim=0 0 0 0,clip]{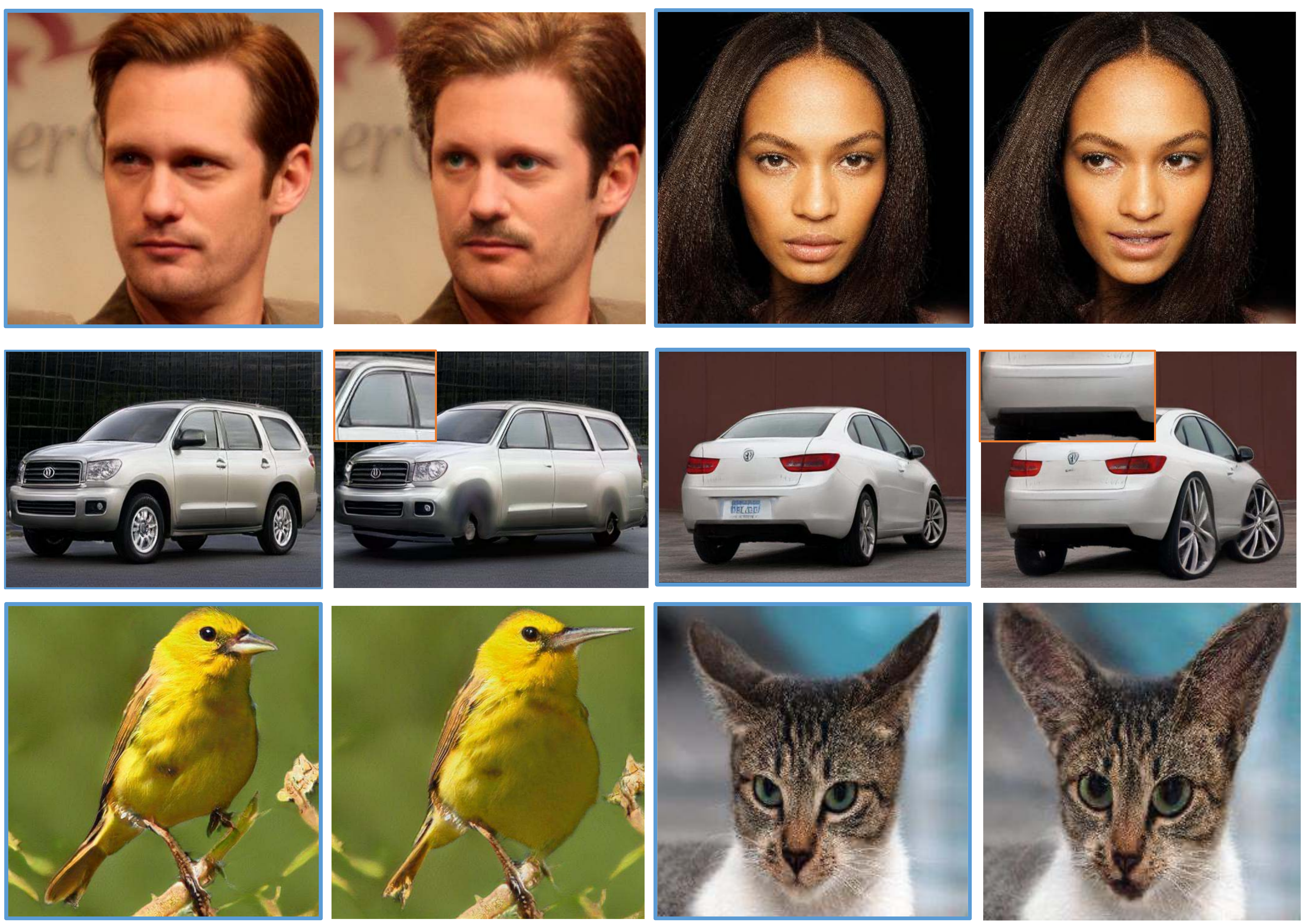}
\end{center}

\vspace{-4.0mm}
\caption{\footnotesize We combine multiple edits. Results are based on editing with editing vectors and 30 steps self-supervised refinement. \textit{Blue boxes}: Original images. Edits in detail: \textit{Second row, first person:} open eyes, add hair, add mustache. \textit{Second person:} smile, look left. \textit{Third row, first car:} remove mirror, remove door handle, shrink wheels. \textit{Second car:} remove license plate, enlarge wheels. \textit{Third row, bird:} longer beak, bigger belly, head up. \textit{Third row, cat:} open mouth, bigger ear, bigger eyes.}	
\label{fig:combine}
\vspace{-5mm}
\end{figure}

\vspace{-3mm}
\section{Experiments} \label{sec:experiments}
\vspace{-2mm}

We extensively evaluate EditGAN on images across four different categories: Cars ($384{\times}512$ spatial resolution), Birds ($512{\times}512$), Cats ($256{\times}256$), and Faces ($1024{\times}1024$). 

\vspace{-2mm}
\paragraph{Implementation} We train our segmentation branch as described in Sec.~\ref{sec:embedding}
using 16, 16, 30, and 30 image-mask pairs as labeled training data
for Faces, Cars, Birds, and Cats, respectively. We utilize very highly-detailed part segmentations from~\cite{zhang2021datasetgan}. The annotation scheme for faces is shown in Fig.~\ref{fig:faceannotation}, all others are presented in the Appendix. When editing is done purely optimization-based or when learning the editing vectors, we always perform 100 steps of optimization using Adam~\cite{kingma2014adam}. For Car, Cat, and Faces, we use real images from DatasetGAN's test set that were not part of GAN training to demonstrate editing functionality. These images are first embedded into EditGAN's latent space via an encoder and optimization as described in Sec.~\ref{sec:embedding}. For Birds, we show editing on GAN-generated images. Model details and hyperparameters are provided in the Appendix.

\vspace{-1.5mm}
\subsection{Qualitative Results}\label{sec:qualitative}
\vspace{-1.5mm}
\paragraph{In-Domain Results} In Fig.~\ref{fig:vis_apply_all}, we demonstrate our EditGAN framework when applying previously learnt editing vectors $\delta \rvw^+_\textrm{edit}$ on novel images and refining with 30 steps of optimization.
Our editing operations preserve high image quality and are well disentangled for all classes. We also show the ability to combine multiple different edits in Fig.~\ref{fig:combine}. To the best of our knowledge, no previous methods can perform as complex and high-precision edits as we do, while preserving image quality and subject identity. In Fig.~\ref{fig:vis_detail_editing}, we demonstrate that we can even perform extremely high-precision edits, such as rotating a car's wheel spoke or dilating pupils. EditGAN can edit semantic parts of objects that consist of only few pixels. At the same time, we can use EditGAN to perform large-scale modifications, too: In Fig.~\ref{fig:large_edits}, we present how we can remove the entire roof of a car or convert it to a station wagon-like vehicle, simply by modifying the segmentation mask accordingly and optimizing. It is worth noting that several of our editing operations generate plausible manipulated images unlike those appearing in the GAN training data. For example, the training data does not include cats with overly large eyes or ears. Nevertheless, we achieve such edits in a high-quality manner.  

The edits in Figs.~\ref{fig:vis_apply_all},~\ref{fig:combine} and~\ref{fig:vis_detail_editing} are based on learnt editing vectors with self-supervised refinement. However, without such refinement usually only very minor artifacts occur, as shown in Fig.~\ref{fig:mask_change_vis}, hence allowing for real-time high-precision semantic image editing (discussed in detail below).

\begin{figure}[t!]

\begin{center}
 \begin{adjustbox}{width=0.8\linewidth}
 \vspace{-3mm}
{\small
\begin{tabular}{cccc}
\includegraphics[width=1\linewidth, trim=0 0 0 0,clip]{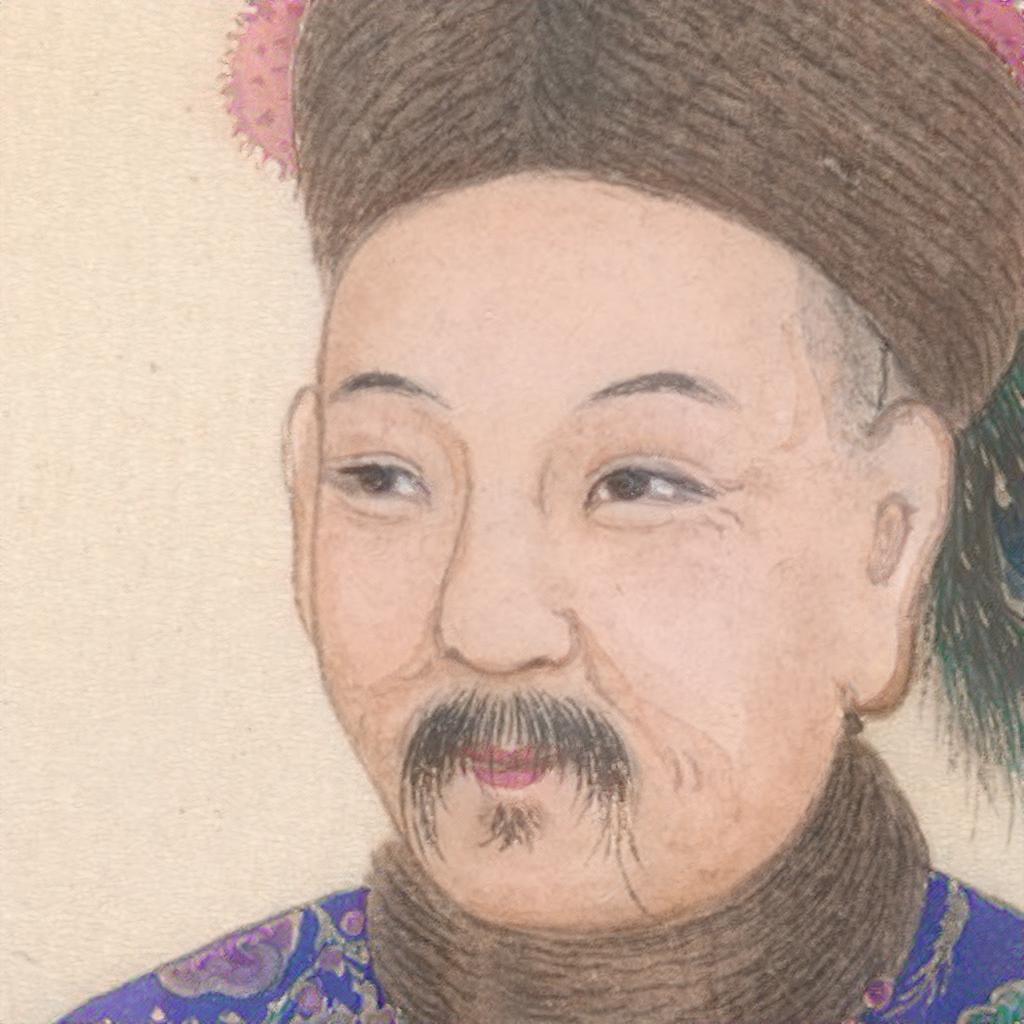}
&  \includegraphics[width=1\linewidth, trim=0 0 0 0,clip]{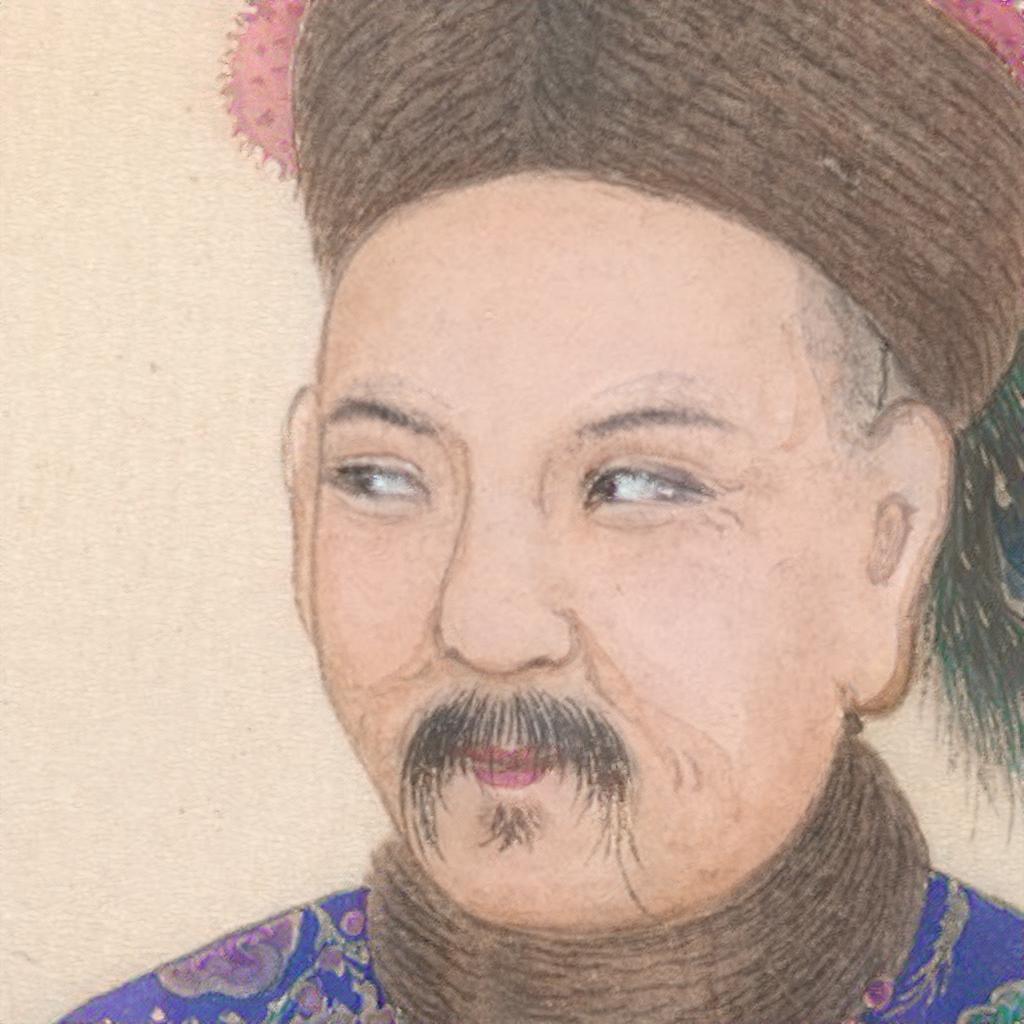} 
& \includegraphics[width=1\linewidth, trim=0 0 0 0,clip]{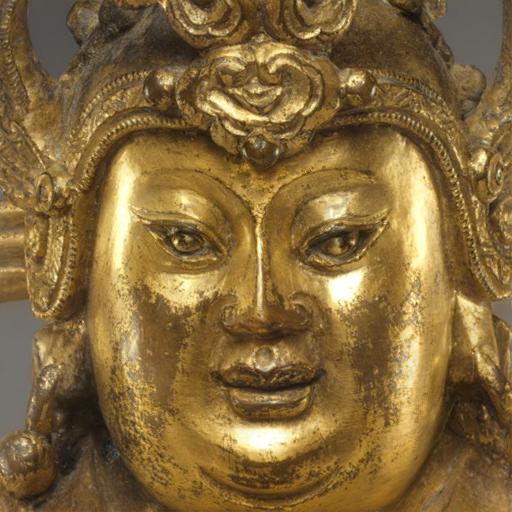}
 & \includegraphics[width=1\linewidth, trim=0 0 0 0,clip]{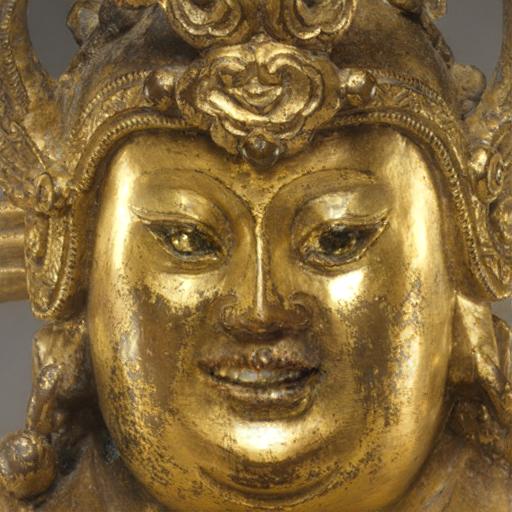}
\\

\includegraphics[width=1\linewidth, trim=0 0 0 0,clip]{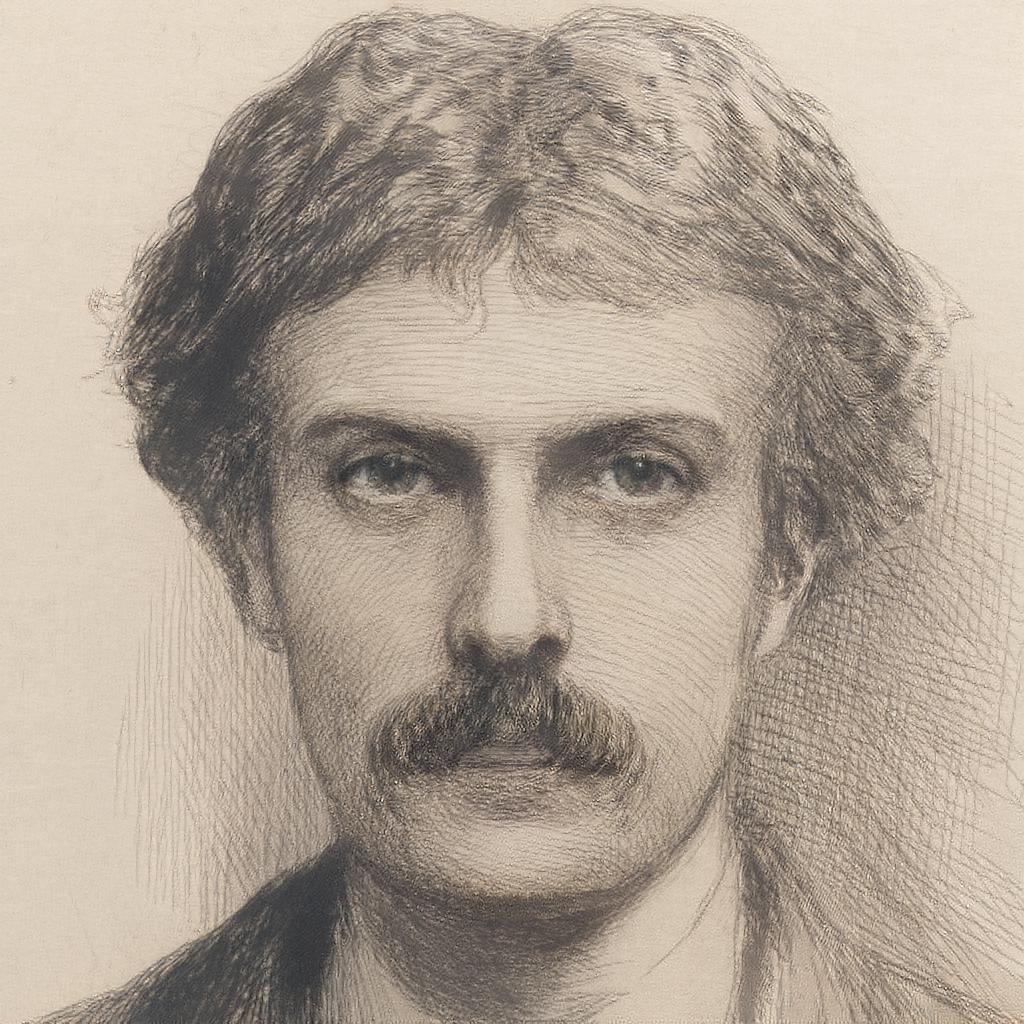}
&  \includegraphics[width=1\linewidth, trim=0 0 0 0,clip]{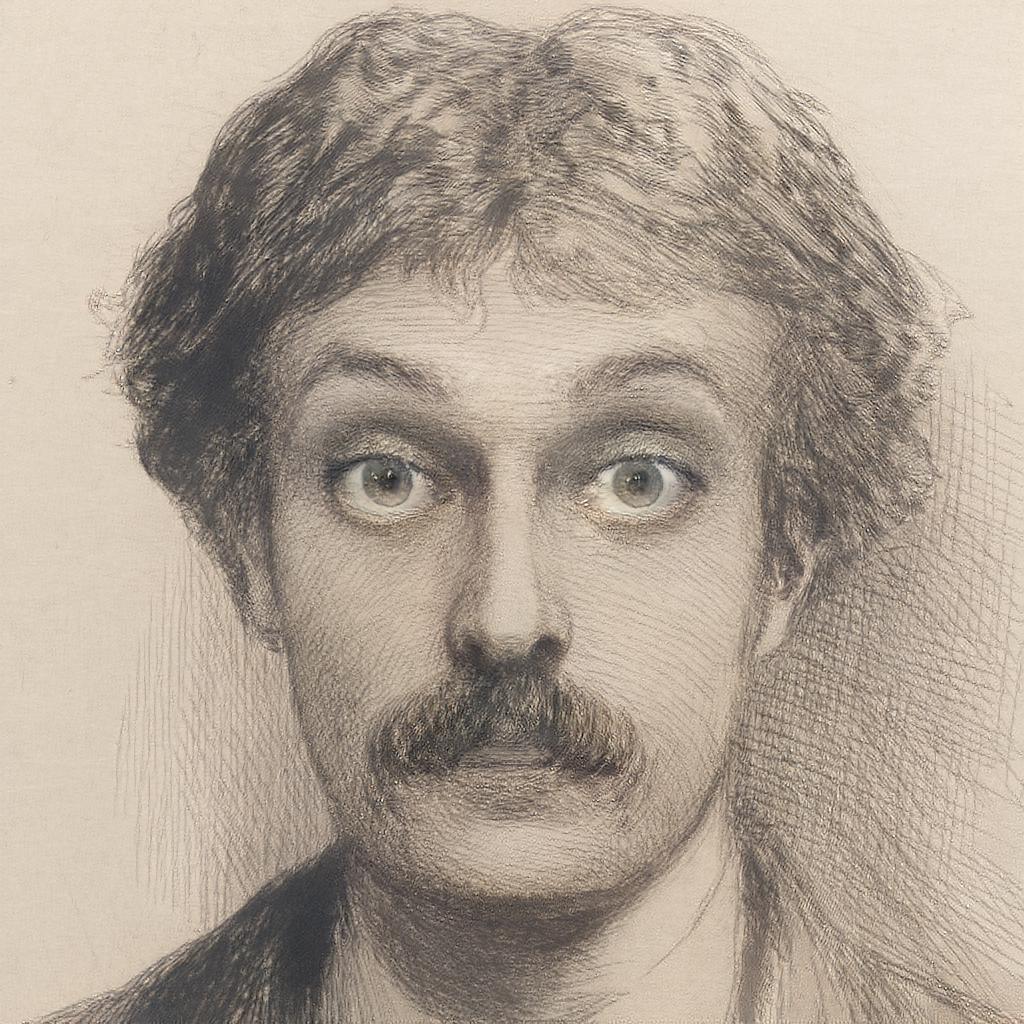} 
& \includegraphics[width=1\linewidth, trim=0 0 0 0,clip]{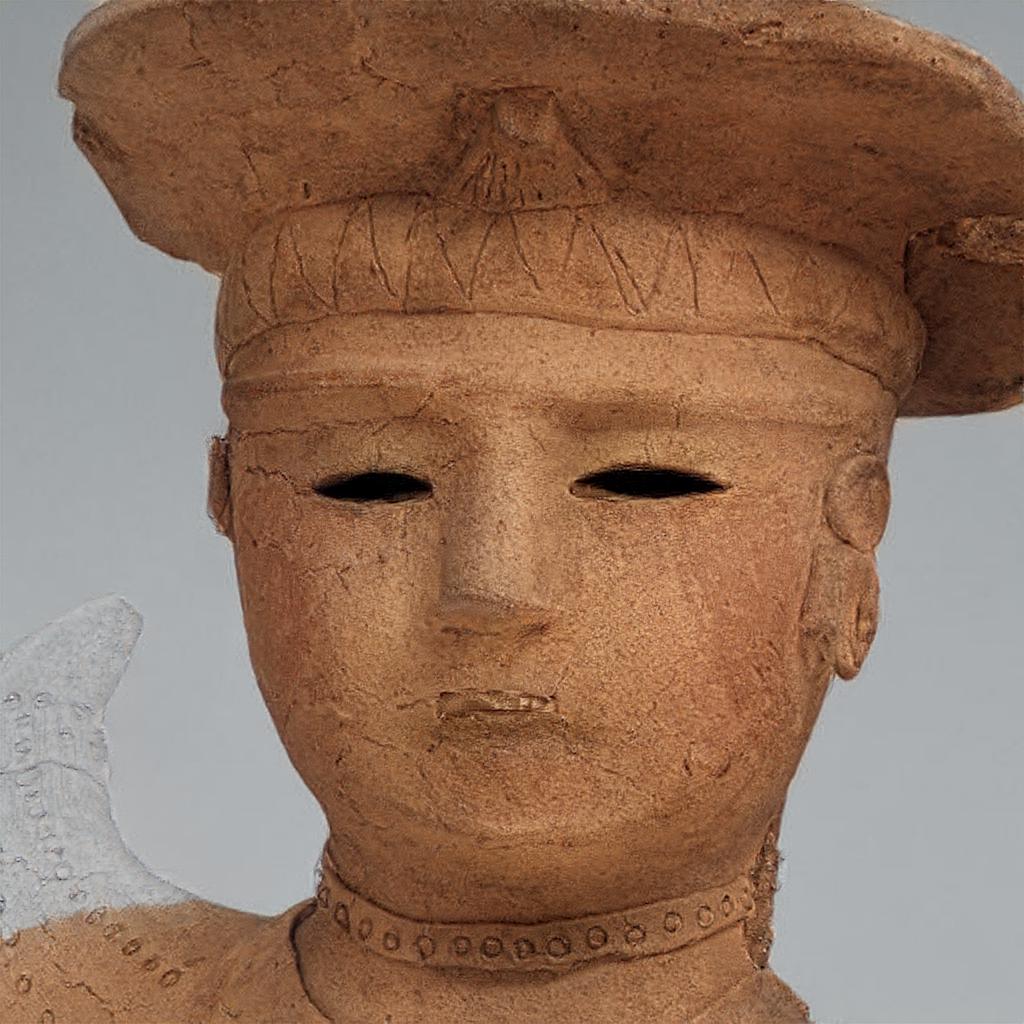}
 & \includegraphics[width=1\linewidth, trim=0 0 0 0,clip]{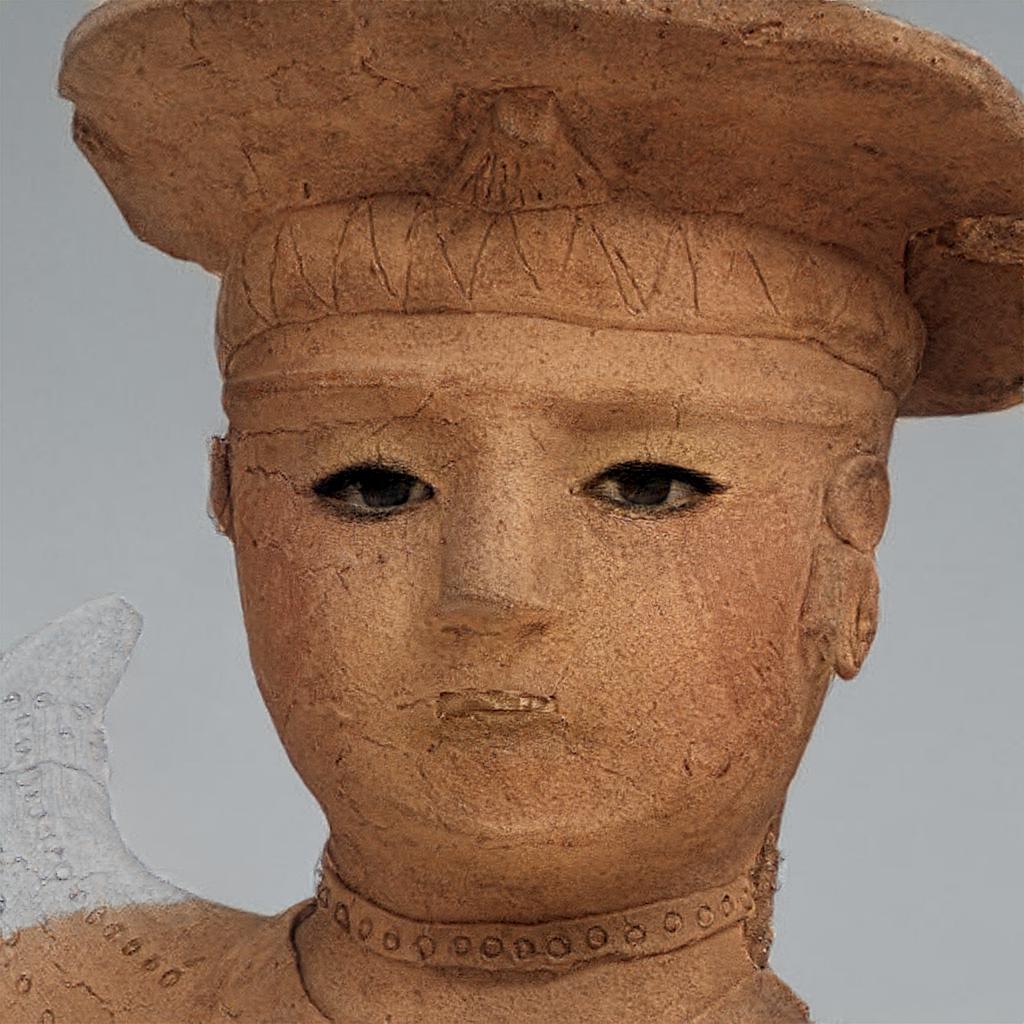}
\\

\end{tabular}}
\end{adjustbox}
\end{center}

\vspace{-4.0mm}
\caption{\footnotesize We combine multiple edits on out-of-domain images. Results are based on editing with editing vectors and 30 steps self-supervised refinement. Edits in detail: \textit{First row, first example:} look left, frown. \textit{Second example:} smile, look right. \textit{Second row, first example:} open eyes, lift eyebrow. \textit{Second example:} open eyes.}	
\label{fig:ood}
\vspace{-5mm}
\end{figure}

\begin{figure*}[t!]
\vspace{-1mm}
\begin{minipage}{.22\linewidth}
\begin{center}
\includegraphics[width=1\linewidth, trim=10 0 8 0,clip]{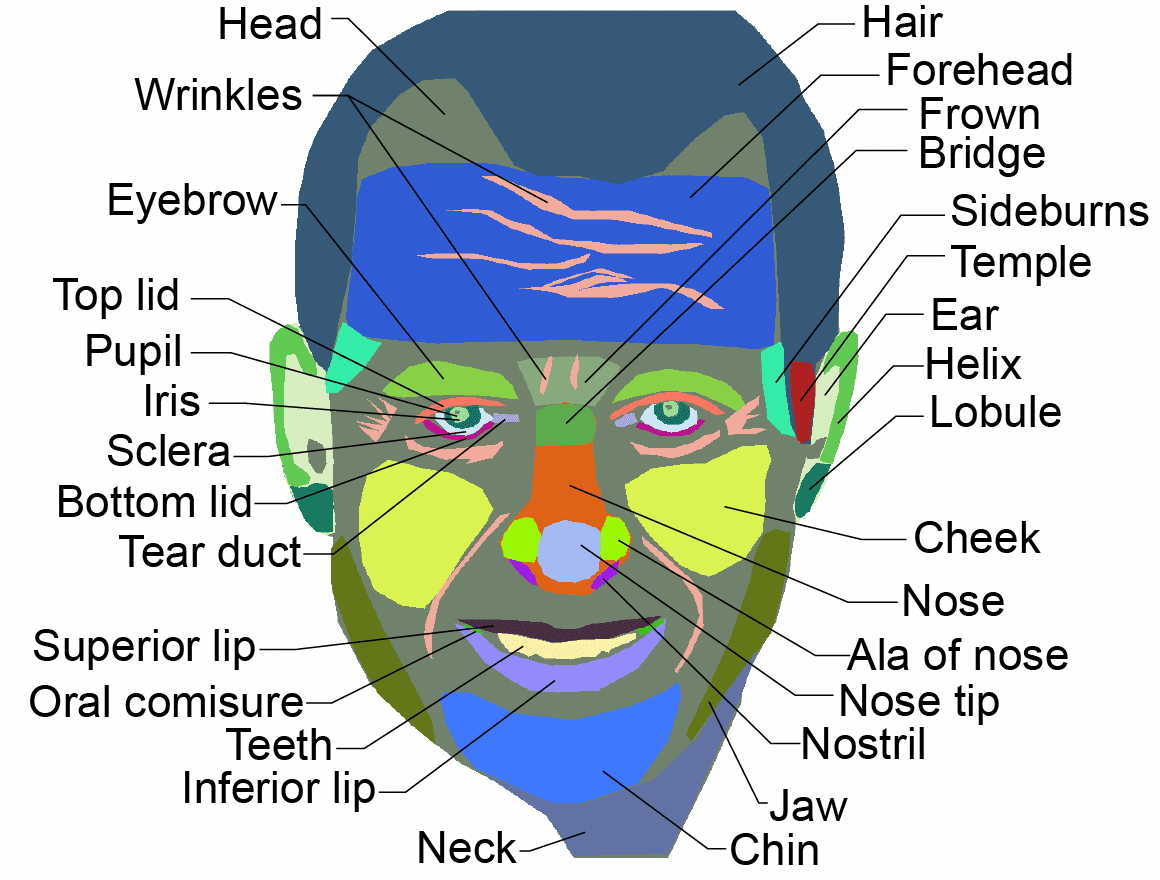}
\vspace{-4mm}
\caption{\footnotesize Face part labeling schema~\cite{zhang2021datasetgan}.}	
\label{fig:faceannotation}
\end{center}
\end{minipage}
\hfill
\begin{minipage}{.75\linewidth}
\begin{center}
\includegraphics[width=1\linewidth, trim=70 100 60 0,clip]{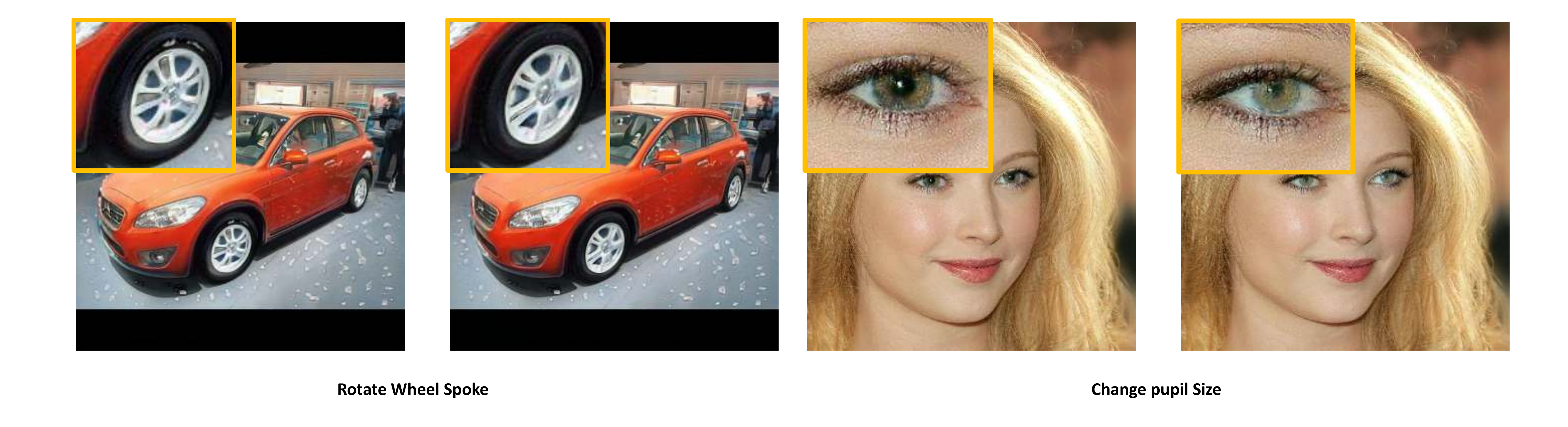}
\vspace{-5mm}
\caption{\footnotesize High-precision editing with EditGAN for extreme details. \textit{Left:} We rotate the spoke. \textit{Right:} We modify pupil size. Results are based on editing with editing vectors and 30 steps self-supervised refinement.}
\label{fig:vis_detail_editing}
\end{center}
\end{minipage}
\vspace{-3mm}
\end{figure*}

\begin{figure*}[t!]
\vspace{-0mm}
\includegraphics[width=1\linewidth, trim=0 140 0 100,clip]{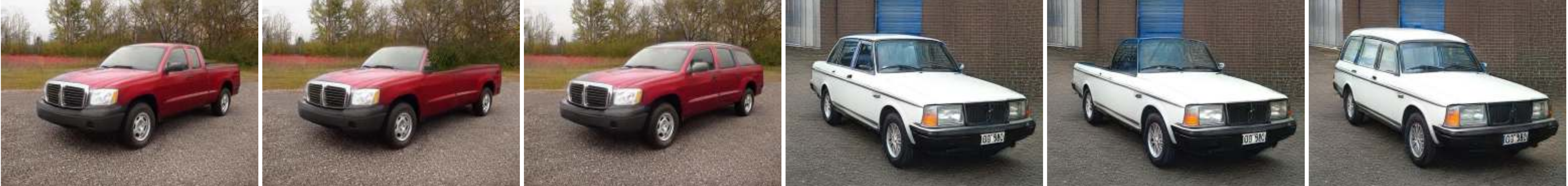}
\vspace{-5mm}
\caption{\footnotesize Pure optimization-based editing. We demonstrate large-scale semantic edits that do not transfer seamlessly to other images via editing vectors. Hence, we perform optimization from scratch.}	
\label{fig:large_edits}
\vspace{-2mm}
\end{figure*}

\begin{figure*}[t!]
\vspace{-3mm}
\includegraphics[width=1\linewidth, trim=0 370 0 0,clip]{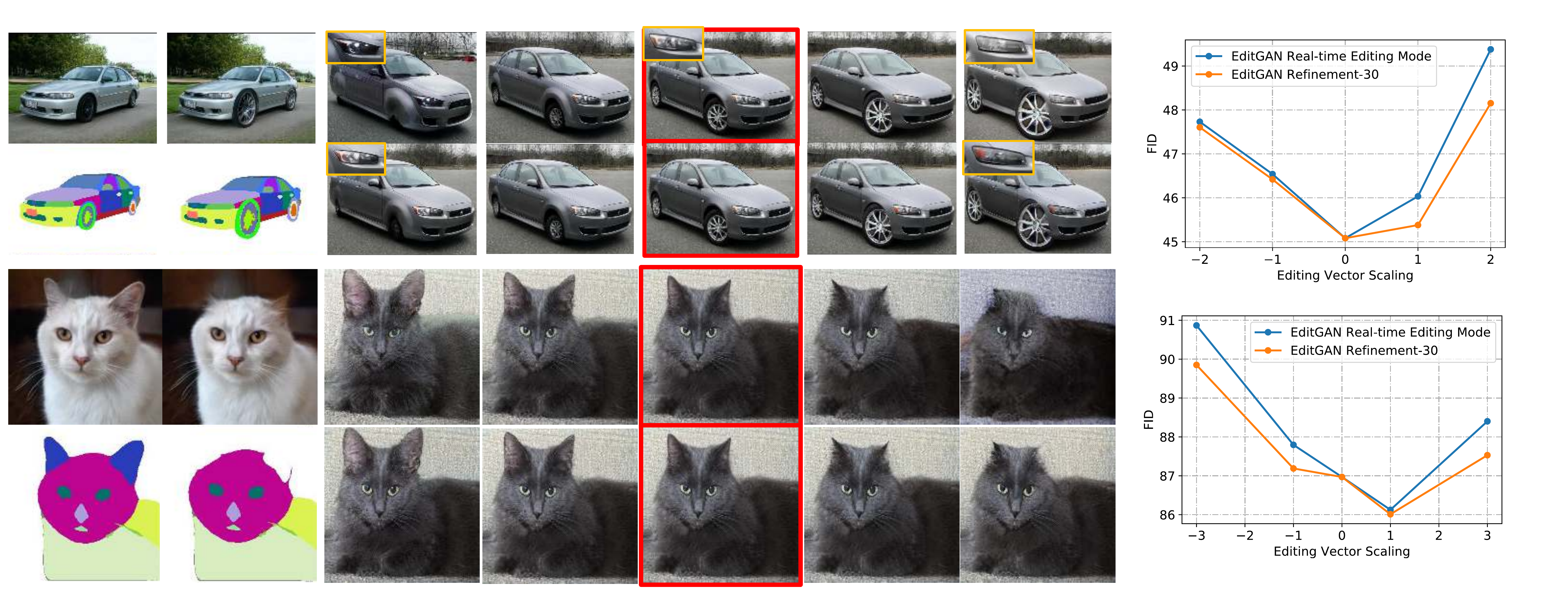}
\vspace{-6mm}
\caption{\footnotesize \textit{Left: }We apply learnt editing vectors with varying scales (see 5 markers in FID plots) both without (top row for each class) and with (bottom row for each class) additional 30-step self-supervised refinement to correct artifacts. Red boxes denote original images. For each class, the leftmost image is the one used to learn the editing vector, with the editing result next to it and orginal and modified segmentations below. \textit{Right:} Visual quality after editing with different scales as measured by FID with and without refinement.}	
\label{fig:mask_change_vis}
\vspace{-5mm}
\end{figure*}

\paragraph{Out-of-Domain Results} We demonstrate the generalization capability of EditGAN to out-of-domain data on the MetFaces~\cite{karras2020analyzing} data set. We use our EditGAN model trained on FFHQ~\cite{karras2020analyzing}, and create editing vectors $\delta \rvw^+_\textrm{edit}$ using in-domain real faces. We then embed out-of-domain MetFaces partraits (with 100 steps optimization) and apply the editing vectors with 30 steps self-supervised refinement. The results are shown in Fig.~\ref{fig:ood}. We find that our editing operations seamlessly translate even to such far out-of-domain examples.


\vspace{-2mm}

\subsection{Quantitative Results}\label{sec:quantitative}
To quantitatively measure EditGAN's image editing capabilities, we use the smile edit benchmark introduced by MaskGAN~\cite{lee2020maskgan}. Faces with neutral expressions are converted into smiling faces and performance is measured by three metrics: \textbf{a. Semantic Correctness:} Using a pre-trained smile attribute classifier, we measure whether the faces show smiling expressions after editing. \textbf{b. Distribution-level Image Quality:} Frechet Inception Distance (FID)~\cite{Seitzer2020FID,NIPS2017_7240} and Kernel Inception Distance (KID)~\cite{binkowski2018demystifying} are calculated between 400 edited test images and the CelebA-HD test dataset. \textbf{c. Identity Preservation:} Using the pretrained ArcFace feature extraction network~\cite{deng2019arcface}, we measure whether the subjects' identity is maintained when applying the edit. Specifically, we report cosine-similiarity between original and edited images. Further details can be found in the Appendix.

\begin{wraptable}[10]{R}{0.55\textwidth}
\vspace{-4mm}
 \begin{adjustbox}{width=1\linewidth}
{\footnotesize
\addtolength{\tabcolsep}{-3.5pt}
\begin{tabular}{lcccccc}
\toprule
Metric &  \# Mask     &  \# Attribute  & Attribute                &  FID $\downarrow$    &  KID $\downarrow$   & ID Score  $\uparrow$ \\
       &  Annot. & Annot.    & Acc.(\%) $\uparrow$  &                   &                      \\
\midrule
MaskGAN~\cite{lee2020maskgan} &  30,000  & - & 77.3 & 46.84 & 0.020 & 0.4611 \\ 
\midrule
LocalEditing~\cite{collins2020editing} & - & -  &   26.0 &  41.26  & 0.012 & 0.5823 \\ 
\midrule
LocalEditing - Encoding4Editing~\cite{tov2021designing} & - & -  &  41.75 &  	48.28	  & 0.016 & 0.6603 \\ 
\midrule
InterFaceGAN~\cite{shen2020interfacegan} & - & 30,000  & 83.5 & \textbf{39.42} & \textbf{0.010} &  0.7295 \\ 
\midrule
EditGAN (ours) & 16 &  - & \textbf{91.5} &  41.74 & 0.013&  0.7047\\ 
\midrule
EditGAN$^+30$ (ours) & 16 &  - &   85.8 & 40.83 & 0.012 &\textbf{0.7452} \\ 
\midrule
\midrule
StyleGAN2 Distillation~\cite{viazovetskyi2020stylegan2} & - & 30,000 &  98.3 &  	45.09		  & 0.013 & 	0.7823 \\ 
\bottomrule
\end{tabular}   
}
\end{adjustbox}
\vspace{-2mm}
\caption{\small Quantitative comparisons to multiple baselines on the smile edit benchmark.} 
\label{tbl:all_results}
\end{wraptable}
For our EditGAN, we simply learn a smiling editing vector $\delta \rvw^+_\textrm{edit}$ using a hold-out neutral expression face image. We embed it into EditGAN, infer its pixel-wise segmentation labels, and manually modify the segmentation towards a smile. Then we perform optimization in latent space, as described above, to learn the editing vector. For the results in Tab.~\ref{tbl:all_results}, it is applied with unit scale $s_\textrm{edit}{=}1$ on new images. We do not use the identity loss (Eq.~\ref{eq:idloss}) in this experiment, since identity preservation is already a target metric itself. We compare our method with three strong baselines: \textbf{(i)} \textit{MaskGAN}\footnote{\url{https://github.com/switchablenorms/CelebAMask-HQ}}~\cite{lee2020maskgan}: It takes non-smiling images, their segmentation masks, and a target smiling segmentation mask as inputs. Note that training MaskGAN requires large annotated datasets, in contrast to us. We also compare to \textbf{(ii)} \textit{LocalEditing}\footnote{\url{https://github.com/IVRL/GANLocalEditing}}~\cite{collins2020editing}: It clusters GAN features to achieve local editing and relies on reference images, in this case images of faces with smiling expressions. Another baseline we use is \textbf{(iii)} \textit{InterFaceGAN}\footnote{\url{https://github.com/genforce/interfacegan}}~\cite{shen2020interfacegan}: Similar to EditGAN, InterFaceGAN aims at finding editing vectors in latent space. However, it uses auxiliary attribute classifiers, relies on large annotated datasets, and can generally not achieve the fine editing control of our EditGAN. Finally, we compare to \textbf{(iv)} \textit{StyleGAN2 Distillation}\footnote{\url{https://github.com/EvgenyKashin/stylegan2-distillation}}~\cite{viazovetskyi2020stylegan2}, which creates an alternative approach that does not require real image embeddings and also relies on an editing-vector model 
to create a training dataset.

\begin{wrapfigure}[12]{R}{0.4\textwidth}
\vspace{-8mm}
\begin{center}
 \begin{adjustbox}{width=1\linewidth}
\includegraphics[width=1\linewidth, trim=5 0 0 30,clip]{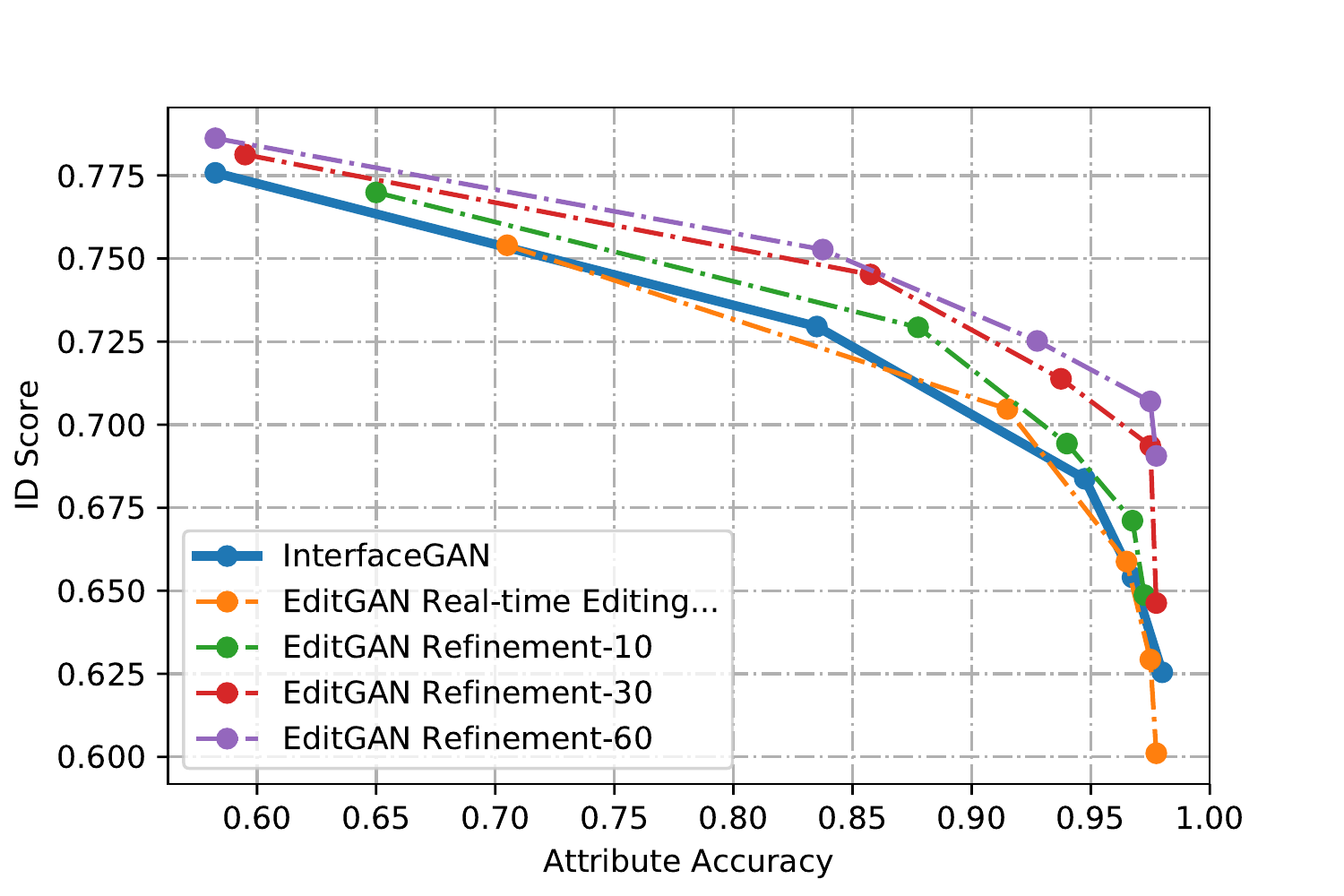}
\end{adjustbox} 
\vspace{-7mm}
\caption{\footnotesize InterFaceGAN's and EditGAN's performance on the smile edit benchmark for different editing vector scalings (scale increases from top-left points towards bottom-right points; see main text and Appendix for details). For EditGAN, we optionally add 10, 30 or 60 additional optimization steps.} 
\label{tbl:smile_acc_vs_id}
\end{center}
\end{wrapfigure}

Results are reported in Tab.~\ref{tbl:all_results}. Using $1,875\times$ less training labels, we outperform MaskGAN on all three metrics. We similarly obtain significantly stronger results than LocalEditing. In our observation, LocalEditing does not work well on real image embeddings. We further exploit a better encoder~\cite{tov2021designing} for the LocalEditing baseline, which leads to a significant improvement in attribute accuracy and ID score, but slightly worse FID \& KID scores. We find that EditGAN outperforms InterFaceGAN on identity preservation and attribute classification accuracy, while InterFaceGAN reaches slightly better FID \& KID scores (for the results in Tab.~\ref{tbl:all_results}, the latent space edits learnt by InterfaceGAN are also applied with unit scale, like for EditGAN). In Fig.~\ref{tbl:smile_acc_vs_id}, we report a more detailed comparison to InterFaceGAN, where we apply the smile editing vectors with different scale coefficients from zero to two. As shown, when the editing vector scale is small, the identity score is high while the smiling attribute score is low, since the modification of the original images is minimal. We find that our real-time editing with editing vectors is on-par with InterFaceGAN. When we perform self-supervised refinement at test time, EditGAN outperforms InterFaceGAN. In Tab.~\ref{tbl:all_results}, we also compare with StyleGAN2 Distillation~\cite{viazovetskyi2020stylegan2}, which achieves strong performance. 
However, StyleGAN2 Distillation relies on pre-trained classifiers, like InterfaceGAN, and only enables relatively high-level editing of image attributes for which large-scale annotations exit. Moreover, it distills edits into separate Pixel2PixelHD networks, such that a new network needs to be trained for each edit, limiting broad, user-interactive applicability. Hence, we consider StyleGAN2 Distillation orthogonal to our EditGAN.

\paragraph{Running Time} We carefully measure the run time of our editing on an NVIDIA Tesla V100 GPU. Conditional optimization, given an edited segmentation mask, with 30 (60) optimization steps takes 11.4 (18.9) seconds. This operation provides us the editing vector. Application of editing vectors is almost instantaneous, taking only 0.4 seconds, therefore allowing for complex real-time interactive editing. A 10 (30) step self-supervised refinement would add an additional 4.2 (9.5) seconds.

\vspace{-2mm}
\subsection{Ablation Studies: Self-Supervised Refinement and Editing Vector Scale}
\vspace{-1mm}

Fig.~\ref{tbl:smile_acc_vs_id} also contains a quantitative ablation study on the number of additional optimization steps done when initializing an edit with a learnt editing vector and refining with additional optimization. Generally, the more refinement steps we perform, the better the performance our model can achieve. As shown in Fig.~\ref{tbl:smile_acc_vs_id}, we find that further optimization can indeed slightly improve performance. Specifically, here we improve the trade-off between maintaining identity and achieving the desired semantic operation when performing editing with different scalings $s_\textrm{edit}$ of the editing vector. However, performing many steps of optimization leads to a run-time vs. performance trade-off, and our results suggest that the improvement beyond 30 additional optimization steps becomes marginal.

In Fig.~\ref{fig:mask_change_vis}, we analyze the editing vector scale and self-supervised refinement visually and with respect to perceptual metrics. As highlighted in the zoom-in areas, small artifacts can appear due to imperfect disentanglement in latent space when applying editing operations with large scales. Self-supervised refinement successfully cleans these editing errors up. We also apply the same edit with different scales on 400 test images and measure FID with respect to 10,000 data from GAN training, inspired by the analyses in~\cite{Navigan_CVPR_2021}. We can clearly see that image quality degrades as measured by FID, the stronger the edit is applied. We also observe small improvements with the iterative refinement on this metric, although the difference is small. Further details are in the Appendix.
We conclude that for most editing operations, real-time editing without iterative refinement already performs very well. However, to clean up artifacts and maintain highest image quality possible, self-supervised refinement with a couple of additional optimization steps is always available.

Additional experiments are presented in the Appendix.

\vspace{-3mm}
\section{Conclusions}\label{sec:conclusions}
\vspace{-2mm}

\paragraph{Limitations} Like all GAN-based image editing methods, EditGAN is limited to images that can be modeled by the GAN. This makes EditGAN's application on, for instance, photos of vivid city scenes challenging. 
Although most of our high-precision edits readily transfer to other images via learnt editing vectors, we also encountered challenging edits that required iterative optimization on each example. Future research therefore includes speeding up the optimization for such edits as well as building better generative models with more disentangled latent spaces. 
\vspace{-2mm}
\paragraph{Summary} We propose EditGAN, a novel method for high-precision, high-quality semantic image editing. It relies on a GAN that jointly models RGB images and their pixel-wise semantic segmentations and that requires only very few annotated data for training. Editing is achieved by performing optimization in latent space while conditioning on edited segmentation masks. 
This optimization can be amortized into editing vectors in latent space, which can be applied on other images directly, allowing for real-time interactive editing without any or only little further optimization. 
We demonstrate a broad variety of editing operations on different kinds of images, achieving an unprecedented level of flexibility and freedom in terms of editing, while preserving high image quality. 

\vspace{-3mm}
\section{Broader Impact} \label{sec:impact}
\vspace{-2mm}
Where previous generative modeling-based image editing methods offer only limited high-level editing capabilities, our method 
provides users unprecedented high-precision semantic editing possibilities.
Our proposed techniques can be used for artistic purposes and creative expression and benefit designers, photographers, and content creators~\cite{Bailey2020thetools}. AI-driven image editing tools like ours promise to democratize high-quality image editing. Related methods have already found their way into everyday applications in the form of neural photo editing filters. On a larger scale, the ability to synthesize data with specific attributes can be leveraged in training and finetuning machine learning models.

At the same time, more precise photo editing also offers opportunities for advanced photo manipulation for nefarious purposes.  
The recent progress of generative models and AI-driven photo editing has profound implications on image authenticity and beyond, which is an area of active debate~\cite{vaccari2020deepfakes}. As one potential way to tackle these challenges, methods for automatically validating real images and detecting manipulated or fake images are being developed by the research community~\cite{nguyen2021deep,mirsky2021deepfakesurvey}. Furthermore, generative models like ours are usually only as good as the data they were trained on. Therefore, biases in the underlying datasets are still present in the synthesized images and preserved even when applying our proposed editing methods. It is therefore important to be aware of such biases in the underlying data and counteract them, for example by actively collecting more representative data or by using bias correction methods, an area of active research~\cite{grover2019bias,choi2020fair,yu2020inclusive,lee2021selfdiagnosing}.

\vspace{-2mm}
\section*{Funding Statement}
\vspace{-1mm}
This work was funded by NVIDIA. Huan Ling and Seung Wook Kim acknowledge additional revenue in the form of student scholarships from University of Toronto and the Vector Institute, which are not in direct support of this work.

\bibliographystyle{unsrt}
{\small
\bibliography{egbib}}

\begin{thebibliography}{10}

\bibitem{zhang2021datasetgan}
Yuxuan Zhang, Huan Ling, Jun Gao, Kangxue Yin, Jean-Francois Lafleche, Adela
  Barriuso, Antonio Torralba, and Sanja Fidler.
\newblock Datasetgan: Efficient labeled data factory with minimal human effort.
\newblock {\em arXiv preprint arXiv:2104.06490}, 2021.

\bibitem{li2021semantic}
Daiqing Li, Junlin Yang, Karsten Kreis, Antonio Torralba, and Sanja Fidler.
\newblock Semantic segmentation with generative models: Semi-supervised
  learning and strong out-of-domain generalization.
\newblock {\em arXiv preprint arXiv:2104.05833}, 2021.

\bibitem{Bailey2020thetools}
J.~Bailey.
\newblock The tools of generative art, from flash to neural networks.
\newblock {\em Art in America}, 2020.

\bibitem{goodfellow2014generative}
Ian Goodfellow, Jean Pouget-Abadie, Mehdi Mirza, Bing Xu, David Warde-Farley,
  Sherjil Ozair, Aaron Courville, and Yoshua Bengio.
\newblock Generative adversarial nets.
\newblock In {\em Advances in neural information processing systems}, pages
  2672--2680, 2014.

\bibitem{radford2015unsupervised}
Alec Radford, Luke Metz, and Soumith Chintala.
\newblock Unsupervised representation learning with deep convolutional
  generative adversarial networks.
\newblock {\em arXiv preprint arXiv:1511.06434}, 2015.

\bibitem{karras2017progressive}
Tero Karras, Timo Aila, Samuli Laine, and Jaakko Lehtinen.
\newblock Progressive growing of gans for improved quality, stability, and
  variation.
\newblock {\em arXiv preprint arXiv:1710.10196}, 2017.

\bibitem{karras2019style}
Tero Karras, Samuli Laine, and Timo Aila.
\newblock A style-based generator architecture for generative adversarial
  networks.
\newblock In {\em Proceedings of the IEEE/CVF Conference on Computer Vision and
  Pattern Recognition}, pages 4401--4410, 2019.

\bibitem{karras2020analyzing}
Tero Karras, Samuli Laine, Miika Aittala, Janne Hellsten, Jaakko Lehtinen, and
  Timo Aila.
\newblock Analyzing and improving the image quality of stylegan.
\newblock In {\em Proceedings of the IEEE/CVF Conference on Computer Vision and
  Pattern Recognition}, pages 8110--8119, 2020.

\bibitem{choi2018stargan}
Yunjey Choi, Minje Choi, Munyoung Kim, Jung-Woo Ha, Sunghun Kim, and Jaegul
  Choo.
\newblock Stargan: Unified generative adversarial networks for multi-domain
  image-to-image translation.
\newblock In {\em Proceedings of the IEEE Conference on Computer Vision and
  Pattern Recognition}, 2018.

\bibitem{lee2020maskgan}
Cheng-Han Lee, Ziwei Liu, Lingyun Wu, and Ping Luo.
\newblock Maskgan: Towards diverse and interactive facial image manipulation.
\newblock In {\em IEEE Conference on Computer Vision and Pattern Recognition
  (CVPR)}, 2020.

\bibitem{wu2020cascade}
Rongliang Wu, Gongjie Zhang, Shijian Lu, and Tao Chen.
\newblock Cascade ef-gan: Progressive facial expression editing with local
  focuses.
\newblock In {\em Proceedings of the IEEE/CVF Conference on Computer Vision and
  Pattern Recognition (CVPR)}, June 2020.

\bibitem{shen2020interpreting}
Yujun Shen, Jinjin Gu, Xiaoou Tang, and Bolei Zhou.
\newblock Interpreting the latent space of gans for semantic face editing.
\newblock In {\em CVPR}, 2020.

\bibitem{shen2020interfacegan}
Yujun Shen, Ceyuan Yang, Xiaoou Tang, and Bolei Zhou.
\newblock Interfacegan: Interpreting the disentangled face representation
  learned by gans.
\newblock {\em TPAMI}, 2020.

\bibitem{alharbi2020disentangled}
Yazeed Alharbi and Peter Wonka.
\newblock Disentangled image generation through structured noise injection.
\newblock In {\em Proceedings of the IEEE/CVF Conference on Computer Vision and
  Pattern Recognition (CVPR)}, June 2020.

\bibitem{hou2020guidedstyle}
Xianxu Hou, Xiaokang Zhang, Linlin Shen, Zhihui Lai, and Jun Wan.
\newblock Guidedstyle: Attribute knowledge guided style manipulation for
  semantic face editing.
\newblock {\em arXiv preprint arXiv:2012.11856}, 2020.

\bibitem{Navigan_CVPR_2021}
Anton Cherepkov, Andrey Voynov, and Artem Babenko.
\newblock Navigating the gan parameter space for semantic image editing.
\newblock In {\em IEEE/CVF Conference on Computer Vision and Pattern
  Recognition (CVPR)}, 2021.

\bibitem{zhang2020image}
Yuxuan Zhang, Wenzheng Chen, Huan Ling, Jun Gao, Yinan Zhang, Antonio Torralba,
  and Sanja Fidler.
\newblock Image gans meet differentiable rendering for inverse graphics and
  interpretable 3d neural rendering.
\newblock {\em arXiv preprint arXiv:2010.09125}, 2020.

\bibitem{collins2020editing}
Edo Collins, Raja Bala, Bob Price, and Sabine S{\"u}sstrunk.
\newblock Editing in style: Uncovering the local semantics of {GANs}.
\newblock In {\em IEEE Conference on Computer Vision and Pattern Recognition
  (CVPR)}, 2020.

\bibitem{zhu2020sean}
Peihao Zhu, Rameen Abdal, Yipeng Qin, and Peter Wonka.
\newblock Sean: Image synthesis with semantic region-adaptive normalization.
\newblock In {\em IEEE/CVF Conference on Computer Vision and Pattern
  Recognition (CVPR)}, June 2020.

\bibitem{lewis2020vogue}
Kathleen~M Lewis, Srivatsan Varadharajan, and Ira Kemelmacher-Shlizerman.
\newblock Vogue: Try-on by stylegan interpolation optimization.
\newblock {\em arXiv preprint arXiv:2101.02285}, 2021.

\bibitem{kim2021stylemapgan}
Hyunsu Kim, Yunjey Choi, Junho Kim, Sungjoo Yoo, and Youngjung Uh.
\newblock Exploiting spatial dimensions of latent in gan for real-time image
  editing.
\newblock In {\em Proceedings of the IEEE Conference on Computer Vision and
  Pattern Recognition}, 2021.

\bibitem{chen2020deepfacedrawing}
Shu-Yu Chen, Wanchao Su, Lin Gao, Shihong Xia, and Hongbo Fu.
\newblock Deepfacedrawing: Deep generation of face images from sketches.
\newblock {\em ACM Trans. Graph.}, 39(4), 2020.

\bibitem{he2019attgan}
Z.~{He}, W.~{Zuo}, M.~{Kan}, S.~{Shan}, and X.~{Chen}.
\newblock Attgan: Facial attribute editing by only changing what you want.
\newblock {\em IEEE Transactions on Image Processing}, 28(11):5464--5478, Nov
  2019.

\bibitem{bau2019gandissect}
David Bau, Jun-Yan Zhu, Hendrik Strobelt, Bolei Zhou, Joshua~B. Tenenbaum,
  William~T. Freeman, and Antonio Torralba.
\newblock Gan dissection: Visualizing and understanding generative adversarial
  networks.
\newblock In {\em Proceedings of the International Conference on Learning
  Representations (ICLR)}, 2019.

\bibitem{bau2019semantic}
David Bau, Hendrik Strobelt, William Peebles, Jonas Wulff, Bolei Zhou, Jun-Yan
  Zhu, and Antonio Torralba.
\newblock Semantic photo manipulation with a generative image prior.
\newblock {\em ACM Trans. Graph.}, 38(4), 2019.

\bibitem{plumerault2020Controlling}
Antoine Plumerault, Hervé~Le Borgne, and Céline Hudelot.
\newblock Controlling generative models with continuous factors of variations.
\newblock In {\em International Conference on Learning Representations}, 2020.

\bibitem{harkonen2020ganspace}
Erik Härkönen, Aaron Hertzmann, Jaakko Lehtinen, and Sylvain Paris.
\newblock Ganspace: Discovering interpretable gan controls.
\newblock In {\em Proc. NeurIPS}, 2020.

\bibitem{bau2020rewriting}
David Bau, Steven Liu, Tongzhou Wang, Jun-Yan Zhu, and Antonio Torralba.
\newblock Rewriting a deep generative model.
\newblock In {\em Proceedings of the European Conference on Computer Vision
  (ECCV)}, 2020.

\bibitem{wolberg1994digital}
George Wolberg.
\newblock {\em Digital Image Warping}.
\newblock IEEE Computer Society Press, Washington, DC, USA, 1st edition, 1994.

\bibitem{efros2001image}
Alexei~A. Efros and William~T. Freeman.
\newblock Image quilting for texture synthesis and transfer.
\newblock SIGGRAPH '01, page 341–346, New York, NY, USA, 2001. Association
  for Computing Machinery.

\bibitem{hertzmann2001image}
Aaron Hertzmann, Charles~E. Jacobs, Nuria Oliver, Brian Curless, and David~H.
  Salesin.
\newblock Image analogies.
\newblock In {\em Proceedings of the 28th Annual Conference on Computer
  Graphics and Interactive Techniques}, SIGGRAPH '01, page 327–340, New York,
  NY, USA, 2001. Association for Computing Machinery.

\bibitem{reinhard2001color}
E.~Reinhard, M.~Adhikhmin, B.~Gooch, and P.~Shirley.
\newblock Color transfer between images.
\newblock {\em IEEE Computer Graphics and Applications}, 21(5):34--41, 2001.

\bibitem{perez2003poisson}
Patrick P\'{e}rez, Michel Gangnet, and Andrew Blake.
\newblock Poisson image editing.
\newblock SIGGRAPH '03, page 313–318, New York, NY, USA, 2003. Association
  for Computing Machinery.

\bibitem{schaefer2006image}
Scott Schaefer, Travis McPhail, and Joe Warren.
\newblock Image deformation using moving least squares.
\newblock {\em ACM Trans. Graph.}, 25(3):533–540, 2006.

\bibitem{barnes2009patchmatch}
Connelly Barnes, Eli Shechtman, Adam Finkelstein, and Dan~B Goldman.
\newblock Patchmatch: A randomized correspondence algorithm for structural
  image editing.
\newblock {\em ACM Trans. Graph.}, 28(3), 2009.

\bibitem{tao2010error}
Michael~W. Tao, Micah~K. Johnson, and Sylvain Paris.
\newblock Error-tolerant image compositing.
\newblock In {\em ECCV}, 2010.

\bibitem{gatys2016imagestyel}
Leon~A. Gatys, Alexander~S. Ecker, and Matthias Bethge.
\newblock Image style transfer using convolutional neural networks.
\newblock In {\em 2016 IEEE Conference on Computer Vision and Pattern
  Recognition (CVPR)}, 2016.

\bibitem{zhu2017unpaired}
Jun-Yan Zhu, Taesung Park, Phillip Isola, and Alexei~A Efros.
\newblock Unpaired image-to-image translation using cycle-consistent
  adversarial networks.
\newblock In {\em Proceedings of the IEEE international conference on computer
  vision}, pages 2223--2232, 2017.

\bibitem{portenier2018faceshop}
Tiziano Portenier, Qiyang Hu, Attila Szab\'{o}, Siavash~Arjomand Bigdeli, Paolo
  Favaro, and Matthias Zwicker.
\newblock Faceshop: Deep sketch-based face image editing.
\newblock {\em ACM Trans. Graph.}, 37(4), 2018.

\bibitem{ling2020variational}
Huan Ling, David Acuna, Karsten Kreis, Seung~Wook Kim, and Sanja Fidler.
\newblock Variational amodal object completion.
\newblock {\em Advances in Neural Information Processing Systems}, 2020.

\bibitem{park2020swapping}
Taesung Park, Jun-Yan Zhu, Oliver Wang, Jingwan Lu, Eli Shechtman, Alexei~A.
  Efros, and Richard Zhang.
\newblock Swapping autoencoder for deep image manipulation.
\newblock In {\em Advances in Neural Information Processing Systems}, 2020.

\bibitem{kim2021grivegan}
Seung~Wook Kim, Jonah Philion, Antonio Torralba, and Sanja Fidler.
\newblock {DriveGAN: Towards a Controllable High-Quality Neural Simulation}.
\newblock In {\em IEEE Conference on Computer Vision and Pattern Recognition
  (CVPR)}, 2021.

\bibitem{kingma2014vae}
Diederik~P Kingma and Max Welling.
\newblock Auto-encoding variational bayes.
\newblock In {\em The International Conference on Learning Representations
  (ICLR)}, 2014.

\bibitem{rezende2014stochastic}
Danilo~Jimenez Rezende, Shakir Mohamed, and Daan Wierstra.
\newblock Stochastic backpropagation and approximate inference in deep
  generative models.
\newblock In {\em International Conference on Machine Learning}, pages
  1278--1286, 2014.

\bibitem{brock2018large}
Andrew Brock, Jeff Donahue, and Karen Simonyan.
\newblock Large scale {GAN} training for high fidelity natural image synthesis.
\newblock In {\em International Conference on Learning Representations}, 2019.

\bibitem{park2019semantic}
Taesung Park, Ming-Yu Liu, Ting-Chun Wang, and Jun-Yan Zhu.
\newblock Semantic image synthesis with spatially-adaptive normalization.
\newblock In {\em Proceedings of the IEEE Conference on Computer Vision and
  Pattern Recognition}, pages 2337--2346, 2019.

\bibitem{goetschalckx2019ganalyze}
Lore Goetschalckx, Alex Andonian, Aude Oliva, and Phillip Isola.
\newblock Ganalyze: Toward visual definitions of cognitive image properties.
\newblock In {\em Proceedings of the IEEE/CVF International Conference on
  Computer Vision (ICCV)}, October 2019.

\bibitem{jahanian2020Osteerability}
Ali Jahanian*, Lucy Chai*, and Phillip Isola.
\newblock On the "steerability" of generative adversarial networks.
\newblock In {\em International Conference on Learning Representations}, 2020.

\bibitem{voynov2020unsupervised}
Andrey Voynov and Artem Babenko.
\newblock Unsupervised discovery of interpretable directions in the gan latent
  space.
\newblock In {\em International Conference on Machine Learning}, pages
  9786--9796. PMLR, 2020.

\bibitem{wang2021geometric}
Binxu Wang and Carlos~R Ponce.
\newblock A geometric analysis of deep generative image models and its
  applications.
\newblock In {\em International Conference on Learning Representations}, 2021.

\bibitem{shen2021closedform}
Yujun Shen and Bolei Zhou.
\newblock Closed-form factorization of latent semantics in gans.
\newblock In {\em CVPR}, 2021.

\bibitem{wang2018high}
Ting-Chun Wang, Ming-Yu Liu, Jun-Yan Zhu, Andrew Tao, Jan Kautz, and Bryan
  Catanzaro.
\newblock High-resolution image synthesis and semantic manipulation with
  conditional gans.
\newblock In {\em Proceedings of the IEEE Conference on Computer Vision and
  Pattern Recognition}, pages 8798--8807, 2018.

\bibitem{luan2017deepphoto}
Fujun Luan, Sylvain Paris, Eli Shechtman, and Kavita Bala.
\newblock Deep photo style transfer.
\newblock In {\em 2017 IEEE Conference on Computer Vision and Pattern
  Recognition (CVPR)}, 2017.

\bibitem{liu2017unsupervised}
Ming-Yu Liu, Thomas Breuel, and Jan Kautz.
\newblock Unsupervised image-to-image translation networks.
\newblock In {\em Advances in neural information processing systems}, pages
  700--708, 2017.

\bibitem{li2018eccv}
Yijun Li, Ming-Yu Liu, Xueting Li, Ming-Hsuan Yang, and Jan Kautz.
\newblock A closed-form solution to photorealistic image stylization.
\newblock In {\em Proceedings of the European Conference on Computer Vision
  (ECCV)}, 2018.

\bibitem{kazemi2019style}
H.~Kazemi, S.~Iranmanesh, and N.~Nasrabadi.
\newblock Style and content disentanglement in generative adversarial networks.
\newblock In {\em 2019 IEEE Winter Conference on Applications of Computer
  Vision (WACV)}, pages 848--856, Los Alamitos, CA, USA, jan 2019. IEEE
  Computer Society.

\bibitem{yoo2019photorealistic}
Jaejun Yoo, Youngjung Uh, Sanghyuk Chun, Byeongkyu Kang, and Jung-Woo Ha.
\newblock Photorealistic style transfer via wavelet transforms.
\newblock In {\em 2019 IEEE/CVF International Conference on Computer Vision
  (ICCV)}, 2019.

\bibitem{perarnau2016invertible}
Guim Perarnau, Joost van~de Weijer, Bogdan Raducanu, and Jose~M. Álvarez.
\newblock Invertible conditional gans for image editing.
\newblock {\em arXiv preprint arXiv:1611.06355}, 2016.

\bibitem{donahue2016adversarial}
Jeff Donahue, Philipp Kr{\"a}henb{\"u}hl, and Trevor Darrell.
\newblock Adversarial feature learning.
\newblock {\em arXiv preprint arXiv:1605.09782}, 2016.

\bibitem{Brock2017neural}
Andrew Brock, Theodore Lim, James~M. Ritchie, and Nick Weston.
\newblock Neural photo editing with introspective adversarial networks.
\newblock In {\em 5th International Conference on Learning Representations,
  {ICLR} 2017, Toulon, France, April 24-26, 2017, Conference Track
  Proceedings}. OpenReview.net, 2017.

\bibitem{dumoulin2017adversarially}
Vincent Dumoulin, Ishmael Belghazi, Ben Poole, Alex Lamb, Mart{\'{\i}}n
  Arjovsky, Olivier Mastropietro, and Aaron~C. Courville.
\newblock Adversarially learned inference.
\newblock In {\em 5th International Conference on Learning Representations,
  {ICLR} 2017, Toulon, France, April 24-26, 2017, Conference Track
  Proceedings}. OpenReview.net, 2017.

\bibitem{richardson2020encoding}
Elad Richardson, Yuval Alaluf, Or~Patashnik, Yotam Nitzan, Yaniv Azar, Stav
  Shapiro, and Daniel Cohen-Or.
\newblock Encoding in style: a stylegan encoder for image-to-image translation.
\newblock {\em arXiv preprint arXiv:2008.00951}, 2020.

\bibitem{zhu2016generative}
Jun-Yan Zhu, Philipp Kr{\"a}henb{\"u}hl, Eli Shechtman, and Alexei~A Efros.
\newblock Generative visual manipulation on the natural image manifold.
\newblock In {\em European conference on computer vision}, pages 597--613.
  Springer, 2016.

\bibitem{Yeh2017}
R.~A. {Yeh}, C.~{Chen}, T.~Y. {Lim}, A.~G. {Schwing}, M.~{Hasegawa-Johnson},
  and M.~N. {Do}.
\newblock Semantic image inpainting with deep generative models.
\newblock In {\em 2017 IEEE Conference on Computer Vision and Pattern
  Recognition (CVPR)}, pages 6882--6890, 2017.

\bibitem{lipton2017precise}
Zachary~C. Lipton and Subarna Tripathi.
\newblock Precise recovery of latent vectors from generative adversarial
  networks.
\newblock {\em arXiv preprint arXiv:1702.04782}, 2017.

\bibitem{abdal2019image2stylegan}
Rameen Abdal, Yipeng Qin, and Peter Wonka.
\newblock Image2stylegan: How to embed images into the stylegan latent space?
\newblock In {\em Proceedings of the IEEE International Conference on Computer
  Vision}, pages 4432--4441, 2019.

\bibitem{huh2020transforming}
Minyoung Huh, Richard Zhang, Jun-Yan Zhu, Sylvain Paris, and Aaron Hertzmann.
\newblock Transforming and projecting images into class-conditional generative
  networks.
\newblock {\em arXiv preprint arXiv:2005.01703}, 2020.

\bibitem{creswell2019inverting}
A.~{Creswell} and A.~A. {Bharath}.
\newblock Inverting the generator of a generative adversarial network.
\newblock {\em IEEE Transactions on Neural Networks and Learning Systems},
  30(7):1967--1974, 2019.

\bibitem{Raj2019gan}
A.~{Raj}, Y.~{Li}, and Y.~{Bresler}.
\newblock Gan-based projector for faster recovery with convergence guarantees
  in linear inverse problems.
\newblock In {\em 2019 IEEE/CVF International Conference on Computer Vision
  (ICCV)}, pages 5601--5610, 2019.

\bibitem{bau2019seeing}
D.~{Bau}, J.~{Zhu}, J.~{Wulff}, W.~{Peebles}, B.~{Zhou}, H.~{Strobelt}, and
  A.~{Torralba}.
\newblock Seeing what a gan cannot generate.
\newblock In {\em 2019 IEEE/CVF International Conference on Computer Vision
  (ICCV)}, pages 4501--4510, 2019.

\bibitem{zhu2020domain}
Jiapeng Zhu, Yujun Shen, Deli Zhao, and Bolei Zhou.
\newblock In-domain gan inversion for real image editing.
\newblock {\em arXiv preprint arXiv:2004.00049}, 2020.

\bibitem{xu2021linear}
Jianjin Xu and Changxi Zheng.
\newblock Linear semantics in generative adversarial networks.
\newblock In {\em Proceedings of the IEEE/CVF Conference on Computer Vision and
  Pattern Recognition}, pages 9351--9360, 2021.

\bibitem{bau2021paint}
David Bau, Alex Andonian, Audrey Cui, YeonHwan Park, Ali Jahanian, Aude Oliva,
  and Antonio Torralba.
\newblock Paint by word.
\newblock {\em arXiv preprint arXiv:2103.10951}, 2021.

\bibitem{radford2021learning}
Alec Radford, Jong~Wook Kim, Chris Hallacy, Aditya Ramesh, Gabriel Goh,
  Sandhini Agarwal, Girish Sastry, Amanda Askell, Pamela Mishkin, Jack Clark,
  et~al.
\newblock Learning transferable visual models from natural language
  supervision.
\newblock {\em arXiv preprint arXiv:2103.00020}, 2021.

\bibitem{zhang2018unreasonable}
Richard Zhang, Phillip Isola, Alexei~A Efros, Eli Shechtman, and Oliver Wang.
\newblock The unreasonable effectiveness of deep features as a perceptual
  metric.
\newblock In {\em Proceedings of the IEEE conference on computer vision and
  pattern recognition}, pages 586--595, 2018.

\bibitem{deng2019arcface}
Jiankang Deng, Jia Guo, Niannan Xue, and Stefanos Zafeiriou.
\newblock Arcface: Additive angular margin loss for deep face recognition.
\newblock In {\em Proceedings of the IEEE/CVF Conference on Computer Vision and
  Pattern Recognition}, pages 4690--4699, 2019.

\bibitem{kingma2014adam}
Diederik~P Kingma and Jimmy Ba.
\newblock Adam: A method for stochastic optimization.
\newblock {\em arXiv preprint arXiv:1412.6980}, 2014.

\bibitem{Seitzer2020FID}
Maximilian Seitzer.
\newblock {pytorch-fid: FID Score for PyTorch}.
\newblock \url{https://github.com/mseitzer/pytorch-fid}, August 2020.
\newblock Version 0.1.1.

\bibitem{NIPS2017_7240}
Martin Heusel, Hubert Ramsauer, Thomas Unterthiner, Bernhard Nessler, and Sepp
  Hochreiter.
\newblock Gans trained by a two time-scale update rule converge to a local nash
  equilibrium.
\newblock In I.~Guyon, U.~V. Luxburg, S.~Bengio, H.~Wallach, R.~Fergus,
  S.~Vishwanathan, and R.~Garnett, editors, {\em Advances in Neural Information
  Processing Systems 30}, pages 6626--6637. Curran Associates, Inc., 2017.

\bibitem{binkowski2018demystifying}
Mikołaj Bińkowski, Danica~J. Sutherland, Michael Arbel, and Arthur Gretton.
\newblock Demystifying {MMD} {GAN}s.
\newblock In {\em International Conference on Learning Representations}, 2018.

\bibitem{tov2021designing}
Omer Tov, Yuval Alaluf, Yotam Nitzan, Or~Patashnik, and Daniel Cohen-Or.
\newblock Designing an encoder for stylegan image manipulation.
\newblock {\em ACM Transactions on Graphics (TOG)}, 40(4):1--14, 2021.

\bibitem{viazovetskyi2020stylegan2}
Yuri Viazovetskyi, Vladimir Ivashkin, and Evgeny Kashin.
\newblock Stylegan2 distillation for feed-forward image manipulation.
\newblock In {\em European Conference on Computer Vision}, pages 170--186.
  Springer, 2020.

\bibitem{vaccari2020deepfakes}
Cristian Vaccari and Andrew Chadwick.
\newblock Deepfakes and disinformation: Exploring the impact of synthetic
  political video on deception, uncertainty, and trust in news.
\newblock {\em Social Media + Society}, 6(1):2056305120903408, 2020.

\bibitem{nguyen2021deep}
Thanh~Thi Nguyen, Quoc Viet~Hung Nguyen, Cuong~M. Nguyen, Dung Nguyen,
  Duc~Thanh Nguyen, and Saeid Nahavandi.
\newblock Deep learning for deepfakes creation and detection: A survey.
\newblock {\em arXiv preprint arXiv:1909.11573}, 2021.

\bibitem{mirsky2021deepfakesurvey}
Yisroel Mirsky and Wenke Lee.
\newblock The creation and detection of deepfakes: A survey.
\newblock {\em ACM Comput. Surv.}, 54(1), 2021.

\bibitem{grover2019bias}
Aditya Grover, Jiaming Song, Ashish Kapoor, Kenneth Tran, Alekh Agarwal, Eric~J
  Horvitz, and Stefano Ermon.
\newblock Bias correction of learned generative models using likelihood-free
  importance weighting.
\newblock In {\em Advances in Neural Information Processing Systems}, 2019.

\bibitem{choi2020fair}
Kristy Choi, Aditya Grover, Trisha Singh, Rui Shu, and Stefano Ermon.
\newblock Fair generative modeling via weak supervision.
\newblock In {\em Proceedings of the 37th International Conference on Machine
  Learning}, 2020.

\bibitem{yu2020inclusive}
Ning Yu, Ke~Li, Peng Zhou, Jitendra Malik, Larry Davis, and Mario Fritz.
\newblock Inclusive {GAN:} improving data and minority coverage in generative
  models.
\newblock In {\em Computer Vision - {ECCV} 2020 - 16th European Conference,
  Glasgow, UK, August 23-28, 2020, Proceedings, Part {XXII}}, 2020.

\bibitem{lee2021selfdiagnosing}
Jinhee Lee, Haeri Kim, Youngkyu Hong, and Hye~Won Chung.
\newblock Self-diagnosing gan: Diagnosing underrepresented samples in
  generative adversarial networks.
\newblock {\em arXiv preprint arXiv:2102.12033}, 2021.

\bibitem{7298658}
G.~{Van Horn}, S.~{Branson}, R.~{Farrell}, S.~{Haber}, J.~{Barry},
  P.~{Ipeirotis}, P.~{Perona}, and S.~{Belongie}.
\newblock Building a bird recognition app and large scale dataset with citizen
  scientists: The fine print in fine-grained dataset collection.
\newblock In {\em Conference on Computer Vision and Pattern Recognition
  (CVPR)}, pages 595--604, 2015.

\bibitem{he2016deep}
Kaiming He, Xiangyu Zhang, Shaoqing Ren, and Jian Sun.
\newblock Deep residual learning for image recognition.
\newblock In {\em Proceedings of the IEEE conference on computer vision and
  pattern recognition}, pages 770--778, 2016.

\end{thebibliography}

 \newpage


\setcounter{section}{0}
\renewcommand{\thesection}{\Alph{section}}

\vspace{-2mm}
\section{Model and Training Details}
We first provide additional details  about our EditGAN.

\subsection{Image GAN}
EditGAN uses StyleGAN2 as a backbone generative model of images. We denote the image generator as $G : \mathcal{Z} \rightarrow \mathcal{W} \rightarrow  \mathcal{X}$, which is trained following standard StyleGAN training, see for more information~\cite{karras2019style, karras2020analyzing}. In particular, we use the pre-trained Car, Face-FFHQ and Cat StyleGAN2 models from the official GitHub repository provided by StyleGAN2\footnote{\url{https://github.com/NVlabs/stylegan2} (Nvidia Source Code License)}. For Bird, we use the StyleGAN2 model trained on NABirds-48k~\cite{7298658}. 

The StyleGAN2 generator maps latent codes $\rvz \in \mathcal{Z}$,
drawn from a multivariate Normal distribution, $\mathcal{N}(\rvz;\bzero, \beye)$, 
into realistic images. A latent code $\rvz$ is first transformed into an intermediate code $\rvw \in \mathcal{W}$ by a non-linear mapping function $m(\rvz)$.
$\rvw$ is then further transformed into $K+1$ independent vectors, $\rvw^0,...,\rvw^K$, through 
$K+1$ learned affine transformations. These $K+1$ 
transformed latent codes are fed into synthesis blocks, sometimes called style layers and denoted as $\{\textrm{Style}^0, \textrm{Style}^1,..,\textrm{Style}^K\}$~\cite{karras2017progressive}. The output of these synthesis blocks are deep feature maps $\{S^0, S^1,..., S^K\}$. These feature maps carry the information for forming the image $\rvx\in\mathcal{X}$, which is achieved by connecting them to a residual image synthesis branch. Further details and visualizations about the StyleGAN2 architecture can be found in~\cite{karras2019style, karras2020analyzing}.

\subsection{Image Encoder}
To embed images into the GAN's latent space, the EditGAN framework relies on optimization, initialized by an encoder. To train this encoder we mainly follow SemanticGAN~\cite{li2021semantic}, which builds on~\cite{richardson2020encoding}, with further improvements.

We start by introducing notation. Let us denote $\mathbb{D}_\rvx$ as a dataset of real images and  $\mathbb{D}_{\rvx, \rvy}$ as a dataset of image-segmentation mask pairs.
Note that the number of images in the unannotated data $\mathbb{D}_\rvx$ is usually much larger than the annotated $\mathbb{D}_{\rvx, \rvy}$. In fact, $\mathbb{D}_{\rvx, \rvy}$ is as small as 16 or 30 image-segmentation pairs for our datasets.
We directly embed the images into $\mathcal{W}^+$ space, where the $K+1$ $\rvw^0,...,\rvw^K$ are modeled independently for each style layer~\cite{abdal2019image2stylegan}. Thus, we can formally define a variation of the generator as $\hat{G} : \mathcal{W}^+ \rightarrow  \mathcal{X} $, which operates on this $\mathcal{W}^+$ space. We follow ~\cite{richardson2020encoding} and train an encoder $ E_\bphi : \mathcal{X} \rightarrow  \mathcal{W}^+ $ with parameters $\bphi$ using the following objective functions:
\begin{gather}
\mathcal{L}_{\textrm{RGB}}(\bphi)  =   \mathbb{E}_{\rvx \in \mathbb{D}_\rvx} \left[\lambda_1L_\textrm{LPIPS}( \rvx, \; \hat{G}(E_\bphi(\rvx)) ) +   \lambda_2 L_\textrm{L2}( \rvx, \; \hat{G}(E_\bphi(\rvx)) )\right]
\label{rgb_loss}
\end{gather}
where $L_\textrm{LPIPS}$ loss is the Learned Perceptual Image Patch Similarity (LPIPS) distance~\cite{zhang2018unreasonable} and $L_\textrm{L2}$ is a standard L2 loss. We also explicitly regularize the encoder output distribution using an additional loss that utilizes samples from the GAN itself:
\begin{gather}
\mathcal{L}_{\textrm{Sampling}}(\bphi) =   \mathbb{E}_{\rvx = G(\rvz), \rvz \sim \mathcal{N}(\rvz;\bzero, \beye) } [\lambda_3 L_\textrm{LPIPS}( \rvx, \; \hat{G}(E_\bphi(\rvx)) )  \\ +  \lambda_4 L_\textrm{L2}( \rvx, \; \hat{G}(E_\bphi(\rvx)) )   +    \lambda_5 L_\textrm{L2}(m(\rvz), \; E_\bphi(\rvx) )]
\label{sampling_loss}
\end{gather}

Here, $m(\rvz)$ is the previously introduced mapping function $m : \mathcal{Z} \rightarrow  \mathcal{W} $ and $\lambda_{1,...,5}$ are hyperparameters. For all classes, we set $\lambda_1=10$, $\lambda_2=1$, $\lambda_3=10$, $\lambda_4=1$, and $\lambda_5=5$. We use the Adam~\cite{kingma2014adam} optimizer with learning rate $3\times 10^{-5}$ and batch size $8$ to train the encoder. Experimentally, for the Car and Cat datasets, we first train only on samples from the GAN itself using Eq.~\ref{sampling_loss} for 20,000 iterations as warm up, and then train jointly using Eq.~\ref{rgb_loss} and Eq.~\ref{sampling_loss} iteratively until the model converges on the training dataset.


After successful encoder training, to embed images we first use the encoder $E_\bphi$ and further iteratively refine the latent code $\rvw^+$ via optimization with respect to the $\mathcal{L}_{\textrm{RGB}}$ objective (without further modifying encoder parameters $\bphi$). We run optimization for 500 steps with $\lambda_1=10$, $\lambda_2=1$. We use the Adam~\cite{kingma2014adam} optimizer with the lookahead technique~\cite{zhang2019lookahead} with a constant learning rate of $0.001$.

\subsection{Segmentation Branch}
Using our encoder together with additional optimization, as described in the previous section, we embed the annotated images $\rvx$ from $\mathbb{D}_{\rvx, \rvy}$ into $\mathcal{W}^+$, formally constructing $\mathbb{D}_{\rvx, \rvy, \rvw^+}$, the annotated dataset augmented with $\rvw^+$ embeddings.

Similar to DatasetGAN~\cite{zhang2021datasetgan}, to generate segmentation maps $\rvy$ alongside images $\rvx$ we then train a segmentation branch $I_\bpsi$ with parameters $\bpsi$. $I_\bpsi$ is a simple three-layer multi-layer perceptron classifier on the layer-wise concatenated and appropriately upsampled feature maps. Specifically, the lower-resolution deep feature maps in $\{S^0, S^1,..., S^K\}$ are first appropriately upsampled, $\hat{S}^k=U_k(S^k)$ for $k\in {0,...,K}$ and upsampling functions $U_k$, so that all feature maps have the same spatial resolution, equal to the highest resolution, and can be concatenated channel-wise. The classifier operates on the layer-wise concatenated feature maps in a per-pixel fashion and predicts the segmentation label of each pixel. It is trained via the objective
\begin{gather}
\mathcal{L}_I(\bpsi) =  \mathbb{E}_{\rvx,\rvy,\rvw^+ \in \mathbb{D}_{\rvx, \rvy, \rvw^+} } \left[ H( \rvy,  I_\bpsi( (\hat{S}^0, \hat{S}^1,..., \hat{S}^K) ) )  \right], \\
\textrm{with}\quad\hat{S}^k = U_k (\textrm{Style}_k (\rvw_k^+)),
\end{gather}
where $I_\bpsi$ takes as input the concatenated and appropriately upsampled feature maps $(\hat{S}^0, \hat{S}^1,..., \hat{S}^K)$. We use bilinear-upsampling operations $U_k$. Furthermore, $H$ denotes the pixel-wise cross-entropy.

To train the segmentation branch $I_\bpsi$ and minimize the objective $\mathcal{L}_I(\bpsi)$, we use the Adam optimizer with learning rate $0.001$. We randomly sample $64$ pixels across all training images for each batch. The segmentation branch is trained until it converges on the training dataset. After training the segmentation branch $I_\bpsi$ we can formally define a generator $\tilde{G} : \mathcal{W}^+ \rightarrow  \mathcal{X}, \mathcal{Y} $ that models the joint distribution $p(\rvx, \rvy)$ of images $\rvx$ and semantic segmentations $\rvy$.

Notice that we defined the segmentation branch here using a new symbol, $I_\bpsi$, opposed to $\tilde{G}^\rvy$ from the main paper. Here, $I_\bpsi$ specifies the specific network that is only part of the segmentation branch and acts on top of the feature maps $(\hat{S}^0, \hat{S}^1,..., \hat{S}^K)$ in a pixel-wise manner. On the other hand, the segmentation generation component $\tilde{G}^\rvy$, defined in the main paper, denotes the complete segmentation generation module, starting from $\mathcal{W}^+$, including the style layers and deep feature maps that are shared between the image and segmentation generation branches.

\subsection{Learning Editing Vectors}
To perform editing and learn editing vectors, we proceed as described in detail in Secs. 3.3 and 3.4 in the main text. The ArcFace feature extraction network checkpoint~\cite{deng2019arcface} is taken from \url{https://github.com/TreB1eN/InsightFace_Pytorch} (MIT License).

\begin{figure*}[t!]
\begin{center}
\includegraphics[width=1\linewidth, trim=10 0 8 0,clip]{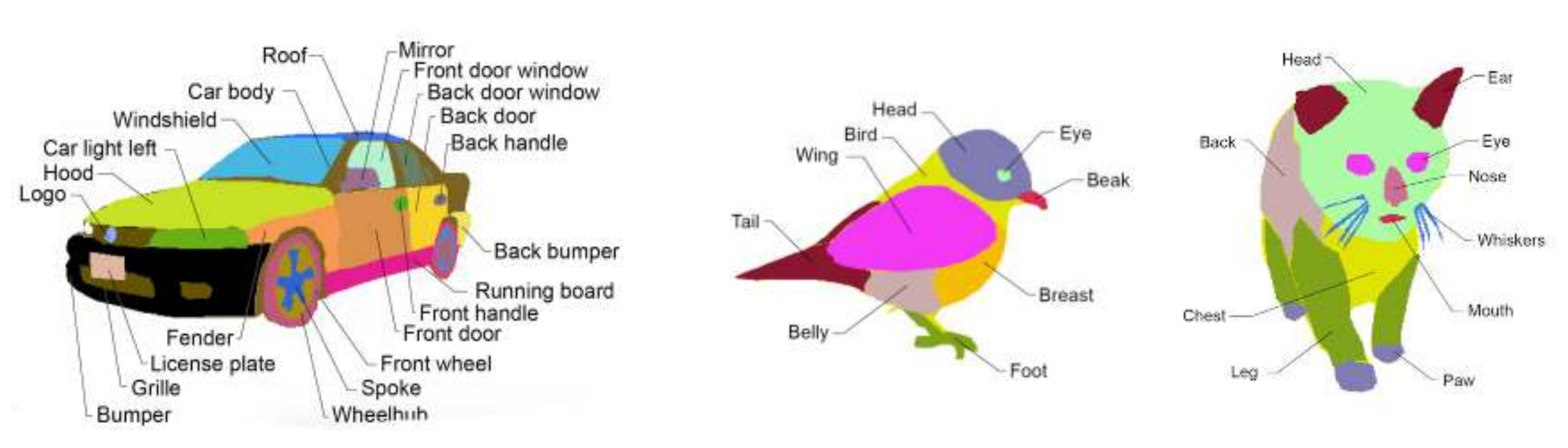}
\vspace{-4mm}
\caption{\footnotesize Car, Bird and Cat part labeling schemes~\cite{zhang2021datasetgan}.}	
\label{fig:annotation}
\end{center}
\end{figure*}

In the main paper, we already provided the label scheme for the Face data (Fig. 6). The further labeling schemes for the Car, Bird, and Cat classes are shown in Fig~\ref{fig:annotation}.  The annotations contain 34, 32, 16, and 11 possible pixel labels for the Face, Car, Bird and Cat data, respectively. When performing optimization to find the editing vectors $\delta \rvw^+_\textrm{edit}$, we use the Adam~\cite{kingma2014adam} optimizer with learning rate $0.02$ and run for $100$ steps. We use the hyperparameters $\lambda^\textrm{editing}_1=15$, $\lambda^\textrm{editing}_2=1$, and $\lambda^\textrm{editing}_3=10$ (Eq. 5 in main paper). When performing optimization for self-supervised refinement after initializing the edit with an editing vector (as described in second bullet point in Sec. 3.4 in main paper), we use the same optimizer and we set hyperparameters $\lambda^\textrm{editing}_1=5$, $\lambda^\textrm{editing}_2=1$, and $\lambda^\textrm{editing}_3=5$.
Hyperparameters are chosen based on visual quality on hold-out examples. We will release the training set $\mathbb{D}_{\rvx, \rvy}$ and learnt editing vectors. 

\section{Experiment Details}
Here, we provide additional experiment details.

\subsection{Smile Edit Benchmark}
In Section \textbf{4.2} of the main paper, we evaluate our model against strong baselines on the smile edit benchmark introduced by MaskGAN~\cite{lee2020maskgan}. Here we provide more details for completeness. \textbf{Semantic Correctness:} To measure whether the faces show smiling expressions after editing, a binary smile attribute classifiers is trained on the CelebA training set, using a ResNet-18~\cite{he2016deep} backbone. The input faces are resized into resolution of $256\times 256$. The classifier achieives 92.2\% accuracy on the CelebA testing dataset. \textbf{Identity Preservation:}  We again use the pretrained ArcFace feature extraction network~\cite{deng2019arcface} with checkpoint from \url{https://github.com/TreB1eN/InsightFace_Pytorch} (MIT License). As pointed out in main paper, we did not use the identity loss in this benchmark experiment when performing face editing. In this benchmark experiment, this facial feature extraction network is used \textit{only} for evaluation purposes.

\begin{figure}
\addtolength{\tabcolsep}{-5pt}
\hspace{1mm}
\begin{tabular}{cccc}
 \includegraphics[ width=0.25\linewidth,trim=0 0 0 0,clip]{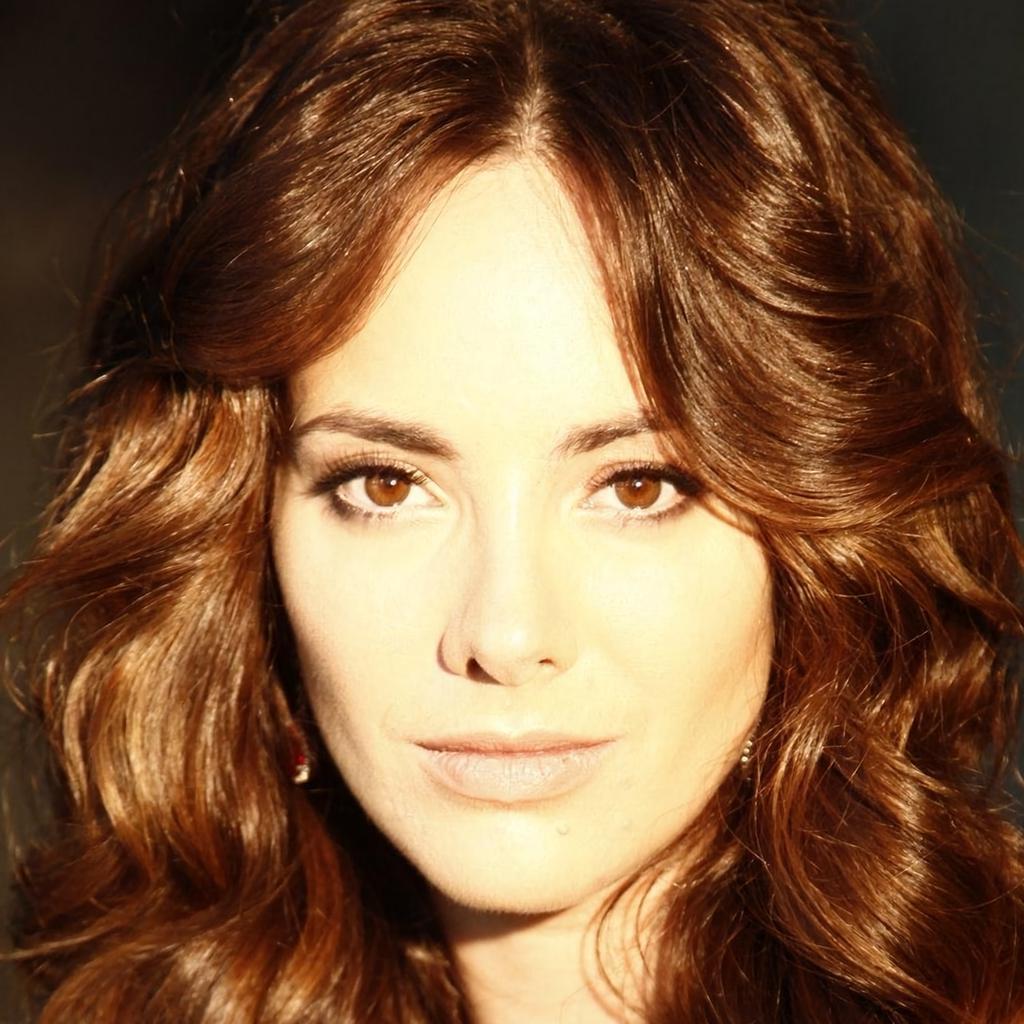} & \includegraphics[width=0.25\linewidth,trim=0 0 0 0,clip]{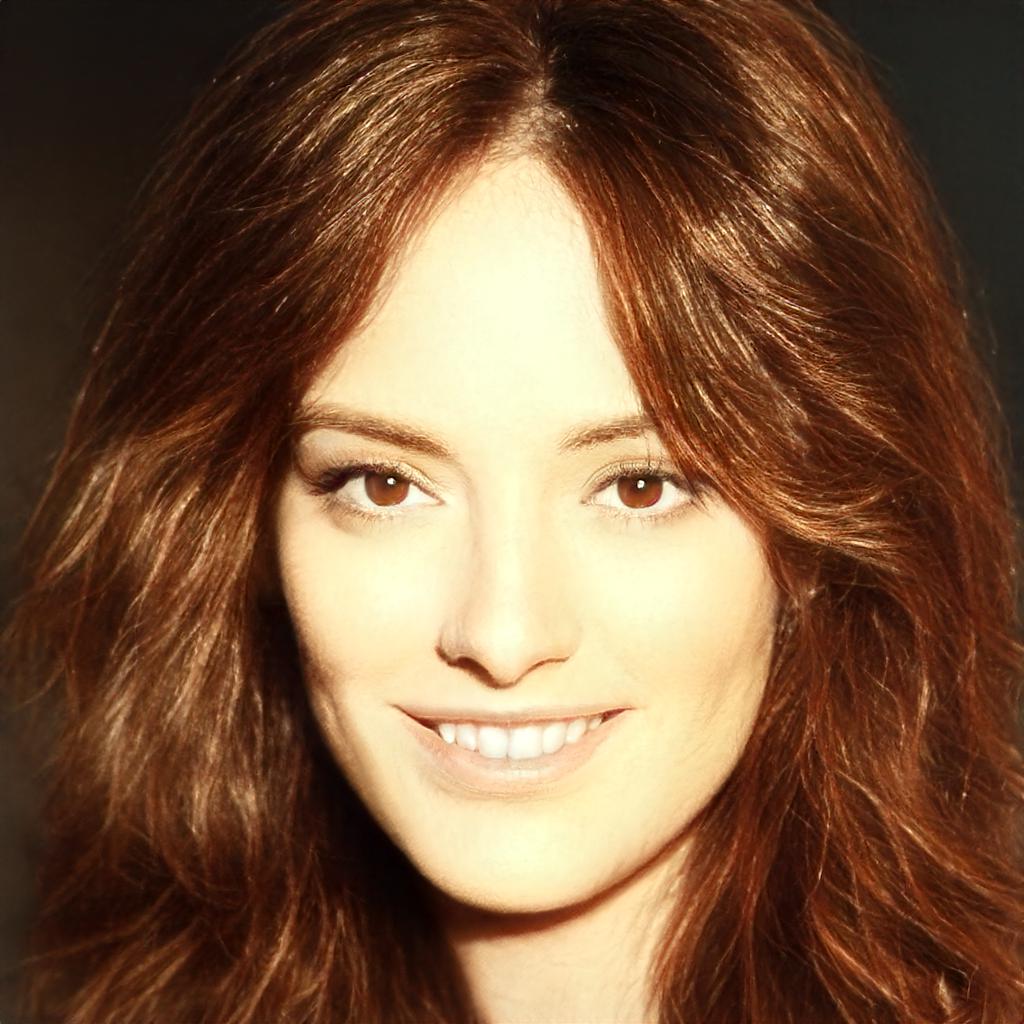}  &
 \includegraphics[width=0.25\linewidth,trim=0 0 0 0,clip]{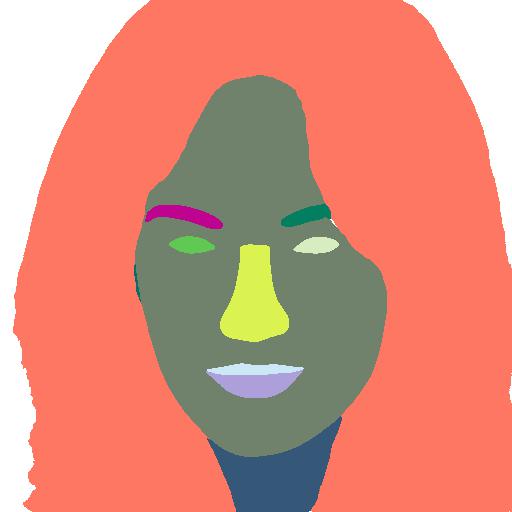} & \includegraphics[width=0.25\linewidth,trim=0 0 0 0,clip]{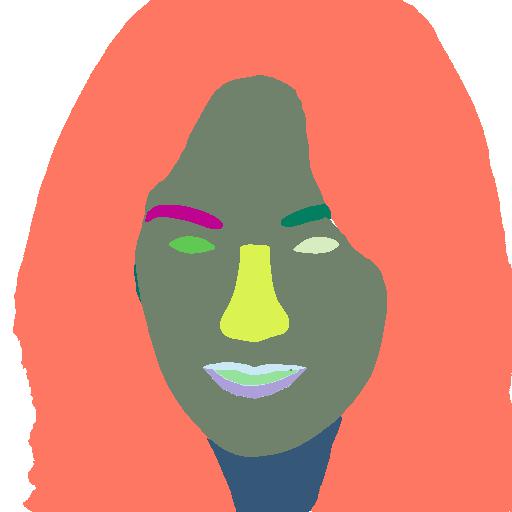}  \\
\end{tabular}

\caption{\footnotesize  {\bf Image and mask pair to learn smile editing vector on CelebA.} Images are face before editing, face after editing, segmentation mask predicted by segmentation branch before editing (after embedding the image into EditGAN's latent space), and target segmentation mask after manual modification.}
\label{fig:celeb_smile}
\vspace{-3mm}
\end{figure}

To compare with the baselines, we took the officially released MaskGAN\footnote{\url{https://github.com/switchablenorms/CelebAMask-HQ}}~\cite{lee2020maskgan} and LocalEditing\footnote{\url{https://github.com/IVRL/GANLocalEditing}}~\cite{collins2020editing} checkpoints. Furthermore, we train an InterFaceGAN~\cite{shen2020interfacegan} smile model using the officially released code\footnote{\url{https://github.com/genforce/interfacegan}} where we replaced the generator with a StyleGAN2 for fair comparison. At inference time when performing editing, we use the same test image embeddings for the InterFaceGAN model as we use for our EditGAN model. As mentioned in the main paper, we also use StyleGAN2 Distillation~\cite{viazovetskyi2020stylegan2} as baseline, for which we rely on the official codebase\footnote{\url{https://github.com/EvgenyKashin/stylegan2-distillation}} and train a Pix2PixHD network for the smile edit using the default hyperparameters as provided in the paper~\cite{viazovetskyi2020stylegan2}. We further show in Fig.~\ref{fig:celeb_smile} the image and initial and modified segmentation masks that were used to learn our smile editing vector. 

Finally, we provide more details for Fig. 10 in the main text: For each curve, we report results with five different editing vector scale coefficients $s_\textrm{edit}\in[0.7, 1, 1.3, 1.5, 1.7]$. 

\begin{figure}
\addtolength{\tabcolsep}{-5pt}
\hspace{1mm}
\begin{tabular}{c}
 \includegraphics[ width=1\linewidth,trim=0 0 0 0,clip]{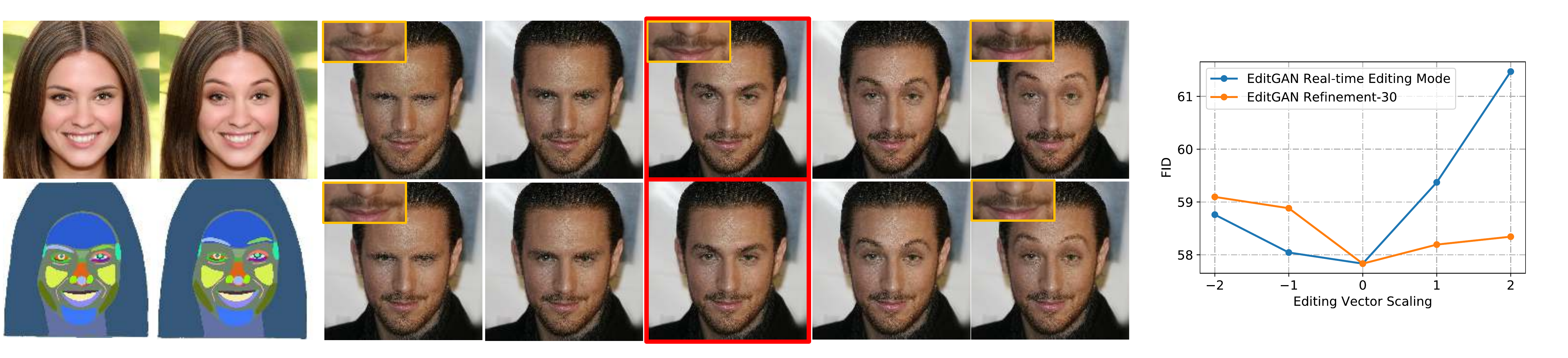}  \\
\end{tabular}
\caption{\footnotesize \textit{Left:} We apply learnt editing vectors with varying scales (see 5 markers in FID plots) both without (top row) and with (bottom row) additional 30-step self-supervised refinement to correct artifacts. Red boxes denote original images. For each class, the leftmost image is the one used to learn the editing vector with the editing result next to it and orginal and modified segmentations below. \textit{Right:} Visual quality after editing with different scales as measured by FID with and without 30-step refinement.}	
\label{fig:face_sacle}
\vspace{-3mm}
\end{figure}

\subsection{Additional Results: Editing Vector Scale Experiment}
In the main paper, in Fig. 9, we presented another ablation study where we studied editing quality when applying edits with different editing vector scales $s_\textrm{edit}$. We analyzed editing quality both visually and quantitatively, both with and without self-supervised refinement. While in the main paper we only presented the results for Car and Cat data, here we additionally show the results on Face images, using an edit that raises eyebrows as example (Fig.~\ref{fig:face_sacle}). Similar to the other results on Car and Cat data, we find that editing by purely applying our learnt editing vector, which can be done at interactive rates, already yields virtually perfect editing results. However, we do observe almost unnoticeable entanglement with the beard. Using self-supervised refinement we can fully remove this editing artifact, if necessary. 

\begin{wraptable}[11]{R}{0.55\textwidth}
\vspace{-3mm}
 \begin{adjustbox}{width=1\linewidth}
{\footnotesize
\addtolength{\tabcolsep}{-3.5pt}
\begin{tabular}{lccccc}
\toprule
Metric &  \# Mask     &  \# Attribute  & Attribute                &  FID $\downarrow$     & ID Score  $\uparrow$ \\
       &  Annot. & Annot.    & Acc.(\%) $\uparrow$  &                   &                      \\
\midrule
MaskGAN~\cite{lee2020maskgan} &  30,000  & - & 65.7 & 	\textbf{ 18.3}  &  0.5229\\ 
\midrule
LocalEditing~\cite{collins2020editing} & - & -  &   23.7 &  20.4 & 0.5726 \\ 
\midrule
InterFaceGAN~\cite{shen2020interfacegan} & - & 30,000  & 	79.5 & 	 34.4  &  0.6560 \\ 
\midrule
EditGAN (ours) & 16 &  - &  \textbf{88.0} &   	 32.1 &  0.6422\\ 
\midrule
EditGAN$^+30$ (ours) & 16 &  - &  81.4 &  	 31.8	 & \textbf{0.6625} \\ 
\bottomrule
\end{tabular}   
}
\end{adjustbox}
\vspace{-2mm}
\caption{\small Quantitative comparisons to multiple baselines on the smile edit 4k benchmark.} 
\label{tbl:all_results}
\end{wraptable}

\subsection{Additional Results: Smile Edit Benchmark with more Test Images}
In Tab. 1 of the main paper, we use MaskGAN's~\cite{lee2020maskgan} smile edit benchmark. The FID scores are calculated between 400 edited test images and the CelebA-HD test database, which enables a fair comparison with existing approaches and directly follows the practice by MaskGAN. Although the estimates may be biased with respect to the true FID due to the limited number of test images~\cite{chong2019effectively}, we expect that they nevertheless provide a fair comparison between the different methods. However, here we re-calculate FID as well as attribute accuracy and ID score using 10 times as many images, i.e. 4000 images, from the training set from MaskGAN. Notice that only MaskGAN uses this data for training, while the GANs of all other baselines, including our EditGAN, are based on the FFHQ faces data and do not use this annotated training data that MaskGAN relies on. Hence, calculating the FID using these 4000 images is advantageous for MaskGAN. 

We show results in Tab.~\ref{tbl:all_results}. Since we use different and much more data for evaluation compared to the evaluation reported in the table in the paper, the numbers are different. However, the rankings and comparisons between the methods remain the same and the conclusions are the same. In particular, EditGAN achieves the best attribute accuracies and ID scores. MaskGAN achieves a relatively low FID, but this is simply due to the unfair comparison, as discussed above. MaskGAN still performs significantly worse than InterfaceGAN and EditGAN in attribute accuracy and ID score.

\section{Computational Resources}
Training of the underlying StyleGAN2, the encoder, and the segmentation branch, as well as optimization for embedding and editing were performed using NVIDIA Tesla V100 GPUs on an in-house GPU cluster. Overall, the project used approximately 14,000 GPU hours (according to internal GPU usage reports), of which around 3,500 GPU hours were used for the final experiments, and the rest for exploration and testing during the earlier stages of the research project.

\vspace{-2mm}

\section{Additional Qualitative Results}
Below, we present further qualitative results. 

We first demonstrate particularly challenging editing operations where we try to disentangle semantically related parts. For example, we want to lift the right eyebrow while keeping the left eyebrow unchanged. We present the results ins Fig.~\ref{fig:disentanglement}. Furthermore, we again demonstrate the ability to combine multiple different edits in Fig~\ref{fig:combine}. We also invite the reader to watch our video, which shows latent code interpolations between the edits. 
Finally, for all edits we perform in the main paper, we first show the image and segmentation mask pairs that were used to learn the latent space editing vectors, and then we present a few more editing results on GAN-generated images (Figs.~\ref{fig:face_gaze}-\ref{fig:cat}).

\newcommand\hhf{4cm}
\newcommand\wwf{4cm}
\newcommand\hhfl{7.7cm}
\newcommand\wwfl{7.7cm}
\newcommand\hhc{3cm}
\newcommand\hhcc{3.5cm}

\newcommand\hhfm{6cm}

\newcommand\hhcl{5.7cm}
\newcommand\hhcm{4.5cm}

\begin{figure}
\addtolength{\tabcolsep}{-6pt}
\hspace{-1.5cm}

\begin{tabular}{cccc}
 \includegraphics[height=\hhcc,width=\hhcc, trim=0 0 0 0,clip]{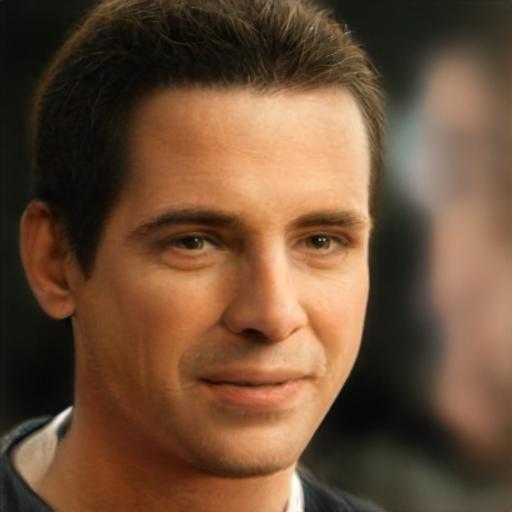} &
  \includegraphics[height=\hhcc,width=\hhcc, trim=0 0 0 0,clip]{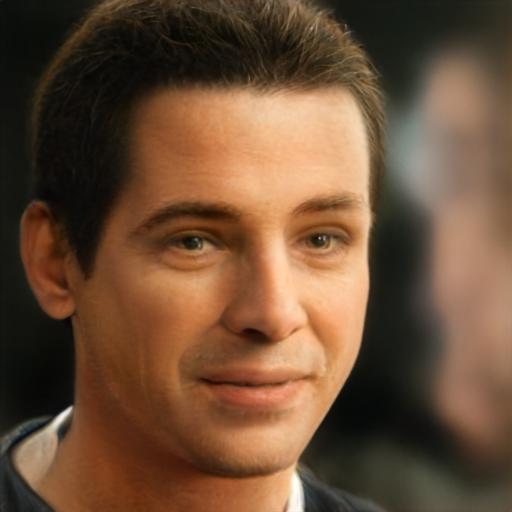}  &
 \includegraphics[height=\hhcc,width=\hhcc, trim=0 0 0 0,clip]{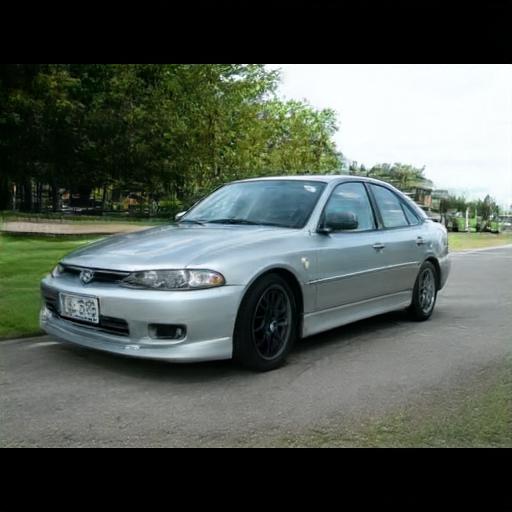} & 
 \includegraphics[height=\hhcc,width=\hhcc, trim=0 0 0 0,clip]{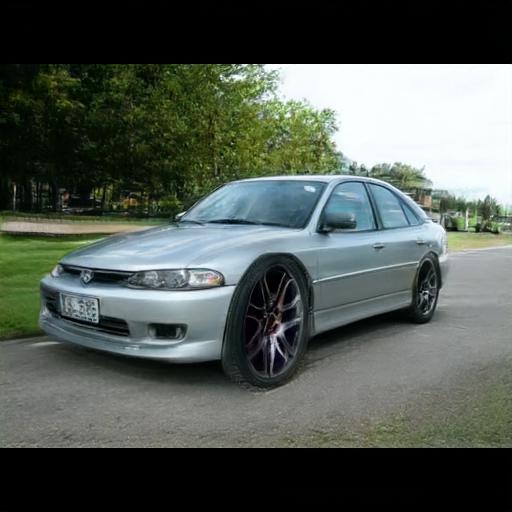}   \\
\end{tabular}

\caption{\footnotesize We demonstrate challenging editing operations where we disentangle semantically related parts. The presented results correspond to pure optimization-based editing. \textit{First example}: Lift right eyebrow while keeping the left eyebrow unchanged. \textit{Second example}: Enlarge the front wheel while keeping the back wheel unchanged.}
\label{fig:disentanglement}
\vspace{-3mm}
\end{figure}

\begin{figure}
\addtolength{\tabcolsep}{-6pt}
\hspace{1.75cm}
\begin{tabular}{cc}
 \includegraphics[height=\hhfm,width=\hhfm, trim=0 0 0 0,clip]{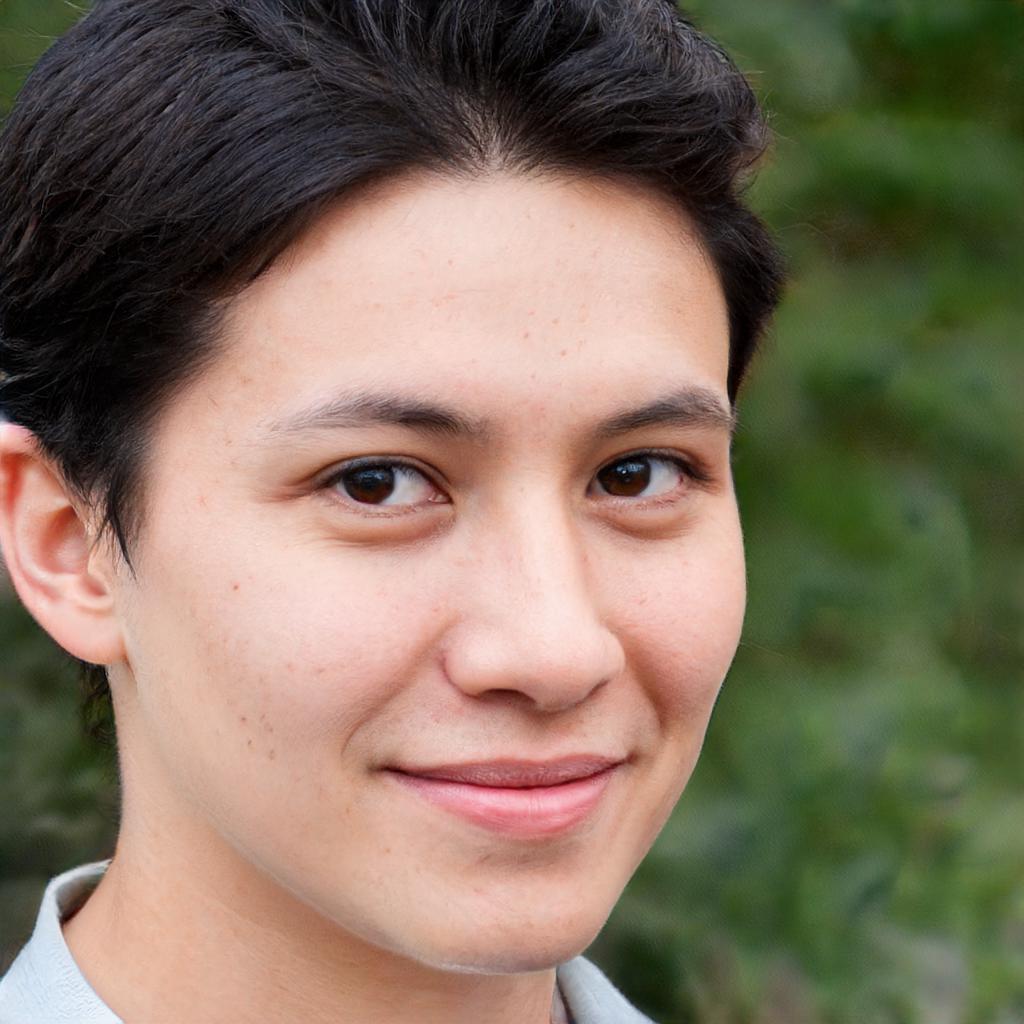} & \includegraphics[height=\hhfm,width=\hhfm, trim=0 0 0 0,clip]{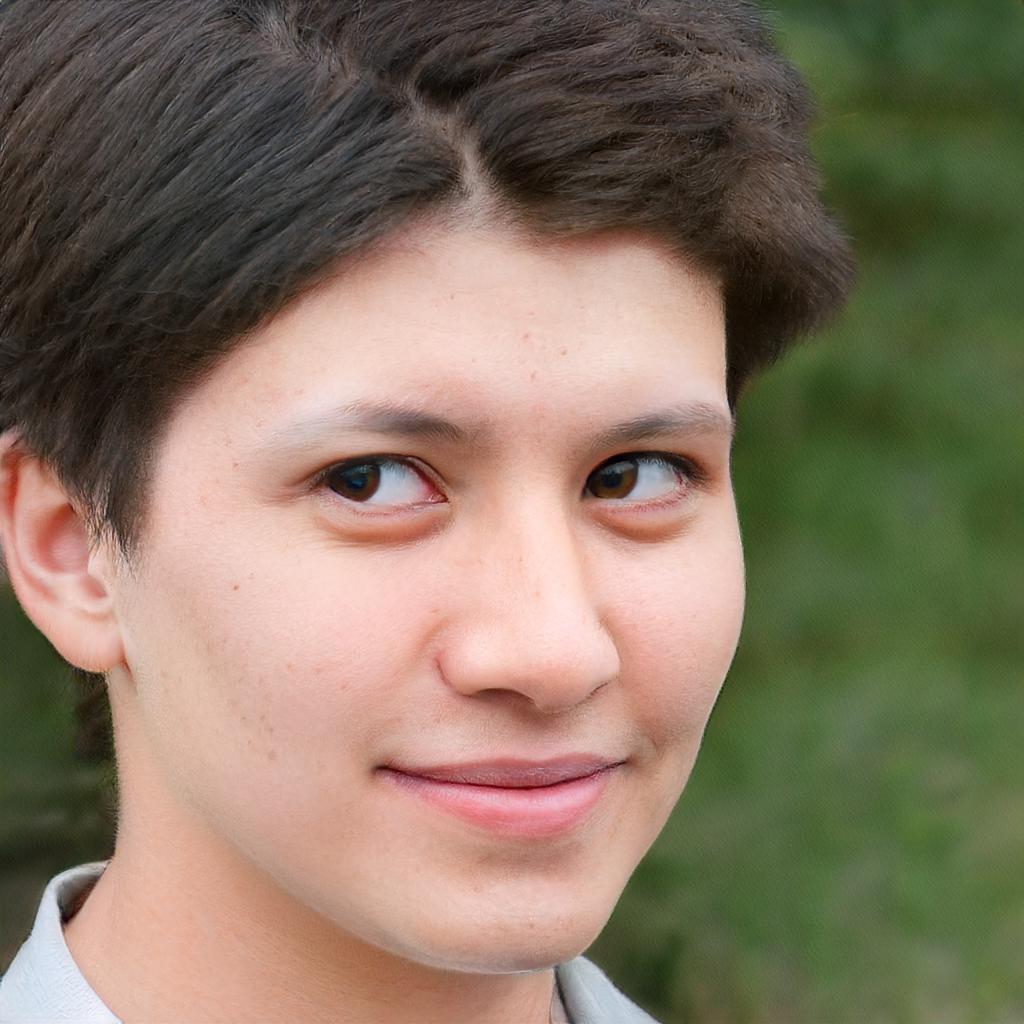}  \\
 \includegraphics[height=\hhfm,width=\hhfm, trim=0 0 0 0,clip]{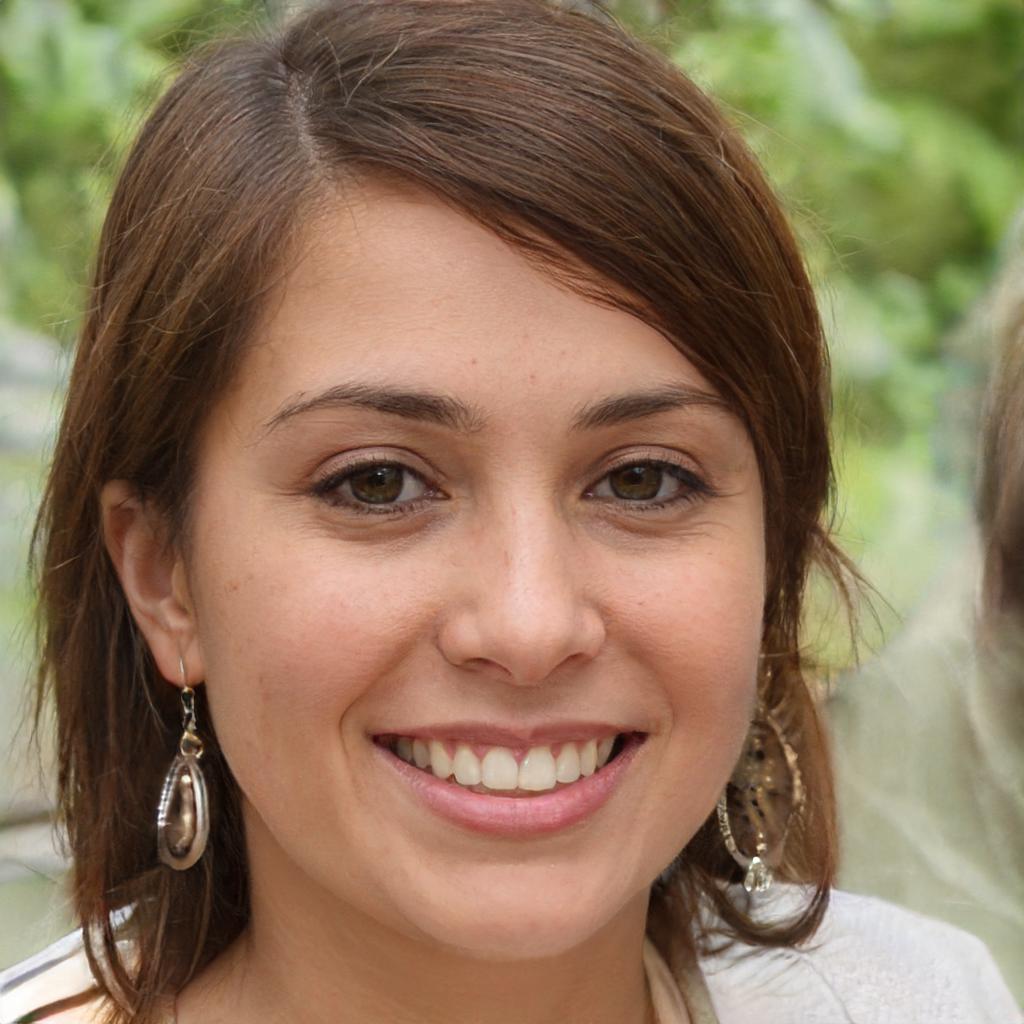} & \includegraphics[height=\hhfm,width=\hhfm, trim=0 0 0 0,clip]{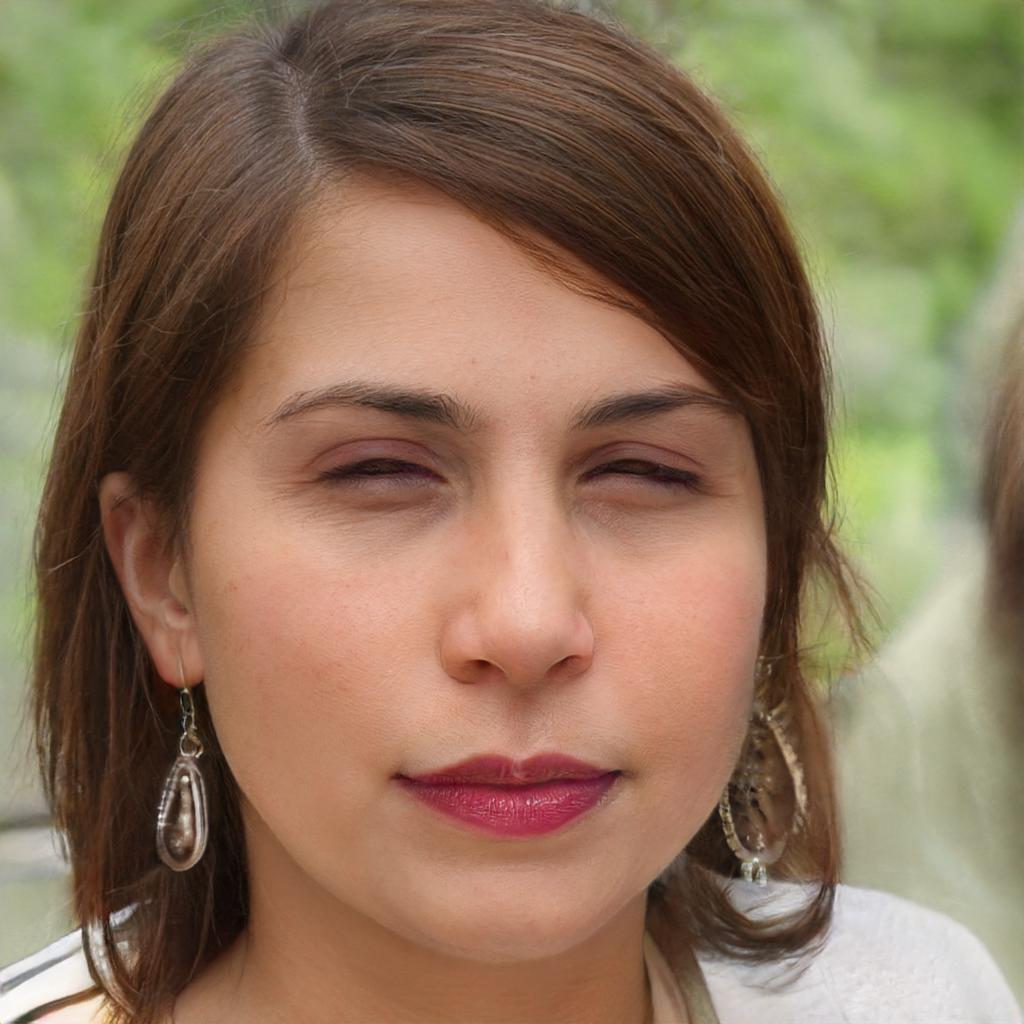}   \\
\end{tabular}

\begin{tabular}{cc}
\hspace{1.75cm}
 \includegraphics[height=\hhcm,width=\hhfm, trim=0 58 0 58,clip]{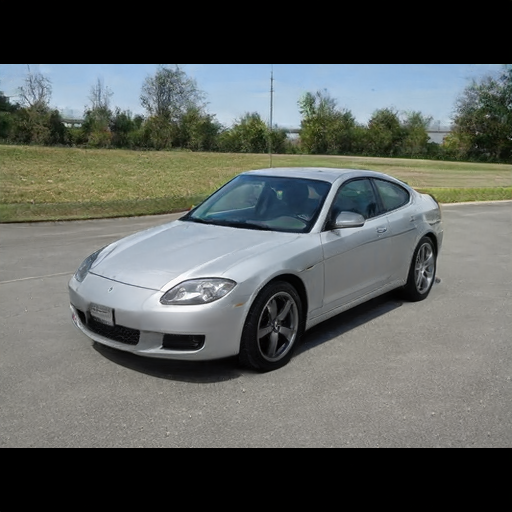} & \includegraphics[height=\hhcm,width=\hhfm, trim=0 58 0 58,clip]{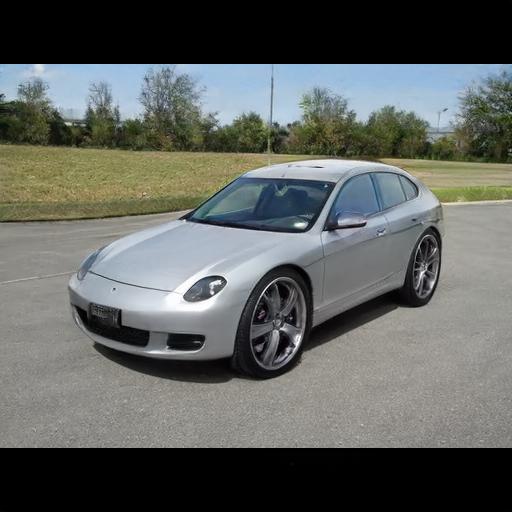}  \\
 \hspace{1.75cm}
 \includegraphics[height=\hhcm,width=\hhfm, trim=0 58 0 58,clip]{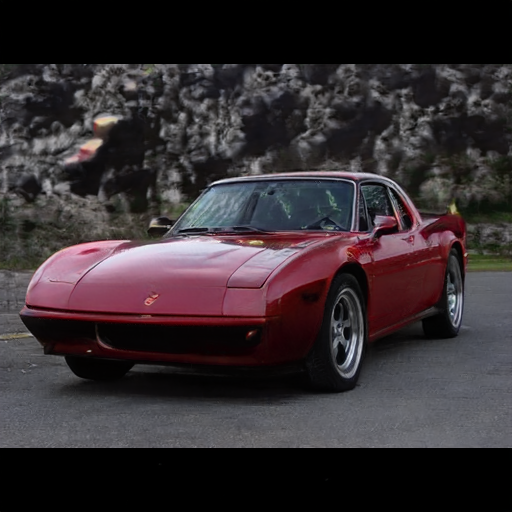} & \includegraphics[height=\hhcm,width=\hhfm, trim=0 58 0 58,clip]{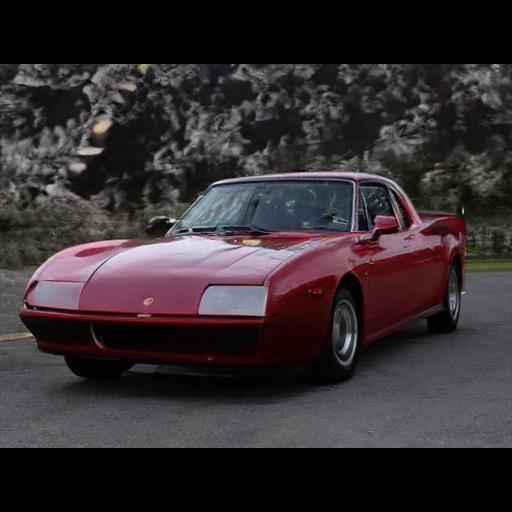}  \\
\end{tabular}

\caption{\footnotesize  {\bf Combining multiple edits.} Results are based on editing with learnt editing vector and 30 steps of self-supervised refinement. Edits in detail: \textit{First row}: Slight frown, look left, add hair, remove smile wrinkle. \textit{Second row}: Close eyes, close mouth, remove smile wrinkle. \textit{Third row}: Lift back of the car, enlarge wheels, shrink front light. \textit{Fourth row}: Enlarge front light, shrink wheels. Please also see attached video which shows latent code interpolations between editing operations.}
\label{fig:combine}
\vspace{-3mm}
\end{figure}

\begin{figure}
\addtolength{\tabcolsep}{-10pt}
\hspace{1mm}
\begin{tabular}{cccc}
 \includegraphics[height=\hhf,width=\wwf, trim=0 0 0 0,clip]{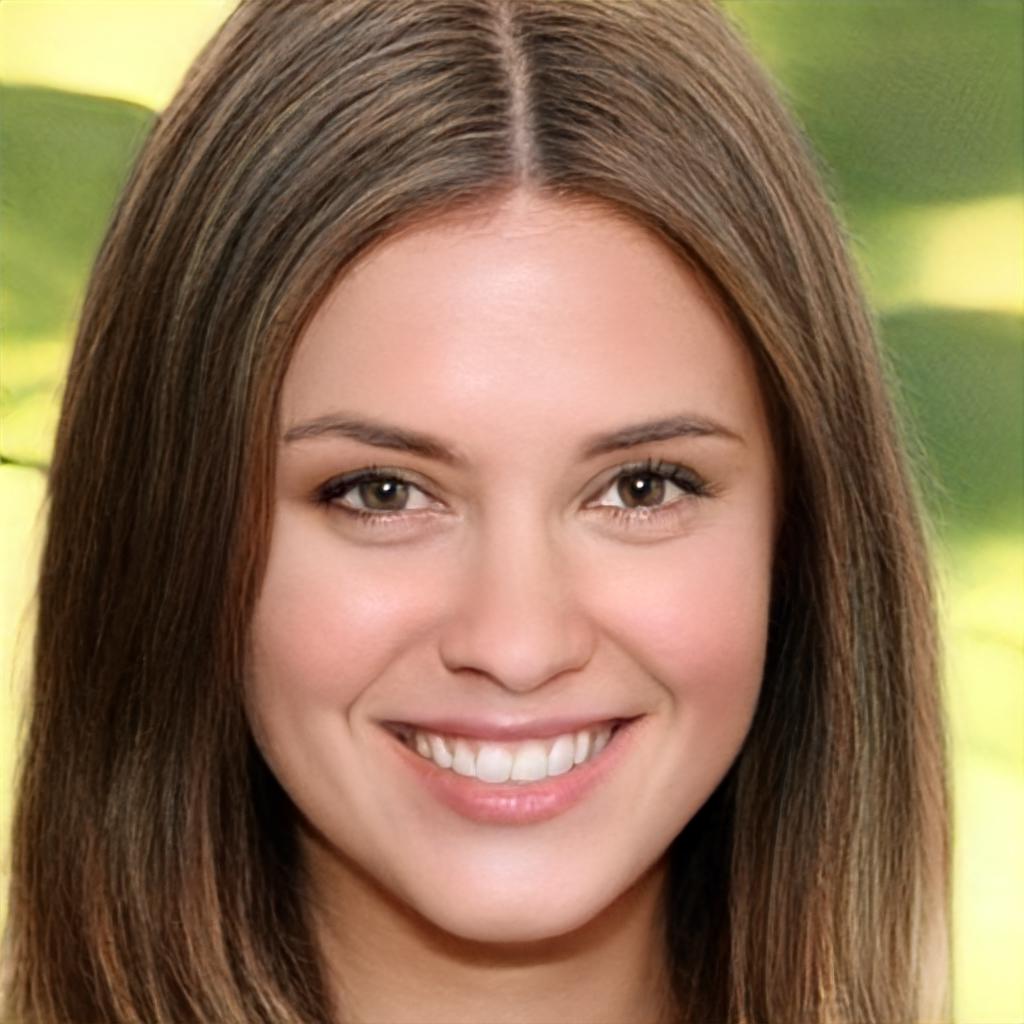} & \includegraphics[height=\hhf,width=\wwf, trim=0 0 0 0,clip]{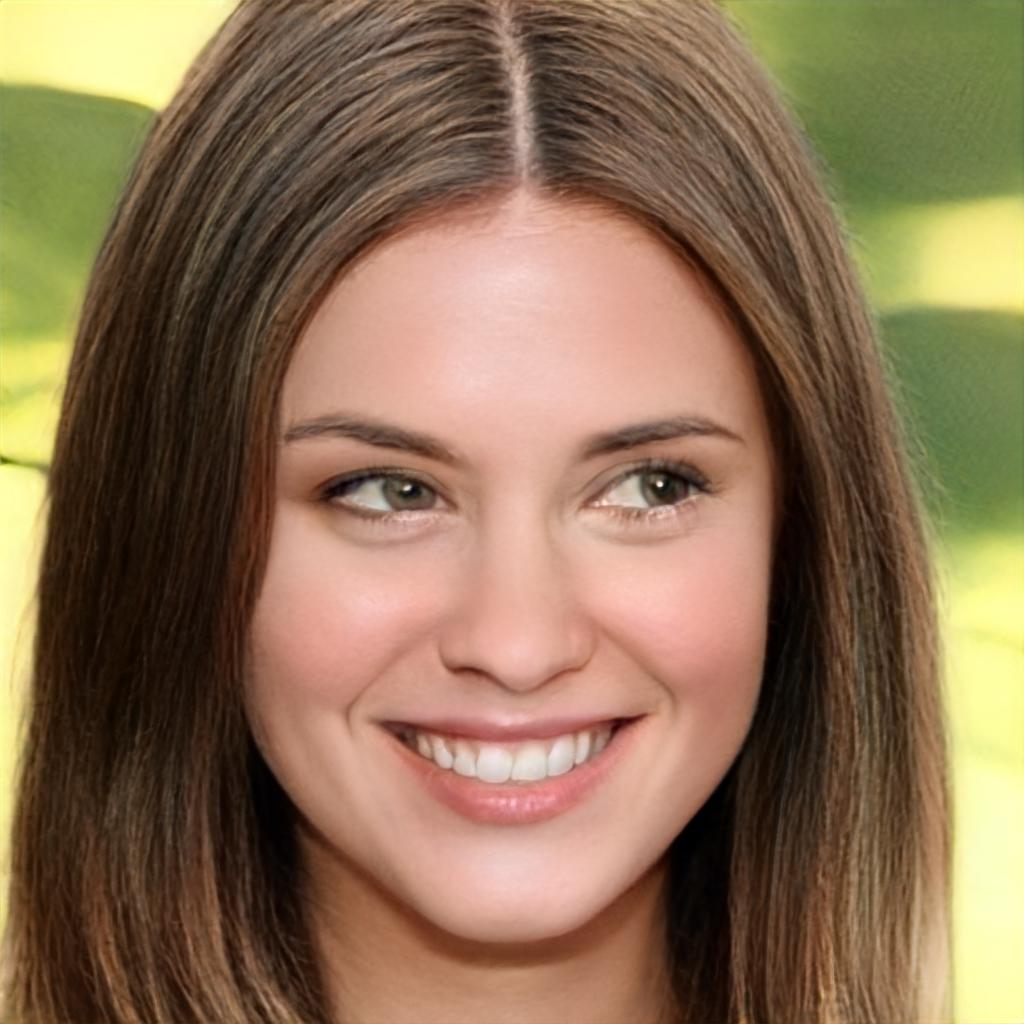}  &
 \includegraphics[height=\hhf,width=\wwf, trim=0 0 0 0,clip]{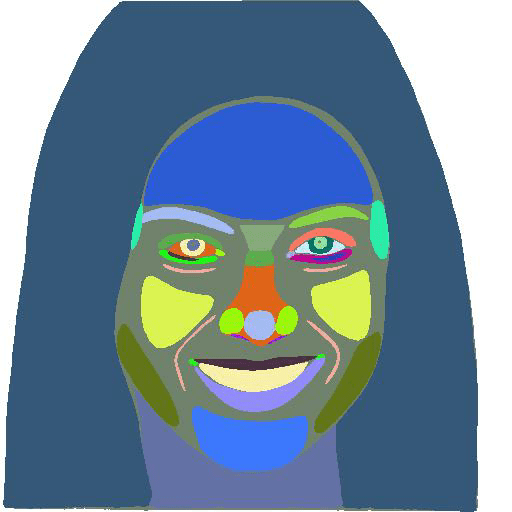} & \includegraphics[height=\hhf,width=\wwf, trim=0 0 0 0,clip]{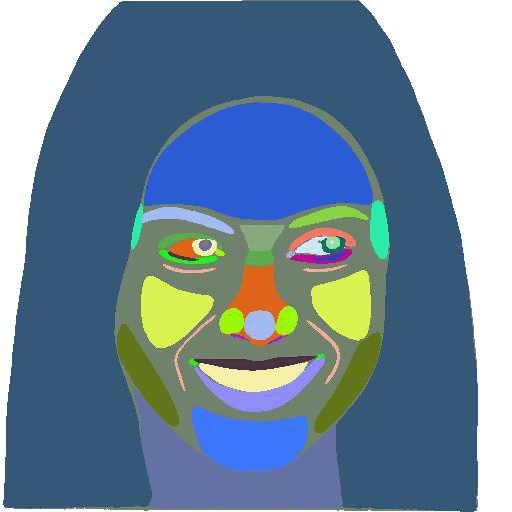}  \\
\end{tabular}

\begin{tabular}{cc}
 \includegraphics[height=\hhfl,width=\wwfl, trim=0 0 0 0,clip]{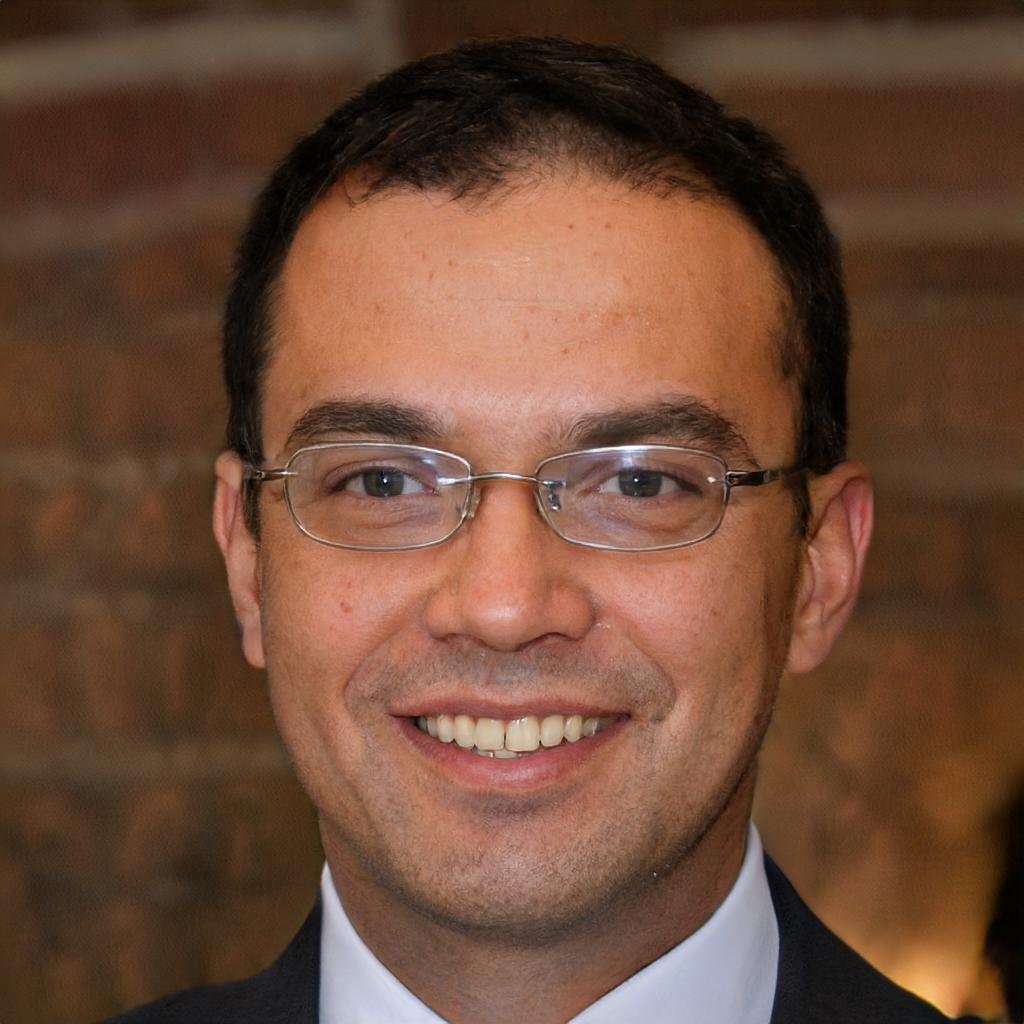} & \includegraphics[height=\hhfl,width=\wwfl, trim=0 0 0 0,clip]{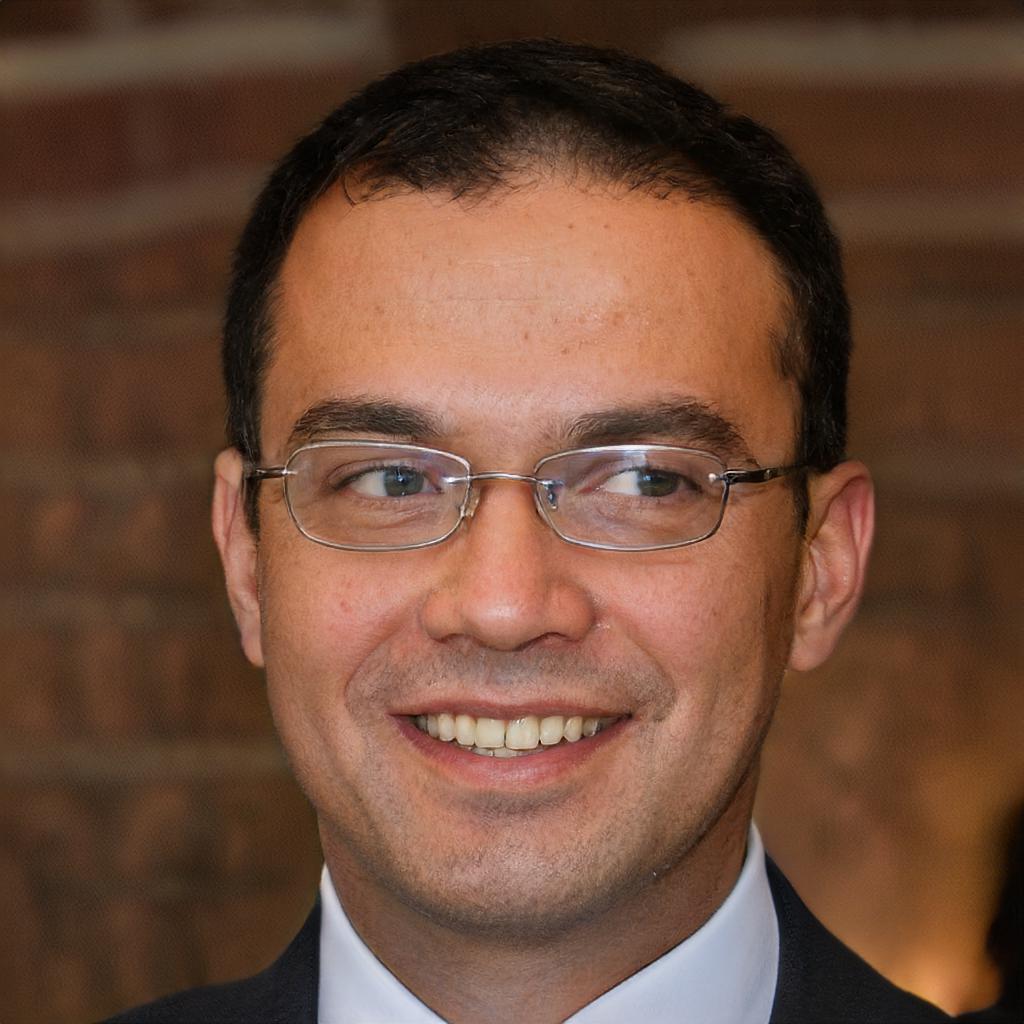}  \\
 \includegraphics[height=\hhfl,width=\wwfl, trim=0 0 0 0,clip]{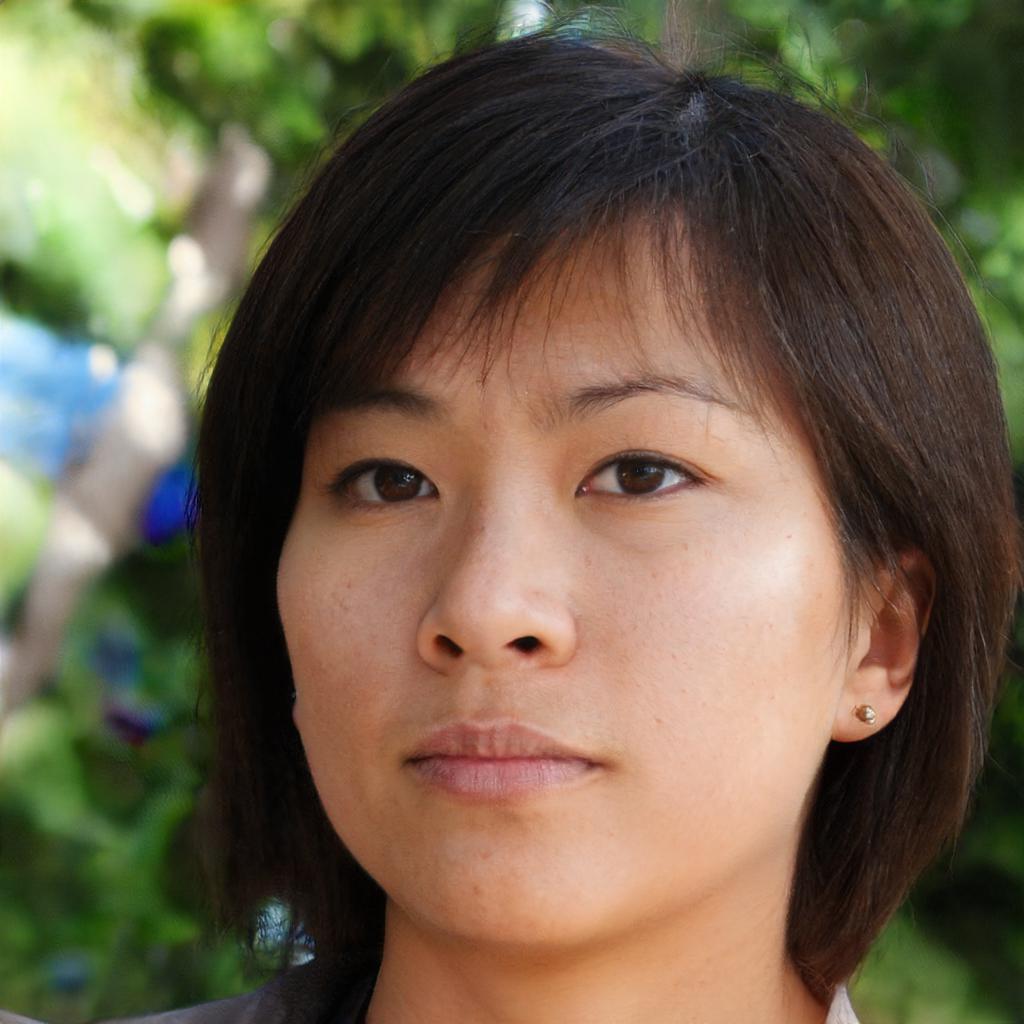} & \includegraphics[height=\hhfl,width=\wwfl, trim=0 0 0 0,clip]{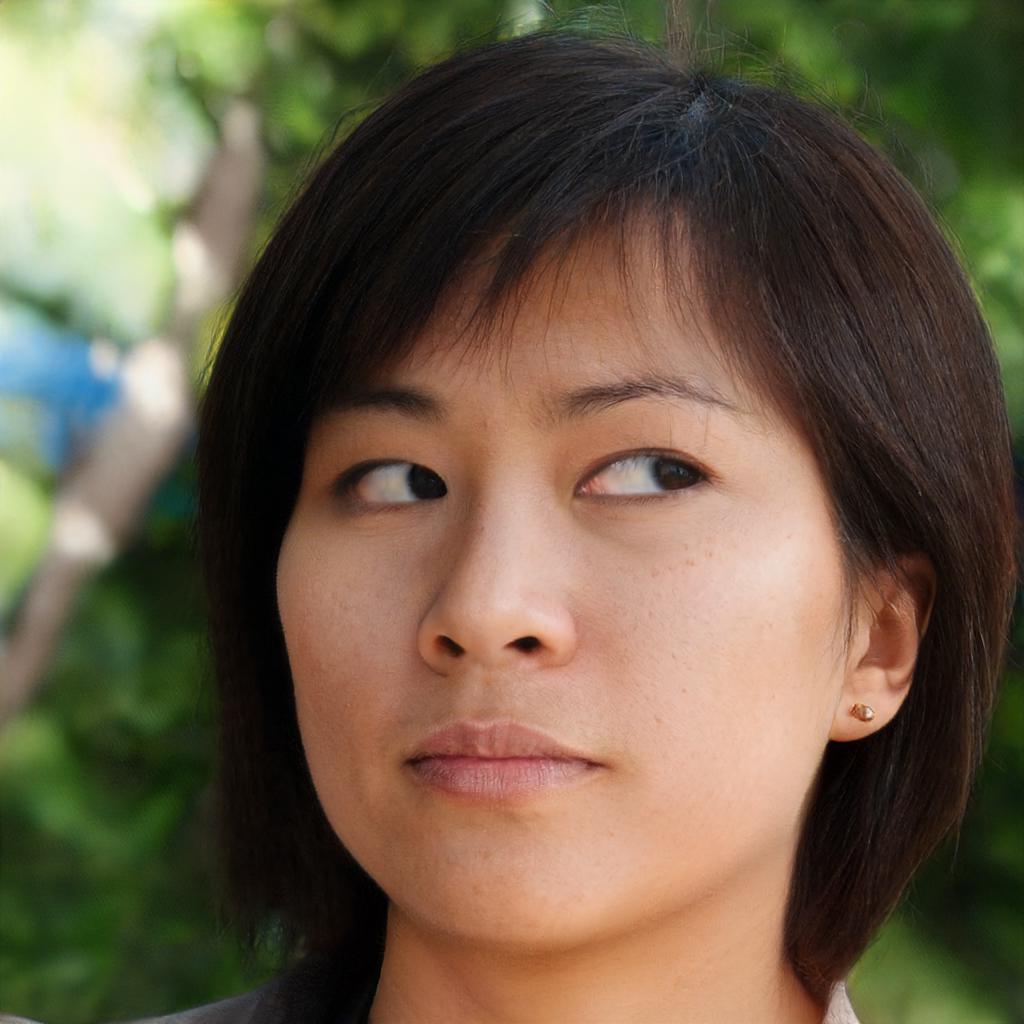}  \\
\end{tabular}

\caption{\footnotesize  {\bf Gaze position editing.} \textit{First row}: Image and mask pair to learn editing vector. Images are images before editing and after editing. Segmentation masks are before editing and target segmentation mask after manual modification. \textit{Second and third rows}: Applying the learnt edit on new images.}
\label{fig:face_gaze}
\vspace{-3mm}
\end{figure}

\begin{figure}
\addtolength{\tabcolsep}{-10pt}
\hspace{1mm}
\begin{tabular}{cccc}
 \includegraphics[height=\hhf,width=\wwf, trim=0 0 0 0,clip]{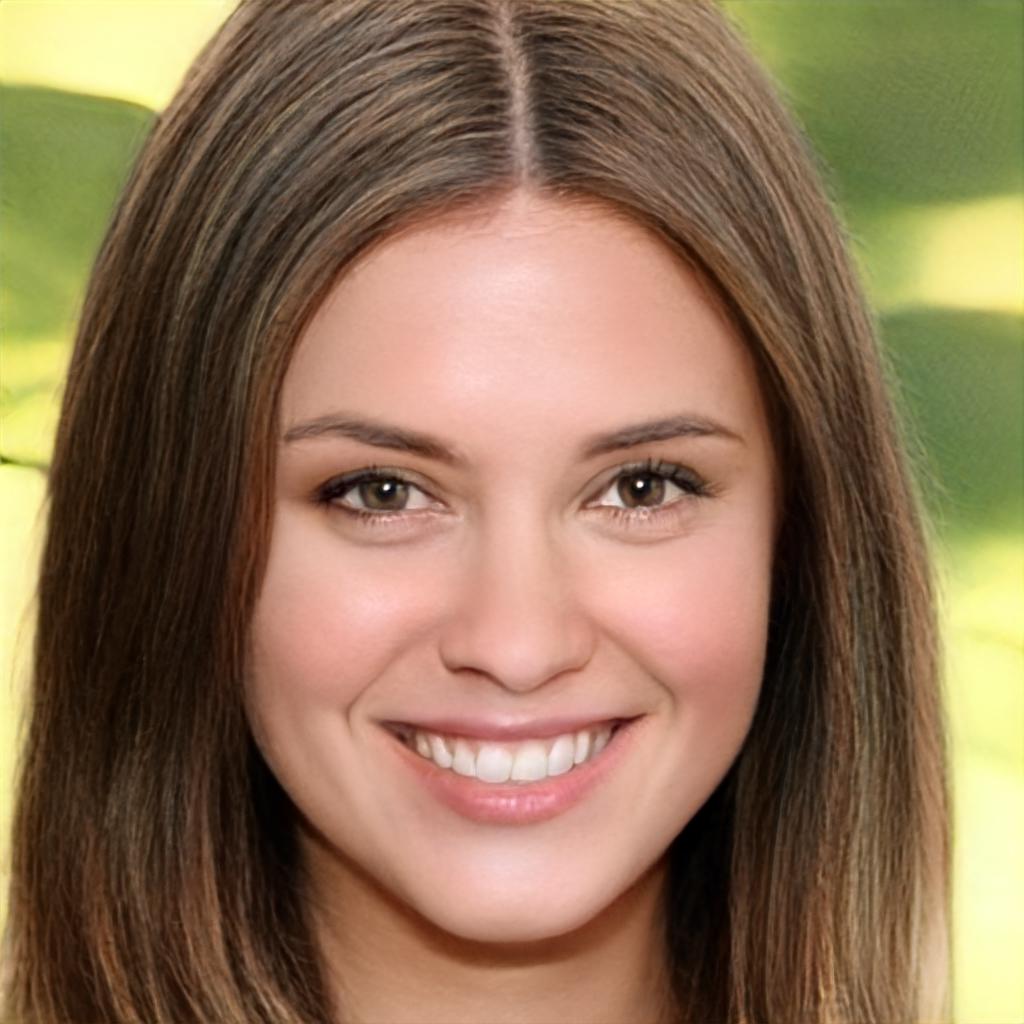} & \includegraphics[height=\hhf,width=\wwf, trim=0 0 0 0,clip]{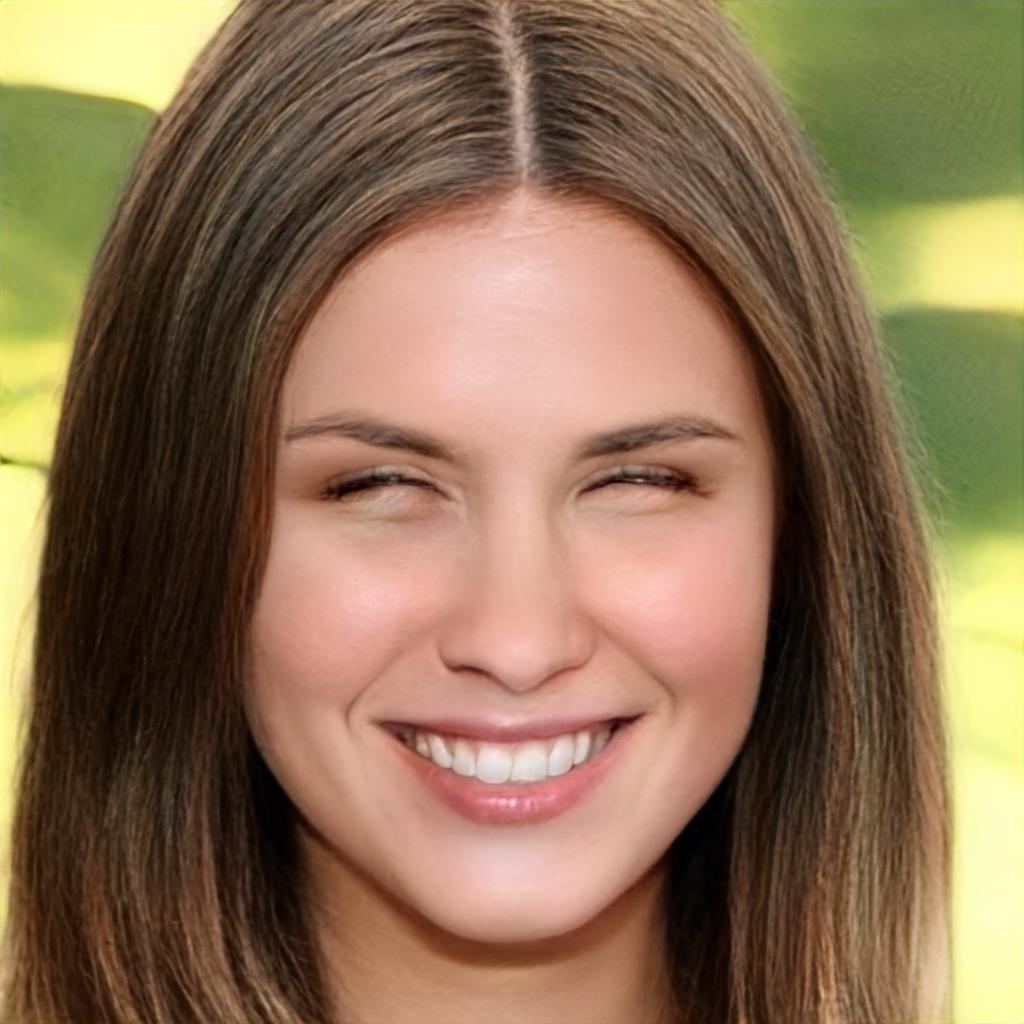}  &
 \includegraphics[height=\hhf,width=\wwf, trim=0 0 0 0,clip]{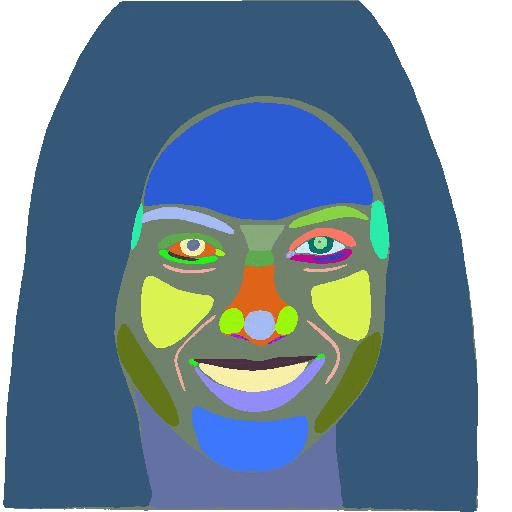} & \includegraphics[height=\hhf,width=\wwf, trim=0 0 0 0,clip]{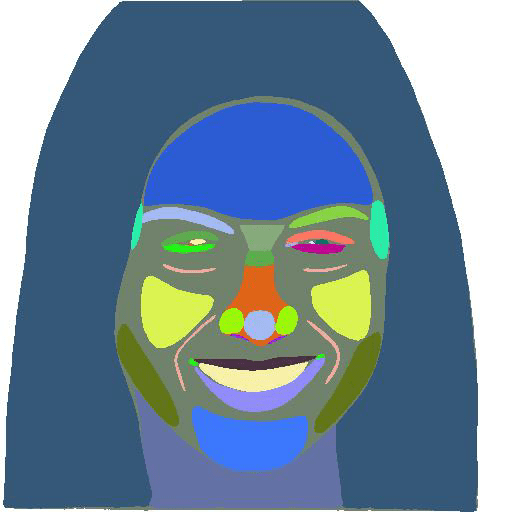}  \\
\end{tabular}

\begin{tabular}{cc}
 \includegraphics[height=\hhfl,width=\wwfl, trim=0 0 0 0,clip]{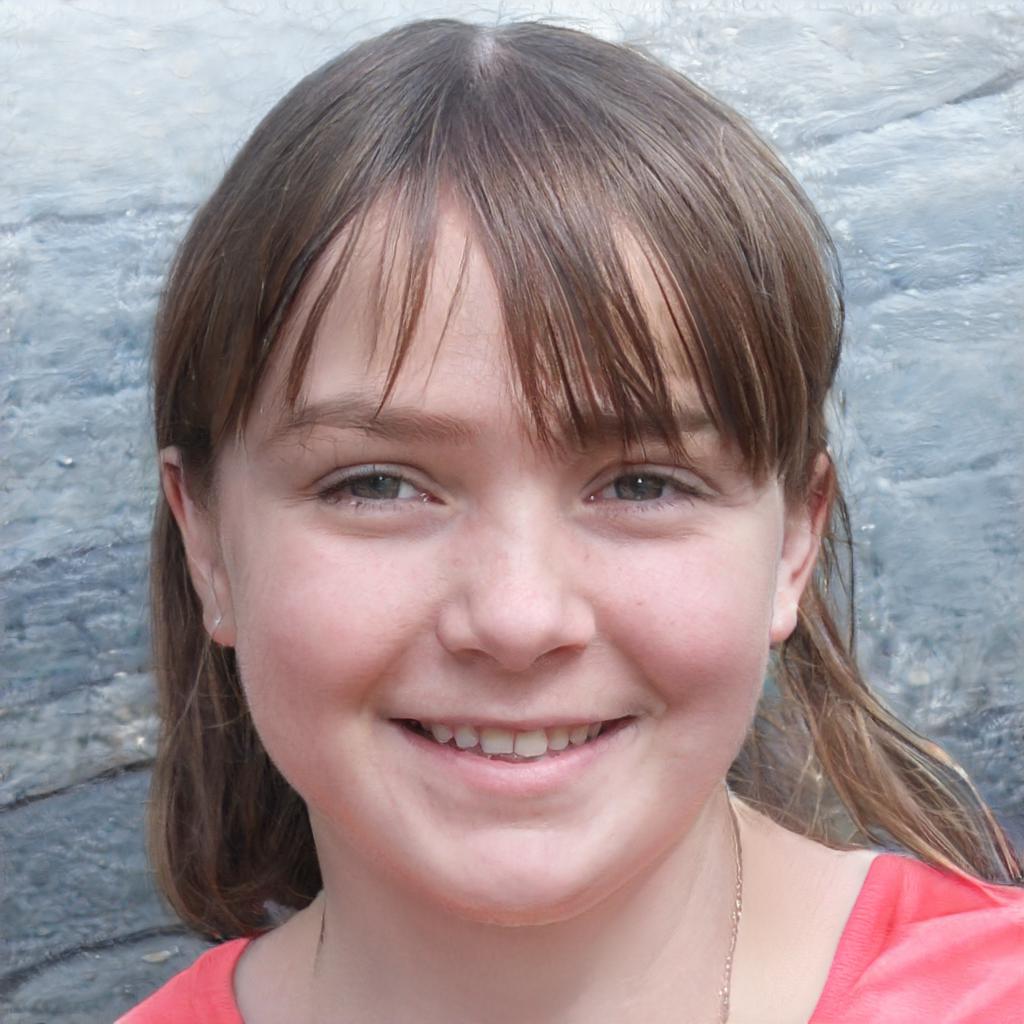} & \includegraphics[height=\hhfl,width=\wwfl, trim=0 0 0 0,clip]{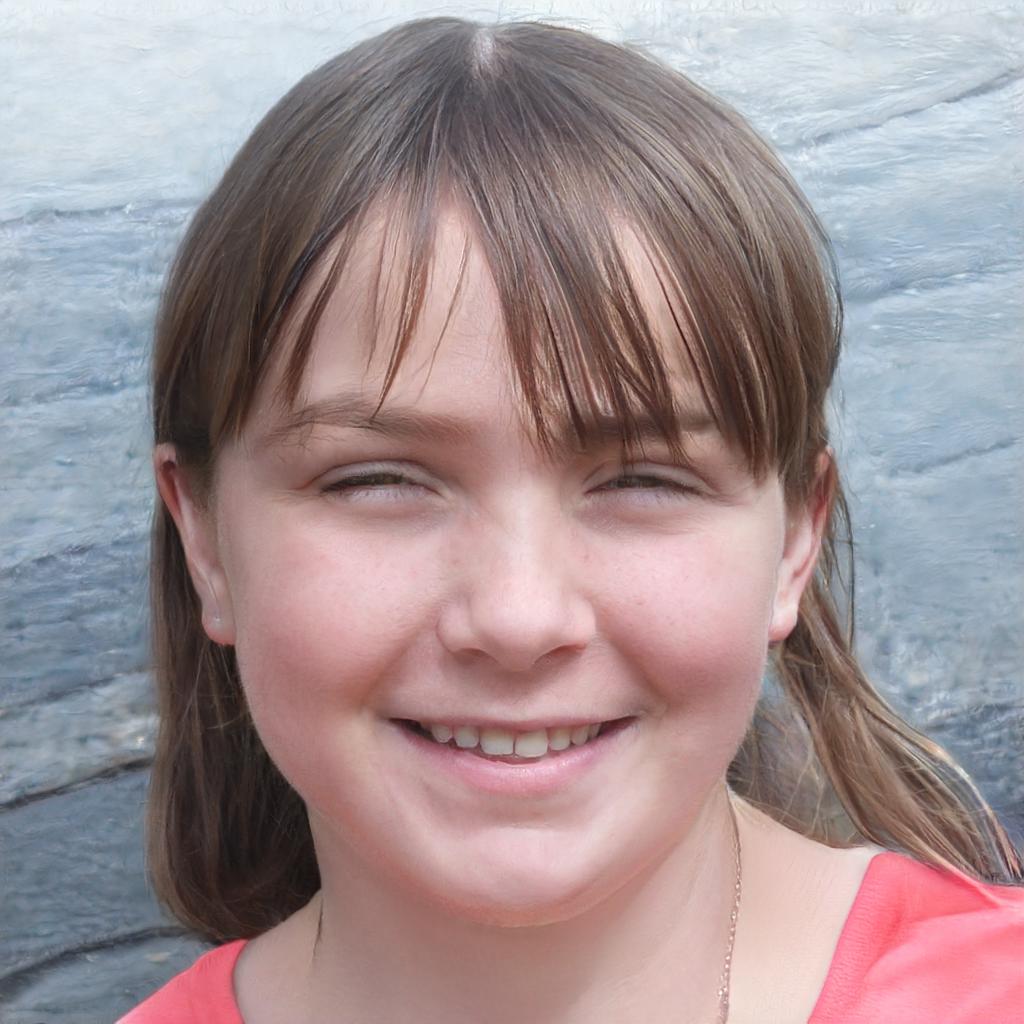}  \\
 \includegraphics[height=\hhfl,width=\wwfl, trim=0 0 0 0,clip]{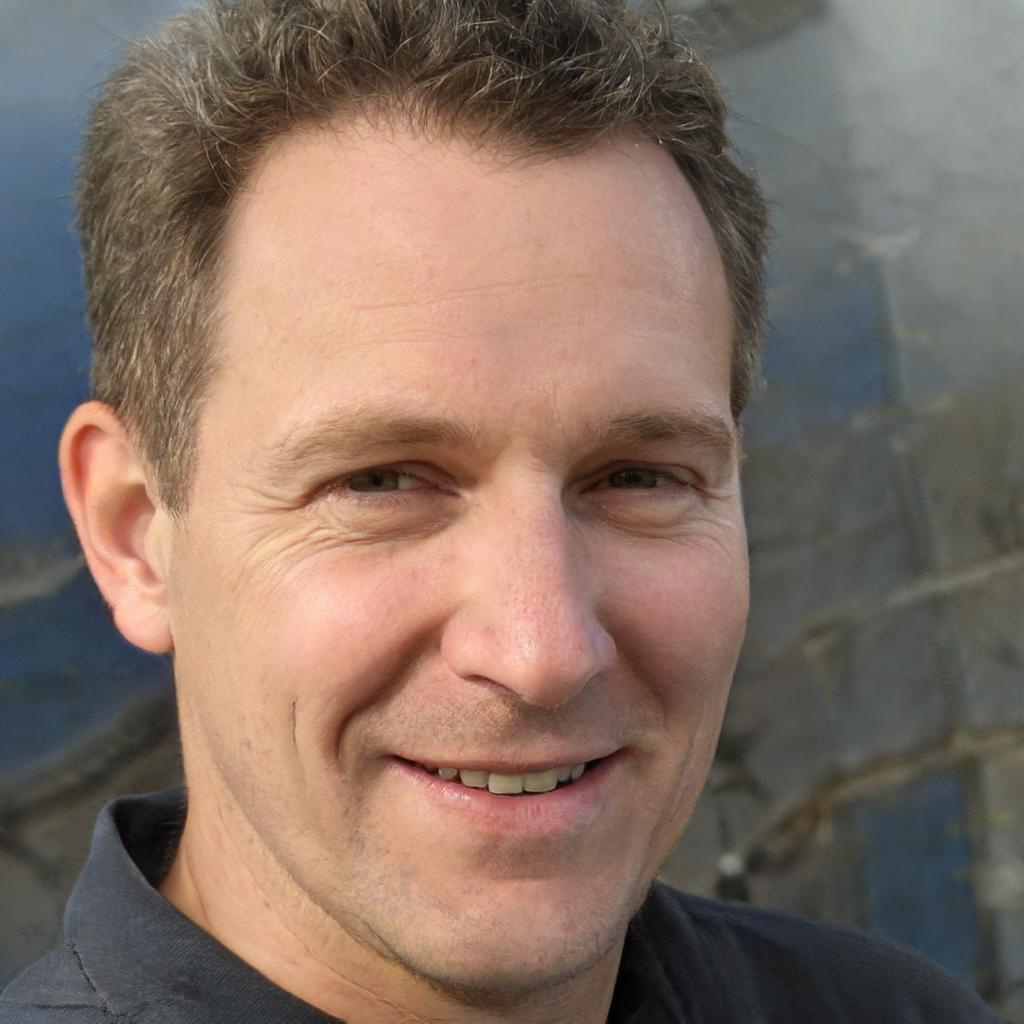} & \includegraphics[height=\hhfl,width=\wwfl, trim=0 0 0 0,clip]{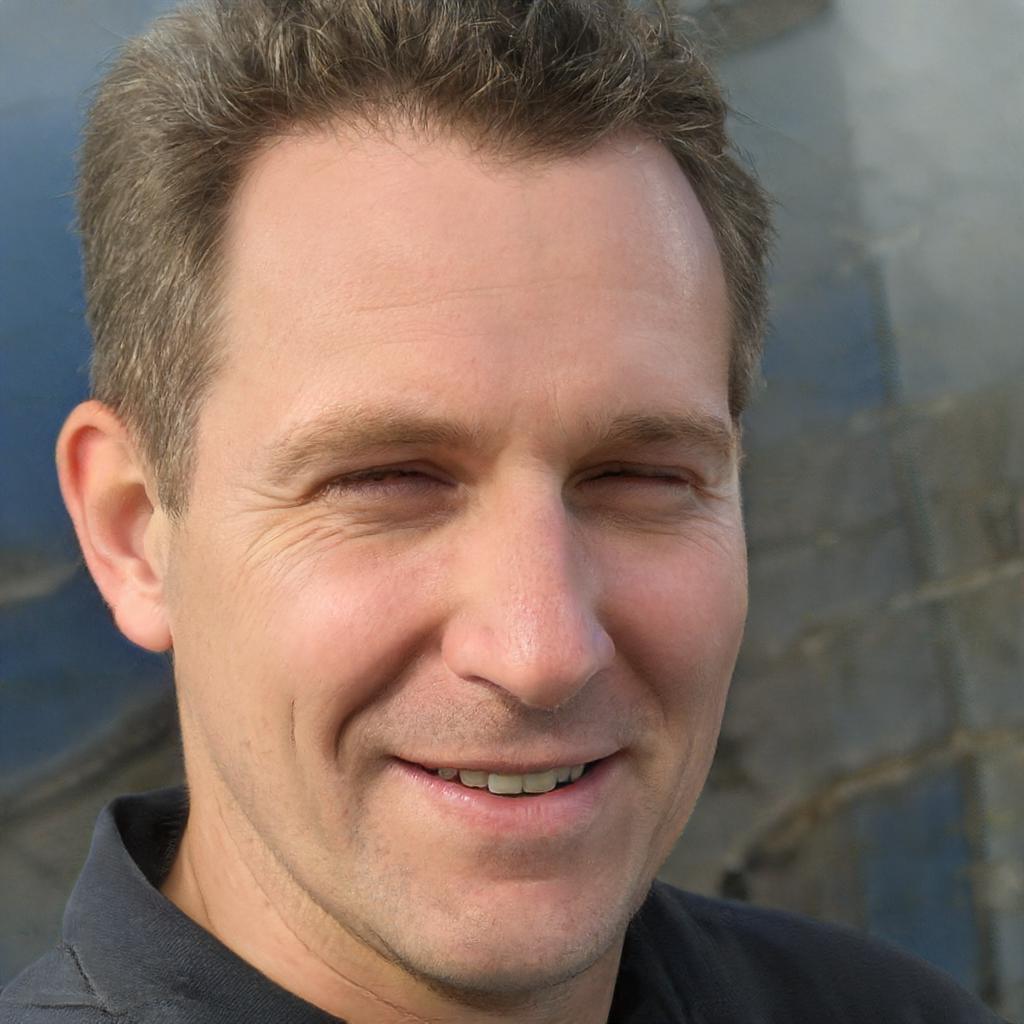}  \\
\end{tabular}

\caption{\footnotesize  {\bf Closing eyes editing.} \textit{First row}: Image and mask pair to learn editing vector. Images are images before editing and after editing. Segmentation masks are before editing and target segmentation mask after manual modification. \textit{Second and third rows}: Applying the learnt edit on new images.}
\label{fig:face_close_eyes}
\vspace{-3mm}
\end{figure}

\begin{figure}
\addtolength{\tabcolsep}{-10pt}
\hspace{1mm}
\begin{tabular}{cccc}
 \includegraphics[height=\hhf,width=\wwf, trim=0 0 0 0,clip]{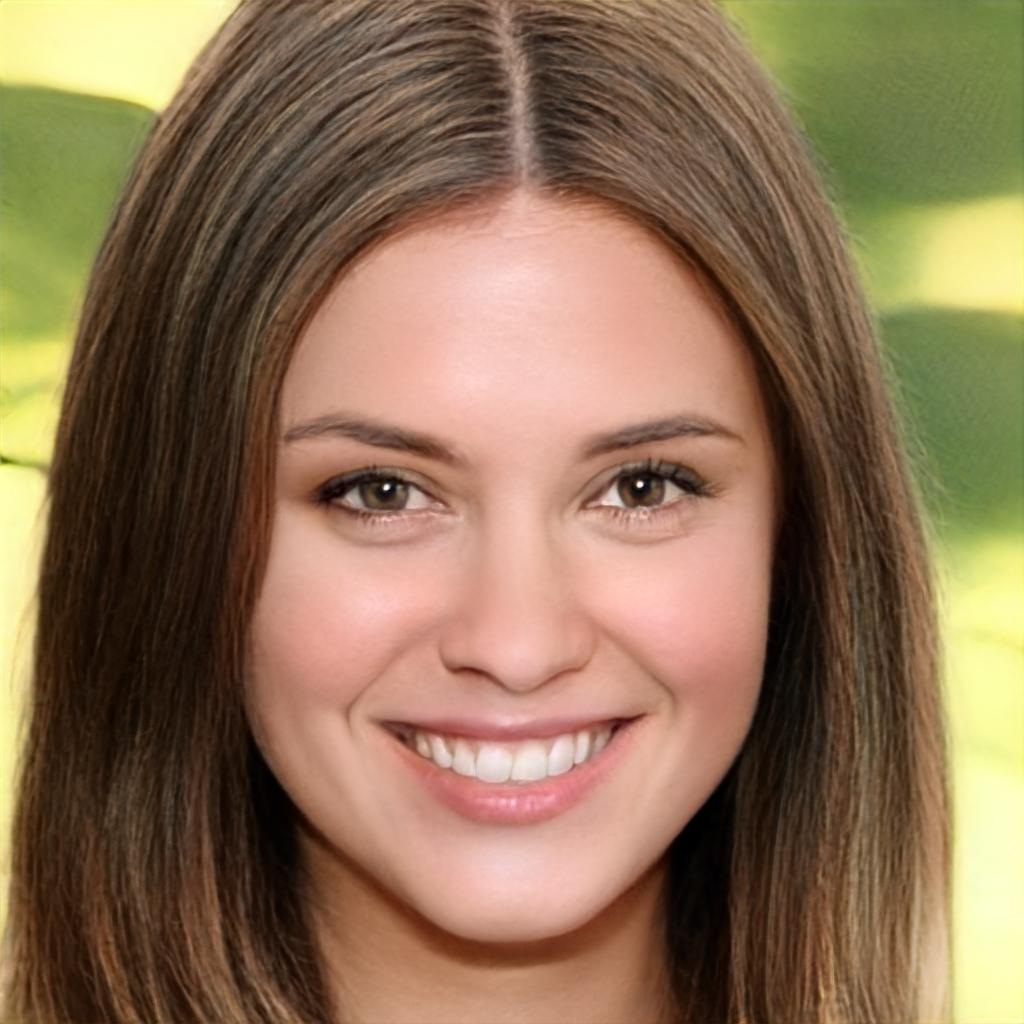} & \includegraphics[height=\hhf,width=\wwf, trim=0 0 0 0,clip]{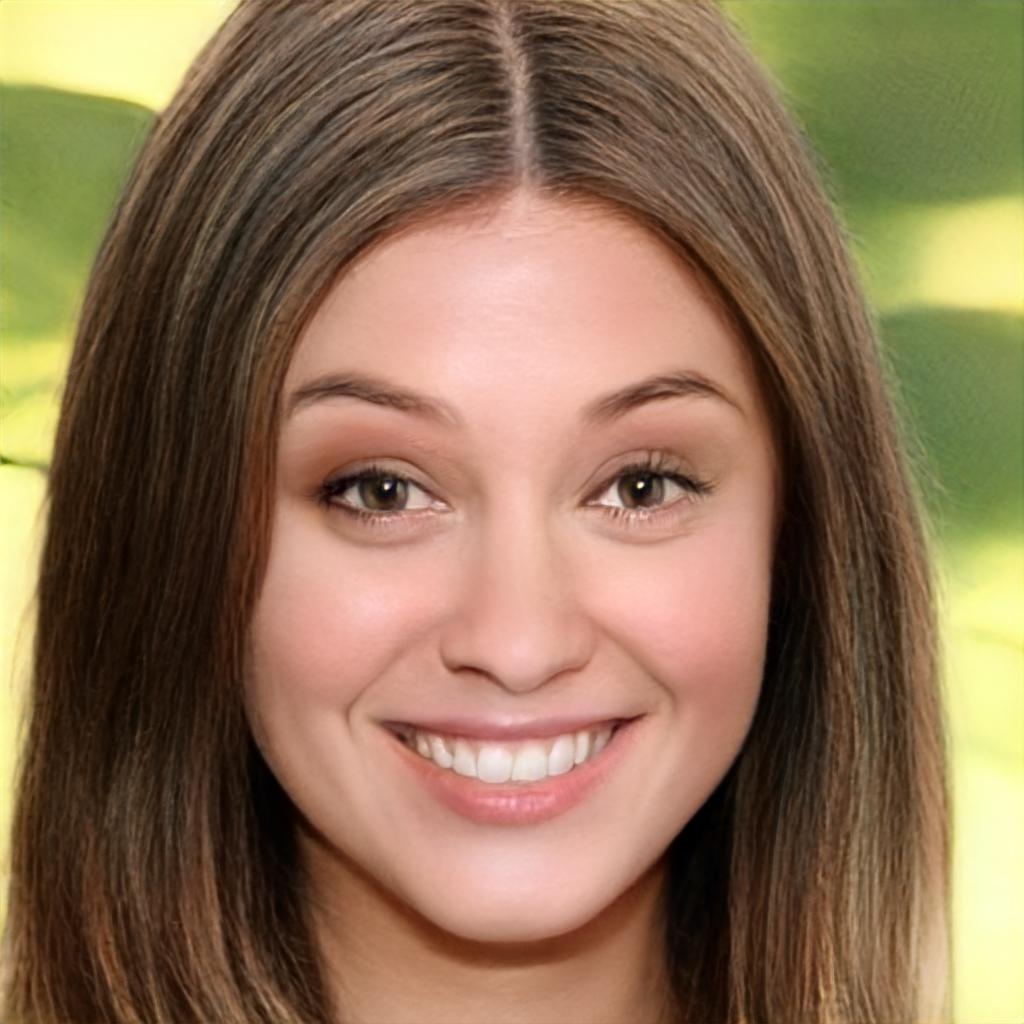}  &
 \includegraphics[height=\hhf,width=\wwf, trim=0 0 0 0,clip]{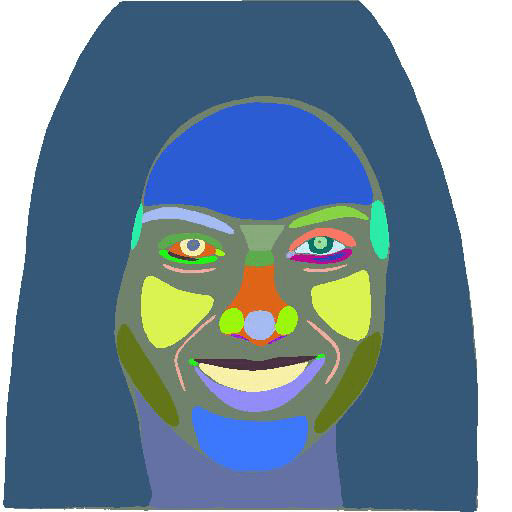} & \includegraphics[height=\hhf,width=\wwf, trim=0 0 0 0,clip]{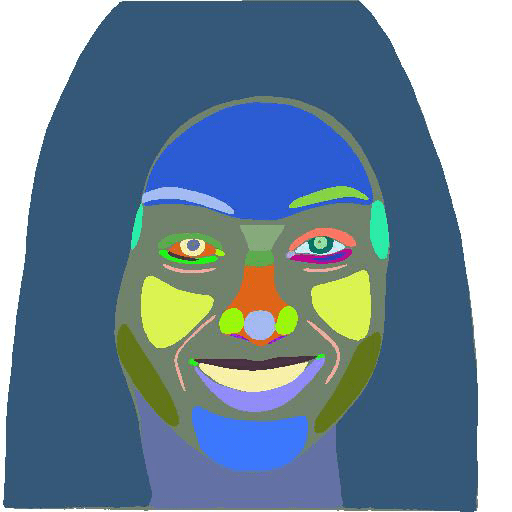}  \\
\end{tabular}

\begin{tabular}{cc}
 \includegraphics[height=\hhfl,width=\wwfl, trim=0 0 0 0,clip]{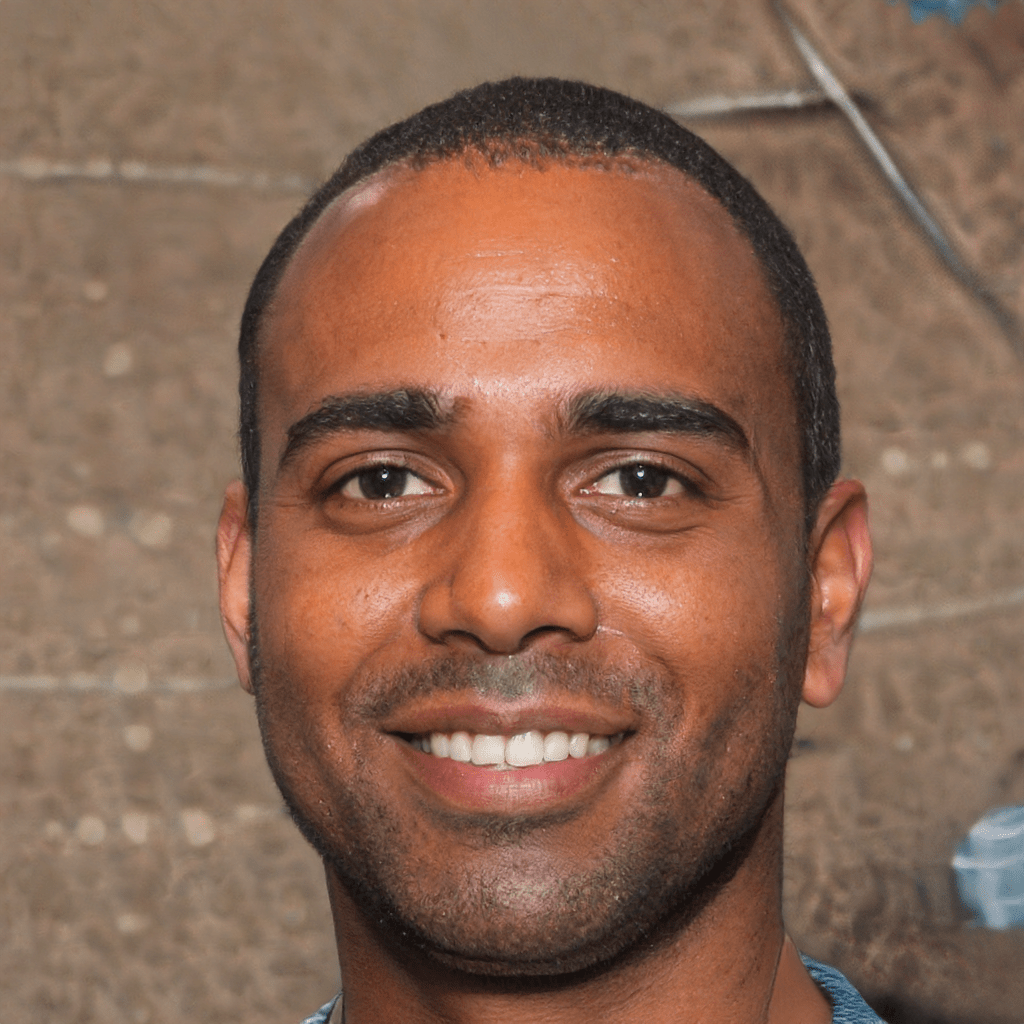} & \includegraphics[height=\hhfl,width=\wwfl, trim=0 0 0 0,clip]{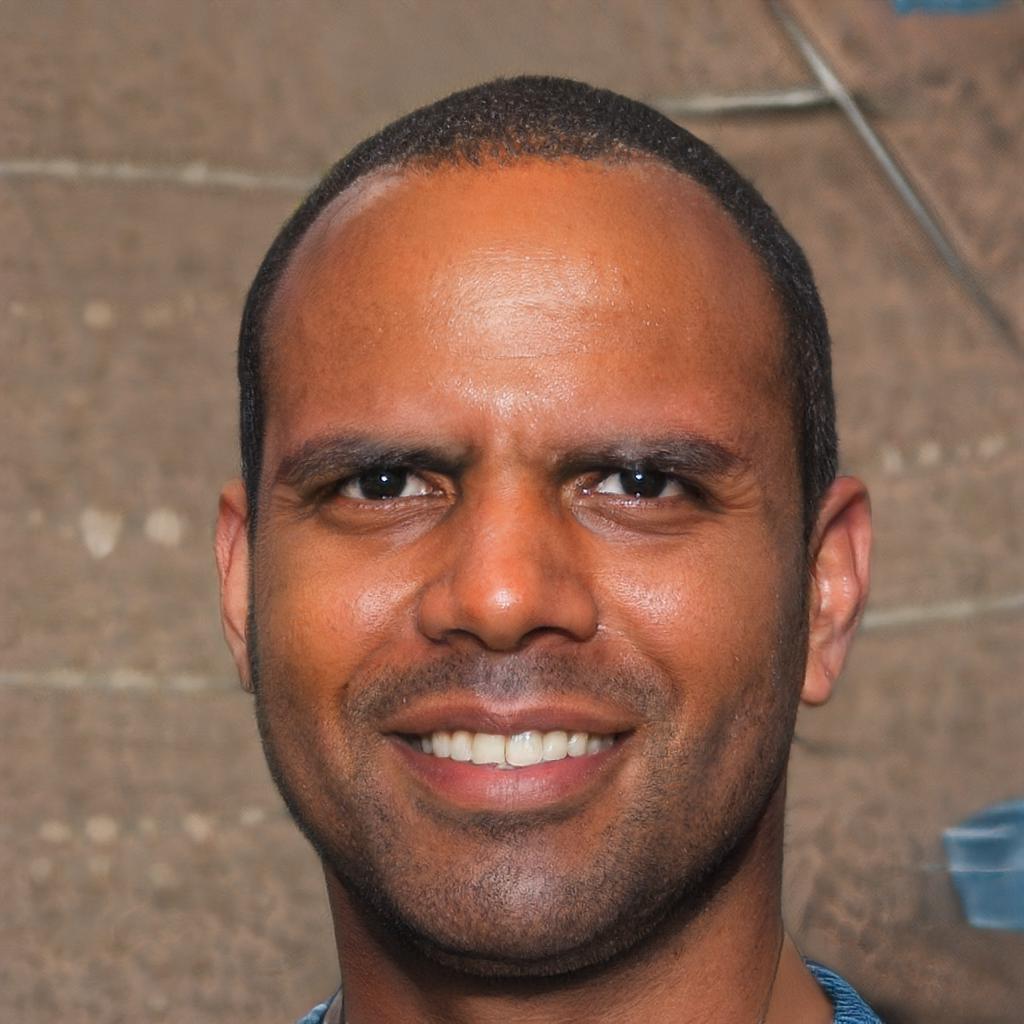}  \\
 \includegraphics[height=\hhfl,width=\wwfl, trim=0 0 0 0,clip]{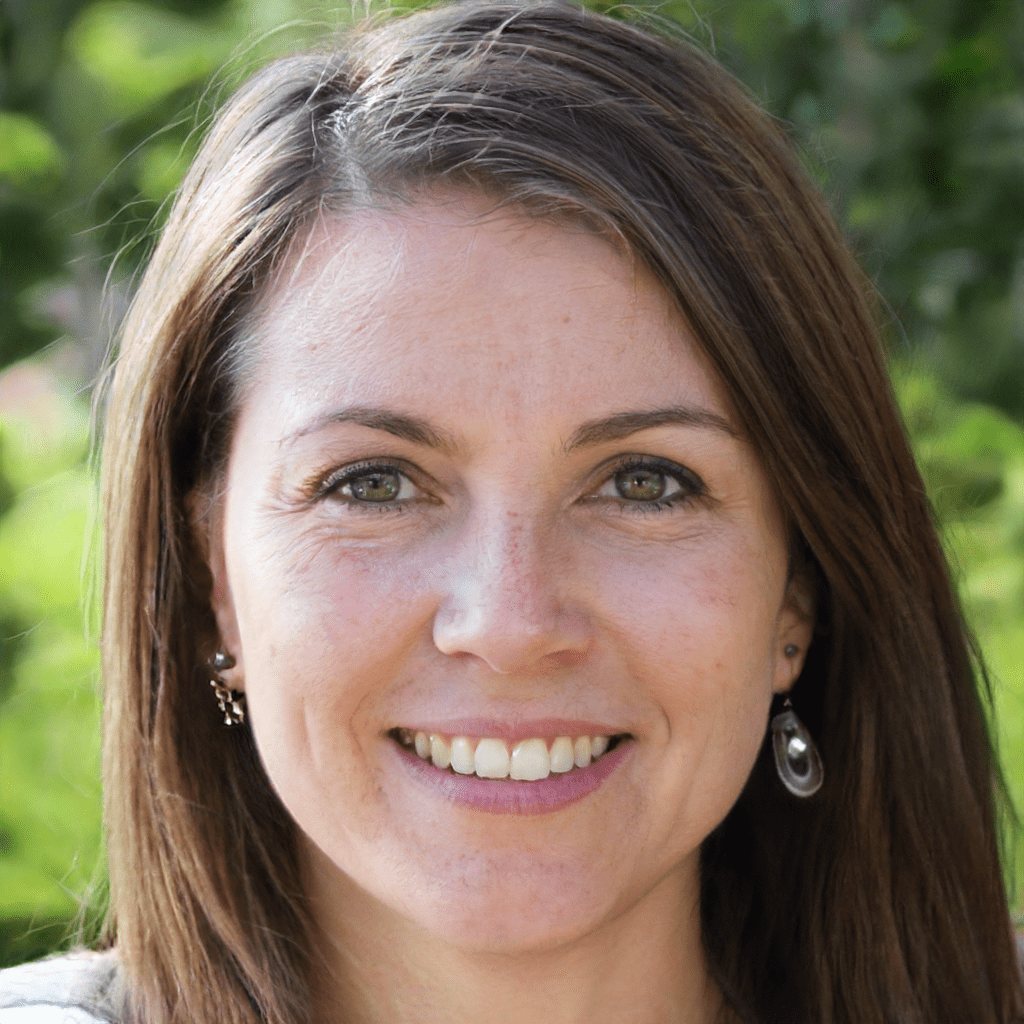} & \includegraphics[height=\hhfl,width=\wwfl, trim=0 0 0 0,clip]{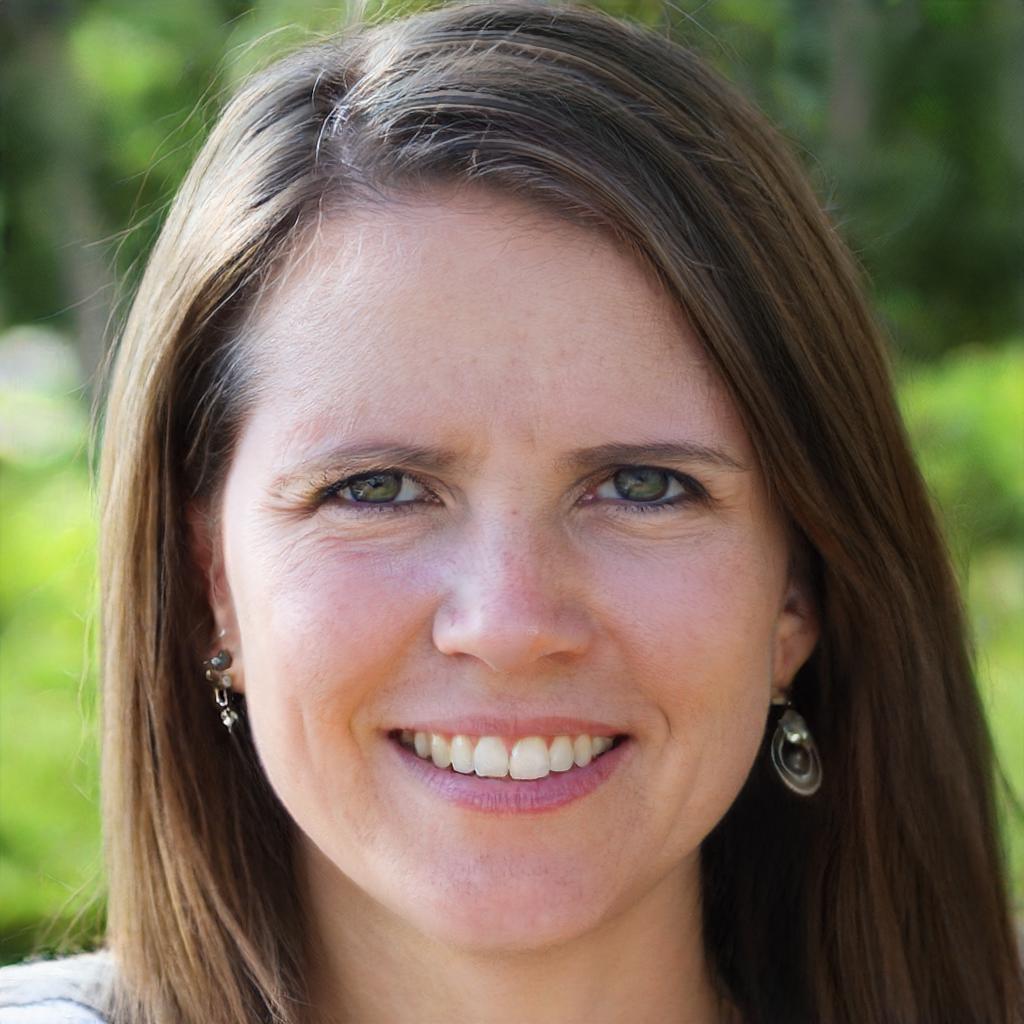}  \\
\end{tabular}

\caption{\footnotesize  {\bf Raising eyebrows editing.} \textit{First row}: Image and mask pair to learn editing vector. Images are images before editing and after editing. Segmentation masks are before editing and target segmentation mask after manual modification. \textit{Second and third rows}: Applying the learnt edit on new images (with flipped direction, i.e. negative editing vector scale $s_\textrm{edit}$.).}
\label{fig:face_eyebrow}
\vspace{-3mm}
\end{figure}

\begin{figure}
\addtolength{\tabcolsep}{-10pt}
\hspace{1mm}
\begin{tabular}{cccc}
 \includegraphics[height=\hhf,width=\wwf, trim=0 0 0 0,clip]{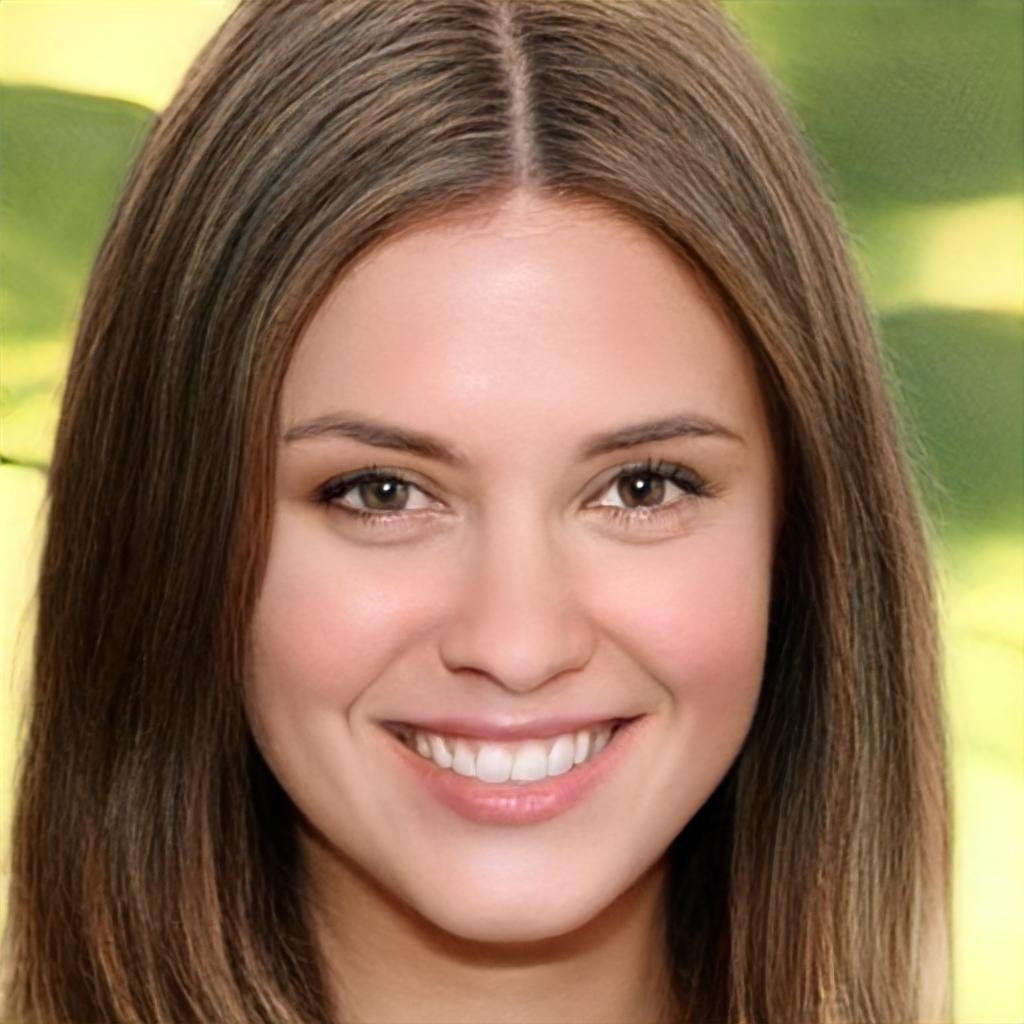} & \includegraphics[height=\hhf,width=\wwf, trim=0 0 0 0,clip]{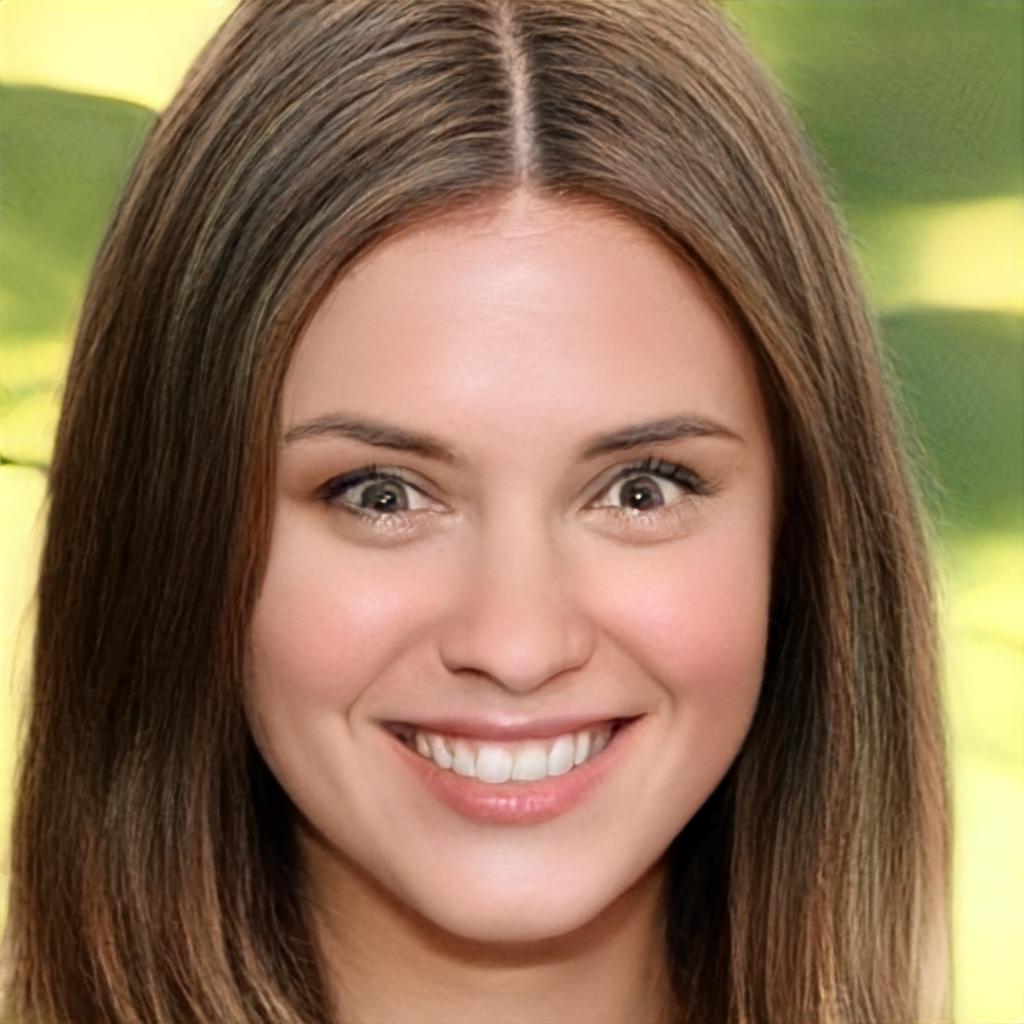}  &
 \includegraphics[height=\hhf,width=\wwf, trim=0 0 0 0,clip]{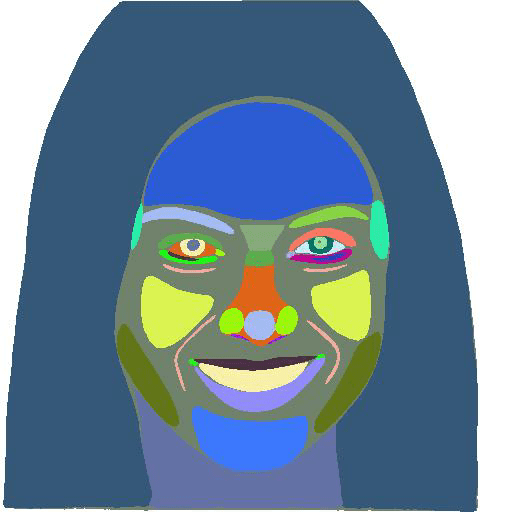} & \includegraphics[height=\hhf,width=\wwf, trim=0 0 0 0,clip]{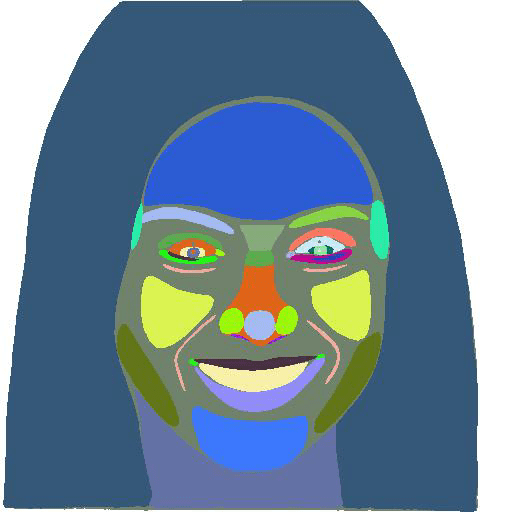}  \\
\end{tabular}

\begin{tabular}{cc}
 \includegraphics[height=\hhfl,width=\wwfl, trim=0 0 0 0,clip]{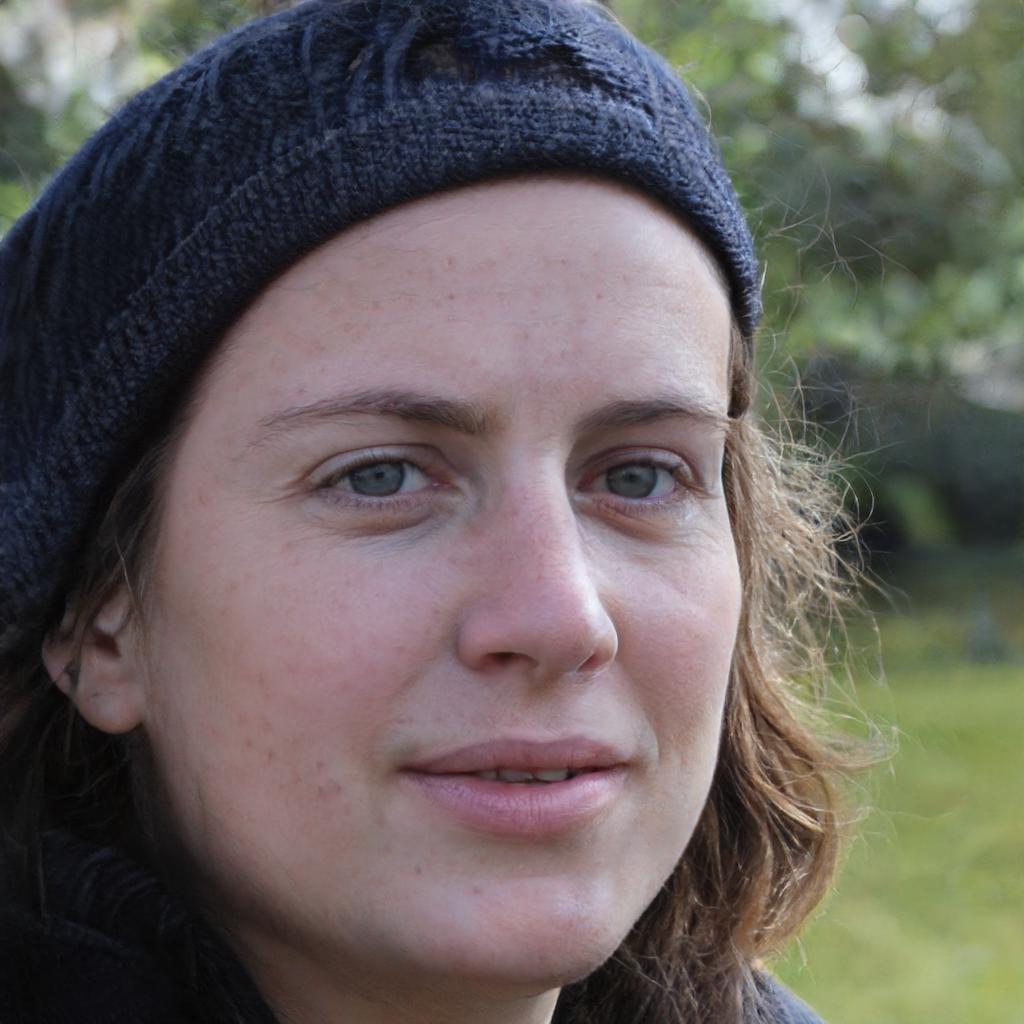} & \includegraphics[height=\hhfl,width=\wwfl, trim=0 0 0 0,clip]{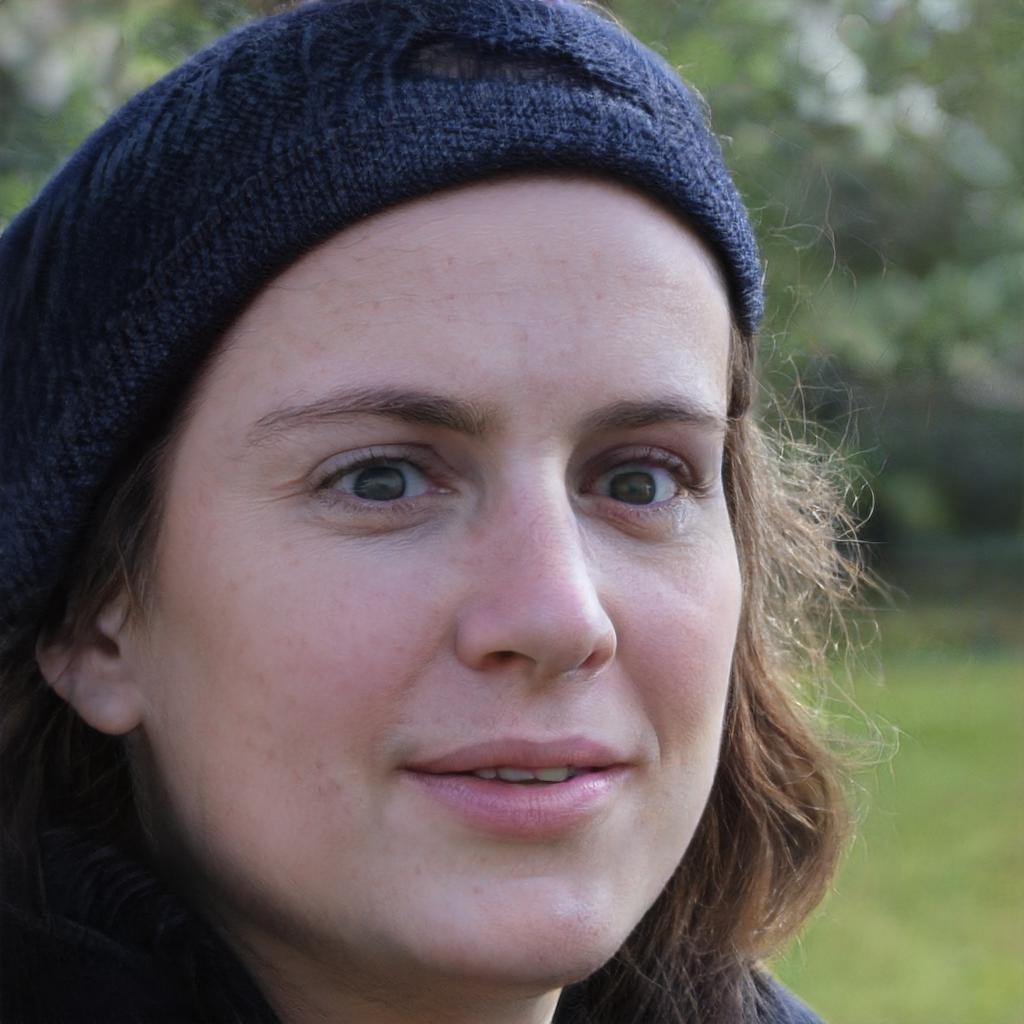}  \\
 \includegraphics[height=\hhfl,width=\wwfl, trim=0 0 0 0,clip]{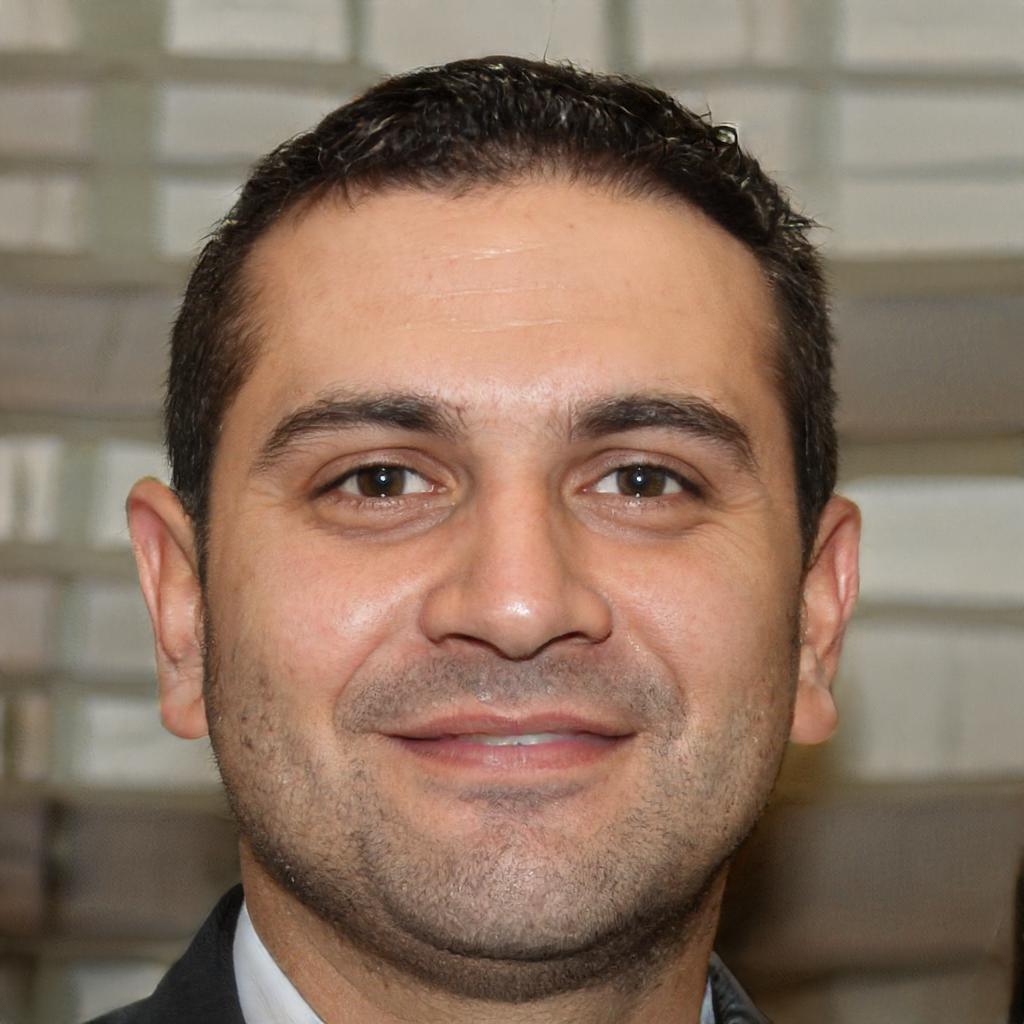} & \includegraphics[height=\hhfl,width=\wwfl, trim=0 0 0 0,clip]{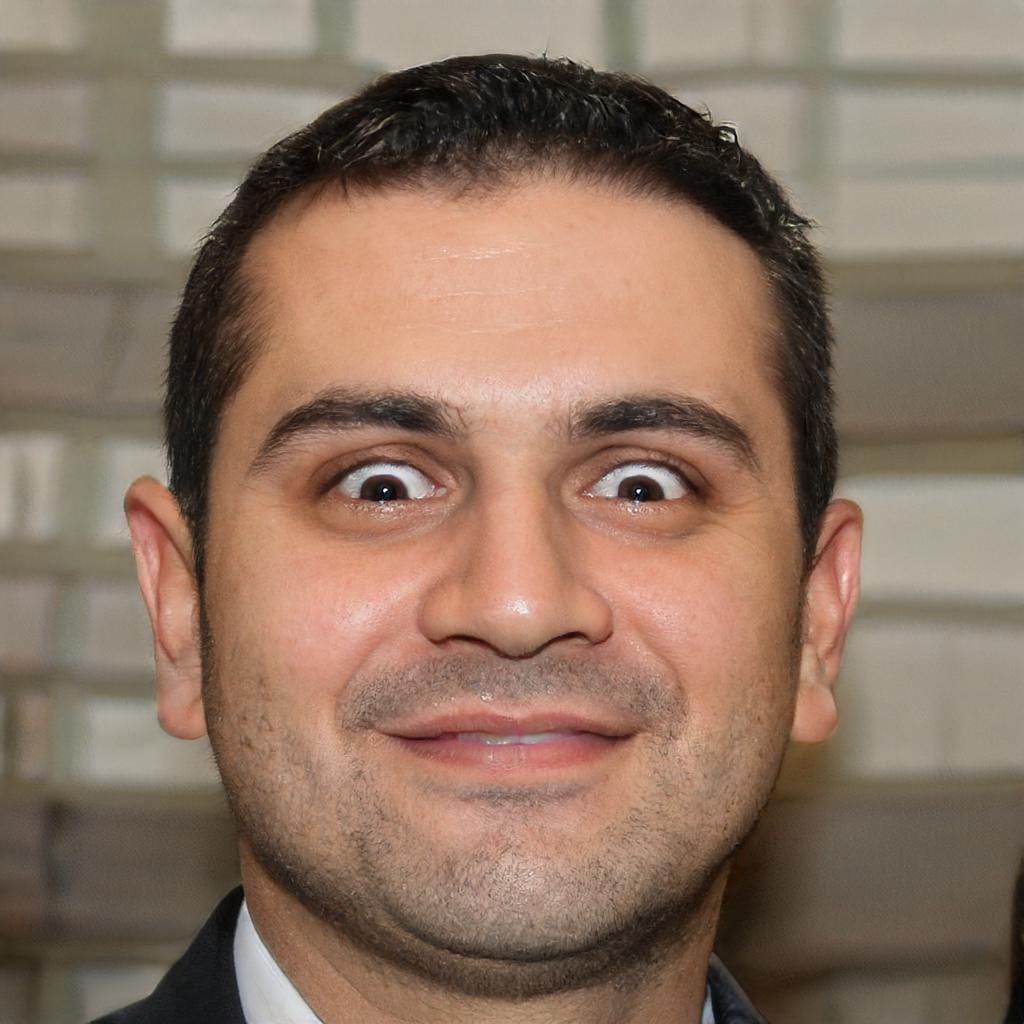}  \\
\end{tabular}

\caption{\footnotesize  {\bf Vertical gaze position editing.} \textit{First row}: Image and mask pair to learn editing vector. Images are images before editing and after editing. Segmentation masks are before editing and target segmentation mask after manual modification. \textit{Second and third rows}: Applying the learnt edit on new images.}
\label{fig:face_gaze2}
\vspace{-3mm}
\end{figure}

\begin{figure}
\addtolength{\tabcolsep}{-10pt}
\hspace{1mm}
\begin{tabular}{cccc}
 \includegraphics[height=\hhf,width=\wwf, trim=0 0 0 0,clip]{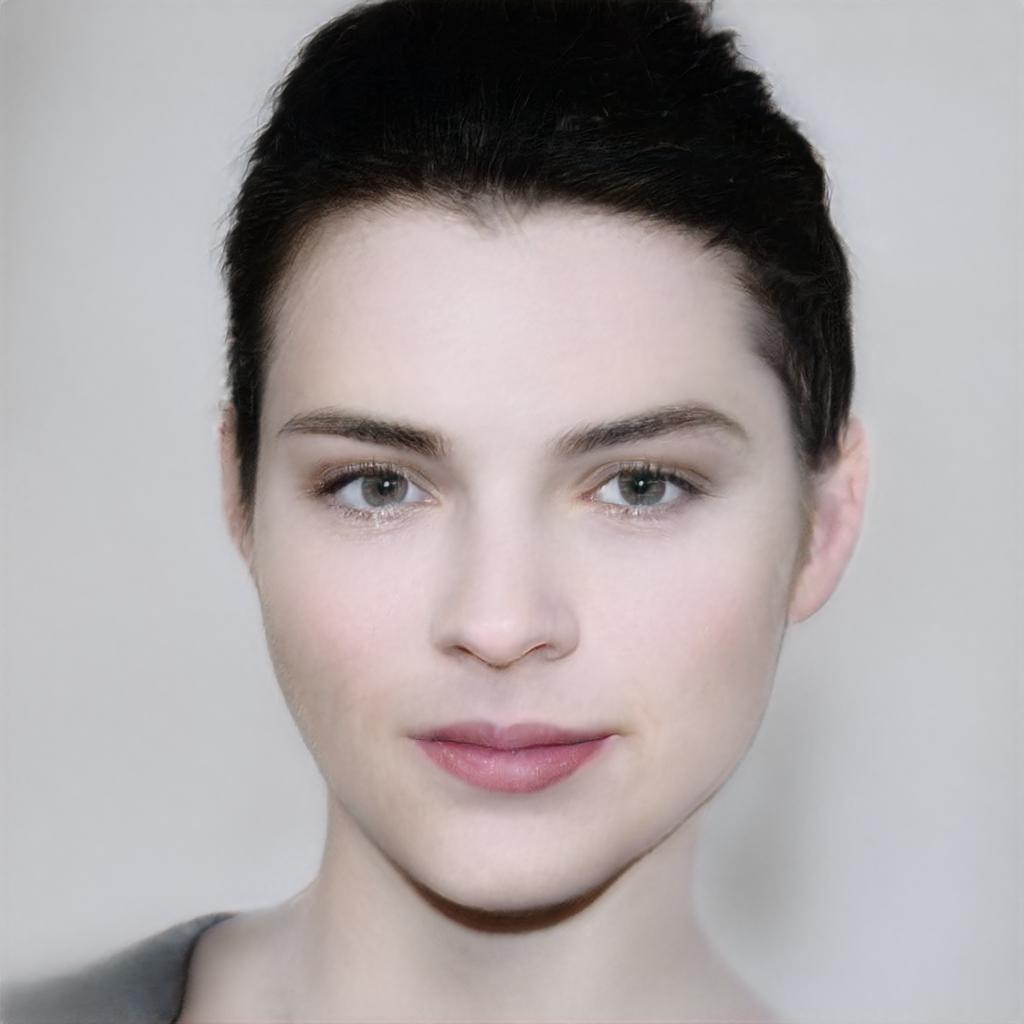} & \includegraphics[height=\hhf,width=\wwf, trim=0 0 0 0,clip]{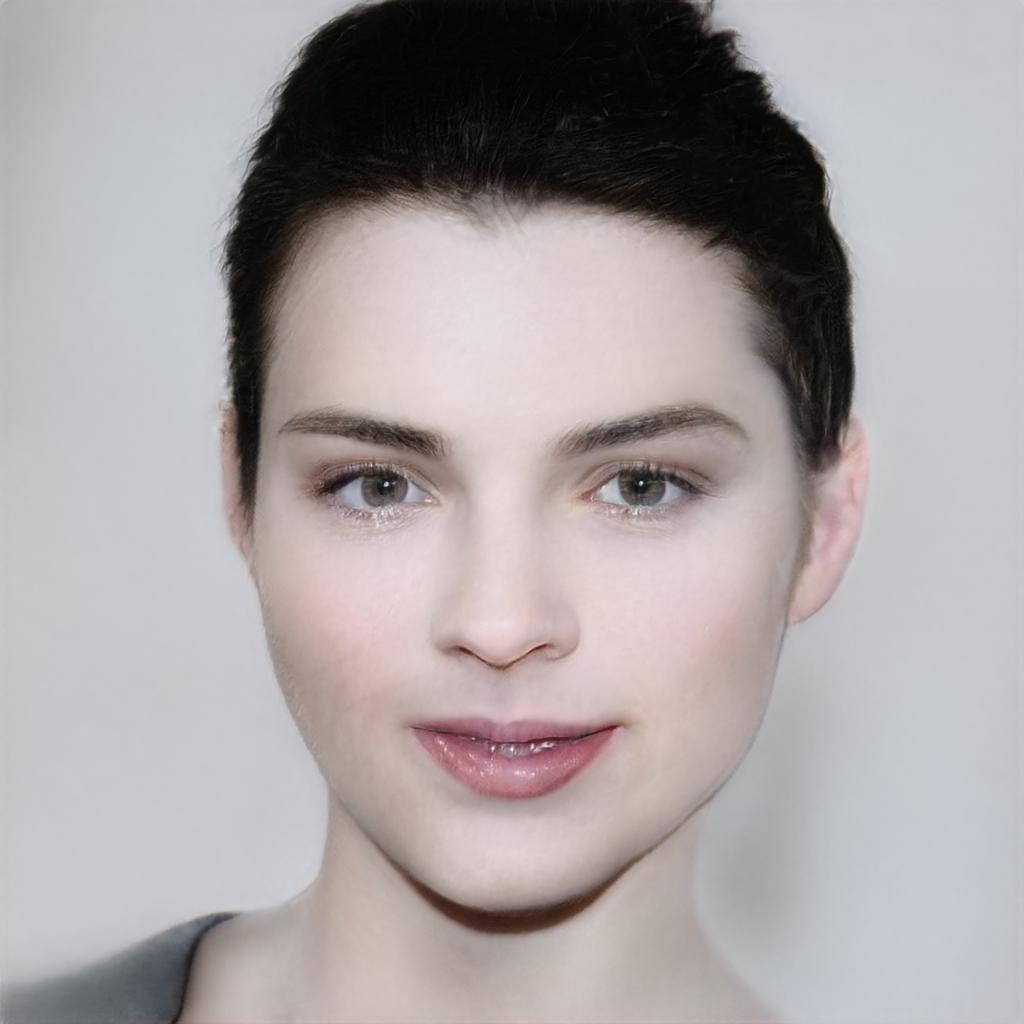}  &
 \includegraphics[height=\hhf,width=\wwf, trim=0 0 0 0,clip]{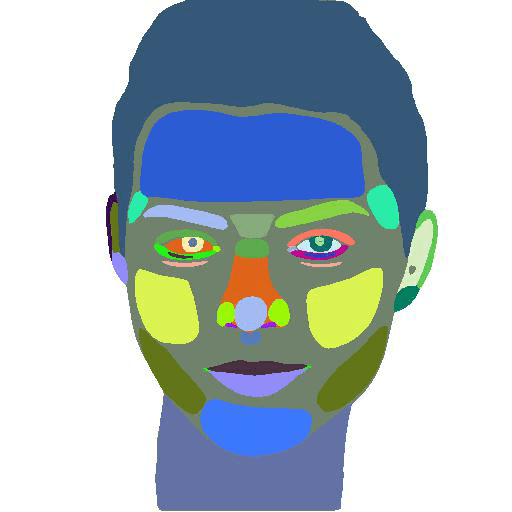} & \includegraphics[height=\hhf,width=\wwf, trim=0 0 0 0,clip]{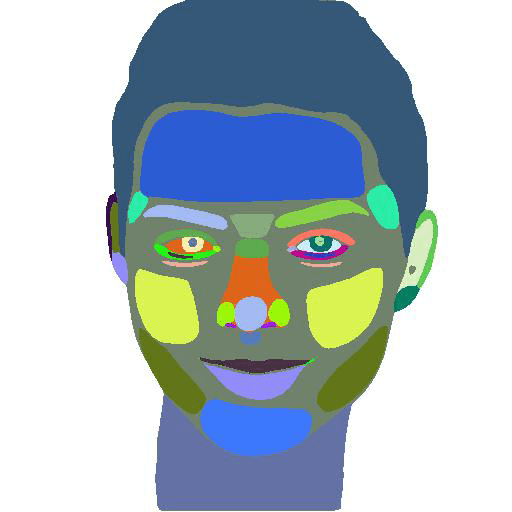}  \\
\end{tabular}

\begin{tabular}{cc}
 \includegraphics[height=\hhfl,width=\wwfl, trim=0 0 0 0,clip]{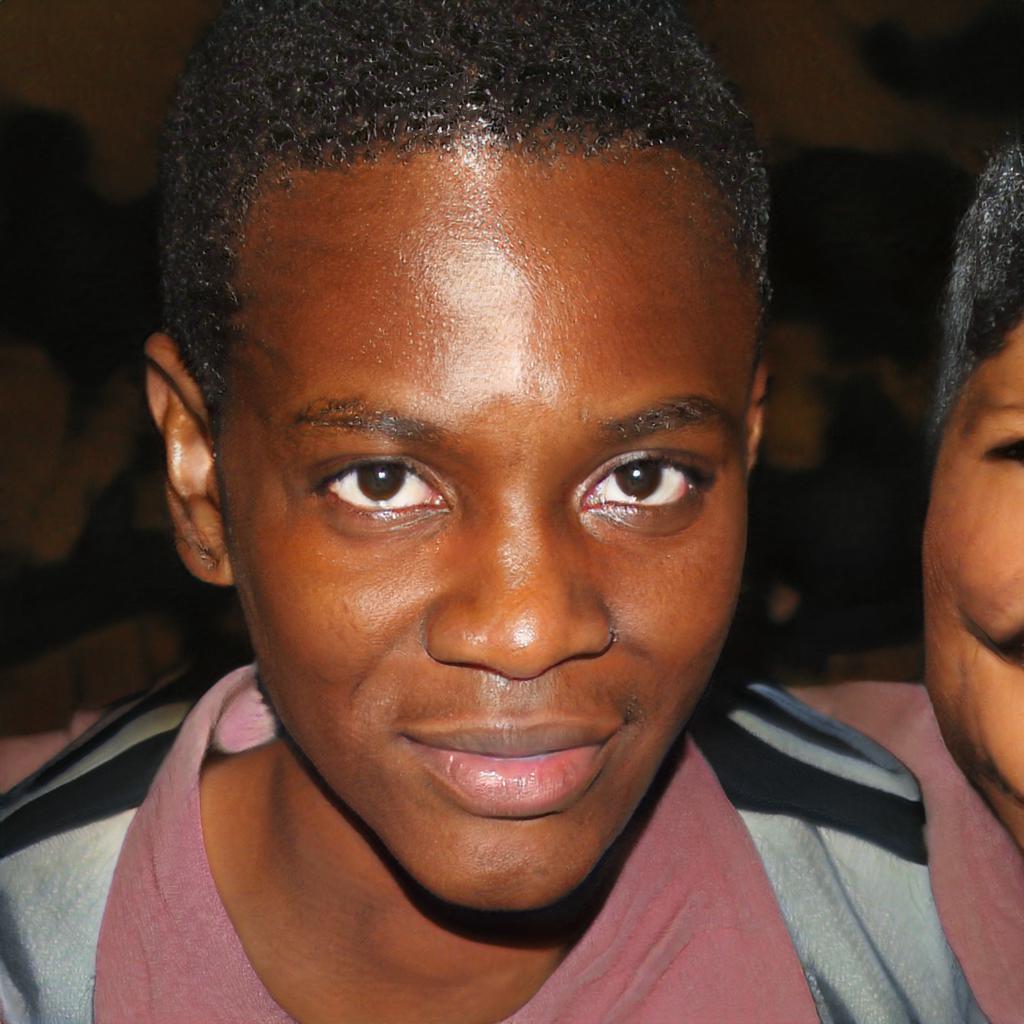} & \includegraphics[height=\hhfl,width=\wwfl, trim=0 0 0 0,clip]{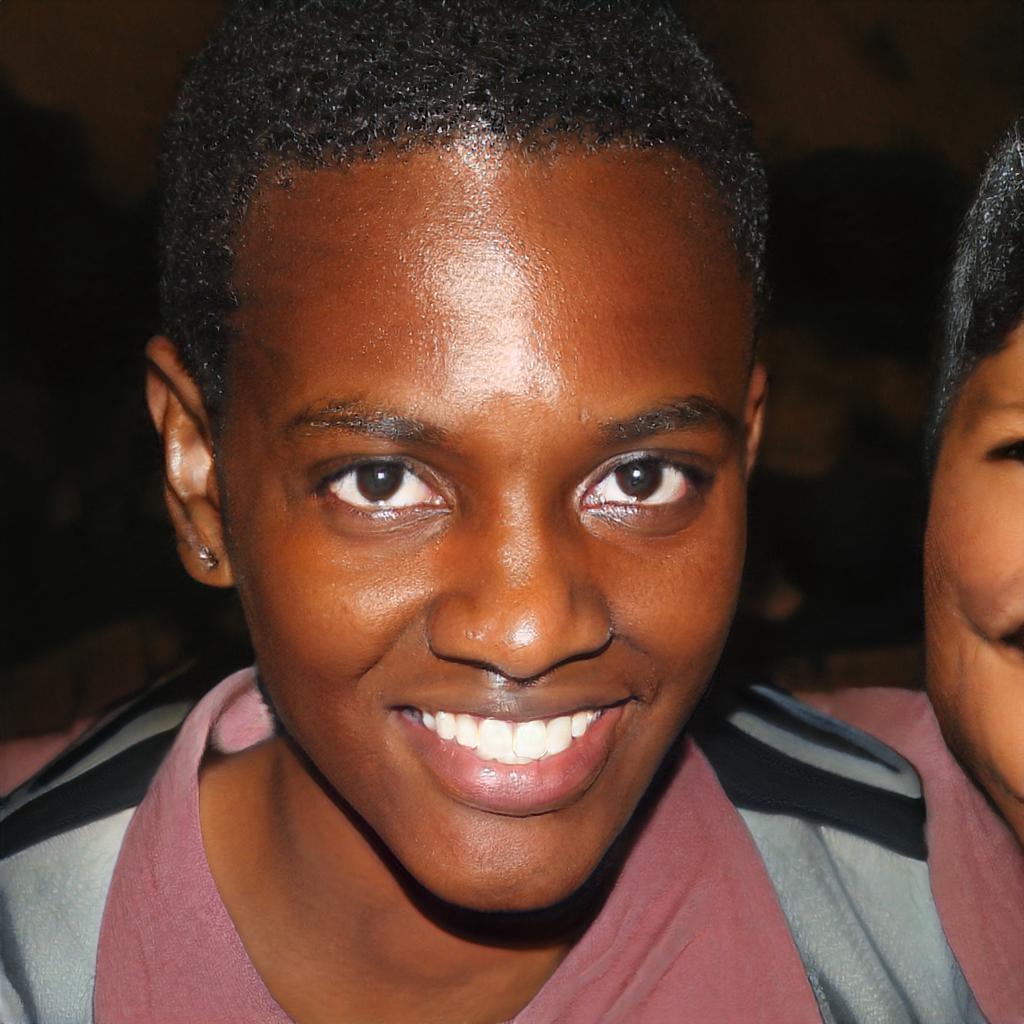}  \\
 \includegraphics[height=\hhfl,width=\wwfl, trim=0 0 0 0,clip]{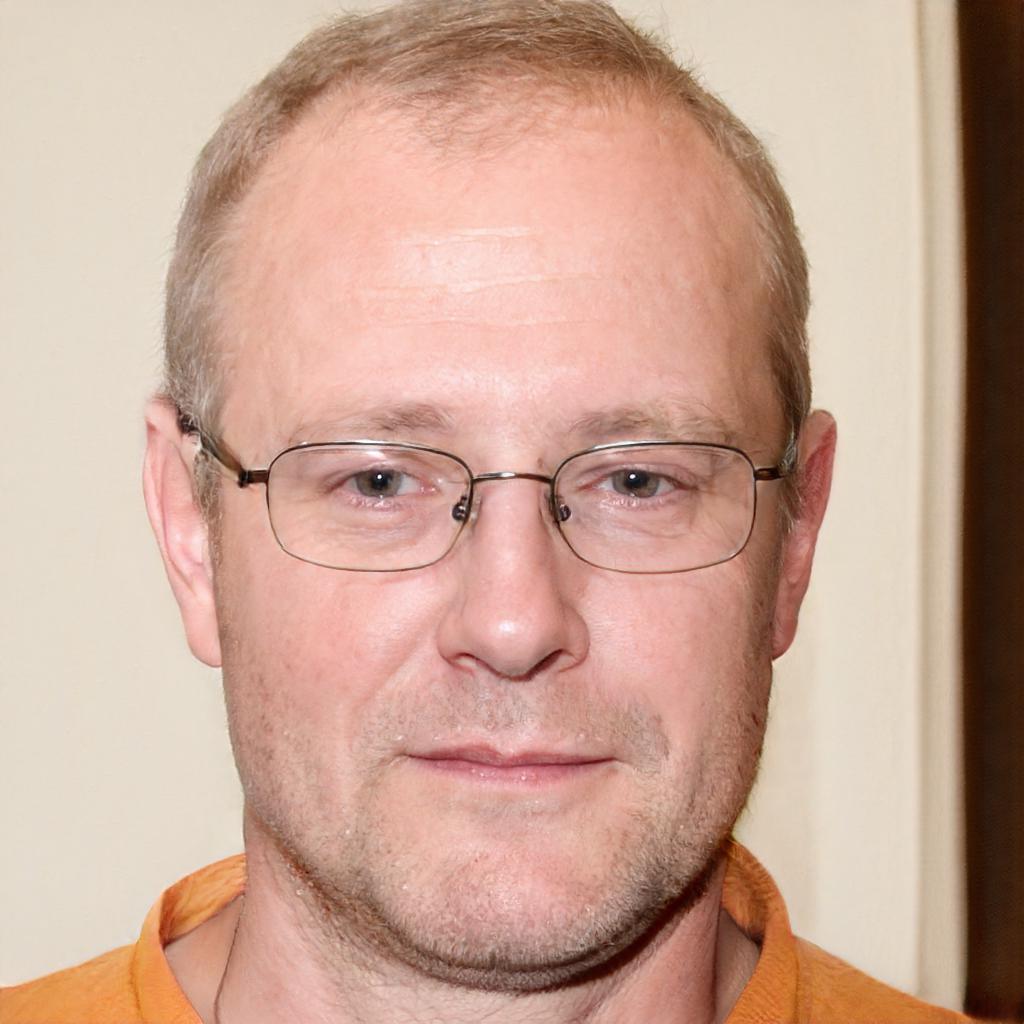} & \includegraphics[height=\hhfl,width=\wwfl, trim=0 0 0 0,clip]{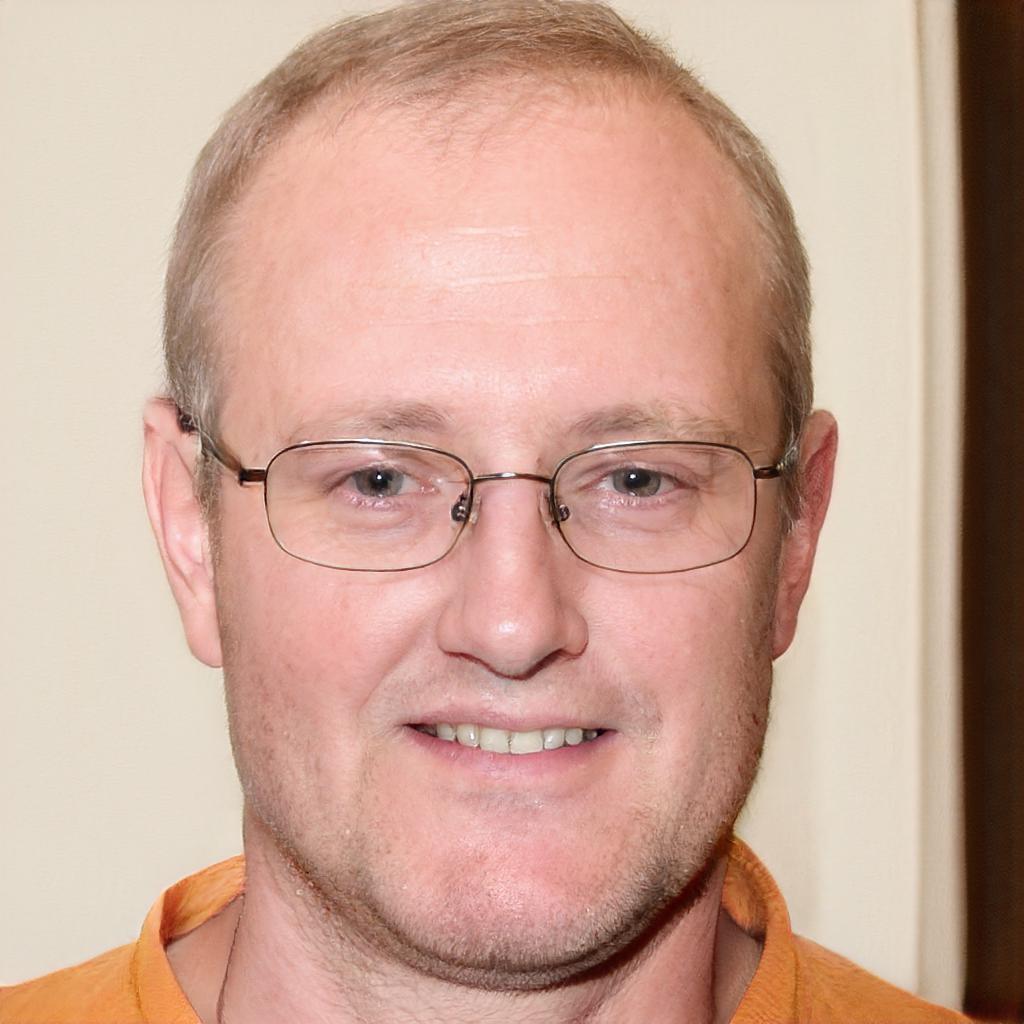}  \\
\end{tabular}

\caption{\footnotesize  {\bf Smile editing.} \textit{First row}: Image and mask pair to learn editing vector. Images are images before editing and after editing. Segmentation masks are before editing and target segmentation mask after manual modification. \textit{Second and third rows}: Applying the learnt edit on new images.}
\label{fig:face_smile}
\vspace{-3mm}
\end{figure}

\begin{figure}
\addtolength{\tabcolsep}{-10pt}
\hspace{1mm}
\begin{tabular}{cccc}
 \includegraphics[height=\hhf,width=\wwf, trim=0 0 0 0,clip]{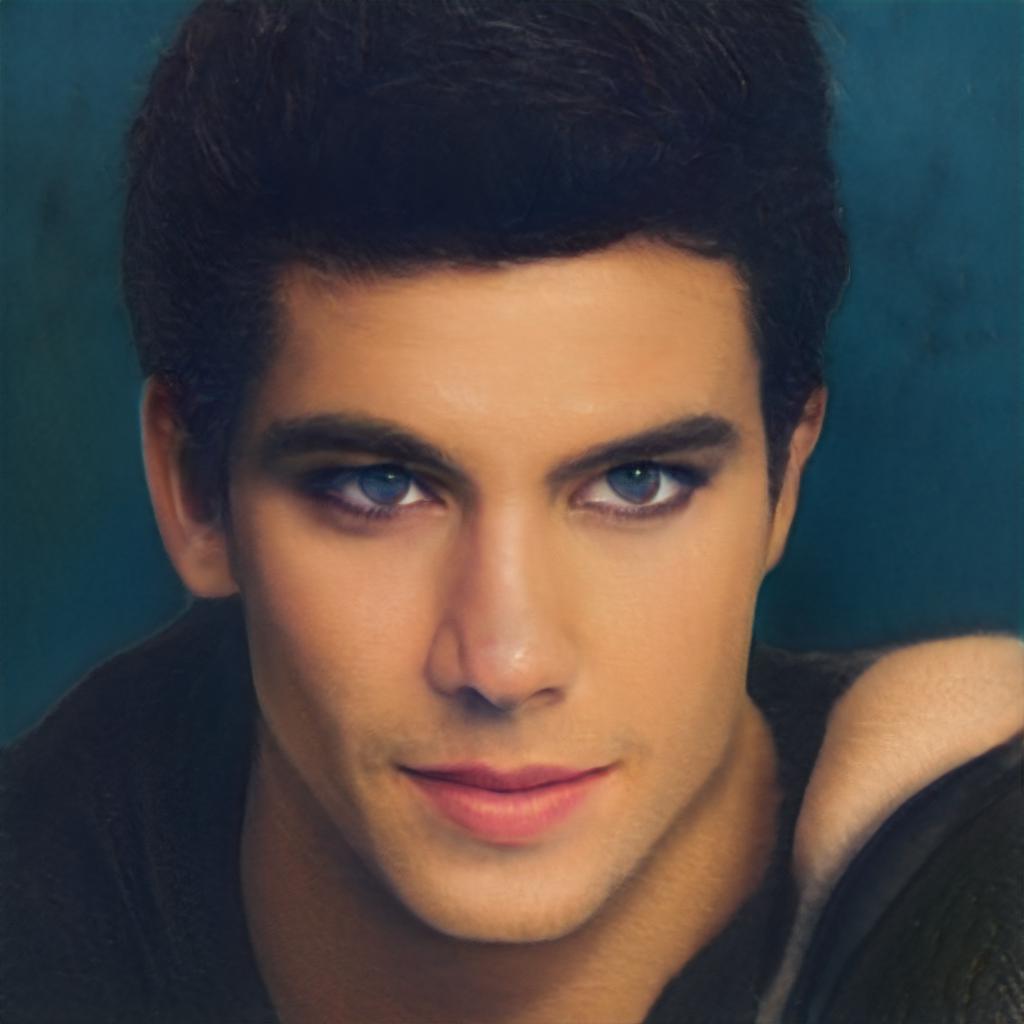} & \includegraphics[height=\hhf,width=\wwf, trim=0 0 0 0,clip]{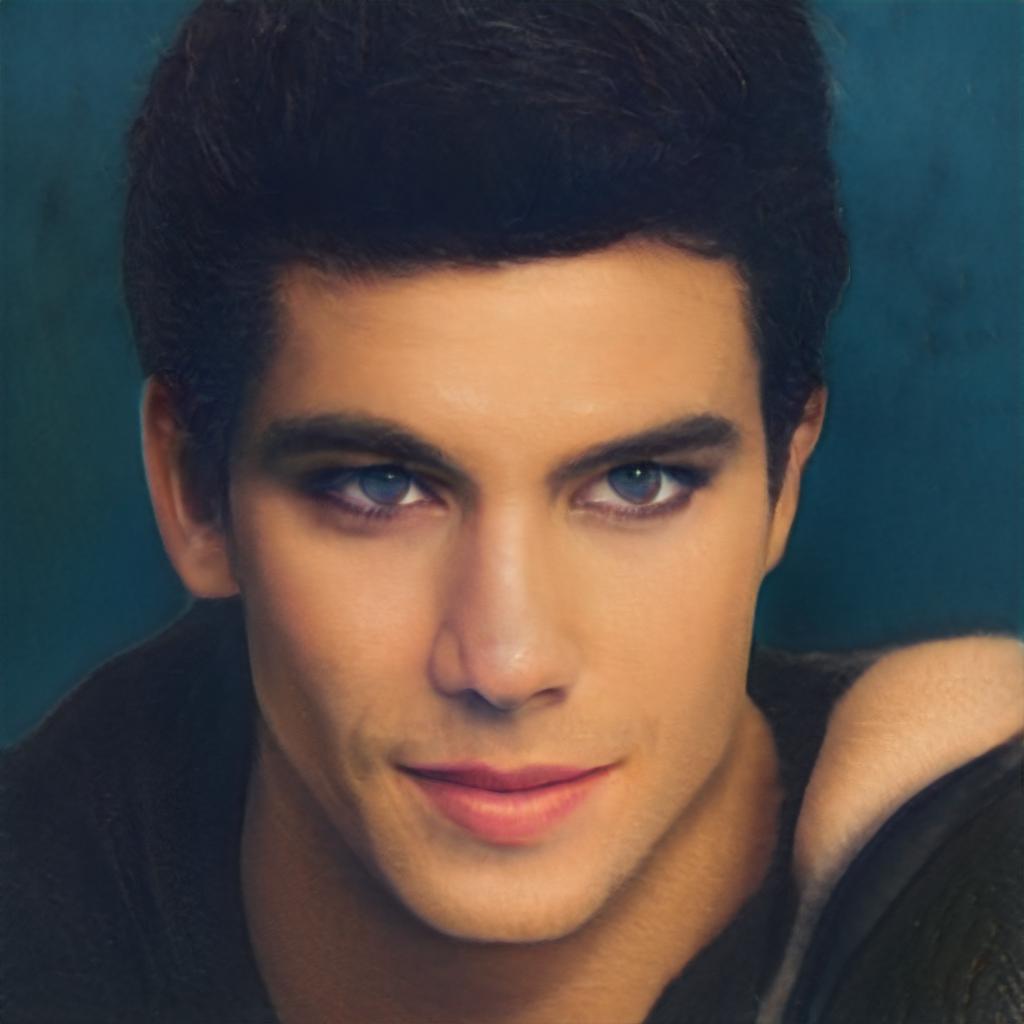}  &
 \includegraphics[height=\hhf,width=\wwf, trim=0 0 0 0,clip]{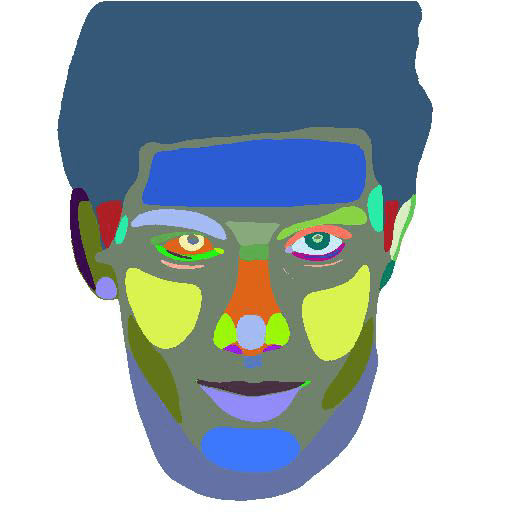} & \includegraphics[height=\hhf,width=\wwf, trim=0 0 0 0,clip]{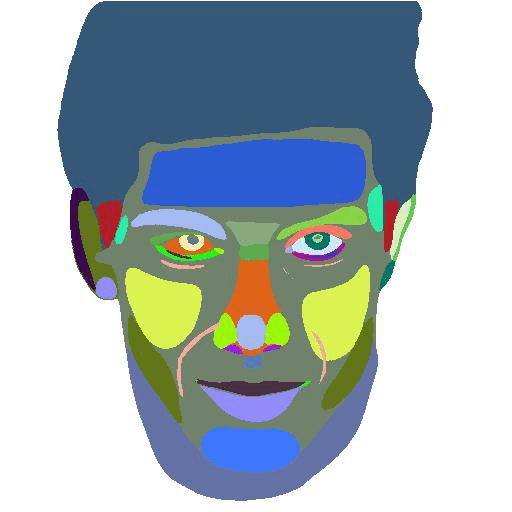}  \\
\end{tabular}

\begin{tabular}{cc}
 \includegraphics[height=\hhfl,width=\wwfl, trim=0 0 0 0,clip]{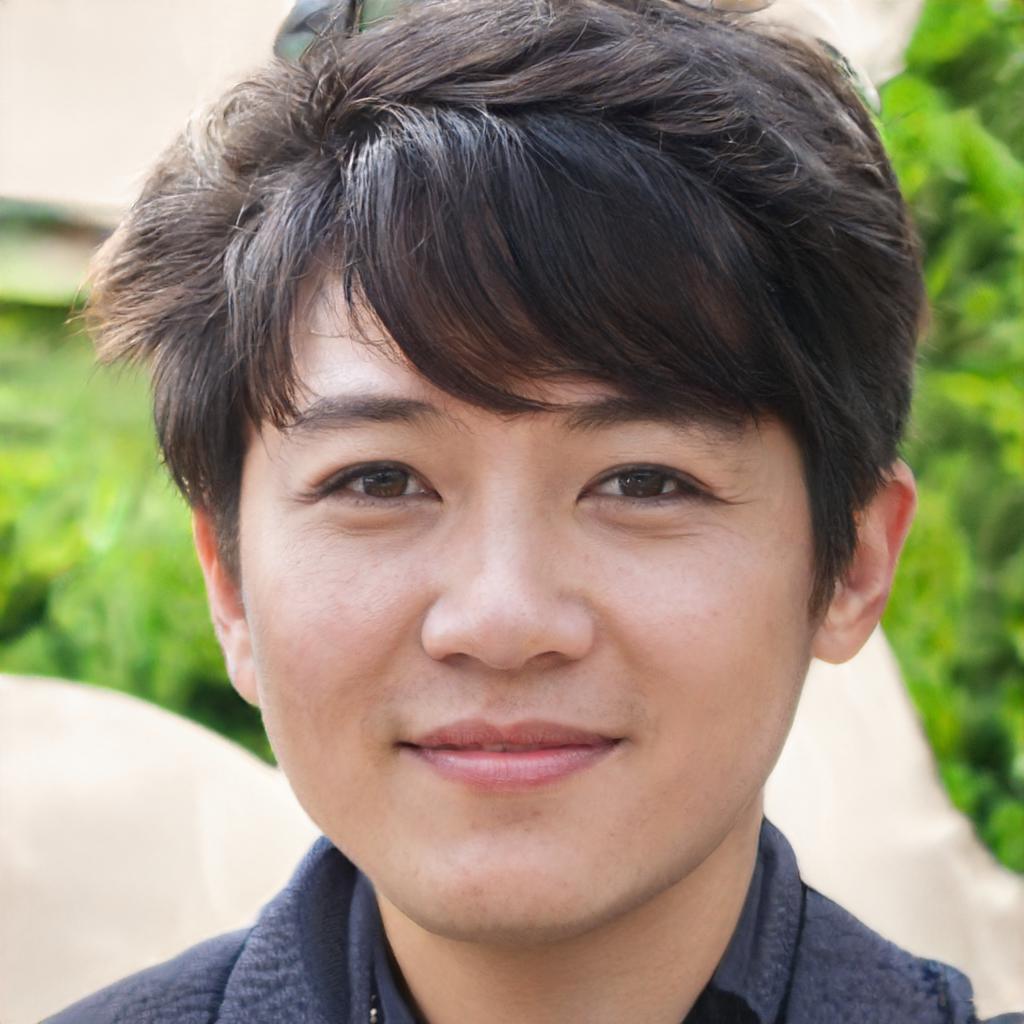} & \includegraphics[height=\hhfl,width=\wwfl, trim=0 0 0 0,clip]{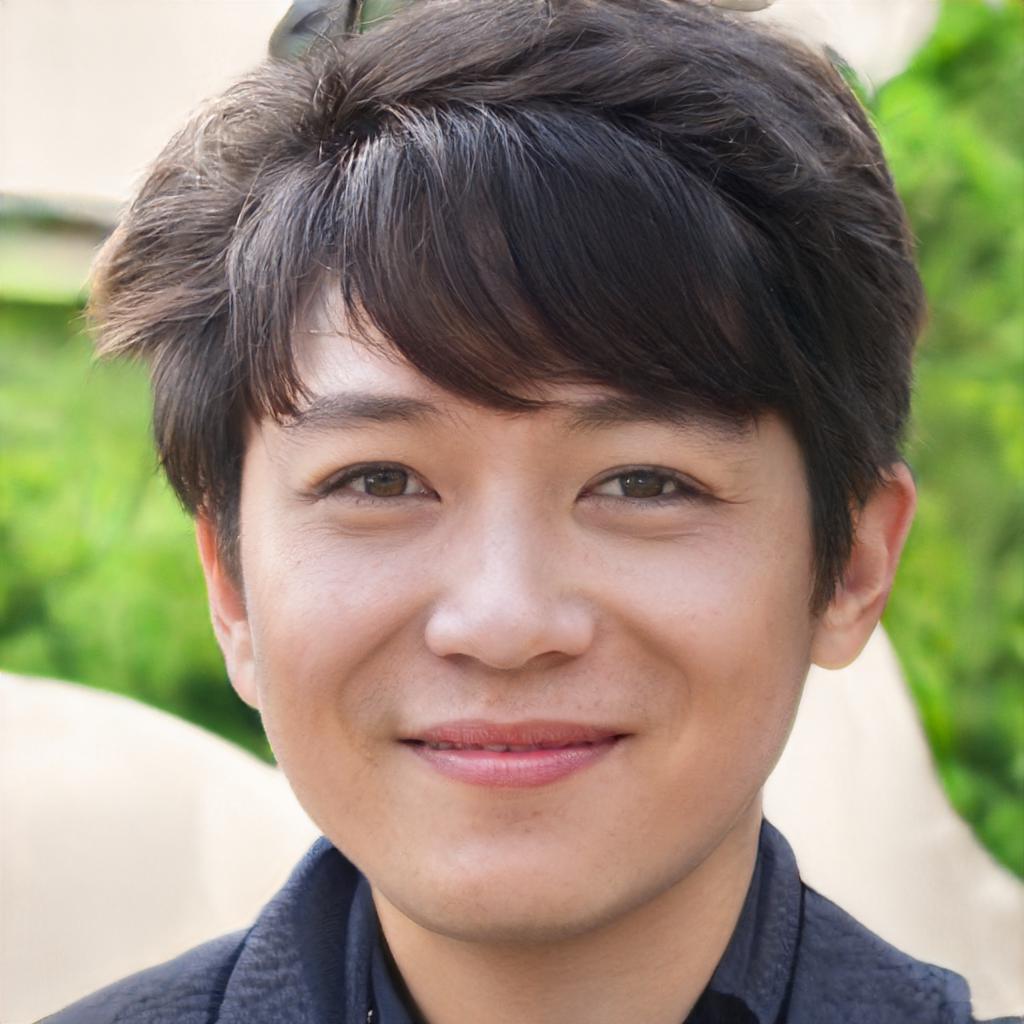}  \\
 \includegraphics[height=\hhfl,width=\wwfl, trim=0 0 0 0,clip]{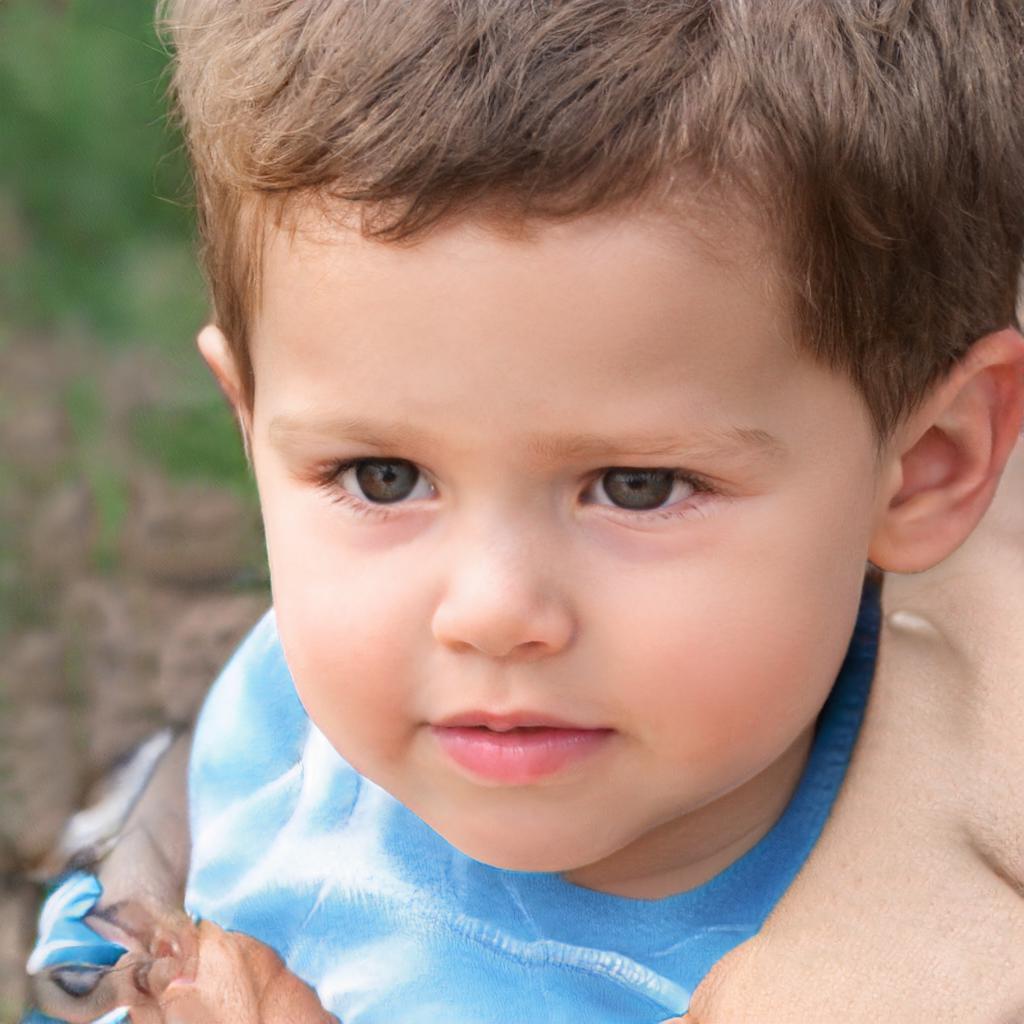} & \includegraphics[height=\hhfl,width=\wwfl, trim=0 0 0 0,clip]{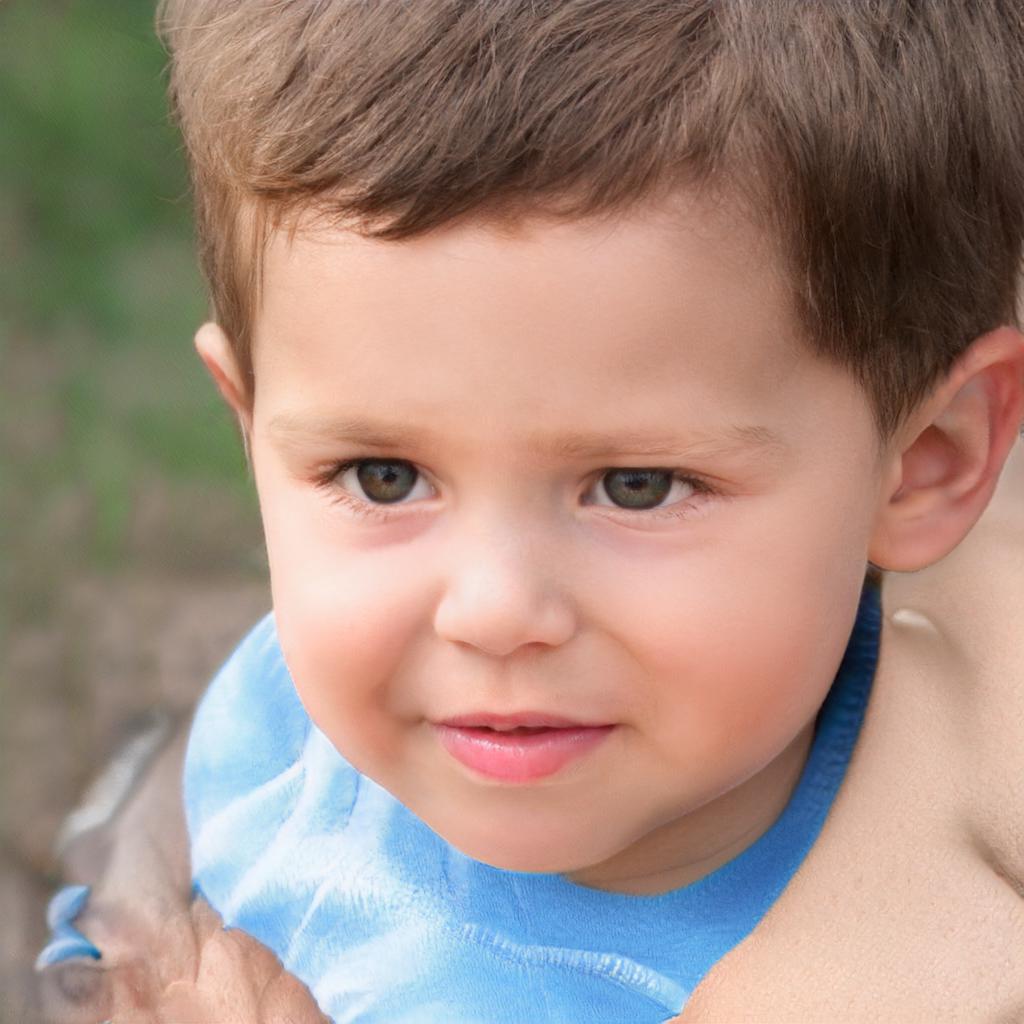}  \\
\end{tabular}

\caption{\footnotesize  {\bf Adding smile wrinkle editing.} \textit{First row}: Image and mask pair to learn editing vector. Images are images before editing and after editing. Segmentation masks are before editing and target segmentation mask after manual modification. \textit{Second and third rows}: Applying the learnt edit on new images.}
\label{fig:face_wrinkle}
\vspace{-3mm}
\end{figure}

\begin{figure}
\addtolength{\tabcolsep}{-10pt}
\hspace{1mm}
\begin{tabular}{cccc}
 \includegraphics[height=\hhf,width=\wwf, trim=0 0 0 0,clip]{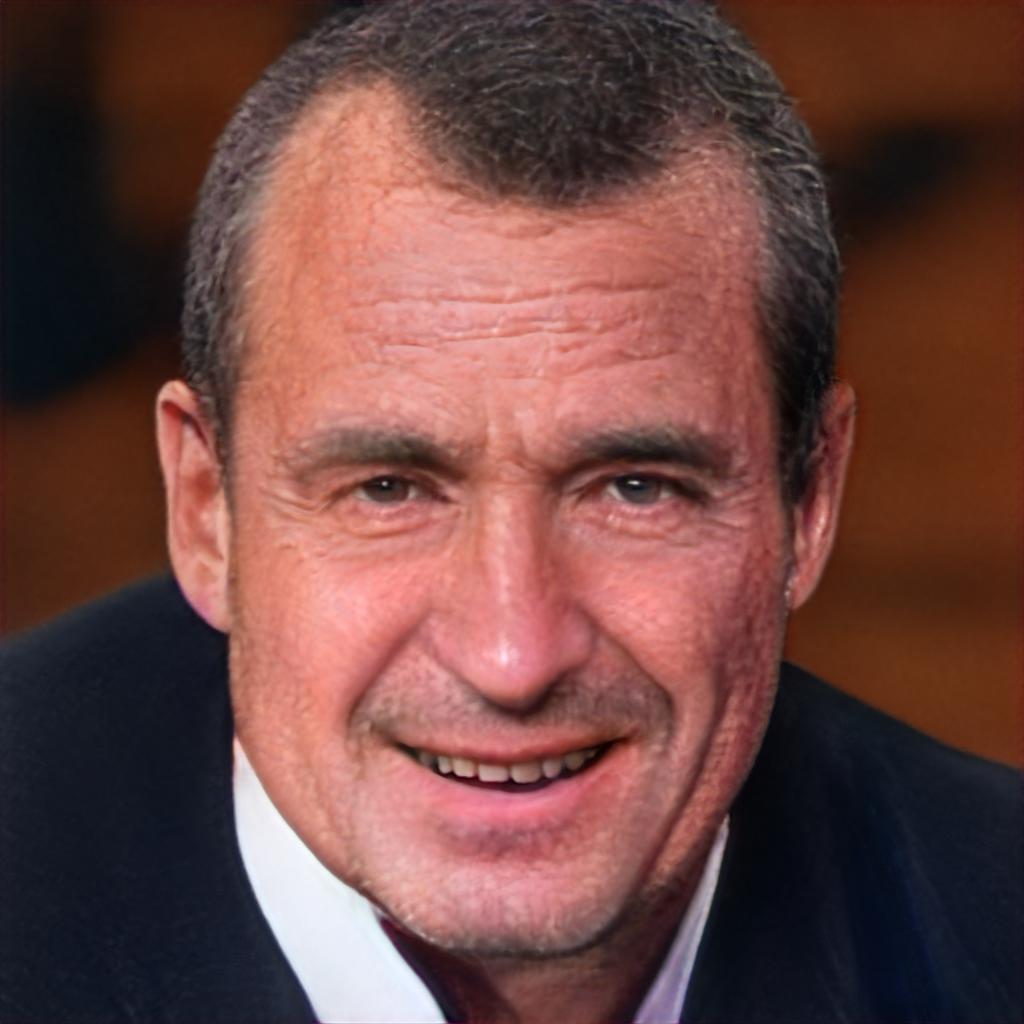} & \includegraphics[height=\hhf,width=\wwf, trim=0 0 0 0,clip]{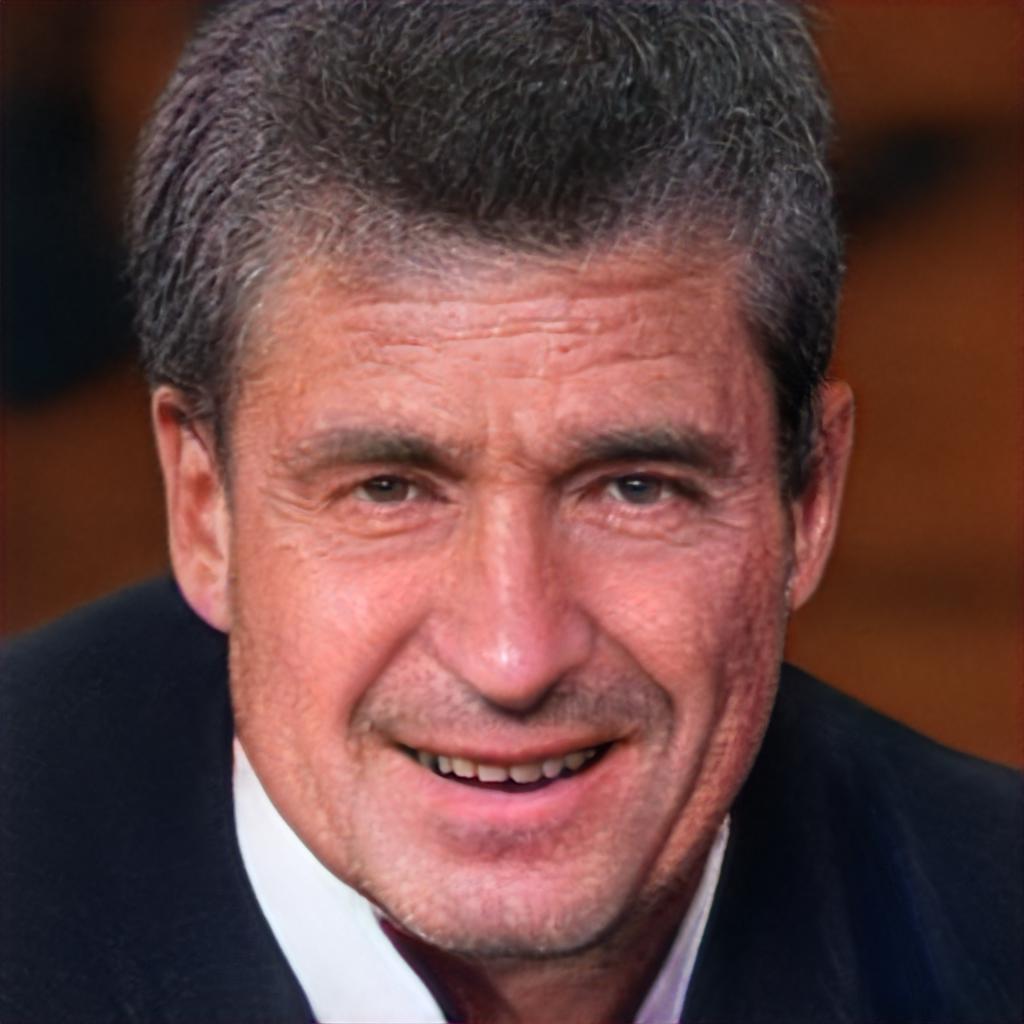}  &
 \includegraphics[height=\hhf,width=\wwf, trim=0 0 0 0,clip]{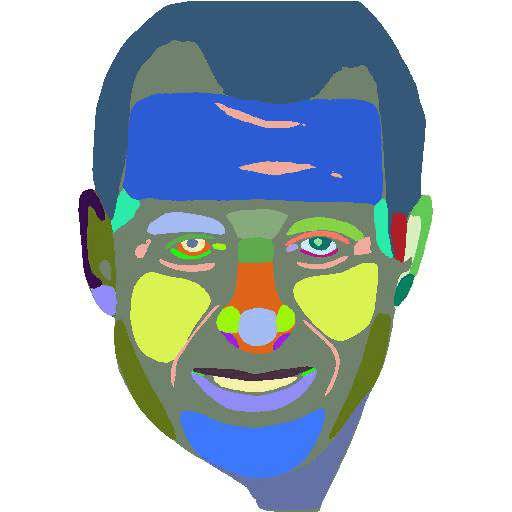} & \includegraphics[height=\hhf,width=\wwf, trim=0 0 0 0,clip]{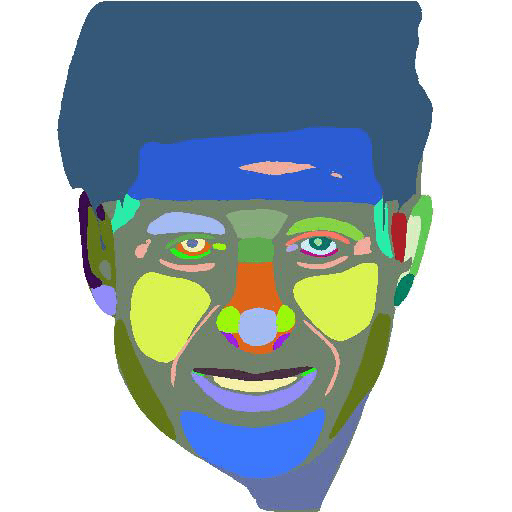}  \\
\end{tabular}

\begin{tabular}{cc}
 \includegraphics[height=\hhfl,width=\wwfl, trim=0 0 0 0,clip]{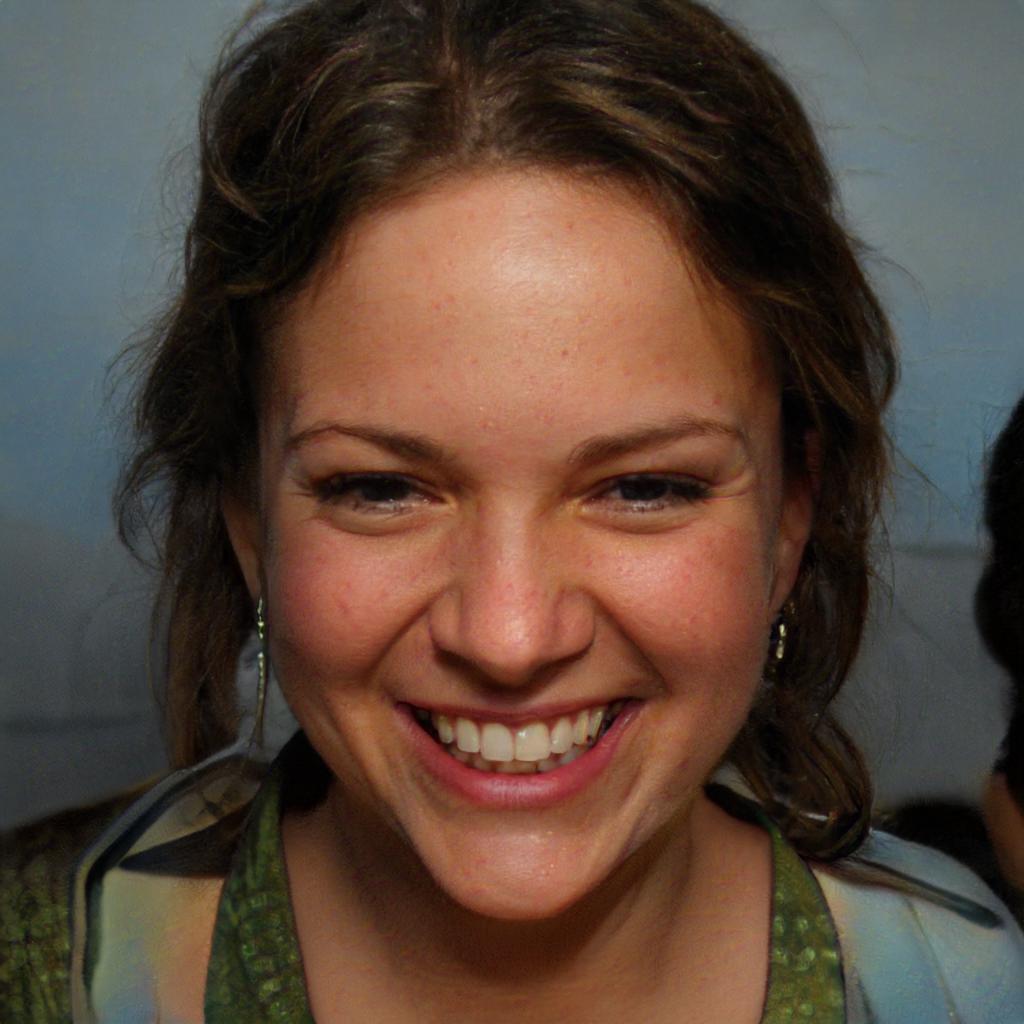} & \includegraphics[height=\hhfl,width=\wwfl, trim=0 0 0 0,clip]{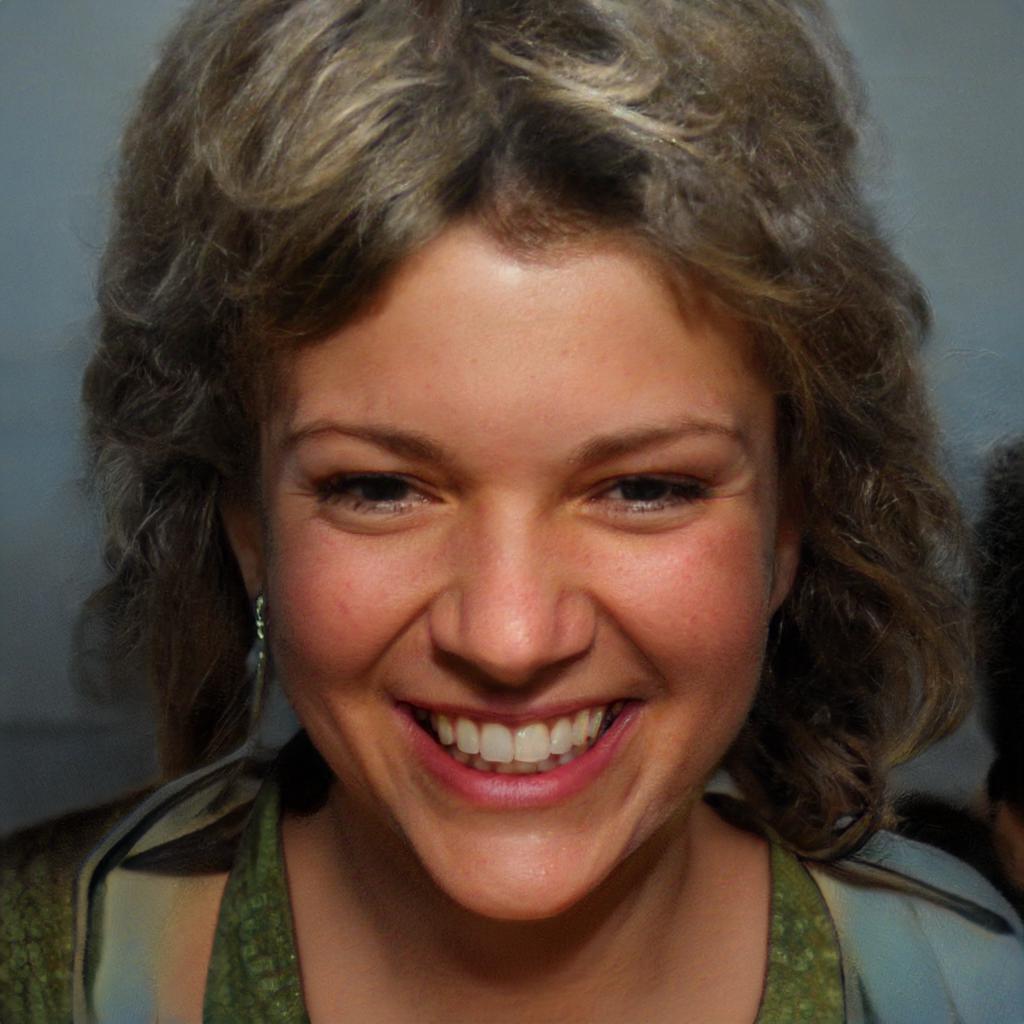}  \\
 \includegraphics[height=\hhfl,width=\wwfl, trim=0 0 0 0,clip]{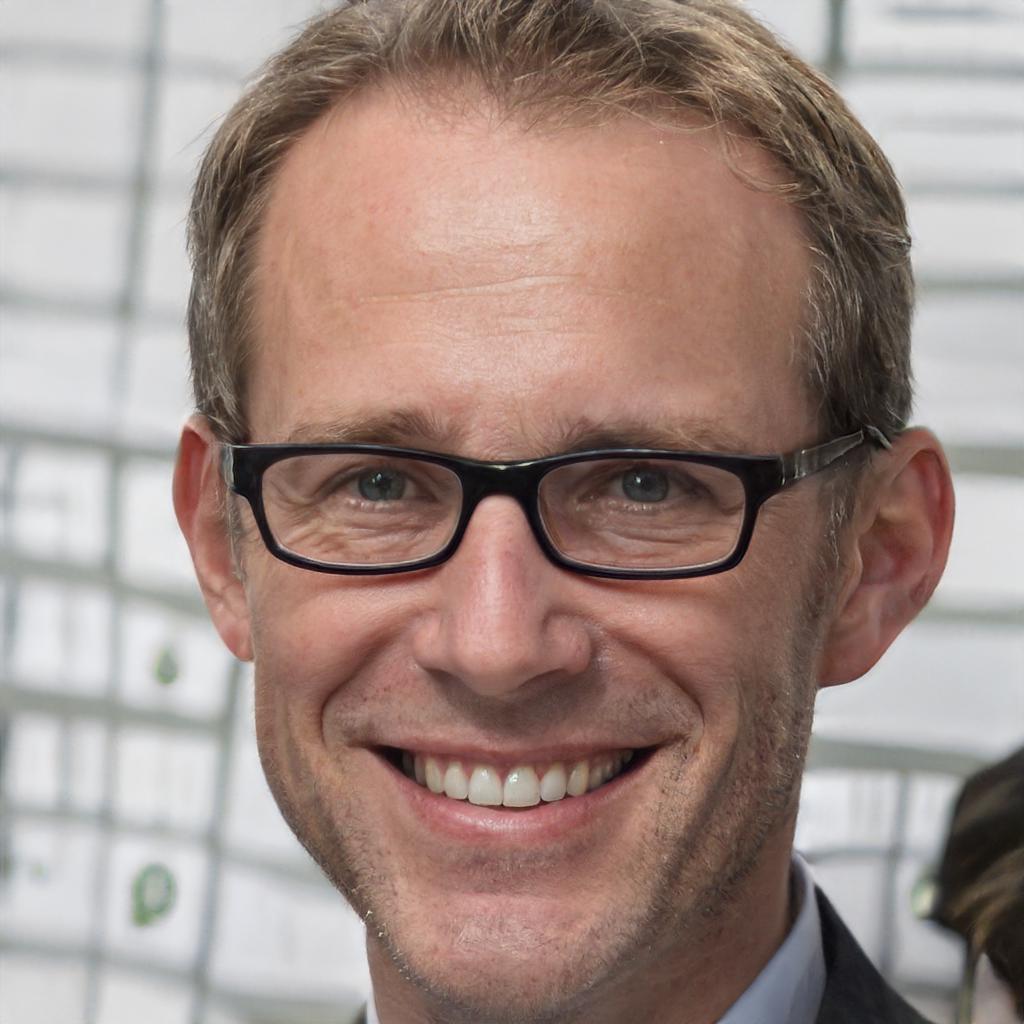} & \includegraphics[height=\hhfl,width=\wwfl, trim=0 0 0 0,clip]{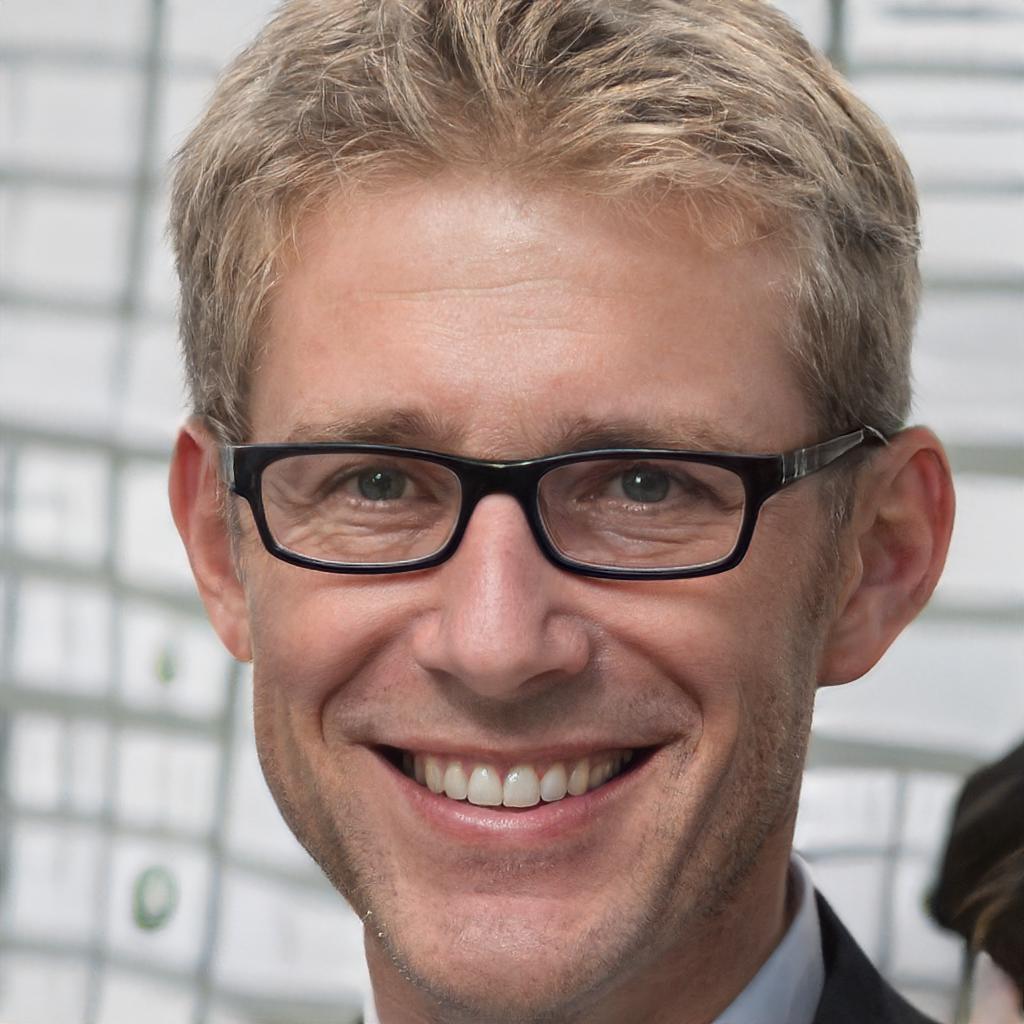}  \\
\end{tabular}

\caption{\footnotesize  {\bf Hairstyle editing.} \textit{First row}: Image and mask pair to learn editing vector. Images are images before editing and after editing. Segmentation masks are before editing and target segmentation mask after manual modification. \textit{Second and third rows}: Applying the learnt edit on new images.}
\label{fig:face_hairstyle}
\vspace{-3mm}
\end{figure}

\begin{figure}
\addtolength{\tabcolsep}{-10pt}
\hspace{1mm}
\begin{tabular}{cccc}
 \includegraphics[height=\hhf,width=\wwf, trim=0 0 0 0,clip]{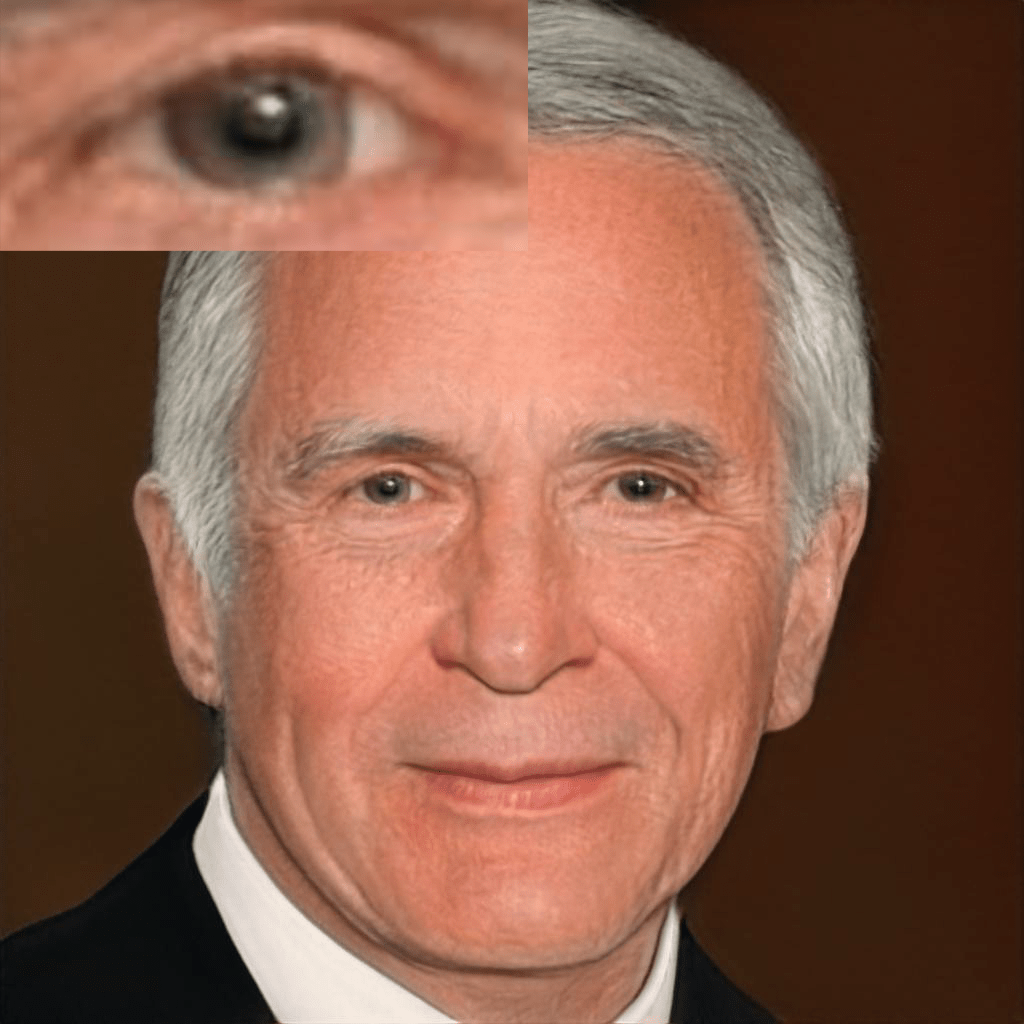} & \includegraphics[height=\hhf,width=\wwf, trim=0 0 0 0,clip]{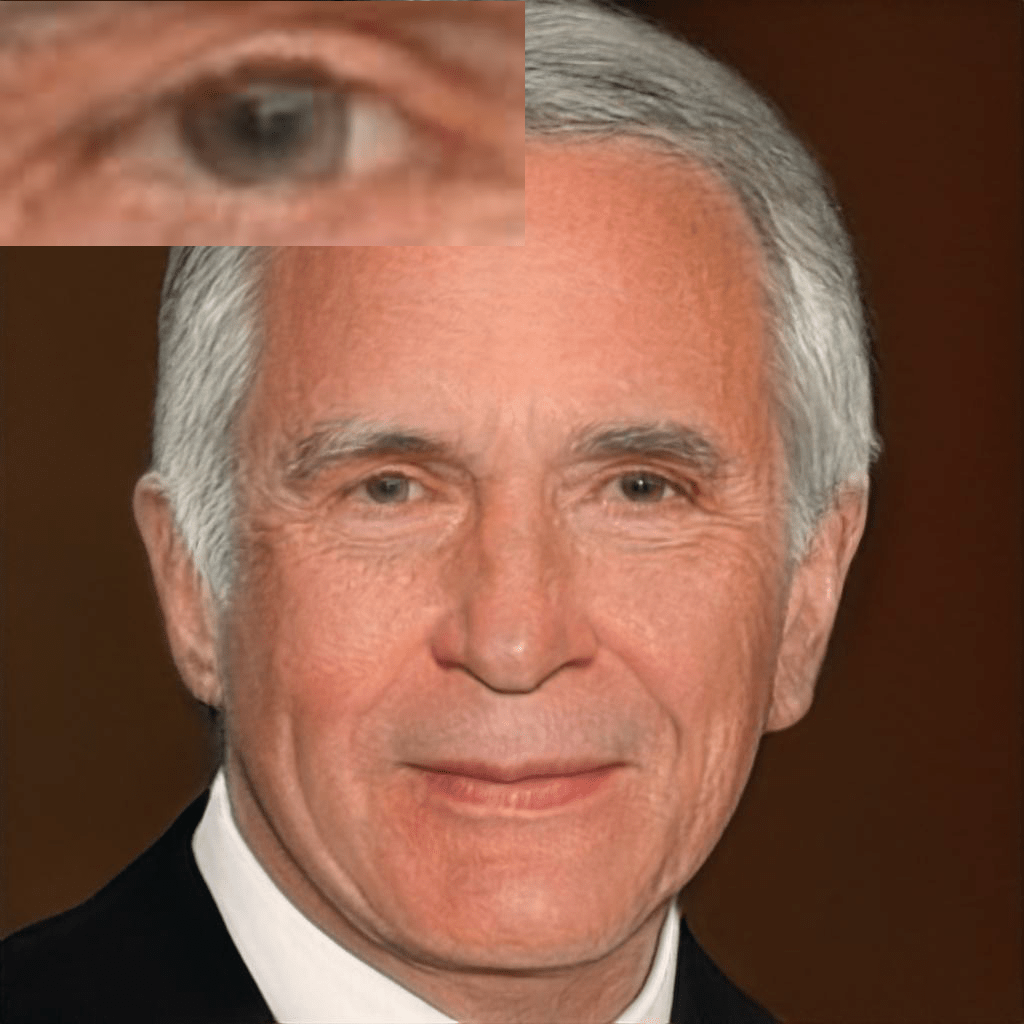}  &
 \includegraphics[height=\hhf,width=\wwf, trim=0 0 0 0,clip]{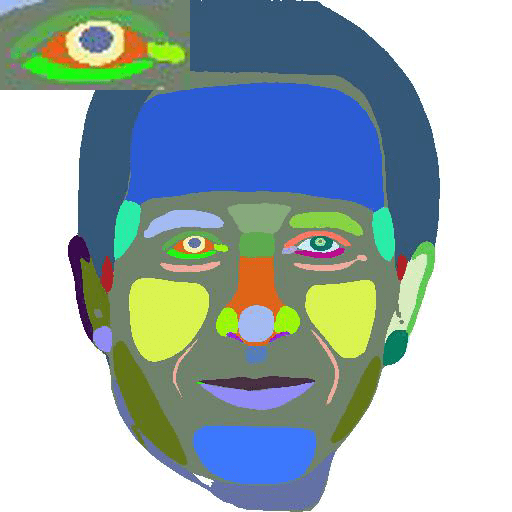} & \includegraphics[height=\hhf,width=\wwf, trim=0 0 0 0,clip]{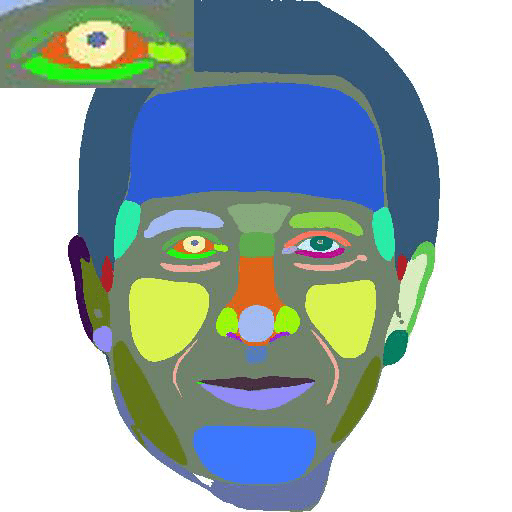}  \\
\end{tabular}

\begin{tabular}{cc}
 \includegraphics[height=\hhfl,width=\wwfl, trim=0 0 0 0,clip]{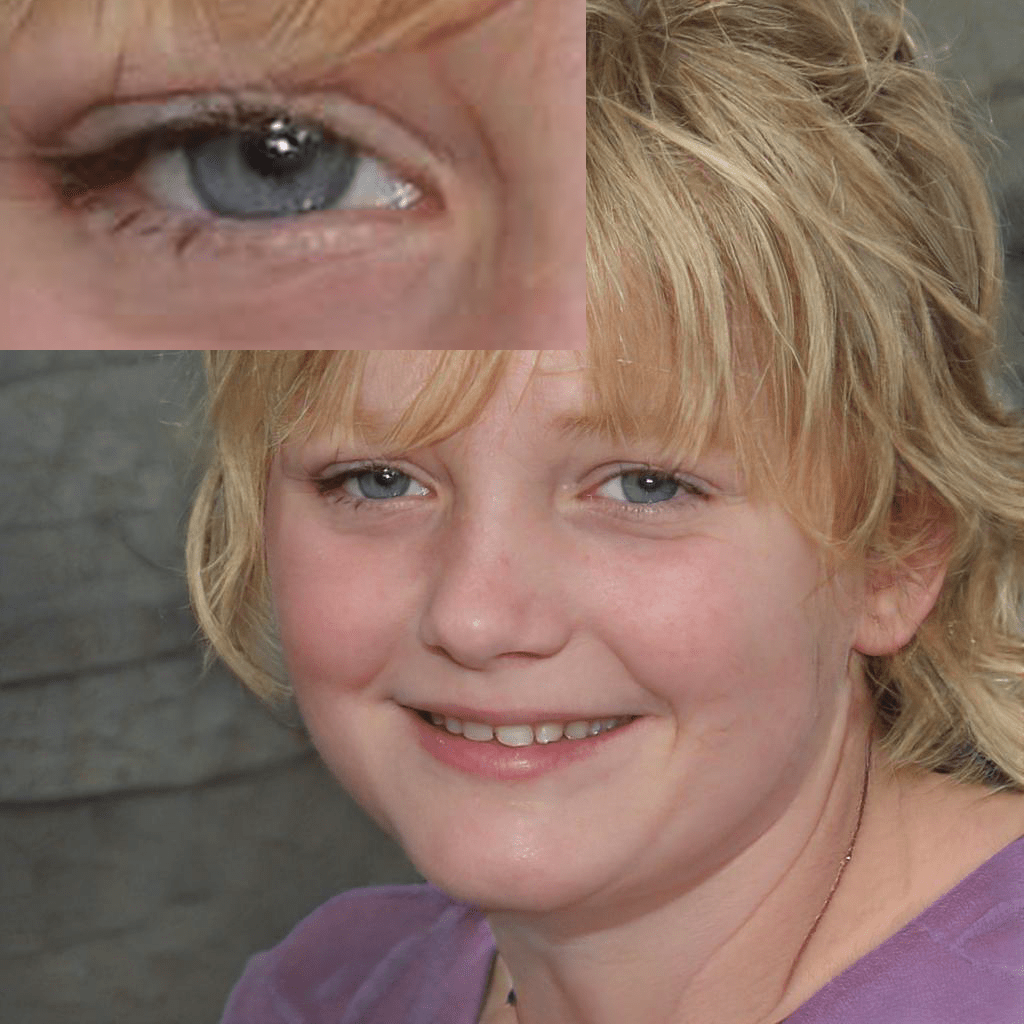} & \includegraphics[height=\hhfl,width=\wwfl, trim=0 0 0 0,clip]{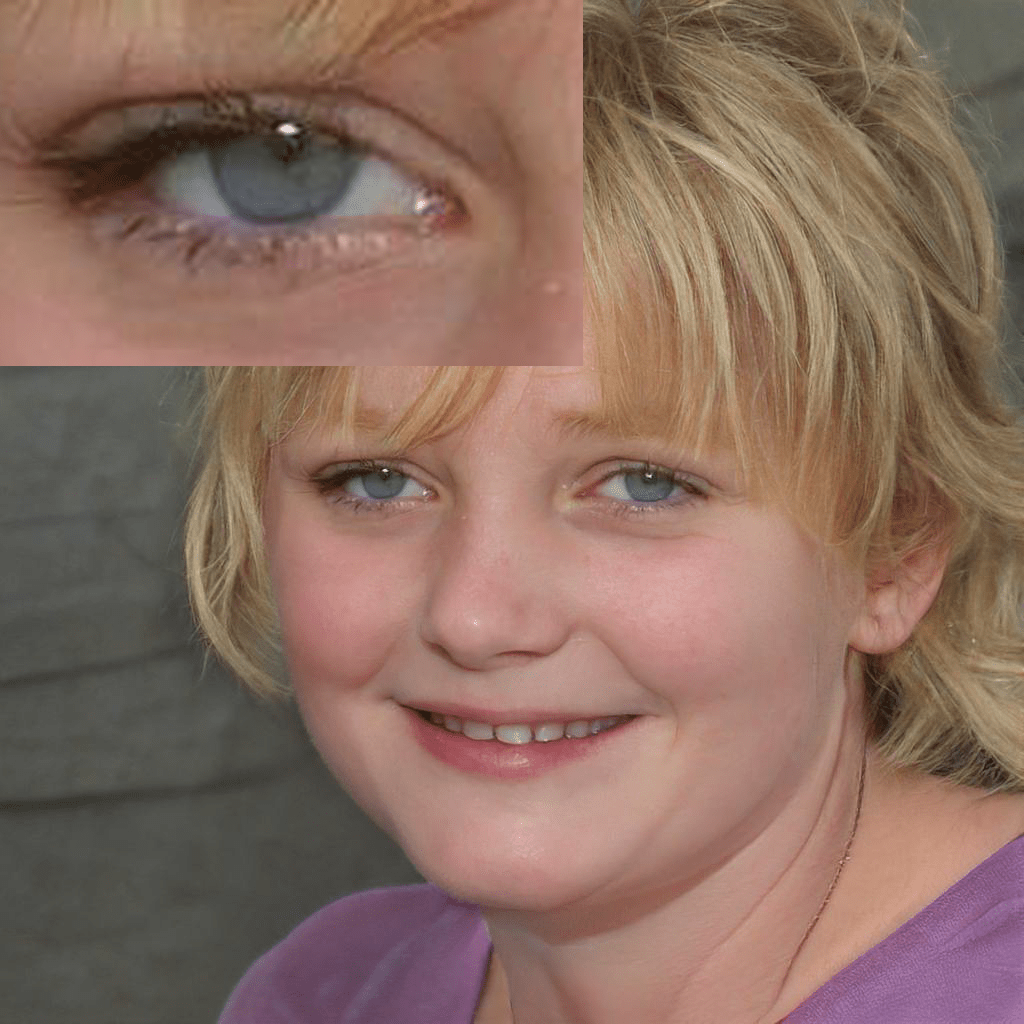}  \\
 \includegraphics[height=\hhfl,width=\wwfl, trim=0 0 0 0,clip]{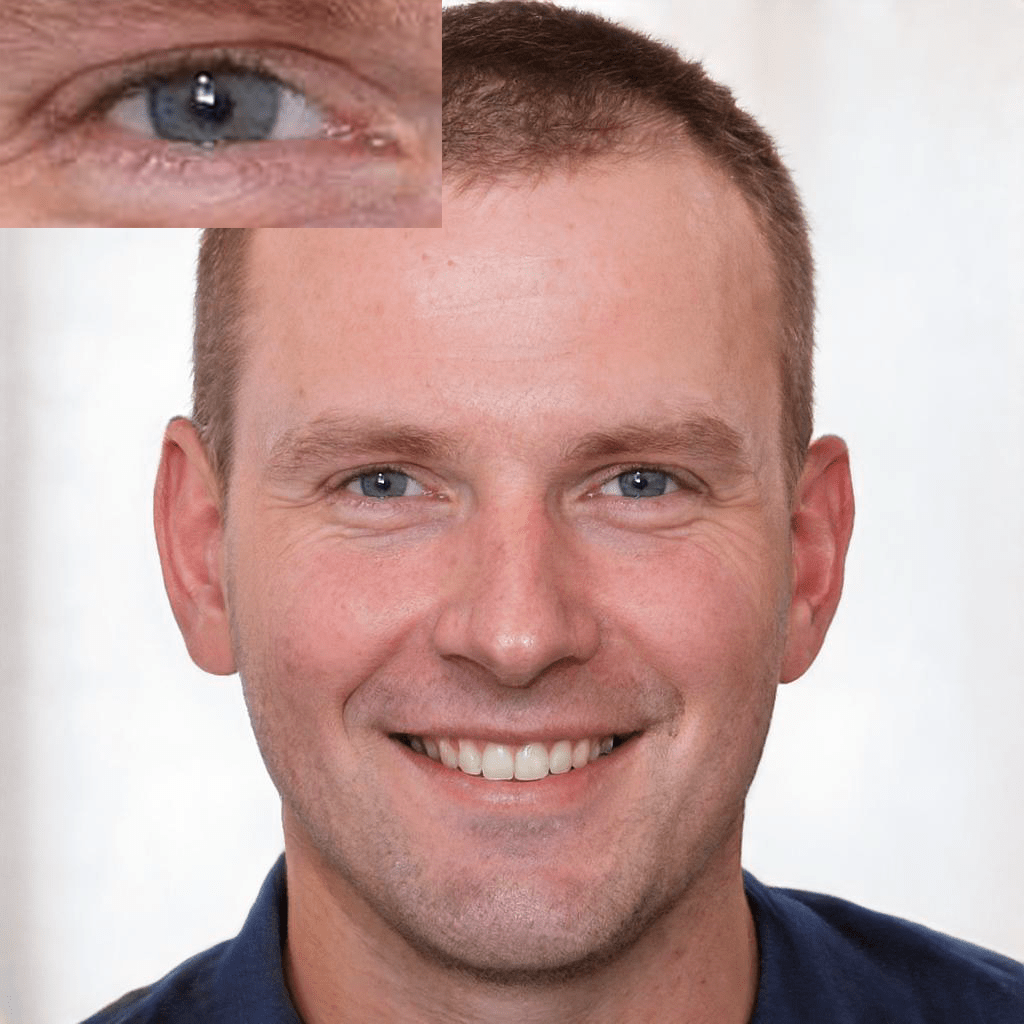} & \includegraphics[height=\hhfl,width=\wwfl, trim=0 0 0 0,clip]{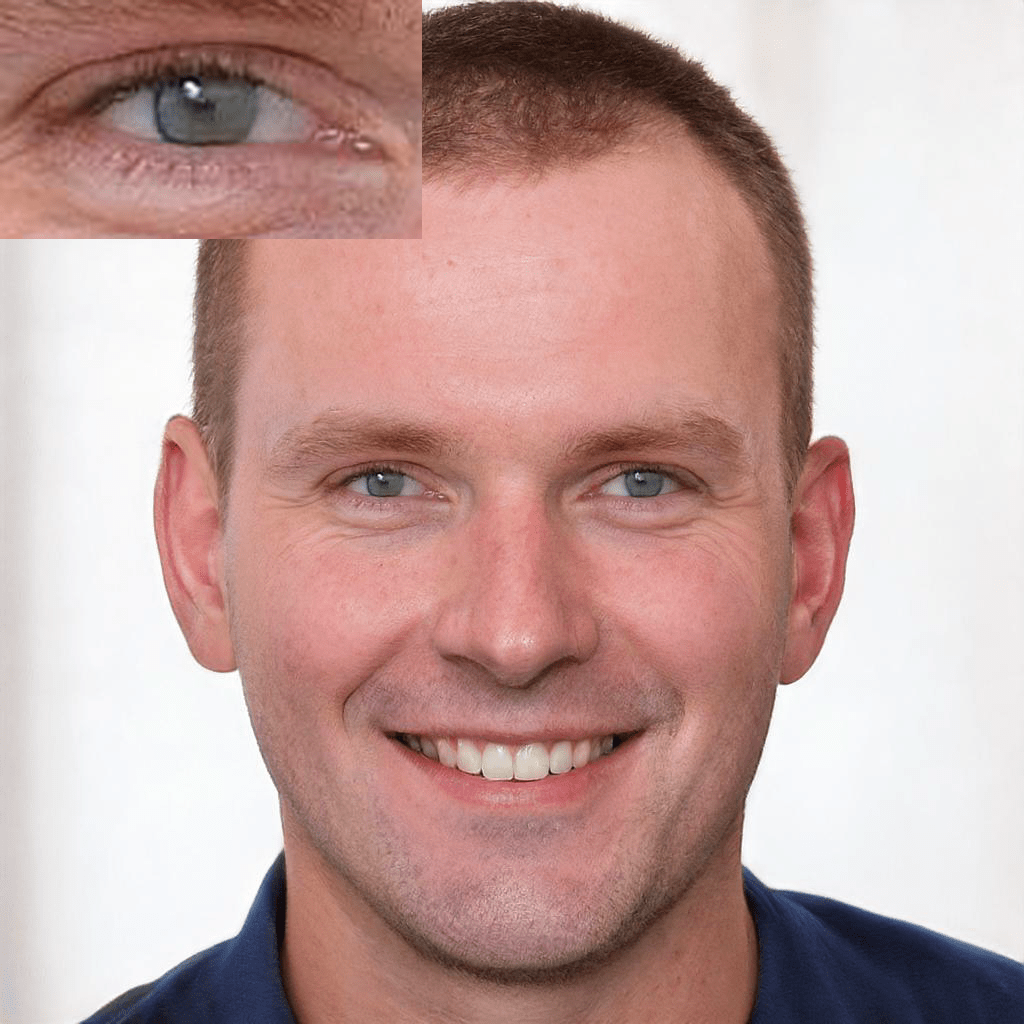}  \\
\end{tabular}

\caption{\footnotesize  {\bf Pupil size editing.} \textit{First row}: Image and mask pair to learn editing vector. Images are images before editing and after editing. Segmentation masks are before editing and target segmentation mask after manual modification. \textit{Second and third rows}: Applying the learnt edit on new images.}
\label{fig:face_pupil}
\vspace{-3mm}
\end{figure}


\begin{figure}
\addtolength{\tabcolsep}{-10pt}
\hspace{1mm}
\begin{tabular}{cccc}
 \includegraphics[height=\hhc,width=\wwf, trim=0 58 0 58,clip]{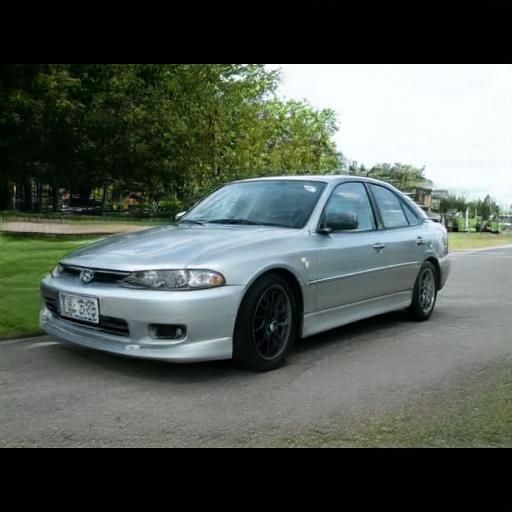} & \includegraphics[height=\hhc,width=\wwf, trim=0 58 0 58,clip]{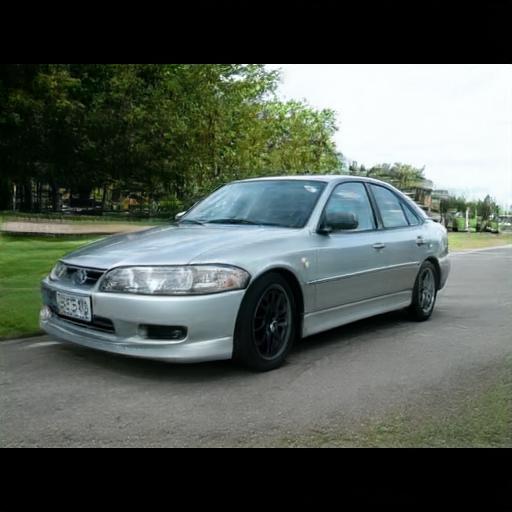}  &
 \includegraphics[height=\hhc,width=\wwf, trim=0 58 0 58,clip]{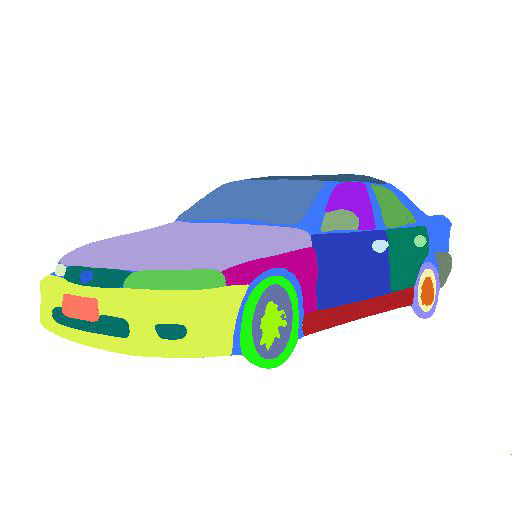} & \includegraphics[height=\hhc,width=\wwf, trim=0 58 0 58,clip]{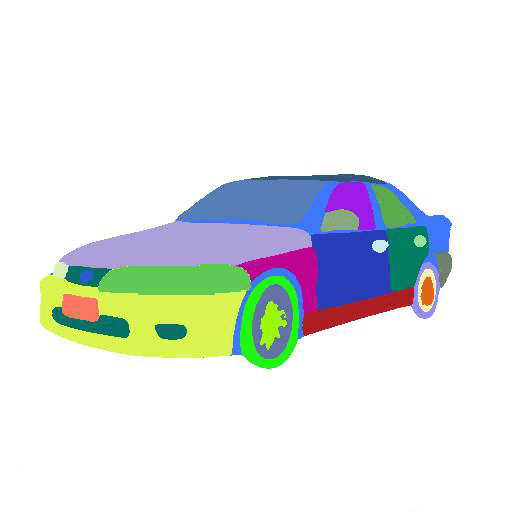}  \\
\end{tabular}

\begin{tabular}{cc}
 \includegraphics[height=\hhcl,width=\wwfl, trim=0 58 0 58,clip]{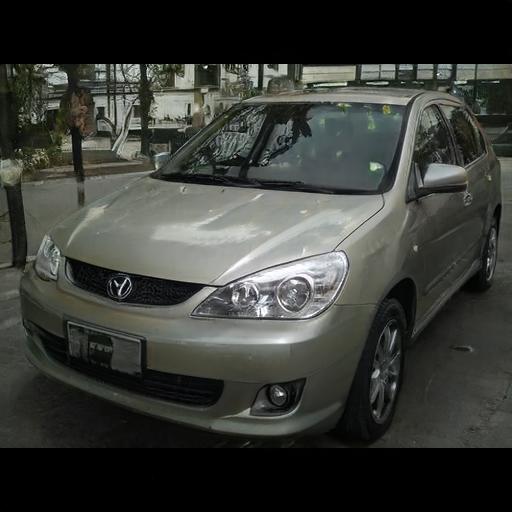} & \includegraphics[height=\hhcl,width=\wwfl, trim=0 58 0 58,clip]{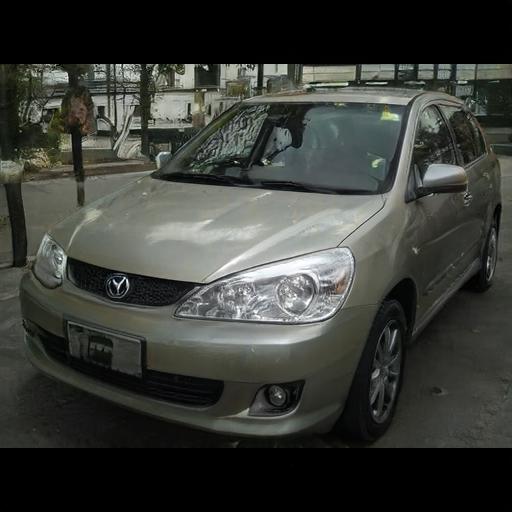}  \\
 \includegraphics[height=\hhcl,width=\wwfl, trim=0 58 0 58,clip]{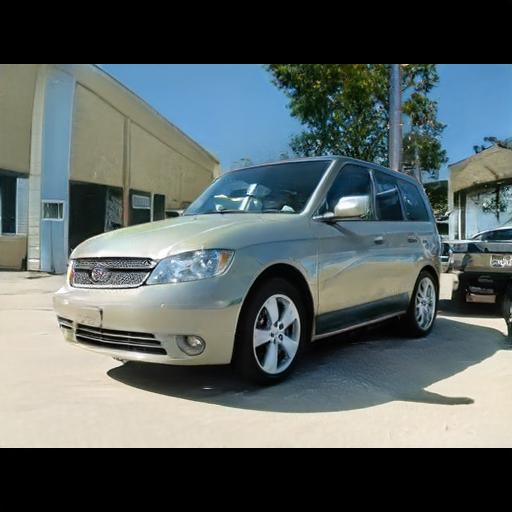} & \includegraphics[height=\hhcl,width=\wwfl, trim=0 58 0 58,clip]{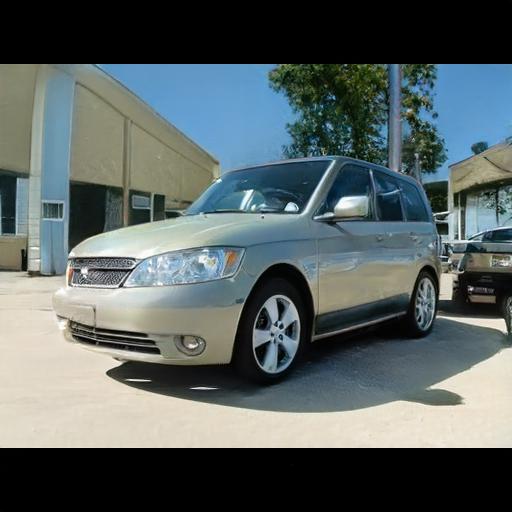}  \\
  \includegraphics[height=\hhcl,width=\wwfl, trim=0 58 0 58,clip]{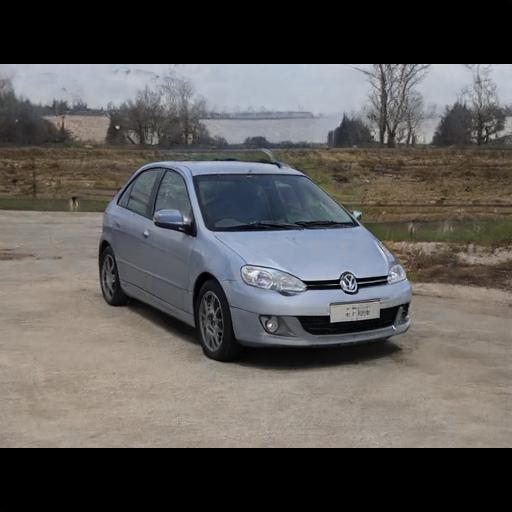} & \includegraphics[height=\hhcl,width=\wwfl, trim=0 58 0 58,clip]{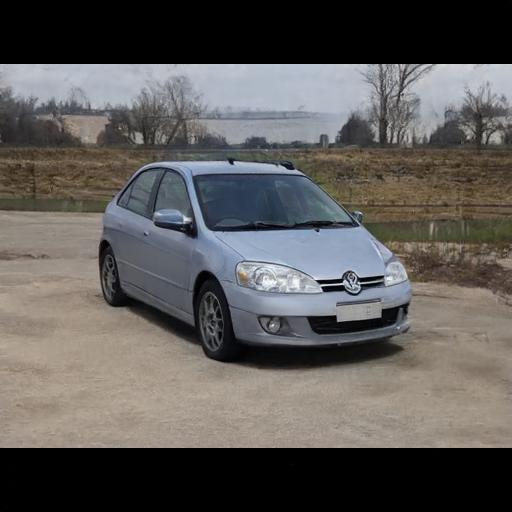}  \\
\end{tabular}

\caption{\footnotesize  {\bf Front light size editing.} \textit{First row}: Image and mask pair to learn editing vector. Images are images before editing and after editing. Segmentation masks are before editing and target segmentation mask after manual modification. \textit{Second to fourth rows}: Applying the learnt edit on new images.}
\label{fig:car_frontlight}
\vspace{-3mm}
\end{figure}

\begin{figure}
\addtolength{\tabcolsep}{-10pt}
\hspace{1mm}
\begin{tabular}{cccc}
 \includegraphics[height=\hhc,width=\wwf, trim=0 58 0 58,clip]{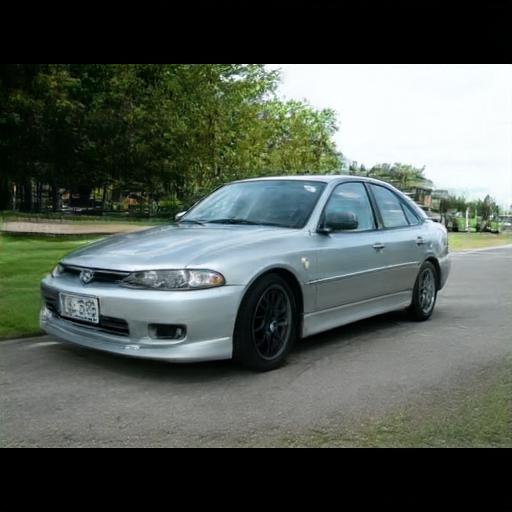} & \includegraphics[height=\hhc,width=\wwf, trim=0 58 0 58,clip]{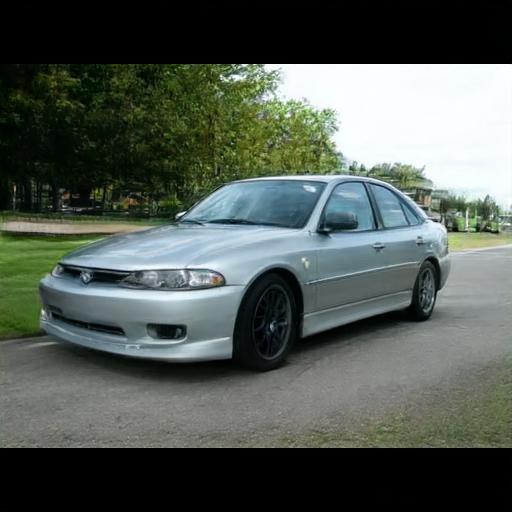}  &
 \includegraphics[height=\hhc,width=\wwf, trim=0 58 0 58,clip]{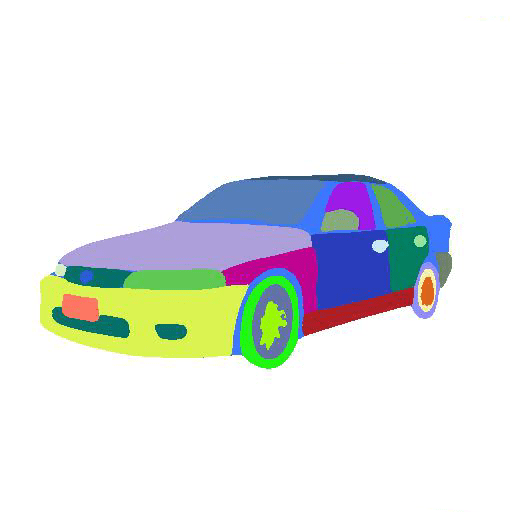} & \includegraphics[height=\hhc,width=\wwf, trim=0 58 0 58,clip]{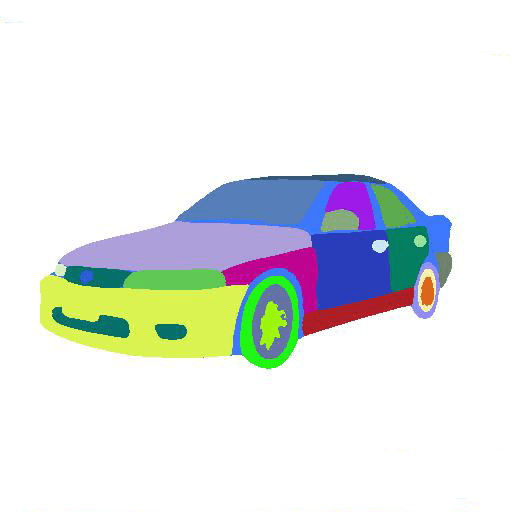}  \\
\end{tabular}

\begin{tabular}{cc}
 \includegraphics[height=\hhcl,width=\wwfl, trim=0 58 0 58,clip]{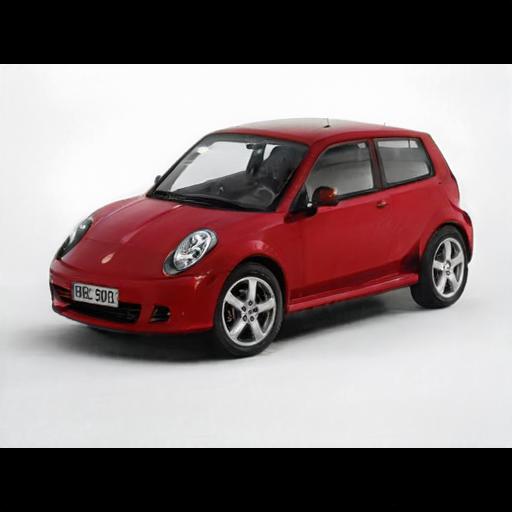} & \includegraphics[height=\hhcl,width=\wwfl, trim=0 58 0 58,clip]{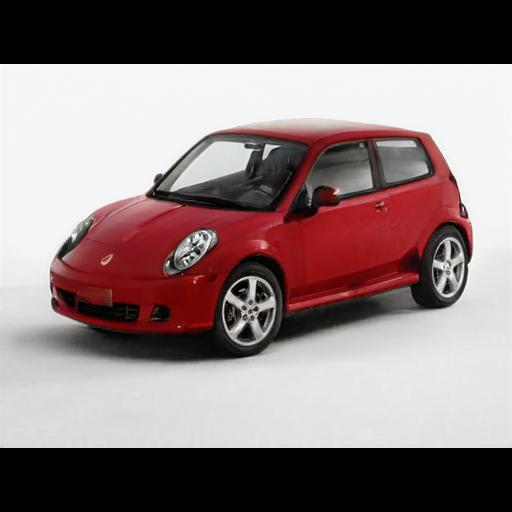}  \\
 \includegraphics[height=\hhcl,width=\wwfl, trim=0 58 0 58,clip]{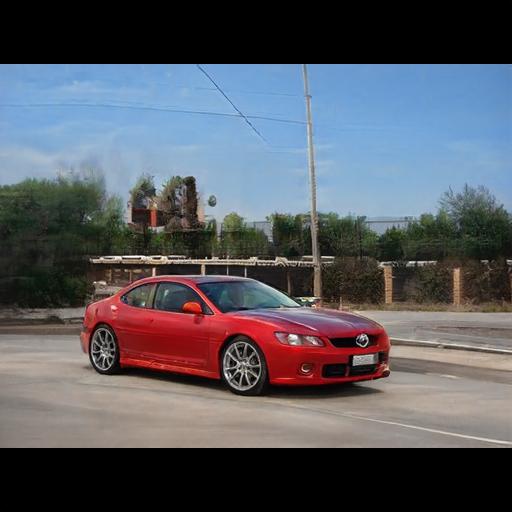} & \includegraphics[height=\hhcl,width=\wwfl, trim=0 58 0 58,clip]{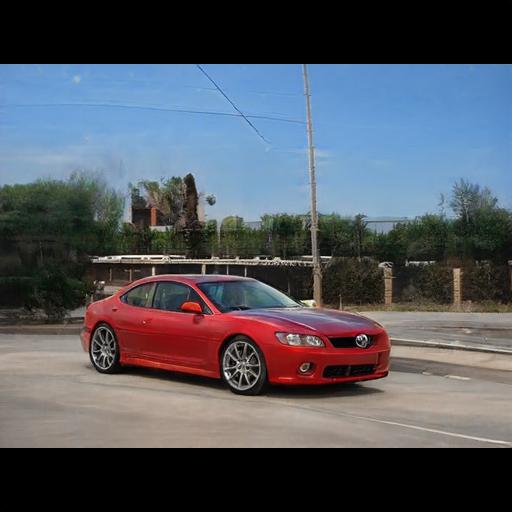}  \\
  \includegraphics[height=\hhcl,width=\wwfl, trim=0 58 0 58,clip]{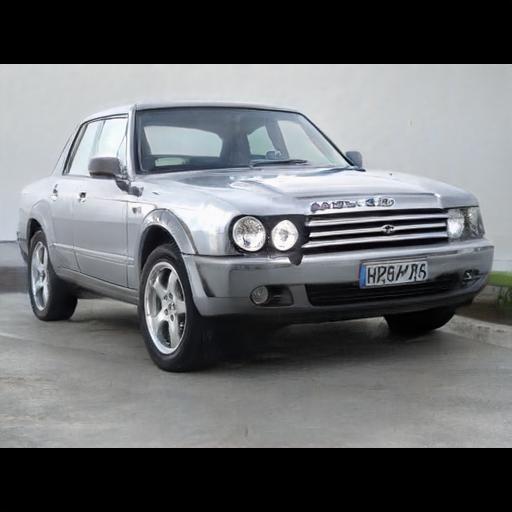} & \includegraphics[height=\hhcl,width=\wwfl, trim=0 58 0 58,clip]{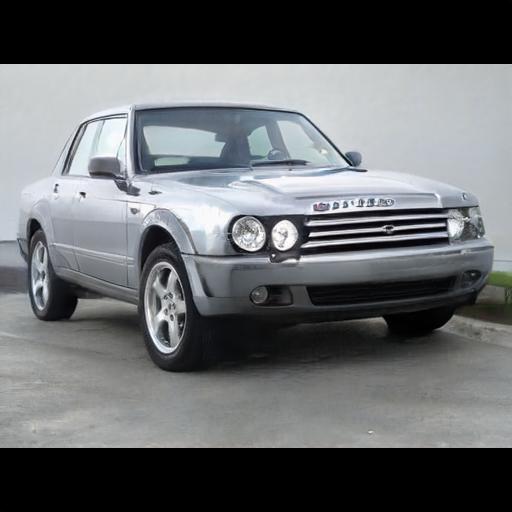}  \\
\end{tabular}

\caption{\footnotesize  {\bf License plate deletion editing.} \textit{First row}: Image and mask pair to learn editing vector. Images are images before editing and after editing. Segmentation masks are before editing and target segmentation mask after manual modification. \textit{Second to fourth rows}: Applying the learnt edit on new images.}
\label{fig:car_license}
\vspace{-3mm}
\end{figure}

\begin{figure}
\addtolength{\tabcolsep}{-10pt}
\hspace{1mm}
\begin{tabular}{cccc}
 \includegraphics[height=\hhc,width=\wwf, trim=0 58 0 58,clip]{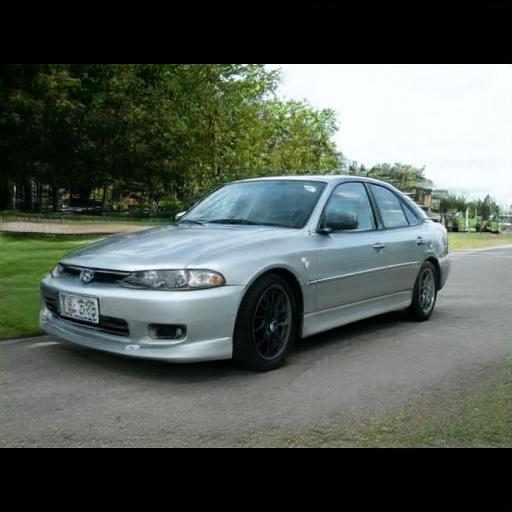} & \includegraphics[height=\hhc,width=\wwf, trim=0 58 0 58,clip]{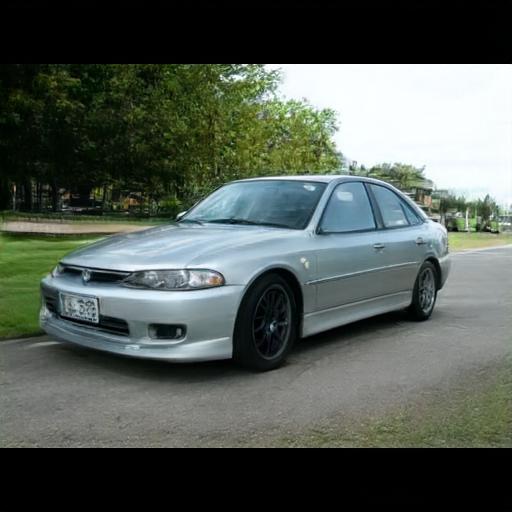}  &
 \includegraphics[height=\hhc,width=\wwf, trim=0 58 0 58,clip]{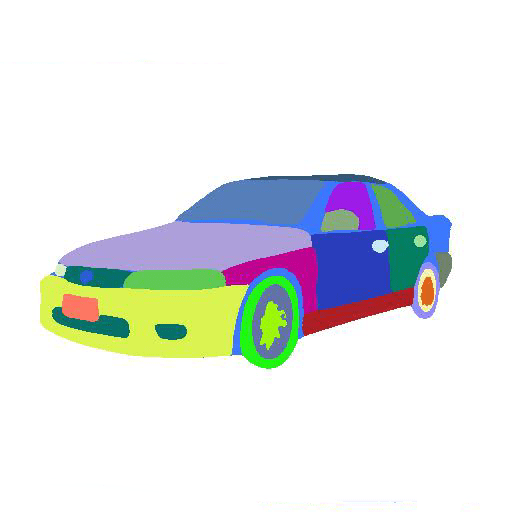} & \includegraphics[height=\hhc,width=\wwf, trim=0 58 0 58,clip]{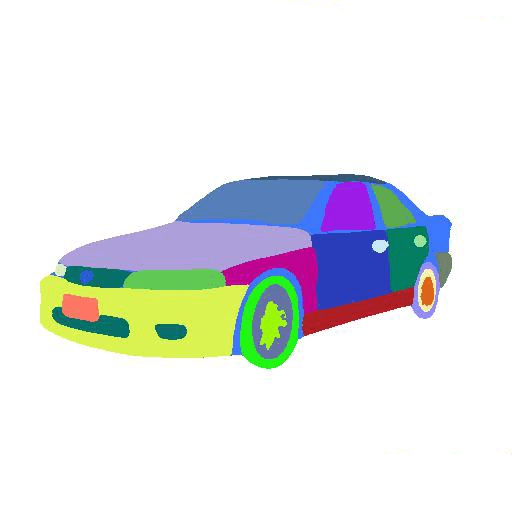}  \\
\end{tabular}

\begin{tabular}{cc}
 \includegraphics[height=\hhcl,width=\wwfl, trim=0 58 0 58,clip]{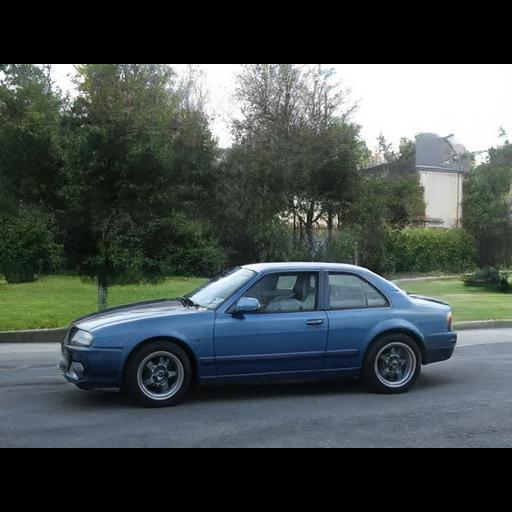} & \includegraphics[height=\hhcl,width=\wwfl, trim=0 58 0 58,clip]{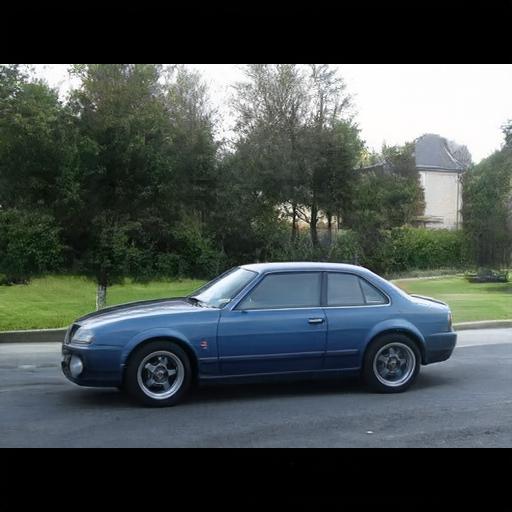}  \\
 \includegraphics[height=\hhcl,width=\wwfl, trim=0 58 0 58,clip]{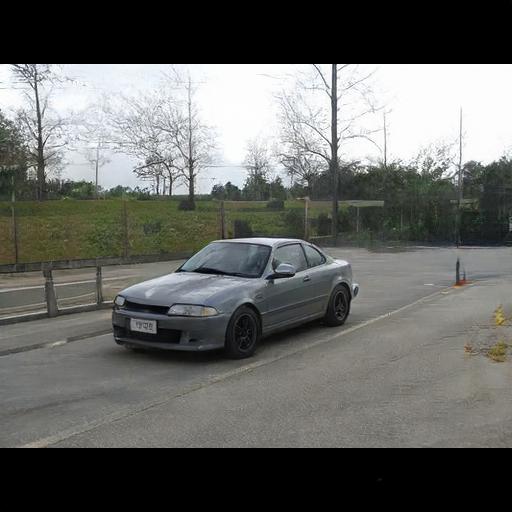} & \includegraphics[height=\hhcl,width=\wwfl, trim=0 58 0 58,clip]{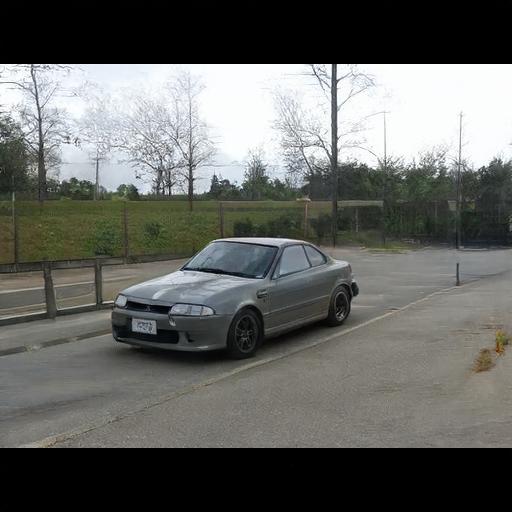}  \\
  \includegraphics[height=\hhcl,width=\wwfl, trim=0 58 0 58,clip]{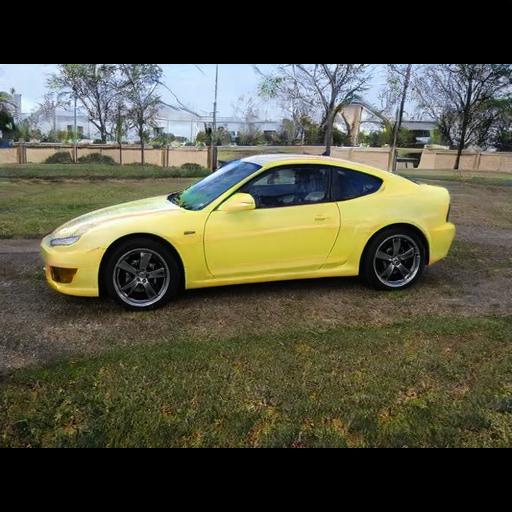} & \includegraphics[height=\hhcl,width=\wwfl, trim=0 58 0 58,clip]{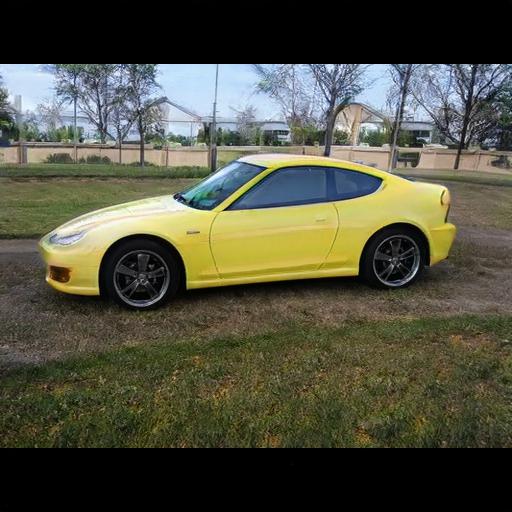}  \\
\end{tabular}

\caption{\footnotesize  {\bf Side mirror deletion editing.} \textit{First row}: Image and mask pair to learn editing vector. Images are images before editing and after editing. Segmentation masks are before editing and target segmentation mask after manual modification. \textit{Second to fourth rows}: Applying the learnt edit on new images.}
\label{fig:car_mirror}
\vspace{-3mm}
\end{figure}

\begin{figure}
\addtolength{\tabcolsep}{-10pt}
\hspace{1mm}
\begin{tabular}{cccc}
 \includegraphics[height=\hhc,width=\wwf, trim=0 58 0 58,clip]{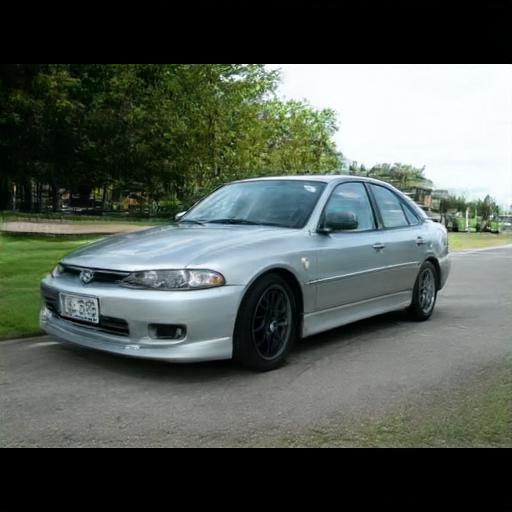} & \includegraphics[height=\hhc,width=\wwf, trim=0 58 0 58,clip]{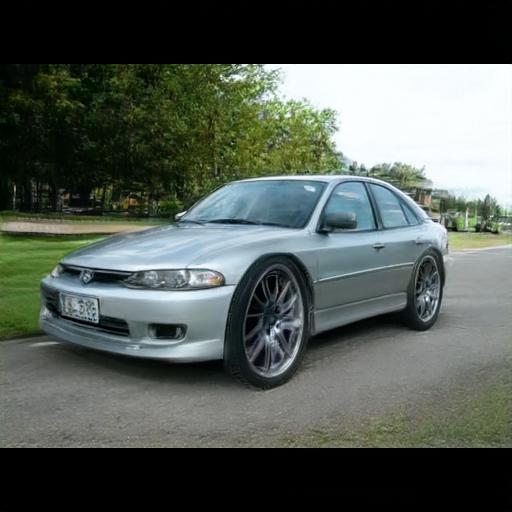}  &
 \includegraphics[height=\hhc,width=\wwf, trim=0 58 0 58,clip]{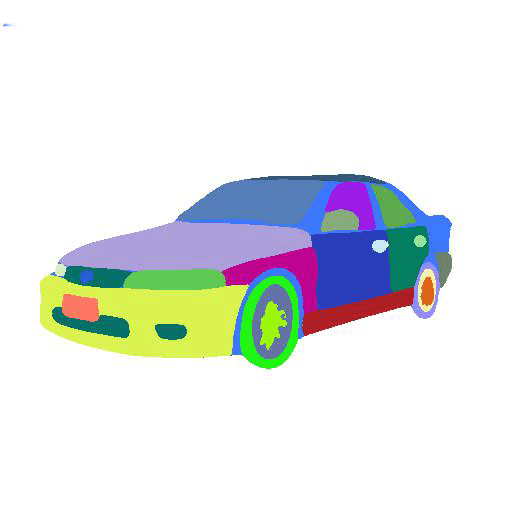} & \includegraphics[height=\hhc,width=\wwf, trim=0 58 0 58,clip]{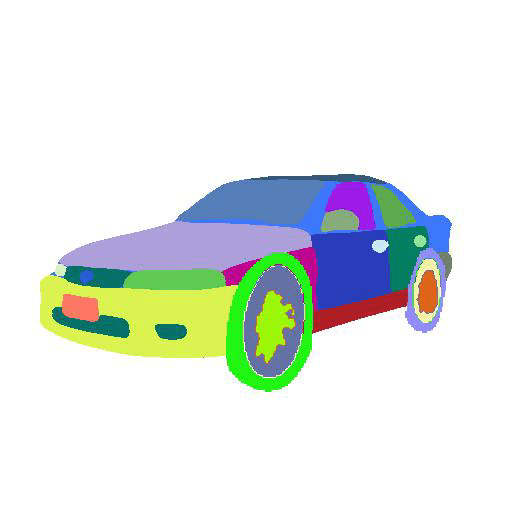}  \\
\end{tabular}

\begin{tabular}{cc}
 \includegraphics[height=\hhcl,width=\wwfl, trim=0 58 0 58,clip]{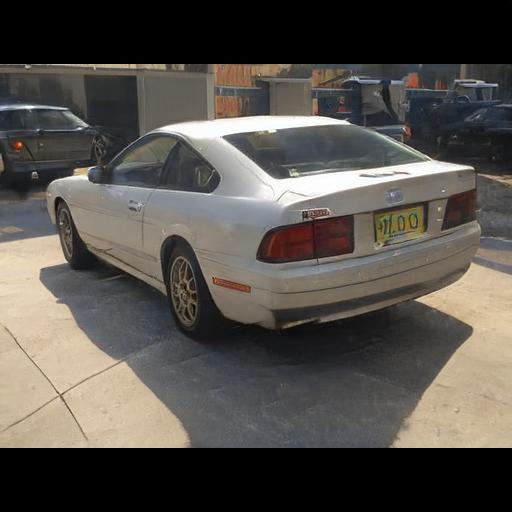} & \includegraphics[height=\hhcl,width=\wwfl, trim=0 58 0 58,clip]{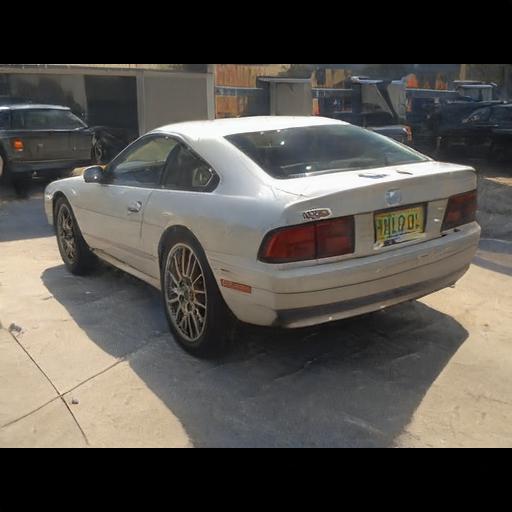}  \\
 \includegraphics[height=\hhcl,width=\wwfl, trim=0 58 0 58,clip]{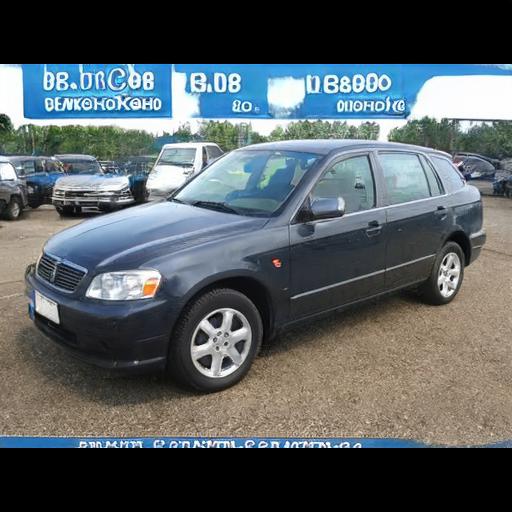} & \includegraphics[height=\hhcl,width=\wwfl, trim=0 58 0 58,clip]{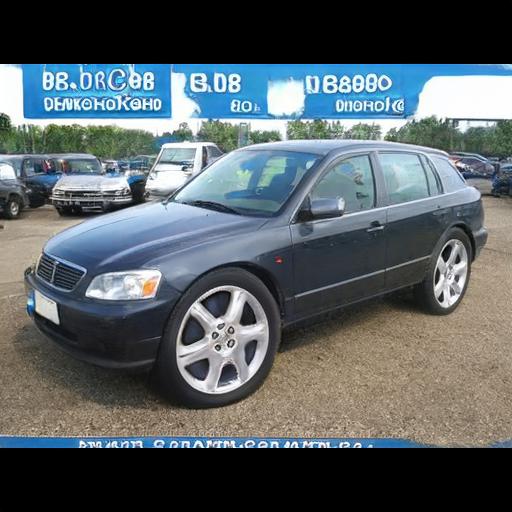}  \\
  \includegraphics[height=\hhcl,width=\wwfl, trim=0 58 0 58,clip]{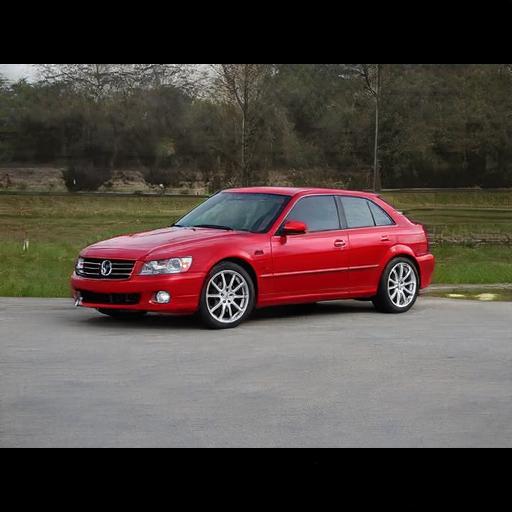} & \includegraphics[height=\hhcl,width=\wwfl, trim=0 58 0 58,clip]{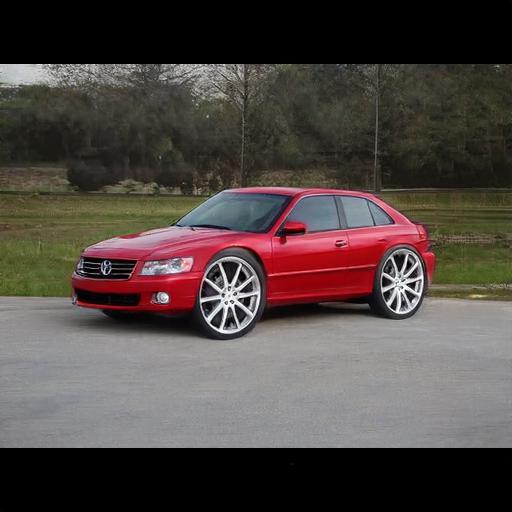}  \\
\end{tabular}

\caption{\footnotesize  {\bf Wheel size editing.} \textit{First row}: Image and mask pair to learn editing vector. Images are images before editing and after editing. Segmentation masks are before editing and target segmentation mask after manual modification. \textit{Second to fourth rows}: Applying the learnt edit on new images.}
\label{fig:car_wheel}
\vspace{-3mm}
\end{figure}

\begin{figure}
\addtolength{\tabcolsep}{-10pt}
\hspace{1mm}
\begin{tabular}{cccc}
 \includegraphics[height=\hhc,width=\wwf, trim=0 58 0 58,clip]{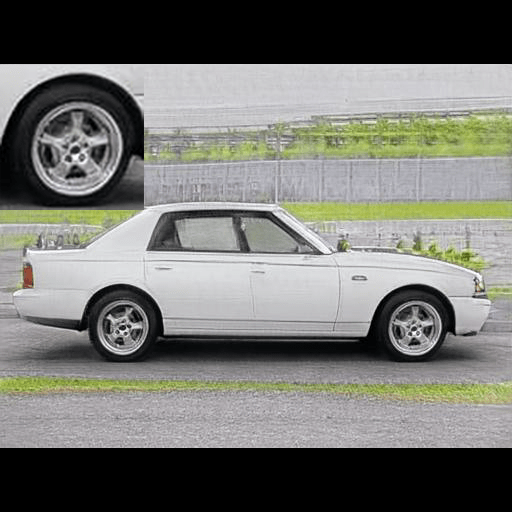} & \includegraphics[height=\hhc,width=\wwf, trim=0 58 0 58,clip]{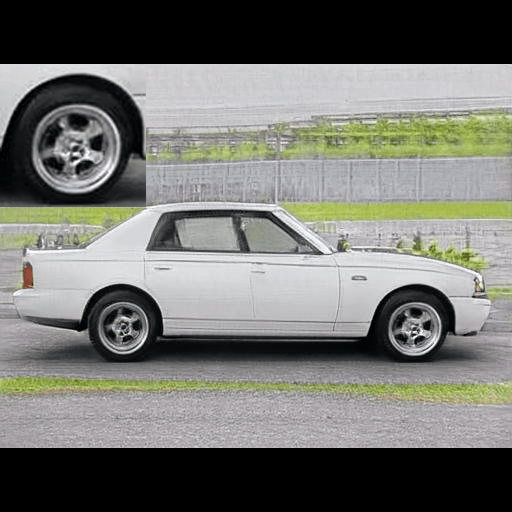}  &
 \includegraphics[height=\hhc,width=\wwf, trim=0 58 0 58,clip]{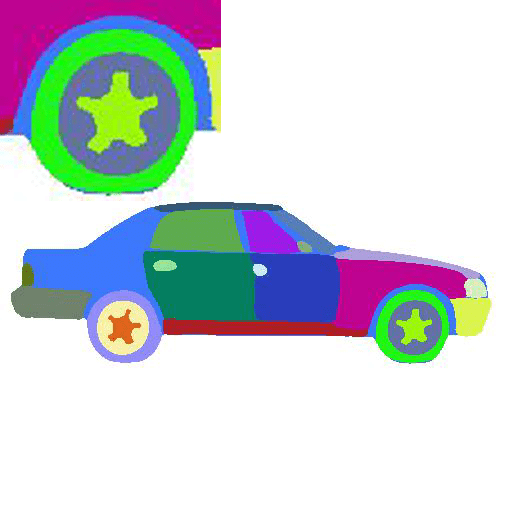} & \includegraphics[height=\hhc,width=\wwf, trim=0 58 0 58,clip]{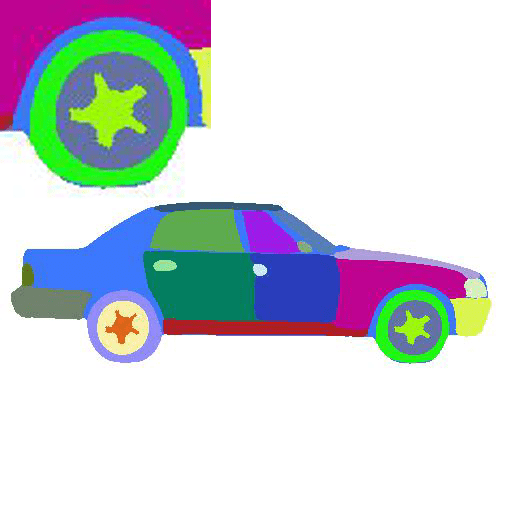}  \\
\end{tabular}

\begin{tabular}{cc}
 \includegraphics[height=\hhcl,width=\wwfl, trim=0 58 0 58,clip]{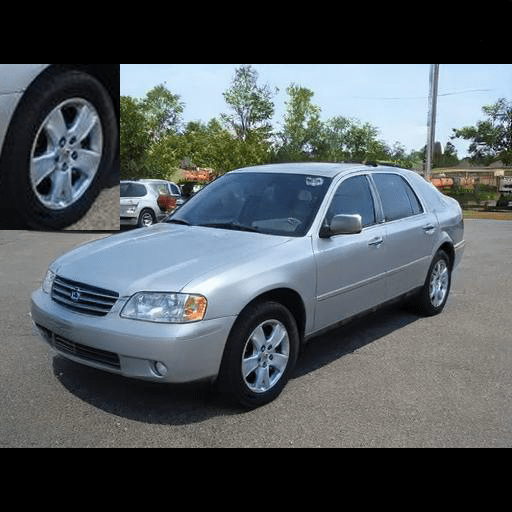} & \includegraphics[height=\hhcl,width=\wwfl, trim=0 58 0 58,clip]{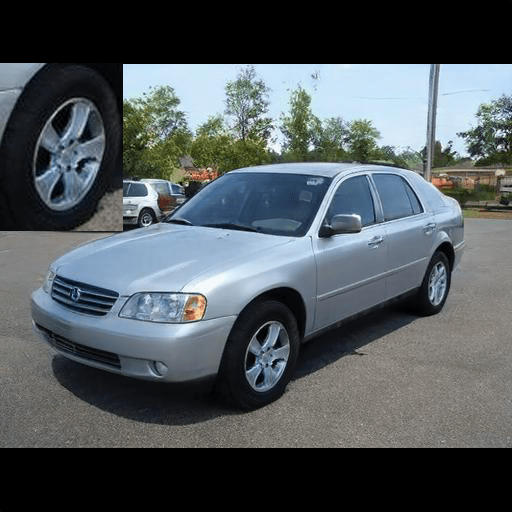}  \\
 \includegraphics[height=\hhcl,width=\wwfl, trim=0 58 0 58,clip]{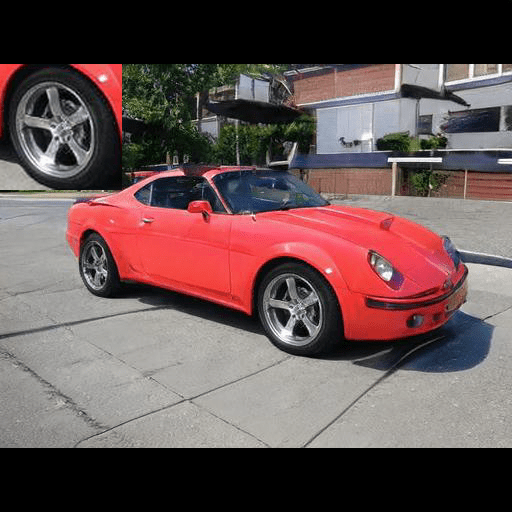} & \includegraphics[height=\hhcl,width=\wwfl, trim=0 58 0 58,clip]{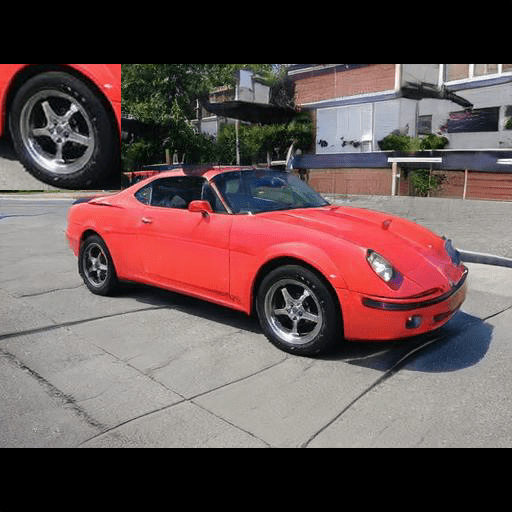}  \\
  \includegraphics[height=\hhcl,width=\wwfl, trim=0 58 0 58,clip]{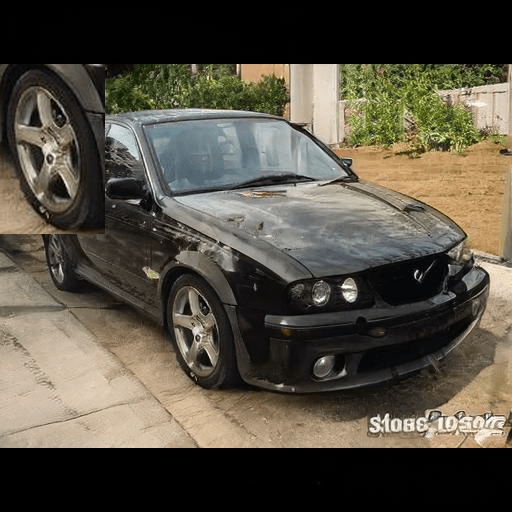} & \includegraphics[height=\hhcl,width=\wwfl, trim=0 58 0 58,clip]{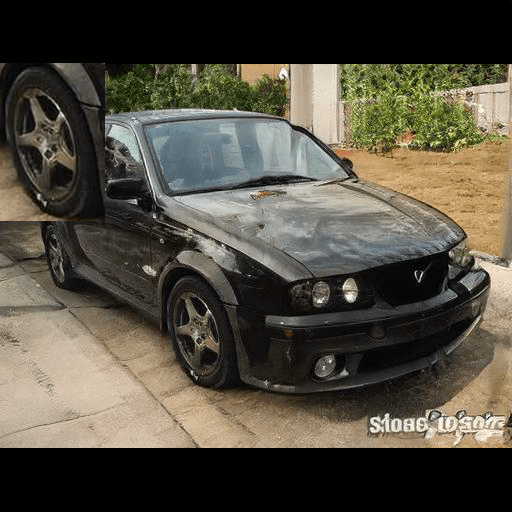}  \\
\end{tabular}

\caption{\footnotesize  {\bf Wheel/spoke rotation editing.} \textit{First row}: Image and mask pair to learn editing vector. Images are images before editing and after editing. Segmentation masks are before editing and target segmentation mask after manual modification. \textit{Second to fourth rows}: Applying the learnt edit on new images.}
\label{fig:car_rotate}
\vspace{-3mm}
\end{figure}


\begin{figure}
\addtolength{\tabcolsep}{-10pt}
\hspace{1mm}
\begin{tabular}{cccc}
 \includegraphics[height=\hhf,width=\wwf, trim=0 0 0 0,clip]{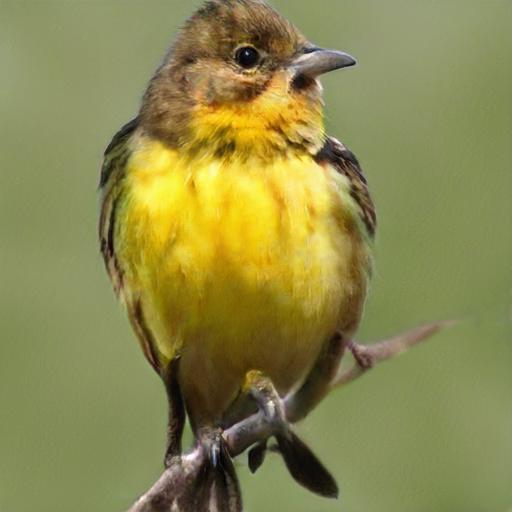} & \includegraphics[height=\hhf,width=\wwf, trim=0 0 0 0,clip]{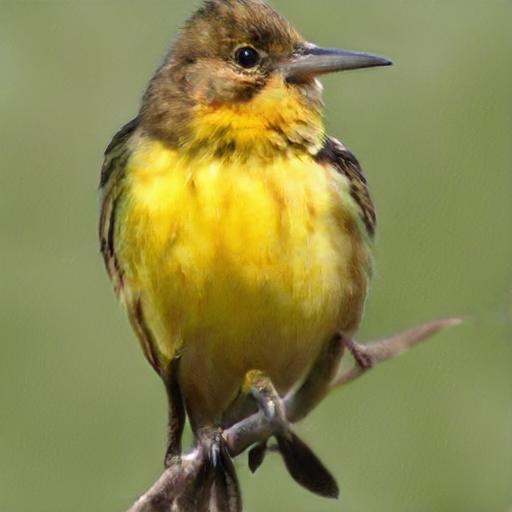}  &
 \includegraphics[height=\hhf,width=\wwf, trim=0 0 0 0,clip]{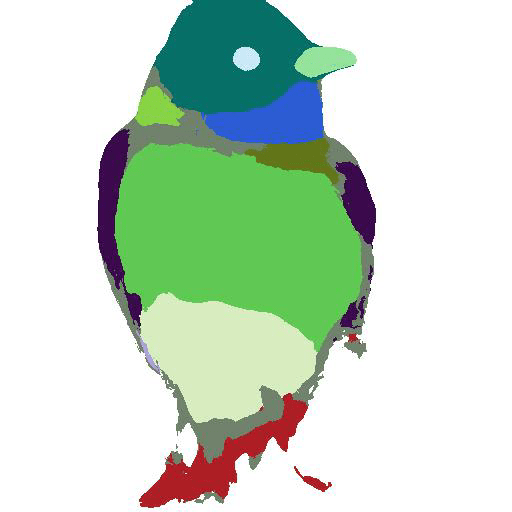} & \includegraphics[height=\hhf,width=\wwf, trim=0 0 0 0,clip]{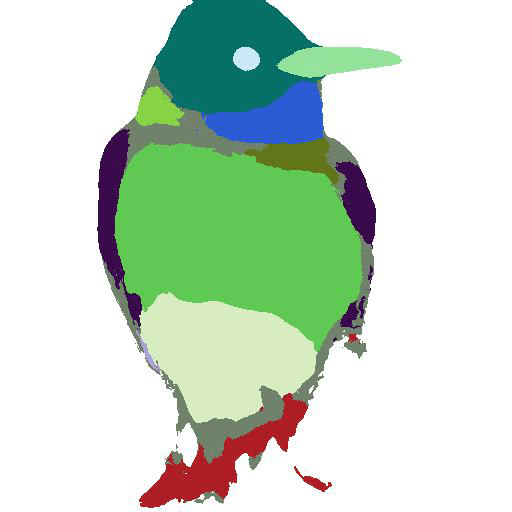}  \\
\end{tabular}

\begin{tabular}{cc}
 \includegraphics[height=\hhfl,width=\wwfl, trim=0 0 0 0,clip]{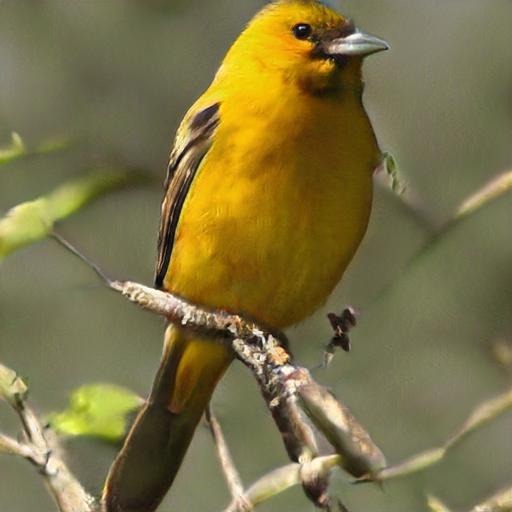} & \includegraphics[height=\hhfl,width=\wwfl, trim=0 0 0 0,clip]{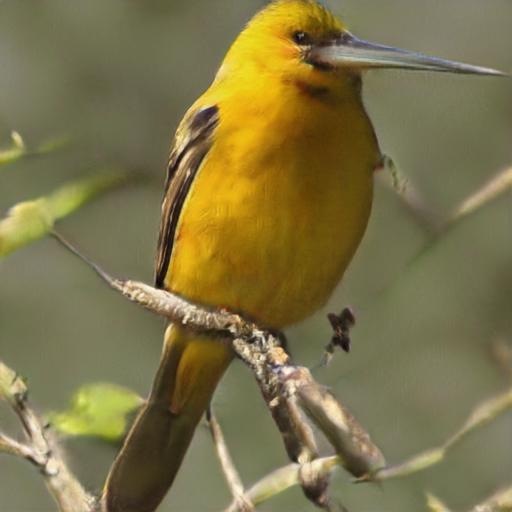}  \\
 \includegraphics[height=\hhfl,width=\wwfl, trim=0 0 0 0,clip]{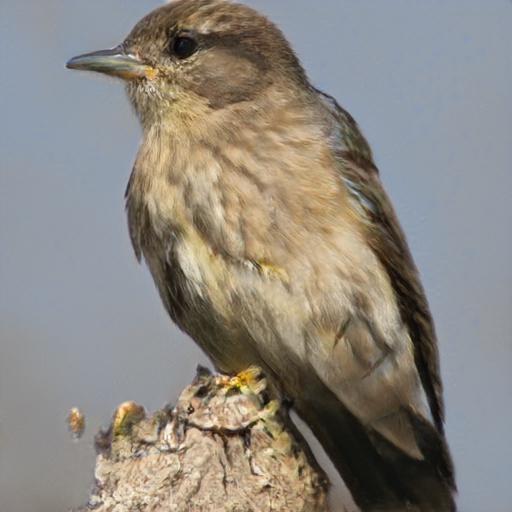} & \includegraphics[height=\hhfl,width=\wwfl, trim=0 0 0 0,clip]{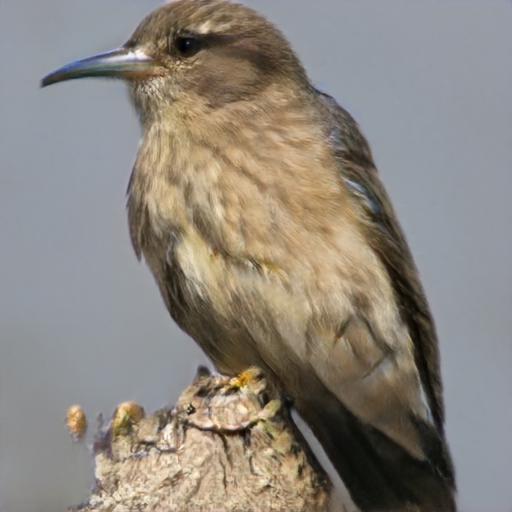}  \\
\end{tabular}

\caption{\footnotesize  {\bf Beak size editing.} \textit{First row}: Image and mask pair to learn editing vector. Images are images before editing and after editing. Segmentation masks are before editing and target segmentation mask after manual modification. \textit{Second and third rows}: Applying the learnt edit on new images.}
\label{fig:bird_beak}
\vspace{-3mm}
\end{figure}

\begin{figure}
\addtolength{\tabcolsep}{-6pt}
\hspace{-3mm}
\begin{tabular}{cccc}
 \includegraphics[height=\hhf,width=\wwf, trim=0 0 0 0,clip]{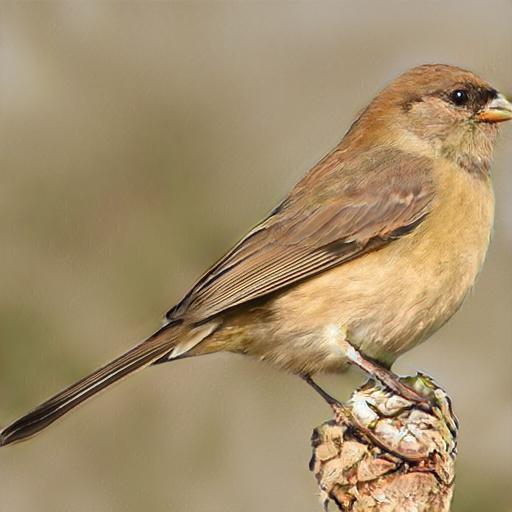} & \includegraphics[height=\hhf,width=\wwf, trim=0 0 0 0,clip]{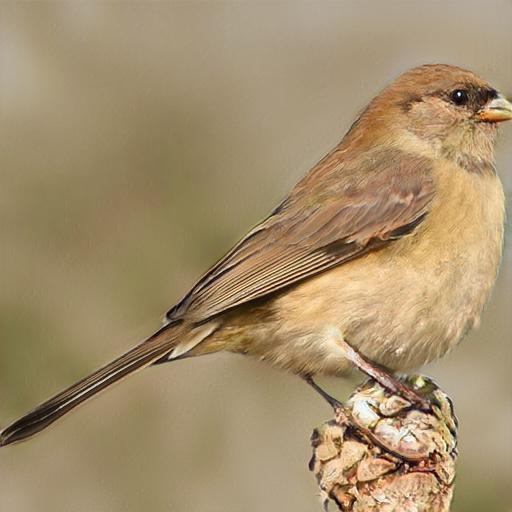}  &
 \includegraphics[height=\hhf,width=\wwf, trim=0 0 0 0,clip]{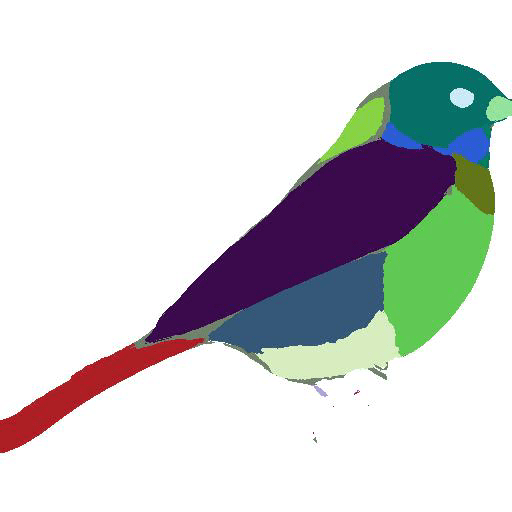} & \includegraphics[height=\hhf,width=\wwf, trim=0 0 0 0,clip]{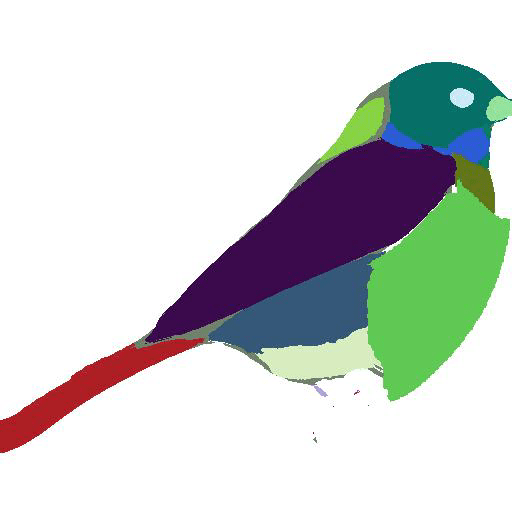}  \\
\end{tabular}

\begin{tabular}{cc}
 \includegraphics[height=\hhfl,width=\wwfl, trim=0 0 0 0,clip]{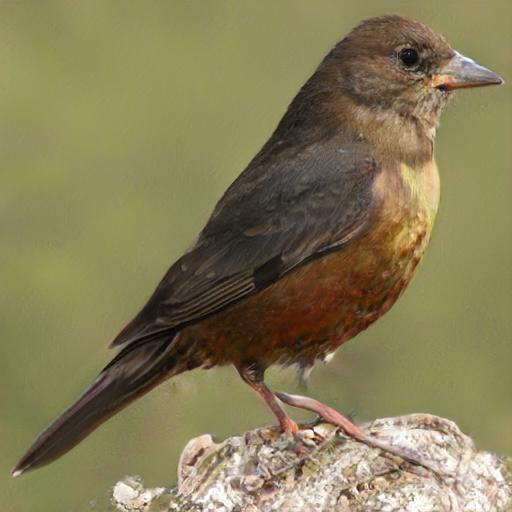} & \includegraphics[height=\hhfl,width=\wwfl, trim=0 0 0 0,clip]{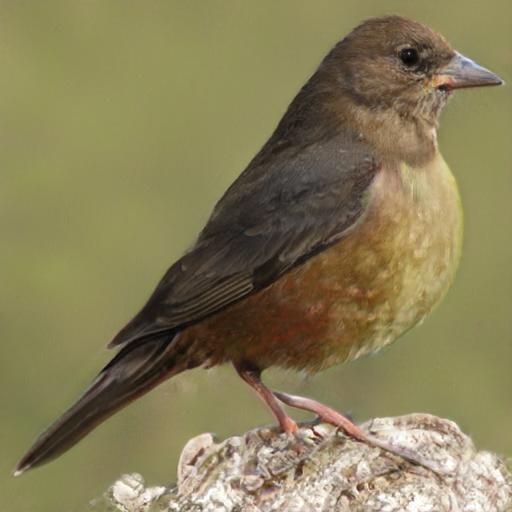}  \\
 \includegraphics[height=\hhfl,width=\wwfl, trim=0 0 0 0,clip]{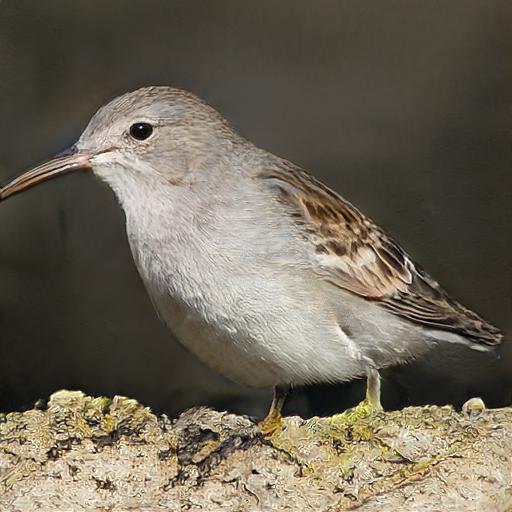} & \includegraphics[height=\hhfl,width=\wwfl, trim=0 0 0 0,clip]{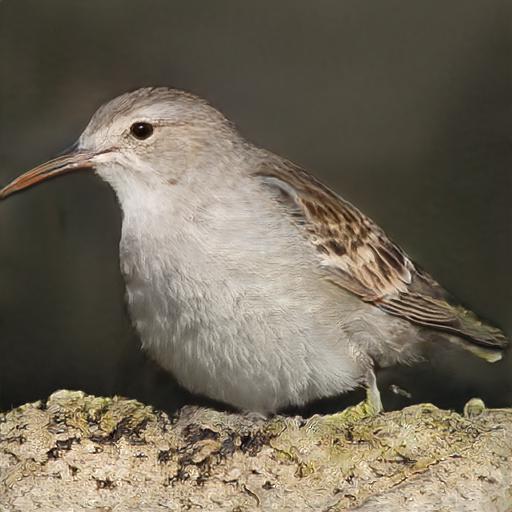}  \\
\end{tabular}

\caption{\footnotesize  {\bf Belly size editing.} \textit{First row}: Image and mask pair to learn editing vector. Images are images before editing and after editing. Segmentation masks are before editing and target segmentation mask after manual modification. \textit{Second and third rows}: Applying the learnt edit on new images.}
\label{fig:bird_belly}
\vspace{-3mm}
\end{figure}

\begin{figure}
\addtolength{\tabcolsep}{-6pt}
\hspace{-3mm}
\begin{tabular}{cccc}
 \includegraphics[height=\hhf,width=\wwf, trim=0 0 0 0,clip]{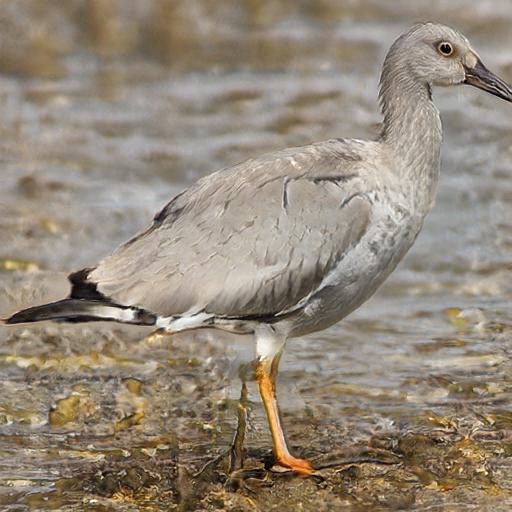} & \includegraphics[height=\hhf,width=\wwf, trim=0 0 0 0,clip]{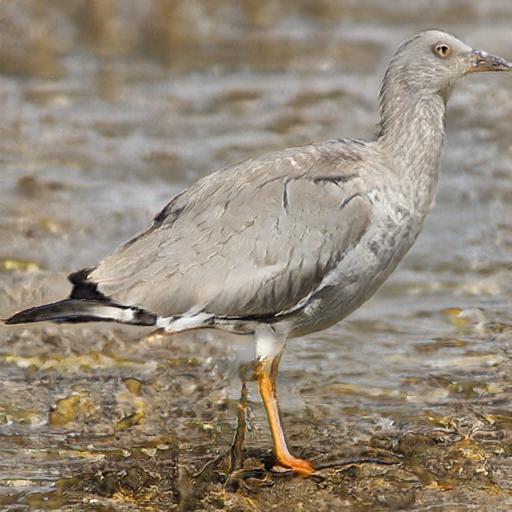}  &
 \includegraphics[height=\hhf,width=\wwf, trim=0 0 0 0,clip]{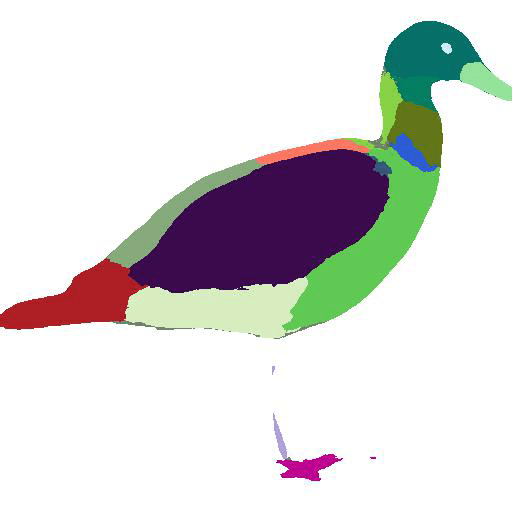} & \includegraphics[height=\hhf,width=\wwf, trim=0 0 0 0,clip]{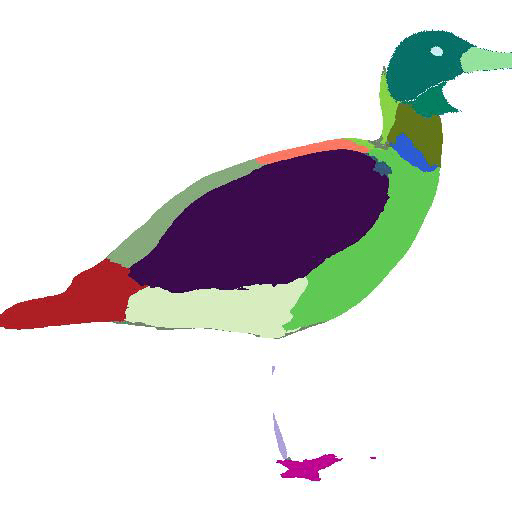}  \\
\end{tabular}

\begin{tabular}{cc}
 \includegraphics[height=\hhfl,width=\wwfl, trim=0 0 0 0,clip]{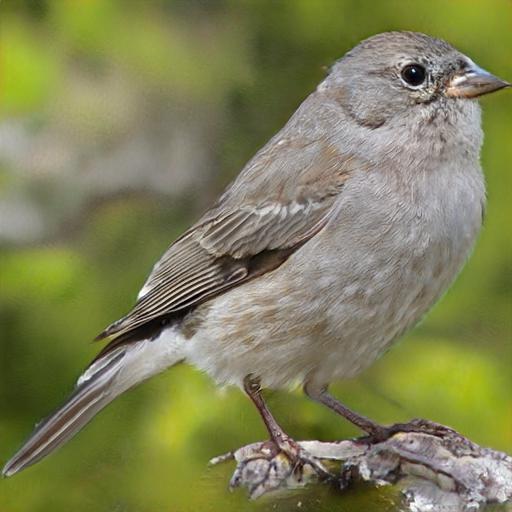} & \includegraphics[height=\hhfl,width=\wwfl, trim=0 0 0 0,clip]{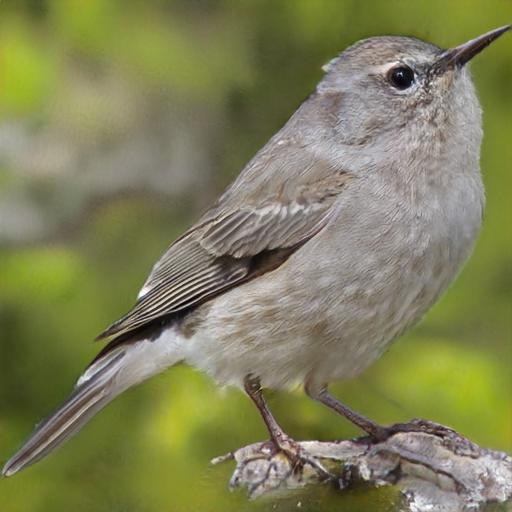}  \\
 \includegraphics[height=\hhfl,width=\wwfl, trim=0 0 0 0,clip]{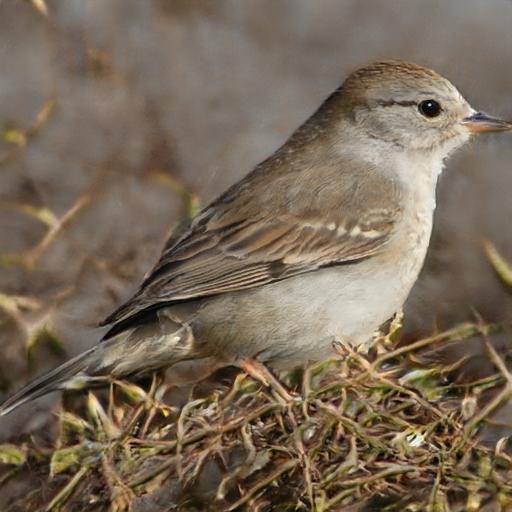} & \includegraphics[height=\hhfl,width=\wwfl, trim=0 0 0 0,clip]{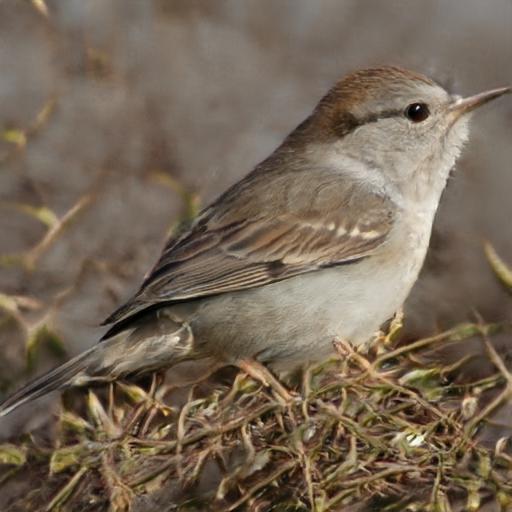}  \\
\end{tabular}

\caption{\footnotesize  {\bf Raising head editing.} \textit{First row}: Image and mask pair to learn editing vector. Images are images before editing and after editing. Segmentation masks are before editing and target segmentation mask after manual modification. \textit{Second and third rows}: Applying the learnt edit on new images.}
\label{fig:bird_headup}
\vspace{-3mm}
\end{figure}

\newcommand\hh{3.7cm}
\newcommand\ww{3.7cm}
\begin{figure}
\vspace{-1.5cm}
\addtolength{\tabcolsep}{-10pt}
\hspace{1mm}
\begin{tabular}{cccc}
 \includegraphics[height=\hh,width=\ww, trim=0 0 0 0,clip]{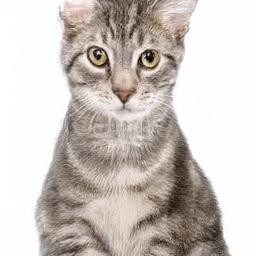} & \includegraphics[height=\hh,width=\ww, trim=0 0 0 0,clip]{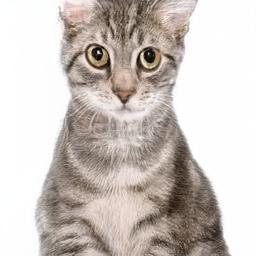}  &
 \includegraphics[height=\hh,width=\ww, trim=0 0 0 0,clip]{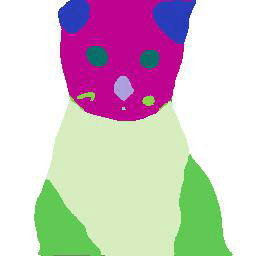} & \includegraphics[height=\hh,width=\ww, trim=0 0 0 0,clip]{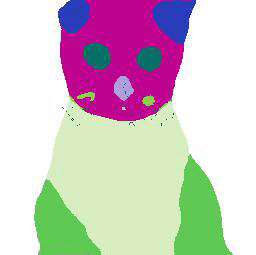} \\

 \includegraphics[height=\hh,width=\ww, trim=0 0 0 0,clip]{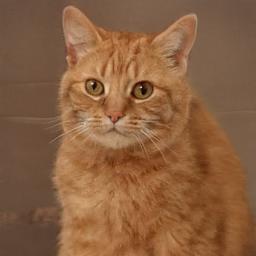} & \includegraphics[height=\hh,width=\ww, trim=0 0 0 0,clip]{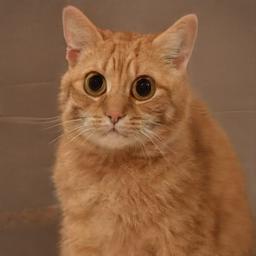}  &
 \includegraphics[height=\hh,width=\ww, trim=0 0 0 0,clip]{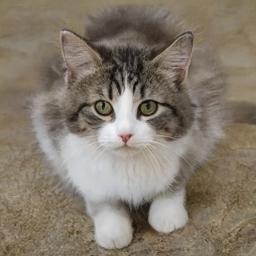} & \includegraphics[height=\hh,width=\ww, trim=0 0 0 0,clip]{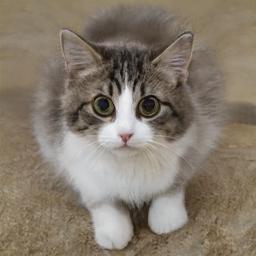} \\

 \includegraphics[height=\hh,width=\ww, trim=0 0 0 0,clip]{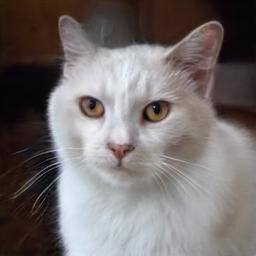} & \includegraphics[height=\hh,width=\ww, trim=0 0 0 0,clip]{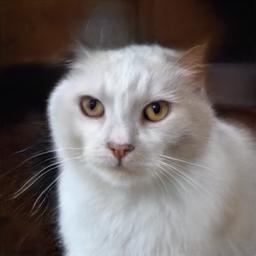}  &
 \includegraphics[height=\hh,width=\ww, trim=0 0 0 0,clip]{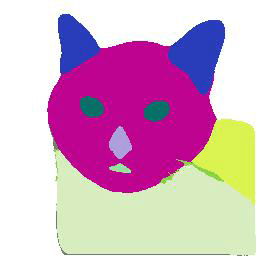} & \includegraphics[height=\hh,width=\ww, trim=0 0 0 0,clip]{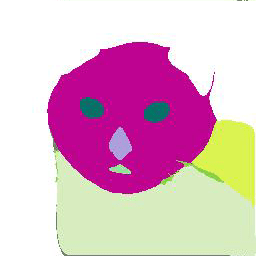} \\

\includegraphics[height=\hh,width=\ww, trim=0 0 0 0,clip]{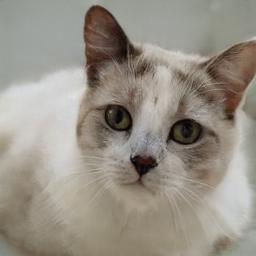} & \includegraphics[height=\hh,width=\ww, trim=0 0 0 0,clip]{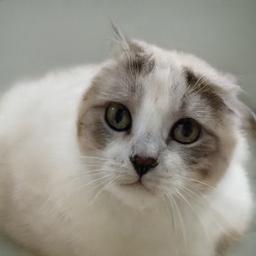}  &
 \includegraphics[height=\hh,width=\ww, trim=0 0 0 0,clip]{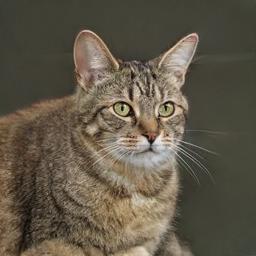} & \includegraphics[height=\hh,width=\ww, trim=0 0 0 0,clip]{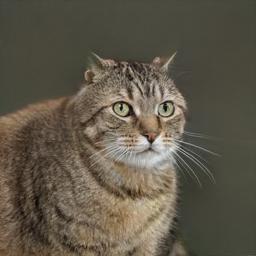} \\

  \includegraphics[height=\hh,width=\ww, trim=0 0 0 0,clip]{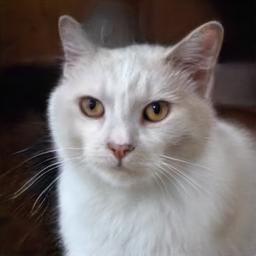} & \includegraphics[height=\hh,width=\ww, trim=0 0 0 0,clip]{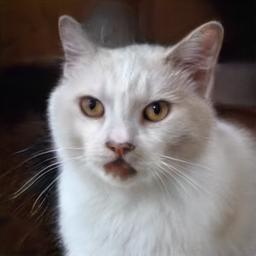}  &
 \includegraphics[height=\hh,width=\ww, trim=0 0 0 0,clip]{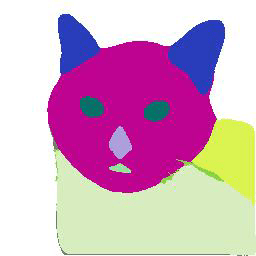} & \includegraphics[height=\hh,width=\ww, trim=0 0 0 0,clip]{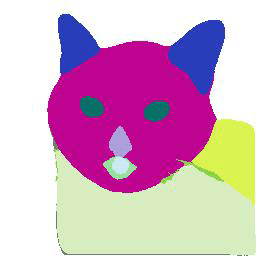} \\

\includegraphics[height=\hh,width=\ww, trim=0 0 0 0,clip]{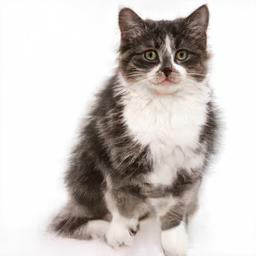} & \includegraphics[height=\hh,width=\ww, trim=0 0 0 0,clip]{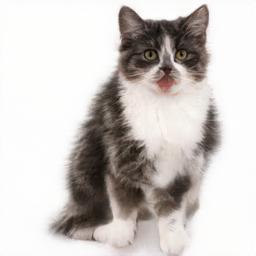}  &
 \includegraphics[height=\hh,width=\ww, trim=0 0 0 0,clip]{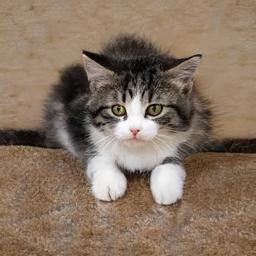} & \includegraphics[height=\hh,width=\ww, trim=0 0 0 0,clip]{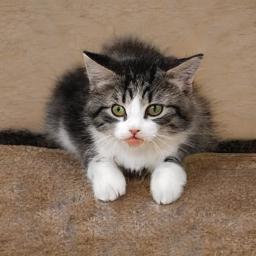} \\

\end{tabular}
\caption{\footnotesize  {\bf Cat image editing.} \textit{First and second rows}: {\bf Eye size editing.} \textit{Third and fourth rows}: {\bf Ear size editing.} \textit{Fifth and sixth rows}: {\bf Open mouth editing.} \textit{First, third, and fifth rows}: Image and mask pair to learn editing vector. Images are images before editing and after editing. Segmentation masks are before editing and target segmentation mask after manual modification. \textit{Second, fourth, and sixth rows}: Applying the learnt edits on new images.}
\label{fig:cat}
\vspace{-3mm}
\end{figure}



\end{document}